\pdfoutput=1
\documentclass[twoside,11pt]{article}

\usepackage{amsmath}
\usepackage{thmtools}
\usepackage[hidelinks]{hyperref}
\usepackage[nohyperref,preprint]{jmlr2e}

\usepackage{cleveref}
\usepackage{microtype}
\usepackage{enumitem}
\setlist[itemize,enumerate]{topsep=2pt, itemsep=1pt}
\usepackage{graphicx}
\usepackage{multicol}
\usepackage{nicematrix}

\makeatletter
\AddToHook{cmd/appendix/before}{\def\cref@section@alias{appendix}\def\cref@subsection@alias{appendix}}
\AddToHook{cmd/appendix/after}{%
  \counterwithin{theorem}{section}%
  \counterwithin{example}{section}%
  \counterwithin{algorithm}{section}%
}
\makeatother

\usepackage{algorithm}
\usepackage{algpseudocode}
\newcommand{\cmark}{\tikz\draw[scale=0.2, line width=1.5pt, color=green!80!black] (0,0) -- (0.3,-0.4) -- (1, 0.5);}
\newcommand{\xmark}{\tikz\draw[scale=0.2, line width=1.5pt, color=red] (0,0) -- (1,-1) (0,-1) -- (1,0);}

\usepackage{xcolor}

\usepackage{centernot}
\usepackage{mathtools}
\usepackage{bm}
\usepackage{tcolorbox}
\usepackage{booktabs}
\usepackage{adjustbox}
\usepackage{subcaption}
\usepackage{ifthen}
\usepackage{amsfonts}
\usepackage{amssymb}
\usepackage{xparse}

\Crefformat{equation}{Equation~#2#1#3}

\newcounter{issuecount}
\newcounter{todocount}
\NewDocumentCommand{\todo}{o m}{%
  \stepcounter{todocount}%
  \par\noindent%
  \textcolor{red!80!white}{%
    \textbf{{To-Do}
      \roman{todocount}%
      \IfNoValueTF{#1}{.}{ (#1).}}} {\textcolor{red!80!white}{#2}}
}
\NewDocumentCommand{\inlinetodo}{o m}{%
  \stepcounter{todocount}%
  \textcolor{blue}{%
    \textbf{{To-Do}
      \roman{todocount}%
      \IfNoValueTF{#1}{.}{ (#1).}}} {\textcolor{blue!80!white}{#2}}
}
\newcounter{notecount}
\NewDocumentCommand{\note}{o m}{%
  \stepcounter{notecount}%
  \par\noindent%
  \textcolor{blue}{%
    \textbf{{Note}
      \roman{notecount}%
      \IfNoValueTF{#1}{.}{ (#1).}}} {\textcolor{blue!80!white}{#2}}
}
\NewDocumentCommand{\inlinenote}{o m}{%
  \stepcounter{notecount}%
  \textcolor{blue}{%
    \textbf{{Note}
      \roman{notecount}%
      \IfNoValueTF{#1}{.}{ (#1).}}} {\textcolor{blue!80!white}{#2}}
}

\newcommand{\model}[1]{\textbf{\texttt{#1}}}
\newcommand{\ie}{i.e.}
\newcommand{\eg}{e.g.}
\renewcommand{\vec}[1]{\ensuremath\bm{{#1}}}

\newcommand{\targets}[2]{\operatorname{AT}(#1, #2)}

\DeclareMathOperator*{\argmin}{arg\,min}

\newcommand{\graph}[1]{\ensuremath\mathcal{#1}}

\DeclarePairedDelimiter\parentheses{\lparen}{\rparen}
\newcommand{\pa}[1]{\operatorname{Pa} \parentheses*{#1}}
\newcommand{\ch}[1]{\operatorname{Ch} \parentheses*{#1}}

\newcommand{\scm}[1]{\ensuremath\mathcal{#1}}

\newcommand{\subscm}[3][]{
  \ifthenelse{\equal{#1}{}}
  {\ensuremath{\mathcal{#2}\left[#3\right]}}
  {\ensuremath{\mathcal{#2}^{#1}\left[#3\right]}}
}

\newcommand{\rel}[1]{\set{R}_{{#1}}}
\newcommand{\relhat}[1]{\hat{\set{R}}_{{#1}}}
\newcommand{\rse}{\model{RSE}}
\newcommand{\rses}{\model{qRSE}}

\newcommand{\mise}[2]{\operatorname{MiSE}(#1,#2)}
\newcommand{\block}[1]{\set{B}_{{#1}}}

\newcommand{\rdirect}{{\xrightarrow{\mbox{\tiny $\rel{}$}}}}
\newcommand{\rpdirect}{{\xrightarrow{\mbox{\tiny $\rel{}'$}}}}

\newcommand{\real}{\ensuremath\mathbb{R}}

\newcommand{\partitions}[1]{\operatorname{Partitions}(#1)}
\newcommand{\set}[1]{\ensuremath\bm{{#1}}}
\newcommand{\size}[1]{\ensuremath{|{#1}|}}

\newcommand{\setdef}[1]{\ensuremath\left\{{#1}\right\}}

\newcommand{\dom}[1]{\ensuremath\mathcal{D} (#1)}

\newcommand{\dist}[1]{\ensuremath{\mathbb{P}_{#1}}}

\newcommand{\mat}[1]{{\ensuremath\boldsymbol{\mathbf{#1}}}}
\newcommand{\norm}[1]{{\ensuremath\|{#1}\|_2}}

\newcommand{\tr}{\ensuremath{\top}}
\newcommand{\inv}{\ensuremath{{-1}}}

\usepackage{tikz}
\usetikzlibrary{calc}
\usetikzlibrary{arrows.meta}
\usetikzlibrary{positioning}
\usetikzlibrary{fit}
\usetikzlibrary{matrix}
\usetikzlibrary{decorations.pathreplacing}
\usetikzlibrary{backgrounds}

\ShortHeadings{Sufficient Extensions of Causal Abstractions}{Massidda et al.}
\firstpageno{1}

\begin{document}

\title{Learning Sufficient Extensions of Linear Causal Abstractions}

\author{%
\name Riccardo Massidda
\email riccardo.massidda@di.unipi.it \\
\addr Dipartimento di Informatica, Università di Pisa
\AND
\name Davide Bacciu
\email davide.bacciu@unipi.it \\
\addr Dipartimento di Informatica, Università di Pisa
\AND
\name Sara Magliacane
\email sara.magliacane@gmail.com \\
\addr
Saarland University \& University of Amsterdam
}

\maketitle

\begin{abstract}
  Causal abstraction summarizes complex low-level causal models into smaller and more interpretable abstract models. Learning abstractions from data usually requires substantial abstract-level knowledge, such as the abstract graph, high-level or interventional samples, or a complete map from low-level to abstract variables. We instead consider abstracting linear Additive Noise Models without high-level data, where only the low-level causal graph and a subset of the abstract variables are known. Since a consistent abstract model might not be causally sufficient even when the low-level model is, we introduce the class of \emph{minimal sufficient extensions} (\model{MiSE}) of the initial abstract variables, which guarantee causal sufficiency without redundant variables. We then propose the Relevant Sufficiency Extender (\model{RSE}) algorithm, which provably enumerates exactly the \model{MiSE}. Its data-driven variant, \model{dataRSE}, uses low-level samples and the abstraction function of the initial variables to recover an extension with a consistent abstract model and the abstraction function of the new variables. We prove that at most one extension satisfies a non-collinearity condition on the abstract coefficients, and that \model{dataRSE} recovers it under exact regression. On synthetic abstractions satisfying this condition, \model{dataRSE} recovers the missing abstract variables and the abstraction function from finite samples, supporting abstract causal effect estimation.
\end{abstract}

\section{Introduction}%
\label{sec:introduction}

In many real-world settings, such as macroeconomics~\citep{stoker_aggregation_2010}, climatology~\citep{chalupka2016unsupervised}, brain imaging~\citep{Dubois2020CausalMO}, and neural network interpretability~\citep{bereska2024mechanistic}, researchers gather data at a lower level of \emph{abstraction} than the actual phenomena they aim to understand. In these scenarios, the causal effects among low-level variables may be uninteresting or too complex to interpret even for domain experts. Researchers are instead usually interested in the causal effects between a smaller number of interpretable high-level causal variables. Thus, defining abstract variables and their relations to the low-level measurements is in these contexts a prerequisite for causal effect estimation.

Causal abstraction~\citep{rubenstein2017causal,beckers2019abstracting} formalizes the relation between high-level and low-level causal models. Intuitively, causal abstraction requires a well-designed high-level variable to aggregate low-level variables while preserving interventional distributions. In other terms, intervening at the lower level and abstracting the realizations must be equivalent to performing an adequate intervention directly on the abstract model. This property is known as \emph{interventional consistency}~\citep{rubenstein2017causal,zennaro2023jointly} and enables reasoning, planning, and estimating causal queries on an interpretable abstract model, with conclusions consistent with the underlying model.

A key open challenge in causal abstraction is learning abstract variables from low-level information. In practice, choosing which low-level variables to aggregate is a non-trivial task that requires in-depth expert knowledge. Yet, existing approaches either rely on ground-truth aggregation information, explicitly~\citep{zennaro2023jointly,dyer2024interventionally,xia2024neural} or implicitly from abstract realizations~\citep{felekis2024causal,massidda2024learning,dacunto2025causal}, or fix the abstract graph a priori and require low-level interventional data~\citep{kekic2024targeted,zhu2024unsupervised}. In all cases, every abstract variable must be specified before learning, either directly or through data.

Compared to previous work, we consider partially specified abstractions. We address the problem of determining an abstract linear causal model from a \emph{strict subset} of the abstract variables and the low-level causal graph, without any high-level data or low-level interventional data. This setting covers, for instance, scenarios where expert knowledge can define the abstract treatment and the abstract outcome of interest, leaving all remaining high-level variables unspecified. However, even when the low-level model is causally sufficient~\citep{spirtes2000causation}, the induced abstract model is typically not, since the unaggregated low-level variables might confound the abstract variables defined by the experts. Causal sufficiency is a fundamental property which allows, for instance, the identification of causal effects through well-defined adjustment sets~\citep{smucler2022efficient}. We therefore address the problem of defining the missing abstract variables, without any prior information on their aggregation, to then ensure abstract causal sufficiency. Overall, our contributions are the following:
\begin{enumerate}
    \item%
    We study the problem of extending a subset of high-level variables to obtain a causally sufficient abstract model. We prove that this problem is  underdetermined, in general, \ie, multiple extensions of the provided variables can entail a causally sufficient model. We therefore introduce a notion of minimality and define the class of extensions of the provided variables that guarantee causal sufficiency and are minimal with respect to inclusion. We name this class \model{MiSE}, for \textbf{Mi}nimal \textbf{S}ufficient \textbf{E}xtension (\Cref{sec:mise}).
    \item%
    We introduce \model{RSE}, for \textbf{R}elevant \textbf{S}ufficiency \textbf{E}xtender, an algorithm that returns all and only minimal sufficient extensions from a strict subset of high-level variables and the low-level graph, and prove its correctness and completeness (\Cref{sec:rsealg}).
    \item%
    We extend \model{RSE} into \model{dataRSE}, which uses low-level samples to filter the \model{MiSE} down to an extension with a consistent abstract model, recovering both the relevant variables and the abstraction function (\Cref{sec:datarse}). We prove that at most one such extension satisfies a non-collinearity condition on the abstract coefficients, and that \model{dataRSE} recovers it under the assumption of exact regression. We evaluate \model{dataRSE} on synthetic abstractions from which we hide every abstract variable except the treatment and the outcome. We show that it recovers both the relevant variables and the abstraction function from a finite number of samples, and that the learned abstraction supports the estimation of abstract causal effects (\Cref{sec:experiments}).
\end{enumerate}

\section{Background}%
\label{sec:background}

We denote each variable~$V$ in plain uppercase and realizations as~$v\in\dom{V}$, where $\dom{V}$ denotes the domain of $V$. Similarly, we denote sets of variables~$\set{V}$ in bold, their realizations as $\vec{v}\in\dom{\set{V}}$, where $\dom{\set{V}}$ is the Cartesian product of the domains of the individual variables, and their cardinality as $\size{\set{V}}$. We employ Structural Causal Models to represent causal relations among random variables~\citep{pearl2000causality}.

\begin{definition}[SCM]
    A Structural Causal Model is a tuple $\scm{M}=(\set{X},\set{E},\set{f},\dist{\set{E}})$, where
    \begin{itemize}
        \item $\set{X}$ is a set of $\size{\set{X}}=d$ distinct endogenous variables,
        \item $\set{E}$ is a set of $\size{\set{E}}=d$ distinct exogenous variables,
        \item ${f_X\colon\dom{\pa{X}\cup\{E_X\}}\to\dom{X}}$
    is
    a function
    that determines the value
    of the variable $X\in\set{X}$
    given its parents $\pa{X}\subset\set{X}$
    and the exogenous noise term $E_X\in\set{E}$,
    \item $\dist{\set{E}}$ is the joint distribution over $\set{E}$.
    \end{itemize}
\end{definition}

We overload the notation and denote by $\scm{M}\colon\dom{\set{E}}\to\dom{\set{X}}$ the push-forward function of the SCM~$\scm{M}$, \ie, the map from the exogenous to the endogenous values. To each SCM~$\scm{M}$, we associate an Acyclic Directed Mixed Graph $\graph{G}_{\set{X}}$ over variables $\set{X}$ that contains an edge $X_i\to X_j$ if and only if $X_i\in\pa{X_j}$. Furthermore, we denote hidden confounding between two variables with a bidirected edge~$X_i\leftrightarrow X_j$ if and only if their exogenous terms are not marginally independent. We say that an SCM is \emph{causally sufficient} if its graph has no bidirected edges, \ie, if every common cause of two or more endogenous variables is itself endogenous~\citep{spirtes2000causation,peters2017elements}. Whenever the functions $\set{f}$ are linear, and the contribution of the exogenous terms is additive, we refer to an SCM as a linear Additive Noise Model (ANM), and we assume that each exogenous term has a finite, non-zero variance. Formally, we denote as $\mat{W}\in\real^{d\times d}$ the coefficient matrix, such that
\begin{equation}
    f_{X_j}(\vec{e}, \vec{x})
    =
    \sum_{X_i \in\pa{X_j}}
    w_{ij} \cdot x_i + e_j.
\end{equation}

We use SCMs to model interventions on a system and estimate causal effects among variables. In particular, we focus on hard interventions, which set the value of a variable to a constant, and on total causal effects over linear ANMs~\citep{henckel2022graphical}.

\begin{definition}[Do Intervention]
    A do intervention~$i=\operatorname{do}(\set{V}\gets\vec{v})$
    on an SCM~$\scm{M}=(\set{X}, \set{E},\set{f},\dist{\set{E}})$
    is a tuple composed of a subset
    of variables~$\set{V}\subseteq\set{X}$
    and values~$\vec{v}\in\dom{\set{V}}$.
    An intervention $i$ induces
    an intervened SCM%
    ~$\scm{M}^i=(\set{X},\set{E},\set{f}^i,\dist{\set{E}})$
    such that, for any $X\in\set{X}$,
    \begin{align}
       f_X^i(\cdot) =
       \begin{cases}
        f_X(\cdot) &\text{if } X \not\in{\set{V}}\\
        v_X &\text{otherwise.}
       \end{cases}
    \end{align}
\end{definition}

\begin{definition}[Total Causal Effect]
    Let $\scm{M}=(\set{X}, \set{E},\set{f},\dist{\set{E}})$
    be an SCM and $X_1, X_2\in\set{X}$ be two endogenous variables.
    Then, the total causal effect of $X_1$ on $X_2$ is defined as
    \begin{equation}
        \operatorname{TCE}(X_1, X_2) =\frac{\partial}{\partial x_1}
        \mathbb{E}\left[
        X_2 \mid \operatorname{do}(X_1 \gets x_1)
        \right]
    \end{equation}
\end{definition}

We define causal abstraction~\citep{beckers2019abstracting} between two linear ANMs as a linear transformation $\mat{T}\in\real^{d \times b}$ from low-level to high-level variables~\citep{massidda2024learning}. In particular, we focus on \emph{strong} $\mat{T}$-abstraction, where the low-level model must implement all possible interventions on the high-level model. We always assume faithfulness and the absence of cancelling paths in all the considered linear ANMs, as we detail in \Cref{app:t-abstraction}.

\begin{definition}[Strong $\mat{T}$-abstraction]
\label{def:tabstraction}
Let $\scm{L}=(\set{X}, \set{E}, \set{f}, \dist{\set{E}})$ and $\scm{H}=(\set{Y}, \set{U}, \set{g}, \dist{\set{U}})$ be two linear ANMs
with intervention sets $\mathcal{I}$ and $\mathcal{J}$,
where $\mathcal{I} = \setdef{\operatorname{do}(\set{V}\gets\vec{v}) \mid \set{V}\subseteq\set{X}, \vec{v}\in\dom{\set{V}}}$
and $\mathcal{J} = \setdef{\operatorname{do}(\set{V}\gets\vec{v}) \mid \set{V}\subseteq\set{Y}, \vec{v}\in\dom{\set{V}}}$.
Then, given a surjective linear transformation $\mat{T}\in\real^{d\times b}$, where $\size{\set{X}}=d$ and $\size{\set{Y}}=b$, $\scm{H}$ is a strong $\mat{T}$-abstraction of $\scm{L}$ if and only if there exists a linear transformation $\mat{S}\in\real^{d \times b}$ from low-level to abstract exogenous terms and, for any abstract intervention $j\in\mathcal{J}$, there exists a low-level intervention $i\in\mathcal{I}$
such that
\begin{equation}\label{eq:consistency}
    \forall \vec{e}\in\dom{\set{E}} . \quad
    \scm{L}^i(\vec{e})\mat{T}
    =
    \scm{H}^{j}(\vec{e}\mat{S}).
\end{equation}
\end{definition}

In a $\mat{T}$-abstraction, the relevant set $\rel{Y}$ is the subset of low-level variables composing the high-level variable $Y$. Formally, it holds $y=\vec{x}_{\rel{Y}}\vec{t}_Y$, where the abstraction vector $\vec{t}_Y\in\real^{\size{\rel{Y}}}$ is the submatrix of $\mat{T}$ containing the non-zero coefficients of the column of $Y$. Relevant sets are necessarily disjoint in strong $\mat{T}$-abstractions~\citep{massidda2024learning}.

\begin{definition}[Relevant Variables]
    Let $\mat{T}\in\real^{d \times b}$ be a linear transformation from low-level variables $\set{X}$ to high-level variables $\set{Y}$. Then, for each high-level variable $Y_j\in\set{Y}$, we define its relevant variables as
    $\rel{Y_j} = \setdef{X_i \mid t_{ij} \neq 0}$,
    and we denote by $\rel{}=\setdef{\rel{Y}}_{Y\in\set{Y}}$ the indexed set collecting the relevant sets of every high-level variable.
\end{definition}

Given a graph $\graph{G}$ over nodes $\set{X}$, we denote a directed path as an ordered sequence of nodes $\vec{p}=(X_{p_1}, X_{p_2}, \ldots, X_{p_n})$ such that $X_{p_i}\to X_{p_{i+1}}\in\graph{G}$ for all $i\in[1,n-1]$. We denote as $\pa{X,\graph{G}}$ and $\ch{X,\graph{G}}$ the parents and children of a node $X$ in $\graph{G}$, omitting the graph when it is clear from the context. We consider every node to be both an ancestor and a descendant of itself, and we refer to the ancestors and descendants of $X$ other than $X$ itself as its \emph{proper} ancestors and descendants. \citet{massidda2024learning} define a $\mat{T}$-direct path as a directed path in the low-level graph that does not cross any relevant variable except the path's source and target. However, relevant variables are induced by the non-zero entries of the abstraction matrix $\mat{T}$, regardless of their values. To highlight the dependence of these paths on the relevant variables~$\rel{}$ only, we refer to $\mat{T}$-direct paths as $\rel{}$-direct.

\begin{definition}[$\rel{}$-directness]
    Let $\scm{L}$ be a linear ANM on variables $\set{X}$
    inducing the graph $\graph{G}_{\set{X}}$
    and let $\rel{}$ be an indexed set of relevant variables
    over a set of abstract variables $\set{Y}$.
    Then,
    we say that a directed path $\vec{p}=(X_{p_1},\ldots, X_{p_n})$ in $\graph{G}_{\set{X}}$
    is $\rel{}$-direct if and only if
    any variable $X_i\in\vec{p}_{2:n-1}$
    in the path
    is not relevant to any abstract variable $Y\in\set{Y}$.
    We denote the existence of a $\rel{}$-direct path from $X_i$ to $X_j$ in the low-level graph as $X_i \rdirect X_j\in\graph{G}_{\set{X}}$.
\end{definition}

By only considering a low-level graph $\graph{G}_{\set{X}}$ and a set of relevant variables $\rel{}$, we can uniquely determine the abstract graph $\graph{G}_{\set{Y}}$ of any linear ANM~$\scm{H}$ abstracting $\scm{L}$ with relevant sets $\rel{}$. A set of relevant variables $\rel{}$ is graphically consistent on a low-level graph $\graph{G}_{\set{X}}$ if and only if the induced abstract graph $\graph{G}_{\set{Y}}$ respects necessary graphical conditions defined by $\rel{}$-direct paths~\citep{massidda2024learning}. If $\rel{}$ is not graphically consistent on $\graph{G}_{\set{X}}$, then no abstract model $\scm{H}$ can be a strong $\mat{T}$-abstraction of $\scm{L}$ with relevant sets $\rel{}$.

\begin{definition}[Graphical Consistency]%
\label{def:wellformed}
    Let $\scm{L}$ be a linear ANM on variables $\set{X}$
    with graph $\graph{G}_{\set{X}}$
    and let $\rel{}$ be an indexed set of relevant variables
    over a set of abstract variables $\set{Y}$.
    We define the abstract graph $\graph{G}_{\set{Y}}$
    as the graph over $\set{Y}$ such that
    \begin{equation}
    Y_1 \to Y_2 \in \graph{G}_{\set{Y}}
    \iff
    Y_1 \neq Y_2 \land
    \exists X_1\in\rel{Y_1}, X_2\in\rel{Y_2} \mathrm{\ s.t.\ } X_1 \rdirect X_2 \in\graph{G}_{\set{X}}.
    \end{equation}
    Then, we define the set of relevant variables $\rel{}$ as graphically consistent on $\graph{G}_{\set{X}}$ if and only if
    \begin{equation}
    Y_1 \to Y_2 \in \graph{G}_{\set{Y}}
    \implies
    \forall X_1\in\rel{Y_1}, \exists X_2\in\rel{Y_2} \mathrm{\ s.t.\ }
    X_1 \rdirect X_2 \in\graph{G}_{\set{X}}.
    \end{equation}
\end{definition}

Given the relevant variables and the low-level graph, we can also determine which low-level variables contribute to each abstract variable. In particular, a relevant variable only contributes to its own abstract variable, while a non-relevant variable contributes to every abstract variable that it reaches before crossing a relevant variable.

\begin{definition}[Blocks]%
\label{def:blocks}
    Let $\rel{}$ be an indexed set of relevant variables over a set of abstract variables $\set{Y}$ on a low-level graph $\graph{G}_{\set{X}}$.
    Then, for each abstract variable $Y\in\set{Y}$, we define its block $\block{Y}$ as the set of low-level variables $X\in\set{X}$ such that either $X\in\rel{Y}$, or $X$ is not relevant to \emph{any} abstract variable and $X\rdirect X'\in\graph{G}_{\set{X}}$ for some $X'\in\rel{Y}$.
    We denote by $\block{}=\setdef{\block{Y}}_{Y\in\set{Y}}$ the indexed set collecting the blocks of every abstract variable.
\end{definition}

Intuitively, the block $\block{Y}$ of an abstract variable $Y$ collects all the low-level variables targeting it. Therefore, we define the \emph{abstract targets} of a low-level variable $X\in\set{X}$ as the set of abstract variables whose blocks contain it, \ie, $\targets{X}{\block{}}=\setdef{Y\in\set{Y}\mid X\in\block{Y}}$. We say that $X$ is \emph{intersected} if it is in two or more blocks, \ie, if it has at least two abstract targets $\size{\targets{X}{\block{}}}\geq 2$. Furthermore, whenever $\scm{H}$ is a strong $\mat{T}$-abstraction of $\scm{L}$ with relevant sets $\rel{}$, the block $\block{Y}$ coincides with the support of the column of $\mat{S}$ corresponding to the abstract exogenous term $U_Y$~\citep[Lemma 3]{massidda2024learning}, \ie, $u_Y=\vec{e}_{\block{Y}}\vec{s}_Y$, where $\vec{s}_Y\in\real^{\size{\block{Y}}}$ has non-zero entries. Hence, blocks identify the low-level exogenous terms that compose each abstract exogenous term.

\section{Theory of Causally Sufficient Abstractions}
\label{sec:mise}

\begin{figure}[t]
  \centering
  \begin{minipage}[b]{0.42\textwidth}
    \centering\small
    \begin{tikzpicture}[
        lnode/.style={draw, circle, thick, minimum size=0.6cm, inner sep=0pt, font=\tiny},
        fbox/.style={draw, very thick, rounded corners=3pt, inner sep=8pt}
    ]
      \node[lnode] (x1) at (0.0, 3.7) {$X_1$};
      \node[lnode] (x2) at (1.0, 3.7) {$X_2$};
      \node[lnode] (x3) at (2.0, 3.7) {$X_3$};
      \node[lnode, fill=blue!30!white] (x4) at (1.0, 2.5) {$X_4$};
      \node[lnode, fill=blue!30!white] (x5) at (2.0, 2.5) {$X_5$};
      \node[lnode, fill=red!40!white]   (x6)  at (0.0, 1.0) {$X_6$};
      \node[lnode]                      (x8)  at (1.5, 1.0) {$X_8$};
      \node[lnode, fill=green!40!white] (x10) at (3.0, 1.0) {$X_{10}$};
      \node[lnode, fill=red!40!white]   (x7)  at (0.0, 0.0) {$X_7$};
      \node[lnode]                      (x9)  at (1.5, 0.0) {$X_9$};
      \node[lnode, fill=green!40!white] (x11) at (3.0, 0.0) {$X_{11}$};
      \begin{pgfonlayer}{background}
        \node[fbox, draw=red!60!black, fill=red!10!white,
              fit=(x6)(x7),
              label={[text=red!60!black]left:$\rel{Y_1}$}] {};
        \node[fbox, draw=green!60!black, fill=green!10!white,
              fit=(x10)(x11),
              label={[text=green!60!black]right:$\rel{Y_2}$}] {};
        \node[fbox, draw=blue!60!black, fill=blue!10!white,
              fit=(x4)(x5),
              label={[text=blue!60!black]right:$\rel{Y_3}$}] {};
      \end{pgfonlayer}
      \draw[-{Latex},gray] (x1) -- (x2);
      \draw[-{Latex},gray] (x1) -- (x4);
      \draw[-{Latex},gray] (x2) -- (x5);
      \draw[-{Latex},gray] (x3) -- (x4);
      \draw[-{Latex},gray] (x4) -- (x7);
      \draw[-{Latex},gray] (x4) to (x11);
      \draw[-{Latex},gray] (x4) -- (x5);
      \draw[-{Latex},gray] (x5) -- (x6);
      \draw[-{Latex},gray] (x5) -- (x11);
      \draw[-{Latex},gray] (x6) -- (x7);
      \draw[-{Latex},gray] (x6) -- (x8);
      \draw[-{Latex},gray] (x8) -- (x10);
      \draw[-{Latex},gray] (x7) -- (x8);
      \draw[-{Latex},gray] (x9) -- (x11);
    \end{tikzpicture}
    \subcaption{Low-level graph $\graph{G}_{\set{X}}$}%
    \label{fig:mise:low}
  \end{minipage}%
  \begin{minipage}[b]{0.27\textwidth}
    \centering\small
    \begin{tikzpicture}[minimum size=0.3cm]
      \node[draw, circle, thick, fill=red!40!white]  (Y1) {$Y_1$};
      \node[draw, circle, thick, fill=green!40!white,
            right=1.4cm of Y1]                       (Y2) {$Y_2$};
      \draw[-{Latex}]                      (Y1) -- (Y2);
      \draw[{Latex}-{Latex}, bend left=45] (Y1) to (Y2);
    \end{tikzpicture}
    \subcaption{Confounded $\graph{G}_{\set{Y}}$}%
    \label{fig:mise:confounded}
  \end{minipage}%
  \begin{minipage}[b]{0.27\textwidth}
    \centering\small
    \begin{tikzpicture}[minimum size=0.3cm]
      \node[draw, circle, thick, fill=blue!30!white] (Y3) {$Y_3$};
      \node[draw, circle, thick, fill=red!40!white,
            below left=0.9cm and 0.5cm of Y3]  (Y1) {$Y_1$};
      \node[draw, circle, thick, fill=green!40!white,
            below right=0.9cm and 0.5cm of Y3] (Y2) {$Y_2$};
      \draw[-{Latex}] (Y3) -- (Y1);
      \draw[-{Latex}] (Y3) -- (Y2);
      \draw[-{Latex}] (Y1) -- (Y2);
    \end{tikzpicture}
    \subcaption{Sufficient $\graph{G}_{\set{Y}'}$}%
    \label{fig:mise:sufficient}
  \end{minipage}
  \caption{
    Illustration of causal sufficiency in abstract graphs.
    \subref{fig:mise:low}.)~Low-level graph $\graph{G}_{\set{X}}$ with relevant sets
    $\rel{Y_1}=\{X_6,X_7\}$ (\textcolor{red!60!black}{red}),
    $\rel{Y_2}=\{X_{10},X_{11}\}$ (\textcolor{green!60!black}{green}), and
    $\rel{Y_3}=\{X_4,X_5\}$ (\textcolor{blue!60!black}{blue}).
    \subref{fig:mise:confounded}.)~Abstracting using only $\rel{Y_1}$ and $\rel{Y_2}$ yields a causally insufficient model since the shared ancestors in $\rel{Y_3}$ confound the causal relation between variables in $\rel{Y_1}$ and $\rel{Y_2}$.
    \subref{fig:mise:sufficient}.)~Including the relevant set $\rel{Y_3}$ induces a causally sufficient abstract model since relevant variables do not share non-relevant ancestors. We report a minimal example in App.~\ref{app:sufficiency}.
    }
  \label{fig:sufficiency}
\end{figure}

In practice, an abstract variable is defined by the set of low-level variables that it aggregates. Therefore, for a practitioner to define a set of abstract variables $\set{Y}$ over a low-level model is to define their corresponding sets of relevant variables $\rel{}=\setdef{\rel{Y}}_{Y\in\set{Y}}$. In this context, since the goal is to summarize a low-level model to perform causal inference at the abstract level, causal sufficiency is a highly desirable property to preserve. In fact, causal effects are always identifiable in \emph{causally sufficient} settings, \ie, when there are no latent confounders or selection bias~\citep{spirtes2000causation,pearl2000causality}. We show that abstract causal sufficiency depends on the overlap of the blocks induced by the relevant variables. Intuitively, whenever two abstract variables share low-level variables in their blocks, these low-level variables will confound the two abstract variables (\Cref{fig:sufficiency}). We formally prove that this condition is both sufficient and necessary, and consequently define a set of relevant variables as \emph{sufficient} whenever it induces non-overlapping blocks. Notably, the \emph{if} direction holds without \Cref{assumption:exogenousfaithfulness}, which only excludes cancellations among the exogenous terms of the abstract variables. We recall that this assumption mirrors our other faithfulness assumptions to avoid cancellations in the endogenous and abstraction functions.

\begin{restatable}[Abstract Causal Sufficiency]{proposition}{causalsufficiencyprop}%
\label{prop:causalsufficiency}
    Let a linear ANM $\scm{H}$ be a $\mat{T}$-abstraction of a causally sufficient linear ANM $\scm{L}$, and $\rel{}$ be the corresponding graphically consistent set of relevant variables. Then, under \Cref{assumption:exogenousfaithfulness}, the abstract ANM $\scm{H}$ is causally sufficient if and only if the relevant variables $\rel{}$ are sufficient, \ie, they induce blocks $\block{}$ of low-level variables such that
    \begin{equation}
    \block{Y_1} \cap \block{Y_2} = \emptyset
    \quad
    \forall Y_1, Y_2 \in\set{Y},\ Y_1 \neq Y_2.
    \end{equation}
\end{restatable}
\begin{proof}
    We report the proof in \Cref{proof:causalsufficiency}.
\end{proof}

Therefore, even if the low-level model is causally sufficient, a practitioner defining a set of abstract variables might still end up with an abstract model that is not causally sufficient. In this context, we study how to define additional abstract variables to preserve causal sufficiency. We recall that $\rel{}=\setdef{\rel{Y}}_{Y\in\set{Y}}$ is the indexed set collecting the relevant variables of every $Y\in\set{Y}$, and thus we define additional abstract variables through an extension~$\rel{}'\supset\rel{}$, \ie, by adding new relevant sets while leaving the existing ones unchanged.

A trivial solution would be to associate a new abstract variable to each low-level variable that is not in a unique block, \ie, the low-level variables that are confounding the existing abstract variables. However, since a non-trivial abstract model should effectively compress the low-level one, we also want to avoid inserting abstract variables that are not needed for sufficiency. Therefore, we define the class of \emph{minimal sufficient extensions}, or \model{MiSE}\footnote{Pronounced \texttt{[miz]}, as in the French phrase \emph{mise en place}.}. Intuitively, a set of relevant variables is in the \model{MiSE} whenever it is sufficient and none of its new abstract variables can be removed without breaking sufficiency.

\begin{definition}[Minimal Sufficient Extensions]%
\label{def:mise}
    Given a graphically consistent set of relevant variables $\rel{}$ on a causal graph $\graph{G}_{\set{X}}$, we define the set of minimal sufficient extensions $\mise{\rel{}}{\graph{G}_{\set{X}}}$ such that, for each set of relevant variables $\rel{}'\in\mise{\rel{}}{\graph{G}_{\set{X}}}$, it holds that
	(i.) $\rel{}'\supseteq\rel{}$,
	(ii.) $\rel{}'$ is sufficient and graphically consistent on $\graph{G}_{\set{X}}$, and
	(iii.) any $\rel{}\subseteq\rel{}''\subset\rel{}'$ is not sufficient on $\graph{G}_{\set{X}}$.
\end{definition}

\begin{example}[\model{MiSE}]%
\label{ex:mise}

Consider the low-level graph in subfigure \emph{(a.)} with relevant sets $\rel{Y_1}=\setdef{X_4}$ and $\rel{Y_2}=\setdef{X_5}$. The abstract graph is not causally sufficient since $X_1, X_2, X_3\in\block{Y_1}\cap\block{Y_2}$. Introducing a new abstract variable $Y_3$ with relevant set $\setdef{X_2, X_3}$ produces a causally sufficient abstract model, since $X_1\in\block{Y_3}'$ only (\emph{b.}). By definition, $\rel{}'\in\mise{\rel{}}{\graph{G}_{\set{X}}}$.\\[1em]
{\small\centering
\begin{tikzpicture}[node distance=1.5cm, minimum size=0.3cm, scale=0.94, transform shape]
\node[draw, circle] (X1) {$X_1$};
\node[draw, circle, right of=X1, yshift=1cm] (X2) {$X_2$};
\node[draw, circle, right of=X1, yshift=-1cm] (X3) {$X_3$};
\node[draw, circle, right of=X2, fill=red!40!white] (X4) {$X_4$};
\node[draw, circle, right of=X3, fill=green!40!white] (X5) {$X_5$};
\draw[-{Latex[length=2mm]}] (X1) -- (X2);
\draw[-{Latex[length=2mm]}] (X1) -- (X3);
\draw[-{Latex[length=2mm]}] (X2) -- (X4);
\draw[-{Latex[length=2mm]}] (X2) -- (X5);
\draw[-{Latex[length=2mm]}] (X3) -- (X4);
\draw[-{Latex[length=2mm]}] (X3) -- (X5);
\draw[-{Latex[length=2mm]}] (X4) -- (X5);
\begin{pgfonlayer}{background}
\node[draw=red!60!black, very thick, rounded corners=3pt, fill=red!10!white, fit=(X4), inner sep=4pt, label={[text=red!60!black]right:{\scriptsize$\rel{Y_1}$}}] {};
\node[draw=green!60!black, very thick, rounded corners=3pt, fill=green!10!white, fit=(X5), inner sep=4pt, label={[text=green!60!black]right:{\scriptsize$\rel{Y_2}$}}] {};
\end{pgfonlayer}
  \node[left of=X1, xshift=-2.0em] {%
    \begin{tikzpicture}
      \node[anchor=west] at (0.0,1.2) {Consistency: \cmark};
      \node[anchor=west] at (0.0,2.8) {\large\emph{(a.)}};
      \node[anchor=west] at (0.0,0.8) {Sufficiency: \xmark};
      \node[anchor=west] at (0.0,0.4) {MiSE: \xmark};
    \end{tikzpicture}
  };
\end{tikzpicture}
\hfill
\begin{tikzpicture}[node distance=1.5cm, minimum size=0.3cm, scale=0.94, transform shape]
\node[draw, circle] (X1) {$X_1$};
\node[draw, circle, right of=X1, yshift=1cm, fill=blue!40!white] (X2) {$X_2$};
\node[draw, circle, right of=X1, yshift=-1cm, fill=blue!40!white] (X3) {$X_3$};
\node[draw, circle, right of=X2, fill=red!40!white] (X4) {$X_4$};
\node[draw, circle, right of=X3, fill=green!40!white] (X5) {$X_5$};
\draw[-{Latex[length=2mm]}] (X1) -- (X2);
\draw[-{Latex[length=2mm]}] (X1) -- (X3);
\draw[-{Latex[length=2mm]}] (X2) -- (X4);
\draw[-{Latex[length=2mm]}] (X2) -- (X5);
\draw[-{Latex[length=2mm]}] (X3) -- (X4);
\draw[-{Latex[length=2mm]}] (X3) -- (X5);
\draw[-{Latex[length=2mm]}] (X4) -- (X5);
\begin{pgfonlayer}{background}
\node[draw=blue!60!black, very thick, rounded corners=3pt, fill=blue!10!white, fit=(X2)(X3), inner sep=4pt, label={[text=blue!60!black, yshift=4pt]above left:{\scriptsize$\rel{Y_3}$}}] {};
\node[draw=red!60!black, very thick, rounded corners=3pt, fill=red!10!white, fit=(X4), inner sep=4pt, label={[text=red!60!black]right:{\scriptsize$\rel{Y_1}$}}] {};
\node[draw=green!60!black, very thick, rounded corners=3pt, fill=green!10!white, fit=(X5), inner sep=4pt, label={[text=green!60!black]right:{\scriptsize$\rel{Y_2}$}}] {};
\end{pgfonlayer}
\node[left of=X1, xshift=-2.0em] {%
  \begin{tikzpicture}
    \node[anchor=west] at (0.0,1.2) {Consistency: \cmark};
    \node[anchor=west] at (0.0,2.8) {\large\emph{(b.)}};
    \node[anchor=west] at (0.0,0.8) {Sufficiency: \cmark};
    \node[anchor=west] at (0.0,0.4) {MiSE: \cmark};
  \end{tikzpicture}
};
\end{tikzpicture}
}\\[1.5em]
On the other hand, a further abstract variable $Y_4$ with relevant set $\rel{Y_4}=\setdef{X_1}$ would still lead to a graphically consistent and causally sufficient relevant set $\rel{}''$ (\emph{c.}). However, since $\rel{}''\supset\rel{}'$, this would not be minimal. Intuitively, we do not need $Y_4$ to adjust the causal effect among the abstract variables introduced by the original relevant set~$\rel{}$. If $X_2$ and $X_3$ are instead assigned to two distinct abstract variables, $X_1$ also needs to become relevant to avoid being in the intersection of the new abstract variables (\emph{d.}).\\[1.5em]
{\small
\begin{tikzpicture}[node distance=1.5cm, minimum size=0.3cm, scale=0.94, transform shape]
\node[draw, circle, fill=orange!40!white] (X1) {$X_1$};
\node[draw, circle, right of=X1, yshift=1cm, fill=blue!40!white] (X2) {$X_2$};
\node[draw, circle, right of=X1, yshift=-1cm, fill=blue!40!white] (X3) {$X_3$};
\node[draw, circle, right of=X2, fill=red!40!white] (X4) {$X_4$};
\node[draw, circle, right of=X3, fill=green!40!white] (X5) {$X_5$};
\draw[-{Latex[length=2mm]}] (X1) -- (X2);
\draw[-{Latex[length=2mm]}] (X1) -- (X3);
\draw[-{Latex[length=2mm]}] (X2) -- (X4);
\draw[-{Latex[length=2mm]}] (X2) -- (X5);
\draw[-{Latex[length=2mm]}] (X3) -- (X4);
\draw[-{Latex[length=2mm]}] (X3) -- (X5);
\draw[-{Latex[length=2mm]}] (X4) -- (X5);
\begin{pgfonlayer}{background}
\node[draw=orange!60!black, very thick, rounded corners=3pt, fill=orange!10!white, fit=(X1), inner sep=4pt, label={[text=orange!60!black]left:{\scriptsize$\rel{Y_4}$}}] {};
\node[draw=blue!60!black, very thick, rounded corners=3pt, fill=blue!10!white, fit=(X2)(X3), inner sep=4pt, label={[text=blue!60!black, yshift=4pt]above left:{\scriptsize$\rel{Y_3}$}}] {};
\node[draw=red!60!black, very thick, rounded corners=3pt, fill=red!10!white, fit=(X4), inner sep=4pt, label={[text=red!60!black]right:{\scriptsize$\rel{Y_1}$}}] {};
\node[draw=green!60!black, very thick, rounded corners=3pt, fill=green!10!white, fit=(X5), inner sep=4pt, label={[text=green!60!black]right:{\scriptsize$\rel{Y_2}$}}] {};
\end{pgfonlayer}
\node[left of=X1, xshift=-2.0em] {%
  \begin{tikzpicture}
    \node[anchor=west] at (0.0,1.2) {Consistency: \cmark};
    \node[anchor=west] at (0.0,2.8) {\large\emph{(c.)}};
    \node[anchor=west] at (0.0,0.8) {Sufficiency: \cmark};
    \node[anchor=west] at (0.0,0.4) {MiSE: \xmark};
  \end{tikzpicture}
};
\end{tikzpicture}\hfill
\begin{tikzpicture}[node distance=1.5cm, minimum size=0.3cm, scale=0.94, transform shape]
\node[draw, circle, fill=orange!40!white] (X1) {$X_1$};
\node[draw, circle, right of=X1, yshift=1cm, fill=yellow!40!white] (X2) {$X_2$};
\node[draw, circle, right of=X1, yshift=-1cm, fill=blue!40!white] (X3) {$X_3$};
\node[draw, circle, right of=X2, fill=red!40!white] (X4) {$X_4$};
\node[draw, circle, right of=X3, fill=green!40!white] (X5) {$X_5$};
\draw[-{Latex[length=2mm]}] (X1) -- (X2);
\draw[-{Latex[length=2mm]}] (X1) -- (X3);
\draw[-{Latex[length=2mm]}] (X2) -- (X4);
\draw[-{Latex[length=2mm]}] (X2) -- (X5);
\draw[-{Latex[length=2mm]}] (X3) -- (X4);
\draw[-{Latex[length=2mm]}] (X3) -- (X5);
\draw[-{Latex[length=2mm]}] (X4) -- (X5);
\begin{pgfonlayer}{background}
\node[draw=orange!60!black, very thick, rounded corners=3pt, fill=orange!10!white, fit=(X1), inner sep=4pt, label={[text=orange!60!black]left:{\scriptsize$\rel{Y_4}$}}] {};
\node[draw=yellow!60!black, very thick, rounded corners=3pt, fill=yellow!10!white, fit=(X2), inner sep=4pt, label={[text=yellow!60!black]left:{\scriptsize$\rel{Y_3}$}}] {};
\node[draw=blue!60!black, very thick, rounded corners=3pt, fill=blue!10!white, fit=(X3), inner sep=4pt, label={[text=blue!60!black]left:{\scriptsize$\rel{Y_5}$}}] {};
\node[draw=red!60!black, very thick, rounded corners=3pt, fill=red!10!white, fit=(X4), inner sep=4pt, label={[text=red!60!black]right:{\scriptsize$\rel{Y_1}$}}] {};
\node[draw=green!60!black, very thick, rounded corners=3pt, fill=green!10!white, fit=(X5), inner sep=4pt, label={[text=green!60!black]right:{\scriptsize$\rel{Y_2}$}}] {};
\end{pgfonlayer}
\node[left of=X1, xshift=-2.0em] {%
  \begin{tikzpicture}
    \node[anchor=west] at (0.0,1.2) {Consistency: \cmark};
    \node[anchor=west] at (0.0,2.8) {\large\emph{(d.)}};
    \node[anchor=west] at (0.0,0.8) {Sufficiency: \cmark};
    \node[anchor=west] at (0.0,0.4) {MiSE: \cmark};
  \end{tikzpicture}
};
\end{tikzpicture}
}
\end{example}

We remark that minimality in \Cref{def:mise} is with respect to inclusion, \ie, it excludes extensions in which some new relevant set can be dropped, as in \Cref{ex:mise} \emph{(c.)}. It does not compare extensions by their number of abstract variables, nor by coarsening of their relevant sets. For instance, in \Cref{ex:mise}, both the extension \emph{(b.)}, with one new abstract variable, and the extension \emph{(d.)}, with three new abstract variables, are minimal sufficient extensions. Selecting among the elements of the \model{MiSE} requires a further criterion. In \Cref{sec:datarse}, we discuss how the parameters of the low-level model restrict this choice.

Not all sets of relevant variables allow for a graphically consistent and sufficient extension. Hence, the \model{MiSE} of a graphically consistent but insufficient set of relevant variables can be empty (\Cref{ex:suffandgraph} in \Cref{app:sufficiency}). However, whenever an extension exists, the \model{MiSE} preserves block assignments for low-level variables already pertaining to a unique abstract variable. Intuitively, if a low-level variable is in a unique block, then, in all minimal sufficient extensions, the low-level variable will still be uniquely in the same block. In other words, the new abstract variables of a minimal sufficient extension are only built over low-level variables that are intersected under $\rel{}$, \ie, where they are needed to resolve confounding between the existing abstract variables.

\begin{restatable}[Target Stability]{lemma}{singletargetlemma}
\label{lemma:singletarget}
    Let $\rel{}$ be a graphically consistent set of relevant variables on a low-level graph $\graph{G}_{\set{X}}$. Then, for any minimal sufficient extension $\rel{}'\in\mise{\rel{}}{\graph{G}_{\set{X}}}$ and any low-level variable $X\in\set{X}$, it holds that
    \begin{equation}
    \size{\targets{X}{\block{}}} = 1
    \implies
    \targets{X}{\block{}} = \targets{X}{\block{}'},
    \end{equation}
    where $\block{}$ and $\block{}'$ are respectively the blocks induced by $\rel{}$ and $\rel{}'$ on $\graph{G}_{\set{X}}$.
\end{restatable}
\begin{proof}
    We report the proof in \Cref{proof:singletarget}.
\end{proof}

A low-level variable that is not an ancestor of any relevant variable can never lie on a $\rel{}$-direct path towards an abstract variable. Therefore, it will never affect block membership and consequently abstract causal sufficiency. Without loss of generality, since we can safely ignore these variables, we can always \emph{clean} a low-level graph by removing every variable that is not an ancestor of some relevant variable in the initial set of abstract variables $\set{Y}$. In other terms, each low-level variable should be in the block of some abstract variable. From here on, we assume that all considered low-level graphs are clean.

\begin{restatable}[Clean Low-Level Graph]{assumption}{cleanassumption}%
\label{assumption:clean}
    Let $\rel{}$ be a set of relevant variables on a low-level graph $\graph{G}_{\set{X}}$ with variables $\set{X}$ and $\block{}$ the corresponding blocks. Then, we assume that $\graph{G}_{\set{X}}$ is clean, \ie, for every $X\in\set{X}$, it holds that $\targets{X}{\block{}} \neq \emptyset$.
\end{restatable}

\section{Computing Minimal Sufficient Extensions}%
\label{sec:rsealg}

In this section, we introduce the Relevant Sufficiency Extender (\model{RSE}, \Cref{alg:rse}), a recursive algorithm that computes the \model{MiSE} of a set of relevant variables. At each recursive call, \model{RSE} identifies a jointly targeted set of abstract variables and introduces new abstract variables to resolve its confounding. The recursion terminates once every branch reaches a graphically consistent and causally sufficient extension, which is provably a minimal sufficient extension.

Practitioners typically provide at least two abstract variables, e.g., a treatment $Y_1$ and an outcome of interest $Y_2$, by specifying their relevant variables $\rel{Y_1}$ and $\rel{Y_2}$. \model{RSE} can however handle the more general setting where $b \geq 2$ abstract variables are specified. We do \emph{not} require the functional form of the abstraction function, only the groups of low-level variables composing each provided abstract variable. We now discuss the main steps of \model{RSE} and prove its correctness and completeness. In \Cref{app:rseexecution}, we report an execution of \model{RSE}.

\begin{algorithm}[t]%
\caption{Relevant Sufficiency Extender (\rse{})}%
\label{alg:rse}
\begin{algorithmic}[1]
\Require Relevant set~$\rel{}$ and low-level graph~$\graph{G}_{\set{X}}$
\Ensure $\set{S}=$ Minimal Sufficient Extensions of $\rel{}$ on $\graph{G}_{\set{X}}$

\If{not $\operatorname{GraphicalConsistency}(\rel{}, \graph{G}_{\set{X}})$}
\Comment{\Cref{alg:graphcons}}
    \State \Return $\emptyset$
\EndIf
\State $\block{} \gets \operatorname{Blocks}(\rel{}, \graph{G}_{\set{X}})$ \Comment{\Cref{alg:blocks}}
\State $\set{I} \gets \operatorname{JointlyTargetedSets}(\block{})$
\Comment{\Cref{alg:intersections}}
\If{$\set{I} = \emptyset$}
    \State \Return $\setdef{\rel{}}$
    \Comment{Causally sufficient solution}
\EndIf
\State $\set{V}_i \gets \argmin_{\set{V}_j\in\set{I}} \size{\set{V}_j}$
\Comment{Select the smallest jointly targeted set}
\State $\set{Z}_i \gets \operatorname{NewRelevants}(\set{V}_i, \set{I}, \block{}, \graph{G}_{\set{X}})$ \Comment{\Cref{alg:newrelevants}}
\State $\set{S} \gets \emptyset$
\For{$\set{P}\in\partitions{\set{Z}_i}$}
\Comment{Iterate partitions of the new relevant variables}
    \State $\set{S} \gets \set{S} \cup \operatorname{RSE}(\rel{}\cup\set{P}, \graph{G}_{\set{X}})$
    \Comment{Collect results from each candidate partition}
\EndFor
\State \Return $\set{S}$
\Comment{Return the \model{MiSE} of $\rel{}$ on $\graph{G}_{\set{X}}$}
\end{algorithmic}
\end{algorithm}

\paragraph{Graphical Consistency Check (ll. 1--3).}
First, \model{RSE} checks whether the set of relevant variables $\rel{}$ is graphically consistent on the low-level graph $\graph{G}_{\set{X}}$, as we detail in \Cref{alg:graphcons}. As we prove in the following lemma, if $\rel{}$ is not graphically consistent on $\graph{G}_{\set{X}}$, then no extension of $\rel{}$ is graphically consistent. Therefore, \model{RSE} can terminate the recursion and return an empty set of solutions from the current branch.

\begin{restatable}[Inconsistency of Extensions]{lemma}{gcextensionlemma}%
\label{lemma:gc-extension}
    Let $\rel{}$ be a set of relevant variables
    that is not graphically consistent
    on a low-level graph $\graph{G}_{\set{X}}$.
    Then, any set of relevant variables
    $\rel{}'\supseteq\rel{}$
    is not graphically consistent on $\graph{G}_{\set{X}}$.
\end{restatable}
\begin{proof}
    We report the proof in \Cref{proof:gc-extension}.
\end{proof}

\paragraph{Blocks and Jointly Targeted Sets (ll. 4--9).}
After checking the graphical consistency of the input $\rel{}$, the algorithm must identify the confounded abstract variables. First, \model{RSE} computes the blocks~$\block{}$ (Line 4) induced by the relevant variables $\rel{}$ on the low-level graph $\graph{G}_{\set{X}}$, as we detail in \Cref{alg:blocks}. Then, it computes the sets of abstract variables that are targeted by the same low-level variable. Intuitively, if a low-level variable $X$ lies in multiple blocks, it confounds its abstract targets $\targets{X}{\block{}}$ in the high-level model.
\begin{definition}[Jointly Targeted Sets]%
\label{def:jointtargets}
    Let $\rel{}$ be a set of relevant variables on a low-level graph $\graph{G}_{\set{X}}$, with blocks $\block{}$ and abstract variables $\set{Y}$.
    A set of abstract variables $\set{V}\subseteq\set{Y}$ is \emph{jointly targeted} if it coincides with the abstract targets of some intersected variable, \ie, $\set{V}=\targets{X}{\block{}}$ for some $X\in\set{X}$ with $\size{\targets{X}{\block{}}}\geq 2$.
    We denote the set of jointly targeted sets as
    \begin{equation}%
    \label{eq:intersections}
        \set{I}=\setdef{\targets{X}{\block{}}\mid X\in\set{X} \land \size{\targets{X}{\block{}}}\geq 2}.
    \end{equation}
\end{definition}
In practice, we compute $\set{I}$ by building a reverse index of the blocks, as we detail in \Cref{alg:intersections}. If there are no jointly targeted sets, then the set of relevant variables is sufficient and \model{RSE} returns it (Lines 6--8). Otherwise, writing $\set{I}=\setdef{\set{V}_1,\ldots,\set{V}_n}$, the algorithm selects a jointly targeted set $\set{V}_i\in\set{I}$ containing the smallest number of abstract variables (Line 9). The completeness of \model{RSE} does not depend on which minimum-size jointly targeted set is selected when multiple are available. Selecting a jointly targeted set of minimum size guarantees that at least one new relevant variable exists (\Cref{lemma:nonemptyz}), so that every recursive call strictly increases the set of relevant variables and \model{RSE} terminates.

\begin{algorithm}%
\caption{New Relevant Variables}%
\label{alg:newrelevants}
\begin{algorithmic}[1]
\Require Selected jointly targeted set~$\set{V}_i\in\set{I}$, all jointly targeted sets~$\set{I}$ (\Cref{def:jointtargets}), blocks~$\block{}$, and low-level graph~$\graph{G}_{\set{X}}$
\Ensure Set of new relevant variables $\set{Z}_i\subseteq\set{X}$

\State Intersected $\gets \emptyset$
\Comment{Collect intersected variables}
\For {$\set{V}_j\in\set{I}$}
\State Intersected $\gets$ Intersected $\cup \bigcap_{Y\in\set{V}_j} \block{Y}$
\EndFor
\State NotIntersected $\gets \set{X}\setminus\text{Intersected}$
\Comment{Variables with a unique abstract target}
\State $\set{Z}_i \gets \emptyset$, $\set{W} \gets \bigcap_{Y\in\set{V}_i} \block{Y}$
\For{$X\in\set{W}$} \Comment{Iterate over low-level variables targeting $\set{V}_i$}
\If{$\ch{X, \graph{G}_{\set{X}}}\cap \text{NotIntersected}\neq\emptyset$}
\State $\set{Z}_i=\set{Z}_i\cup \setdef{X}$ \Comment{$X$ is intersected and has a non-intersected child}
\EndIf
\EndFor
\State \Return $\set{Z}_i$
\end{algorithmic}
\end{algorithm}

\paragraph{New Relevant Variables (l. 10).}
In the previous step, \model{RSE} selected a jointly targeted set of confounded abstract variables $\set{V}_i\in\set{I}$. Therefore, the algorithm must now identify which low-level variables must become relevant to avoid confounding between these variables in the abstract graph. To this end, we prove the existence of a set of low-level variables $\set{Z}_i$ that must be relevant in any minimal sufficient extension of $\rel{}$. We call this subset $\set{Z}_i$ the set of \emph{new relevant variables} on $\set{V}_i$. Intuitively, a low-level variable $X\in\set{X}$ is a new relevant variable if it is in the blocks of all the variables in $\set{V}_i$ and has at least one outgoing edge to a variable that is not intersected. Since the low-level graph is clean (\Cref{assumption:clean}), every variable has at least one abstract target, hence a variable that is not intersected has exactly one abstract target. In \Cref{alg:newrelevants}, we report how to compute the new relevant variables.
\begin{definition}[New Relevant Variable]
\label{def:newrelevants}
    Let $\rel{}$ be a graphically consistent set of relevant variables
    on a low-level graph $\graph{G}_{\set{X}}$ and
    $\set{V}_i\in\set{I}$ a jointly targeted set (\Cref{def:jointtargets}).
    Then, we define the new relevant variables on $\set{V}_i$ as
\begin{equation}%
\label{eq:newrelevants}
    \set{Z}_i
    =
    \setdef{
    X \in \bigcap_{Y \in \set{V}_i} \block{Y}
    \mid
    \exists X'\in\ch{X, \graph{G}_{\set{X}}}.
    \size{\targets{X'}{\block{}}} = 1
    }.
\end{equation}
\end{definition}
Each new relevant variable $X\in\set{Z}_i$ confounds the variables in $\set{V}_i$ in a way that no minimal sufficient extension can resolve without making $X$ itself relevant. Indeed, $X$ has a child $X'$ with a unique abstract target $Y$, which remains its unique target in every minimal sufficient extension (\Cref{lemma:singletarget}), so that $X$ stays in the block of $Y$. At the same time, $X$ also reaches the relevant variables of another abstract variable in $\set{V}_i$, and thus lies in the block of some abstract variable other than $Y$, unless it becomes relevant. Since extensions do not alter the existing relevant sets, $X$ must then be relevant to a new abstract variable.

\begin{restatable}[Necessity of New Relevant Variables]{lemma}{newrelevantslemma}%
\label{lemma:newrelevants}
    Let $\rel{}$ be a graphically consistent set of relevant variables on a low-level graph $\graph{G}_{\set{X}}$ and $\set{Z}_i$ be the set of new relevant variables on a jointly targeted set $\set{V}_i\in\set{I}$.
    Then, for any minimal sufficient extension
    $\rel{}'\in\mise{\rel{}}{\graph{G}_{\set{X}}}$
    inducing abstract variables $\set{Y}'\supseteq\set{Y}$,
    it holds that
    $\set{Z}_i\subseteq\bigcup_{Y\in\set{Y}'\setminus\set{Y}}\rel{Y}'$.
\end{restatable}
\begin{proof}
We report the proof in \Cref{proof:newrelevants}.
\end{proof}

\paragraph{New Abstract Variables (ll. 11--14).}
Finally, \model{RSE} partitions the new relevant variables~$\set{Z}_i$ to form new abstract variables (Lines 11--14). Indeed, in every minimal sufficient extension, no variable in $\set{Z}_i$ shares a relevant set with a variable outside $\set{Z}_i$. Building on this intuition, we prove that, by recursively exploring the partitions, \model{RSE} returns the set of solutions corresponding to the minimal sufficient extensions of $\rel{}$ on $\graph{G}_{\set{X}}$.

\begin{restatable}[RSE Correctness and Completeness]{theorem}{rsethm}%
\label{thm:rse}
    Let $\rel{}$ be a
    graphically consistent
    set of relevant variables
    on a low-level graph~$\graph{G}_{\set{X}}$.
    Then,
    $\model{RSE}(\rel{},\graph{G}_{\set{X}})
    =
    \mise{\rel{}}{\graph{G}_{\set{X}}}$.
\end{restatable}
\begin{proof}
    We report the proof in \Cref{app:rse}.
\end{proof}

\subsection{Optimized Partitioning}

Checking graphical consistency requires multiple visits of the low-level graph. However, we can anticipate graphical inconsistencies by inspecting the children of each new relevant variable $X\in\set{Z}_i$ that are not intersected. In fact, by target stability (\Cref{lemma:singletarget}), each of these children keeps its unique abstract target in every minimal sufficient extension, so that $X$ still reaches the relevant variables of their targets after the extension. Therefore, for each $X\in\set{Z}_i$, we define the set $\set{Q}_X$ of abstract variables that $X$ targets through these children as
\begin{equation}
\label{eq:abstargets}
    \set{Q}_X = \setdef{Y\in\set{Y}\mid\exists X'\in\ch{X}.\targets{X'}{\block{}}=\setdef{Y}}.
\end{equation}

Since graphical consistency requires all the relevant variables of an abstract variable to reach the same abstract variables, two new relevant variables with different sets $\set{Q}_X$ cannot be relevant to the same abstract variable in any minimal sufficient extension.

\begin{restatable}[$Q_X$ Constraint]{lemma}{qxconstraintlemma}%
\label{lemma:qxconstraint}
    Let $\rel{}$ be a graphically consistent set of relevant variables on a low-level graph $\graph{G}_{\set{X}}$ and $\set{Z}_i$ be the new relevant variables on a jointly targeted set $\set{V}_i\in\set{I}$. Then, for any minimal sufficient extension $\rel{}'\in\mise{\rel{}}{\graph{G}_{\set{X}}}$ inducing abstract variables $\set{Y}'$, any abstract variable $Y\in\set{Y}'$, and any $X_1, X_2\in\set{Z}_i$, it holds
        $X_1, X_2\in\rel{Y}' \implies \set{Q}_{X_1}=\set{Q}_{X_2}$.
\end{restatable}
\begin{proof}
    We report the proof in \Cref{app:qxconstraint}.
\end{proof}

Building on \Cref{lemma:qxconstraint}, we can restrict the partitions explored by \model{RSE} at each recursive call. Instead of iterating over all the partitions of $\set{Z}_i$, we group the new relevant variables sharing the same set $\set{Q}_X$, and only iterate over the Cartesian product of the partitions of each group (\Cref{alg:rse_star}). We refer to this variant as \model{qRSE}. Every partition explored by \model{qRSE} is also a partition of $\set{Z}_i$, hence \model{qRSE} explores a subset of the recursive calls of \model{RSE} and is correct. Conversely, \model{qRSE} only discards partitions that cannot lead to any minimal sufficient extension of $\rel{}$, thus the completeness argument for \model{RSE} still holds (\Cref{app:rsestar}).

\begin{algorithm}%
\caption{Optimized Partitioning for the \rse{} algorithm (\rses{})}%
\label{alg:rse_star}
\begin{algorithmic}[1]
\Require New relevant variables~$\set{Z}_i$, blocks~$\block{}$, relevant set~$\rel{}$, and low-level graph~$\graph{G}_{\set{X}}$
\Ensure Minimal sufficient extensions $\set{S}$ reached through the partitions of $\set{Z}_i$
\State $\set{S} \gets \emptyset$
\For{$X\in\set{Z}_i$}
\State $\set{Q}_X \gets \emptyset$;
\Comment{Initialize empty targets}
\For{$X'\in\ch{X}$}
    \If{$\size{\targets{X'}{\block{}}}=1$}
    \Comment{$X'$ is in a unique block}
        \State $\set{Q}_X \gets \set{Q}_X \cup \targets{X'}{\block{}}$
        \Comment{Add the block of $X'$ to the targets of $X$}
    \EndIf
\EndFor
\EndFor
\For{$(j, \set{q}) \in \operatorname{Enumerate}(\operatorname{Unique}(\set{Q}))$}
    \State $\set{M}_j \gets \{X \in \set{Z}_i : \set{Q}_X = \set{q}\}$
    \Comment{Group variables with the same target set}
\EndFor
\For{$\set{P}\in\partitions{\set{M}_1}\times\cdots\times\partitions{\set{M}_k}$}
\Comment{Iterate grouped partitions}
    \State $\set{S} \gets \set{S} \cup \operatorname{qRSE}(\rel{}\cup\set{P}, \graph{G}_{\set{X}})$
\EndFor
\State \Return $\set{S}$
\end{algorithmic}
\end{algorithm}

\section{Data-Driven Filtering of Parametric Inconsistencies}%
\label{sec:datarse}

Since all the relevant sets in the \model{MiSE} are graphically consistent and sufficient, we might be tempted to only select the ones with the fewest abstract variables and discard the remaining ones. However, this does not ensure the existence of a consistent $\mat{T}$-abstraction built over the selected set of relevant variables, as we show in \Cref{ex:miseparams}. We define a set of relevant variables as \emph{parametrically consistent} if it leads to at least one consistent $\mat{T}$-abstraction. Therefore, while graphical consistency and sufficiency only relate to the low-level graph, parametric consistency also depends on the low-level structural mechanisms.
\begin{definition}[Parametric Consistency]%
\label{def:paramconsistency}
  Given a linear ANM $\scm{L}$, a set of relevant variables $\rel{}$ is parametrically consistent on $\scm{L}$ if there exists a linear ANM $\scm{H}$ and an abstraction function $\mat{T}$ with relevant variables $\rel{}$ such that $\scm{H}$ is a strong $\mat{T}$-abstraction of $\scm{L}$.
\end{definition}
\begin{example}[Parametric Inconsistency]%
\label{ex:miseparams}
  Given the depicted low-level linear ANM~$\scm{L}$ and a graphically consistent but not sufficient set of relevant variables $\rel{}$, we report two minimal sufficient extensions that satisfy sufficiency and graphical consistency. We can show that there exists a choice of the parameters of the low-level model such that we can define an abstract model built over $\rel{}'\supset\rel{}$, but not over $\rel{}''\supset\rel{}$, despite both being in the \model{MiSE} of $\rel{}$ on $\scm{L}$. Intuitively, this property depends on the rank of the low-level weights (\Cref{app:miseparams}).%
\begin{center}
\small
\begin{tikzpicture}[%
node distance=1.5cm,
scale=0.82, transform shape,
minimum size=0.3cm,
]
\node[draw, circle] (X1) {$X_1$};
\node[draw, circle, right of=X1] (X2) {$X_2$};
\node[draw, circle, right of=X2] (X3) {$X_3$};
\node[draw, circle, below of=X1, xshift=0.5cm, fill=blue!40!white] (X4) {$X_4$};
\node[draw, circle, below of=X3, xshift=-0.5cm, fill=red!40!white] (X5) {$X_5$};
\node[below of=X2, yshift=-1cm] (label) {\emph{a.} $\rel{}$ on $\scm{L}$};

\draw[-{Latex}] (X1) -- (X4);
\draw[-{Latex}] (X1) -- (X5);
\draw[-{Latex}] (X2) -- (X4);
\draw[-{Latex}] (X2) -- (X5);
\draw[-{Latex}] (X3) -- (X4);
\draw[-{Latex}] (X3) -- (X5);
\draw[-{Latex}] (X4) -- (X5);
\begin{pgfonlayer}{background}
\node[draw=blue!60!black, very thick, rounded corners=3pt, fill=blue!10!white, fit=(X4), inner sep=4pt, label={[text=blue!60!black]left:{\scriptsize$\rel{Y_T}$}}] {};
\node[draw=red!60!black, very thick, rounded corners=3pt, fill=red!10!white, fit=(X5), inner sep=4pt, label={[text=red!60!black]right:{\scriptsize$\rel{Y_O}$}}] {};
\end{pgfonlayer}
\end{tikzpicture}
\hspace{2em}
\begin{tikzpicture}[%
node distance=1.5cm,
scale=0.92, transform shape,
minimum size=0.3cm,
]
\node[draw, circle, fill=yellow!40!white] (X1) {$X_1$};
\node[draw, circle, right of=X1, fill=green!40!white] (X2) {$X_2$};
\node[draw, circle, right of=X2, fill=green!40!white] (X3) {$X_3$};
\node[draw, circle, below of=X1, xshift=0.5cm, fill=blue!40!white] (X4) {$X_4$};
\node[draw, circle, below of=X3, xshift=-0.5cm, fill=red!40!white] (X5) {$X_5$};
\node[below of=X2, yshift=-1cm] (label) {\emph{b.} $\rel{}'$ on $\scm{L}$};

\draw[-{Latex}] (X1) -- (X4);
\draw[-{Latex}] (X1) -- (X5);
\draw[-{Latex}] (X2) -- (X4);
\draw[-{Latex}] (X2) -- (X5);
\draw[-{Latex}] (X3) -- (X4);
\draw[-{Latex}] (X3) -- (X5);
\draw[-{Latex}] (X4) -- (X5);
\begin{pgfonlayer}{background}
\node[draw=yellow!60!black, very thick, rounded corners=3pt, fill=yellow!10!white, fit=(X1), inner sep=4pt, label={[text=yellow!60!black]above:{\scriptsize$\rel{Y_1}$}}] {};
\node[draw=green!60!black, very thick, rounded corners=3pt, fill=green!10!white, fit=(X2)(X3), inner sep=4pt, label={[text=green!60!black]above:{\scriptsize$\rel{Y_2}$}}] {};
\node[draw=blue!60!black, very thick, rounded corners=3pt, fill=blue!10!white, fit=(X4), inner sep=4pt, label={[text=blue!60!black]left:{\scriptsize$\rel{Y_T}$}}] {};
\node[draw=red!60!black, very thick, rounded corners=3pt, fill=red!10!white, fit=(X5), inner sep=4pt, label={[text=red!60!black]right:{\scriptsize$\rel{Y_O}$}}] {};
\end{pgfonlayer}
\end{tikzpicture}
\hspace{2em}
\small
\begin{tikzpicture}[%
node distance=1.5cm,
scale=0.92, transform shape,
minimum size=0.3cm,
]
\node[draw, circle, fill=orange!40!white] (X1) {$X_1$};
\node[draw, circle, right of=X1, fill=orange!40!white] (X2) {$X_2$};
\node[draw, circle, right of=X2, fill=orange!40!white] (X3) {$X_3$};
\node[draw, circle, below of=X1, xshift=0.5cm, fill=blue!40!white] (X4) {$X_4$};
\node[draw, circle, below of=X3, xshift=-0.5cm, fill=red!40!white] (X5) {$X_5$};
\node[below of=X2, yshift=-1cm] (label) {\emph{c.} $\rel{}''$ on $\scm{L}$};

\draw[-{Latex}] (X1) -- (X4);
\draw[-{Latex}] (X1) -- (X5);
\draw[-{Latex}] (X2) -- (X4);
\draw[-{Latex}] (X2) -- (X5);
\draw[-{Latex}] (X3) -- (X4);
\draw[-{Latex}] (X3) -- (X5);
\draw[-{Latex}] (X4) -- (X5);
\begin{pgfonlayer}{background}
\node[draw=orange!60!black, very thick, rounded corners=3pt, fill=orange!10!white, fit=(X1)(X2)(X3), inner sep=4pt, label={[text=orange!60!black]above:{\scriptsize$\rel{Y_3}$}}] {};
\node[draw=blue!60!black, very thick, rounded corners=3pt, fill=blue!10!white, fit=(X4), inner sep=4pt, label={[text=blue!60!black]left:{\scriptsize$\rel{Y_T}$}}] {};
\node[draw=red!60!black, very thick, rounded corners=3pt, fill=red!10!white, fit=(X5), inner sep=4pt, label={[text=red!60!black]right:{\scriptsize$\rel{Y_O}$}}] {};
\end{pgfonlayer}
\end{tikzpicture}
\end{center}
\end{example}

\subsection{Necessary Conditions for Parametric Consistency}

Unlike \model{RSE}, which solves a single jointly targeted set at a time, we now consider all the jointly targeted sets $\set{I}=\setdef{\set{V}_1,\ldots,\set{V}_n}$ induced by $\rel{}$ at once. Therefore, in the following, we consider the union of the jointly targeted sets, which we denote as $\set{V}=\bigcup_{i=1}^n\set{V}_i$, and the union of the corresponding sets of new relevant variables, which we denote as $\set{Z}=\bigcup_{i=1}^n\set{Z}_i$.

\model{RSE} extended an insufficient set of relevant variables while preserving graphical consistency. We now study how to also ensure parametric consistency. Therefore, we start by writing the necessary form for a structural equation of an abstract variable in a parametrically consistent extension. To this end, given a set of relevant variables $\rel{}$, for each confounded variable $Y_j\in\set{V}$, we denote as $\set{Z}^{(j)}=\setdef{X\in\set{Z}\mid Y_j\in\targets{X}{\block{}}}$ the subset of new relevant variables targeting $Y_j$. Then, we prove that in any parametrically consistent extension, each confounded variable $Y_j$ is a linear function of its existing abstract parents and of $\set{Z}^{(j)}$, with an additive exogenous term that is independent of both.

\begin{restatable}[Parametric Consistency]{theorem}{paramconsistencythm}
\label{thm:paramconsistency}
    Let $\rel{}$ be a graphically consistent set of relevant variables on a low-level ANM $\scm{L}$, inducing abstract variables $\set{Y}$, confounded variables $\set{V}\subset\set{Y}$, and new relevant variables $\set{Z}$. Then, any minimal sufficient extension $\rel{}'$, inducing abstract variables $\set{Y}'\supseteq\set{Y}$, is parametrically consistent on $\scm{L}$ only if there exist a coefficient matrix $\mat{M}\in\real^{\size{\set{Y}'}\times\size{\set{Y}'}}$ and an abstraction function $\mat{T}$ such that, for each $Y_j\in\set{V}$,
    \begin{align}%
    \label{eq:paramconsistency}
        Y_j &= \sum_{Y_i\in\pa{Y_j,\graph{G}_{\set{Y}}}} m_{ij} Y_i + \sum_{X_l\in\set{Z}^{(j)}} X_l a_{lj} + U_j,\\
        a_{lj} &= \sum_{Y_h\in\set{Y}'\setminus\set{Y}} t_{lh}m_{hj},
        \label{eq:composedabstraction}
    \end{align}
    where $\mat{A}=[a_{lj}]\in\real^{\size{\set{Z}}\times\size{\set{V}}}$, $\graph{G}_{\set{Y}}$ is the abstract graph induced by $\rel{}$ on $\scm{L}$, $m_{ij}$ might be zero for some $Y_i\in\pa{Y_j,\graph{G}_{\set{Y}}}$, and $U_j$ is independent of $\pa{Y_j,\graph{G}_{\set{Y}}}$ and of the low-level variables $\set{Z}^{(j)}$.
    Furthermore, it holds $a_{lj}=0$ for every $X_l\in\set{Z}\setminus\set{Z}^{(j)}$.
\end{restatable}
\begin{proof}
    We report the proof in \Cref{app:paramconsistency}.
\end{proof}

\subsection{Composed Abstraction Matrix}

We refer to the matrix $\mat{A}$ in \Cref{eq:composedabstraction} as the \emph{composed abstraction matrix}. Since $U_j$ is independent of the regressors of $Y_j$, the coefficients in \Cref{eq:paramconsistency} coincide with those of the linear regression of $Y_j$ on $\pa{Y_j,\graph{G}_{\set{Y}}}$ and $\set{Z}^{(j)}$, which is unique since these regressors are non-zero linear combinations of disjoint sets of low-level variables. Therefore, the composed abstraction matrix depends only on the low-level distribution, on $\rel{}$, and on the abstraction of $\set{Y}$, and it is the same for every parametrically consistent $\rel{}'$ in the \model{MiSE}.

Intuitively, the composed abstraction matrix captures the linear contribution of each new relevant variable~$X_l\in\set{Z}$ to each confounded variable~$Y_j\in\set{V}$. Therefore, since a new relevant variable only contributes through the abstract variable it is relevant to, two low-level variables can be relevant to the same abstract variable only if their rows in the composed abstraction matrix are collinear.

\begin{restatable}[Collinearity of Relevant Variables]{corollary}{collinearcor}%
\label{cor:collinear}
    Let $\rel{}$ be a graphically consistent but not sufficient set of relevant variables on a low-level ANM $\scm{L}$, with confounded variables $\set{V}$ and new relevant variables $\set{Z}$, and let $\mat{A}\in\real^{\size{\set{Z}}\times\size{\set{V}}}$ be the corresponding composed abstraction matrix.
    Then, any minimal sufficient extension $\rel{}'\in\mise{\rel{}}{\graph{G}_{\set{X}}}$ is parametrically consistent on $\scm{L}$ only if, for each new abstract variable $Y\in\set{Y}'\setminus\set{Y}$ with $\rel{Y}'\subseteq\set{Z}$ and each pair of relevant variables $X_l, X_{l'}\in\rel{Y}'$, it holds that
    $\mat{A}_{l,:}\propto\mat{A}_{l',:}$.
\end{restatable}
\begin{proof}
    We report the proof in \Cref{app:collinear}.
\end{proof}

In practice, assuming access to the composed abstraction matrix, we can filter out non-parametrically consistent extensions by checking the collinearity of its rows. Therefore, we consider any extension $\rel{}'$ satisfying these \emph{collinearity constraints}. Once an extension is selected, we can show that, for any new abstract variable defined by the extension, the composed abstraction matrix also determines its abstraction vector and the outgoing coefficients towards the confounded variables, up to scale and sign.

\begin{restatable}[Rank-One Decomposition]{proposition}{rankoneprop}
\label{prop:rankone}
    Let $\rel{}$ be a graphically consistent but not sufficient set of relevant variables on a low-level ANM $\scm{L}$, with confounded variables $\set{V}$ and new relevant variables $\set{Z}$, and let $\mat{A}\in\real^{\size{\set{Z}}\times\size{\set{V}}}$ be the corresponding composed abstraction matrix. Let $\rel{}'\in\mise{\rel{}}{\graph{G}_{\set{X}}}$ be parametrically consistent on $\scm{L}$ with abstraction function $\mat{T}$ and abstract coefficients $\mat{M}$.
    Then, for each new abstract variable $Y_h\in\set{Y}'\setminus\set{Y}$ with $\rel{Y_h}'\subseteq\set{Z}$, it holds that
    \begin{equation}
        \mat{A}_{\rel{Y_h}',\set{V}} = \vec{t}_h\vec{m}_{h,\set{V}}^\top\in\real^{\size{\rel{Y_h}'}\times \size{\set{V}}},
    \end{equation}
    where $\vec{t}_h\in\real^{\size{\rel{Y_h}'}}$ and $\vec{m}_{h,\set{V}}\in\real^{\size{\set{V}}}$ are non-zero vectors, hence, $\mat{A}_{\rel{Y_h}',\set{V}}$ has rank one.
\end{restatable}
\begin{proof}
    We report the proof in \Cref{app:rankone}.
\end{proof}

Since the composed abstraction matrix depends only on $\scm{L}$, $\rel{}$, and the abstraction of $\set{Y}$, the rank-one submatrix $\mat{A}_{\rel{Y_h}',\set{V}}=\vec{t}_h\vec{m}_{h,\set{V}}^\top$ is the same for every parametrically consistent extension $\rel{}'\in\mise{\rel{}}{\graph{G}_{\set{X}}}$ that contains the relevant set $\rel{Y_h}'$. Hence, all these extensions must agree on the abstraction vector $\vec{t}_h$ and on the outgoing coefficients $\vec{m}_{h,\set{V}}$ up to scale and sign, which we can then recover from the composed abstraction matrix. In practice, we fix scale and sign by computing the SVD of $\mat{A}_{\rel{Y_h}',\set{V}}$. The first left singular vector $\vec{u}_1$ gives the normalized abstraction vector $\vec{t}_h/\norm{\vec{t}_h}$, and the first right singular vector $\vec{v}_1$ gives the normalized outgoing coefficients $\vec{m}_{h,\set{V}}/\norm{\vec{m}_{h,\set{V}}}$. By convention, we assign a positive sign to the entry of $\vec{u}_1$ with the largest absolute value and assign the magnitude $\sigma_1$ of the first singular value to the outgoing coefficients, \ie, $\vec{m}_{h,\set{V}}=\sigma_1\vec{v}_1$.

\subsection{Maximal Partitioning}%
\label{subsec:maximalpartitioning}

Applying the rank-one decomposition of the composed abstraction matrix (\Cref{prop:rankone}) requires selecting an extension of the relevant variables composed by partitioning the new relevant variables $\set{Z}$. As we discussed, the collinearity constraints can effectively discard some extensions, but do not determine a unique partition. Therefore, we define the \emph{maximal partitioning} $\set{P}^\ast$ as the coarsest partition of $\set{Z}$ that satisfies the collinearity constraints, \ie, the one that groups together all and only the new relevant variables whose rows in $\mat{A}$ are collinear. We now study under which conditions the relevant sets in the maximal partitioning are contained in a parametrically consistent minimal sufficient extension.

In general, the entire maximal partitioning $\set{P}^\ast$ need not be contained in a parametrically consistent minimal sufficient extension. In fact, the composed abstraction matrix only captures the direct effects of the new relevant variables on the confounded variables $\set{V}$, and not the effects that a new abstract variable might have on \emph{other} new abstract variables. Hence, the maximal partitioning will merge the relevant variables of two new abstract variables that have collinear effects on $\set{V}$ but different effects on another new abstract variable. As we report in \Cref{app:mpinconsistency}, such a merge can lead to a set of relevant variables that is graphically consistent, sufficient, and minimal, but parametrically inconsistent.

Therefore, we focus on the relevant sets in $\set{P}^\ast$ inducing abstract variables whose children are all confounded variables, which we call \emph{unmediated} relevant sets. Intuitively, the effects of these abstract variables on $\set{V}$ are not mediated by other new abstract variables in any consistent extension. We then prove that, under a mild assumption on the outgoing coefficients of an extension, the unmediated relevant sets must be contained in that extension.

\begin{definition}[Unmediated Relevant Sets]%
\label{def:unmediated}
    Let $\rel{}$ be a set of relevant variables with confounded variables $\set{V}$ and maximal partitioning $\set{P}^\ast$, and let $\rel{}''=\rel{}\cup\set{P}^\ast$ induce the abstract graph $\graph{G}''$, where each $\set{P}\in\set{P}^\ast$ is the relevant set of a new abstract variable $Y_{\set{P}}$. Then, we define the \emph{unmediated} relevant sets as the relevant sets of the induced abstract variables whose children are all in $\set{V}$, \ie, $\set{P}^\ast_{\set{V}} = \setdef{\set{P}\in\set{P}^\ast \mid \ch{Y_{\set{P}}, \graph{G}''}\subseteq\set{V}}$.
\end{definition}

\begin{restatable}[Maximal Partitioning]{theorem}{maximalpartitioningthm}%
\label{thm:maximalpartitioning}
Let $\rel{}$ be a graphically consistent but not sufficient set of relevant variables on a low-level ANM $\scm{L}$, with confounded variables $\set{V}$, new relevant variables $\set{Z}$, maximal partitioning $\set{P}^\ast$, and unmediated relevant sets $\set{P}^\ast_{\set{V}}$.
Let $\rel{}'\in\mise{\rel{}}{\graph{G}_{\set{X}}}$ be parametrically consistent on $\scm{L}$, with $\scm{H}$ a strong $\mat{T}$-abstraction of $\scm{L}$ with relevant sets $\rel{}'$ and abstract coefficients $\mat{M}$, and let $\set{N}_{\set{V}}=\setdef{Y\in\set{Y}'\setminus\set{Y}\mid\ch{Y,\graph{G}_{\set{Y}'}}\subseteq\set{V}}$ be the new abstract variables whose children in $\graph{G}_{\set{Y}'}$ are all in $\set{V}$.
If the outgoing coefficients $\vec{m}_{h,\set{V}}\in\real^{\size{\set{V}}}$ of the new abstract variables towards $\set{V}$ satisfy
\begin{equation}
\label{eq:mpcondition}
\vec{m}_{s,\set{V}} \not\propto \vec{m}_{h,\set{V}}
\quad\text{for every } Y_s\in\set{N}_{\set{V}} \text{ and every new } Y_h\neq Y_s \text{ with } \rel{Y_h}'\subseteq\set{Z},
\end{equation}
then $\set{P}^\ast_{\set{V}}$ is non-empty and it holds $\set{P}^\ast_{\set{V}}=\setdef{\rel{Y}'\mid Y\in\set{N}_{\set{V}}}$. Hence, $\rel{}\cup\set{P}^\ast_{\set{V}}\subseteq\rel{}'$.
\end{restatable}
\begin{proof}
    We report the proof in \Cref{app:maximalpartitioning}.
\end{proof}

Notably, \Cref{thm:maximalpartitioning} does not assume knowledge of $\rel{}'$, since $\set{P}^\ast_{\set{V}}$ does not depend on it. Rather, it guarantees that, whenever some parametrically consistent minimal sufficient extension satisfies \Cref{eq:mpcondition}, the sets in $\set{P}^\ast_{\set{V}}$ are relevant sets of that extension. In particular, all such extensions share the relevant sets in $\set{P}^\ast_{\set{V}}$. In our experiments, we enforce \Cref{eq:mpcondition} with a numerical margin when sampling the abstract coefficients (\Cref{app:datagen}).

\subsection{Data-Driven RSE}%
\label{subsec:datarse}

Building on the necessary conditions for parametric consistency and on \model{RSE}, we introduce the \model{dataRSE} algorithm to search for a parametrically consistent minimal sufficient extension of $\rel{}$ (\Cref{alg:datarse}). Unlike \model{RSE}, which enumerates the whole \model{MiSE}, \model{dataRSE} returns a single extension by greedily aggregating new relevant variables in the maximal partitioning we defined in \Cref{subsec:maximalpartitioning}. In addition to the low-level graph and the relevant sets of $b\geq 2$ abstract variables of interest, as in \model{RSE}, \model{dataRSE} requires (i.) a dataset $\mat{X}$ of \emph{low-level} realizations and (ii.) the abstraction vector $\vec{t}_Y$ of each provided abstract variable $Y\in\set{Y}$, collected in the abstraction coefficients $\mat{T}$.

Therefore, \model{dataRSE} returns either a set of relevant variables $\rel{}'$ that extends $\rel{}$ and is graphically consistent and sufficient, together with its abstraction coefficients $\mat{T}'$, or $\bot$. Furthermore, given the exact composed abstraction matrix and its maximal partitioning at each recursive call, \model{dataRSE} recovers a parametrically consistent minimal sufficient extension of $\rel{}$, together with its abstraction coefficients up to scale. This holds whenever such an extension satisfies a non-collinearity condition on its abstract coefficients, as we formalize in \Cref{cor:datarsecorrectness}. We now describe each step of \model{dataRSE} in detail.

\begin{algorithm}[t]%
\caption{Data-Driven Relevant Sufficiency Extender (\model{dataRSE})}%
\label{alg:datarse}
\begin{algorithmic}[1]
\Require Low-level graph~$\graph{G}_{\set{X}}$, data~$\mat{X}$, set of relevant variables~$\rel{}$, abstraction coefficients~$\mat{T}$
\Ensure Graphically consistent and sufficient $\rel{}'\supseteq\rel{}$ and its abstraction coefficients $\mat{T}'$
\If{not $\operatorname{GraphicalConsistency}(\rel{}, \graph{G}_{\set{X}})$}
    \State \Return $\bot$
\EndIf
\State $\block{} \gets \operatorname{Blocks}(\rel{}, \graph{G}_{\set{X}})$
\State $\set{I} \gets \operatorname{JointlyTargetedSets}(\block{})$
\If{$\set{I} = \emptyset$}
    \State \Return $(\rel{}, \mat{T})$
    \Comment{Sufficient: no jointly targeted sets}
\EndIf
\For{$\set{V}_i\in\set{I}$}
\Comment{Iterate over all jointly targeted sets}
    \State $\set{Z}_i \gets \operatorname{NewRelevants}(\set{V}_i, \set{I}, \block{}, \graph{G}_{\set{X}})$
\EndFor
\State $\set{Z} \gets \cup_{i=1}^{\size{\set{I}}}\set{Z}_{i}$, $\set{V} \gets \cup_{i=1}^{\size{\set{I}}}\set{V}_{i}$
\State $\graph{G}_{\set{Y}} \gets \operatorname{AbstractGraph}(\rel{}, \graph{G}_{\set{X}})$ \Comment{Abstract graph on $\set{Y}$ (Alg.~\ref{alg:absgraph})}
\State $\set{K} \gets \left(\cup_{Y_j\in\set{V}}\pa{Y_j, \graph{G}_{\set{Y}}}\right)\setminus\set{V}$ \Comment{Known parents of $\set{V}$ outside $\set{V}$}
\State $\mat{Y}_{\set{V}} \gets [\mat{X}_{\rel{Y_1}} \vec{t}_{Y_1}\mid \ldots \mid \mat{X}_{\rel{Y_q}} \vec{t}_{Y_q}]\in\real^{n \times q}$ \Comment{Realizations of the confounded variables $\set{V}$}
\State $\mat{Y}_{\set{K}} \gets [\mat{X}_{\rel{Y}}\vec{t}_Y]_{Y\in\set{K}}\in\real^{n\times\size{\set{K}}}$ \Comment{Realizations of the known parents $\set{K}$}
\For{$Y_j\in\set{V}$}
\Comment{Estimate the composed abstraction matrix column by column}
    \State $\set{Z}^{(j)} \gets \setdef{X\in\set{Z}\mid Y_j\in\targets{X}{\block{}}}$
    \Comment{New relevant variables targeting $Y_j$}
    \State $\hat{\vec{a}}_{\set{Z}^{(j)},j}, \hat{\vec{m}}_j \gets \operatorname{Regress}(\vec{y}_j \sim [\mat{X}_{\set{Z}^{(j)}} \mid \mat{Y}_{\pa{Y_j}}])$
    \Comment{$\vec{y}_j$ is the $j$-th column of $\mat{Y}_{\set{V}}$}
    \State $\hat{a}_{lj} \gets 0$ for every $X_l\in\set{Z}\setminus\set{Z}^{(j)}$
    \Comment{No direct effect on $Y_j$}
\EndFor
\State $\set{P}, \setdef{\vec{t}_h}_{h=1}^{k} \gets \operatorname{MaximalPartitioning}(\hat{\mat{A}}, \mat{X}_{\set{Z}}, \mat{Y}_{\set{K}}, \mat{Y}_{\set{V}}, \setdef{\set{Z}^{(j)}}_{Y_j\in\set{V}}, \graph{G}_{\set{Y}})$
\Comment{Alg.~\ref{alg:maximalpartitioning}}
\State $\graph{G}'' \gets \operatorname{AbstractGraph}(\rel{}\cup\set{P}, \graph{G}_{\set{X}})$ \Comment{$Y_{\set{P}_h}$ has relevant set $\set{P}_h$}
\State $\set{P}_{\set{V}} \gets \setdef{\set{P}_h\in\set{P} \mid \ch{Y_{\set{P}_h},\graph{G}''}\subseteq\set{V}}$ \Comment{Unmediated relevant sets}
\If{$\set{P}_{\set{V}}=\emptyset$}
    \State \Return $\bot$
\EndIf
\State $\rel{}' \gets \rel{} \cup \set{P}_{\set{V}}$
\State $\mat{T}' \gets \left[\mat{T}\mid\vec{t}_h\right]_{\set{P}_h\in\set{P}_{\set{V}}}\in\real^{d \times (\size{\set{Y}}+\size{\set{P}_{\set{V}}})}$ \Comment{$\vec{t}_h$ padded with zeros outside $\set{P}_h$}
\State \Return $\model{dataRSE}(\graph{G}_{\set{X}}, \mat{X}, \rel{}', \mat{T}')$
\Comment{Recurse on the extended relevant sets}
\end{algorithmic}
\end{algorithm}

\paragraph{Graphical Consistency and Sufficiency (ll. 1--8).}
As in \model{RSE}, \model{dataRSE} first checks for graphical consistency and sufficiency of the input set of relevant variables $\rel{}$. If the input set is graphically inconsistent, it returns $\bot$ (Lines 1--3). Then, it computes the blocks and the jointly targeted sets (Lines 4--5). If the input set is graphically consistent and sufficient, it returns it together with the corresponding abstraction coefficients $\mat{T}$ (Lines 6--8).

\paragraph{Confounded Variables and New Relevant Variables (ll. 9--14).}
\model{dataRSE} computes the new relevant variables $\set{Z}_i$ of every jointly targeted set $\set{V}_i\in\set{I}$ (Lines 9--11) and takes their unions $\set{Z}$ and $\set{V}$ (Line 12). Then, given the abstract graph $\graph{G}_{\set{Y}}$ induced by $\rel{}$ (Line 13), it selects the set $\set{K}$ of abstract variables outside $\set{V}$ that are parents of some confounded variable (Line 14). Together with $\set{V}$, the set $\set{K}$ contains every parent of a confounded variable in $\graph{G}_{\set{Y}}$, which will be used as regressors for the confounded variables, as in \Cref{eq:paramconsistency}.

\paragraph{Abstract Realizations (ll. 15--16).}%
Given the low-level dataset $\mat{X}$ and the abstraction coefficients $\mat{T}$ of the known abstract variables $\set{Y}$, \model{dataRSE} computes the abstract realizations of the confounded abstract variables $\set{V}$ and of their known parents $\set{K}$ by abstracting each low-level relevant set. Formally, for each $Y\in\set{V}\cup\set{K}$ it computes the $n$ realizations $\mat{X}_{\rel{Y}}\vec{t}_Y \in\real^{n}$, where $\vec{t}_Y\in\real^{\size{\rel{Y}}}$ is the abstraction vector of $Y$. Indexing the confounded variables as $Y_1,\ldots,Y_q\in\set{V}$, it collects their realizations in $\mat{Y}_{\set{V}}$ and those of $\set{K}$ in $\mat{Y}_{\set{K}}$.

\paragraph{Composed Abstraction Matrix (ll. 17--21).}%
If $\rel{}$ admits a parametrically consistent minimal sufficient extension, then, by \Cref{thm:paramconsistency}, the exogenous term $U_j$ of any confounded variable $Y_j\in\set{V}$ is independent of both its parents $\pa{Y_j,\graph{G}_{\set{Y}}}$ and the new relevant variables $\set{Z}^{(j)}$ targeting $Y_j$, while the remaining new relevant variables have no direct effect on $Y_j$. Hence, \model{dataRSE} estimates the composed abstraction matrix $\hat{\mat{A}}$ column by column (Line 17). For each $Y_j\in\set{V}$, it computes the new relevant variables $\set{Z}^{(j)}$ targeting $Y_j$ (Line 18), regresses $Y_j$ on $\set{Z}^{(j)}$ and on its parents $\pa{Y_j,\graph{G}_{\set{Y}}}\subseteq\set{K}\cup\set{V}$ by least squares (Line 19), with a small ridge penalty in practice (\Cref{app:hyperparams}), and sets $\hat{a}_{lj}=0$ for every $X_l\in\set{Z}\setminus\set{Z}^{(j)}$ (Line 20). Parent coefficients $\hat{\vec{m}}_j$ are then discarded.

\paragraph{Maximal Partitioning (l. 22).}
In practice, \model{dataRSE} only has a noisy estimate $\hat{\mat{A}}$ of the composed abstraction matrix, which in general does not exhibit exact row collinearity. Therefore, it estimates the maximal partitioning of the new relevant variables $\set{Z}$ by clustering the rows of $\hat{\mat{A}}$ by their absolute cosine distance, with agglomerative clustering \citep{hastie2009elements,mullner2011modern}, into $k=1,\ldots,\size{\set{Z}}$ clusters. With the exact composed abstraction matrix, agglomerative clustering recovers $\set{P}^\ast$ for $k=\size{\set{P}^\ast}$, since rows in the same set of $\set{P}^\ast$ have zero distance, while rows in different sets have positive distance. Each $k$-cluster partition induces $k$ new abstract variables. Their abstraction vectors are fitted jointly with the regression of each confounded variable $Y_j$ on its parents and on the induced variables targeting it, as in \Cref{eq:paramconsistency}. Then, \model{dataRSE} selects the number of clusters $k^\ast$ by the Bayesian Information Criterion (BIC) of the fitted models, and keeps the corresponding partition $\set{P}$ and abstraction vectors $\setdef{\vec{t}_h}_{h=1}^{k^\ast}$. We report the procedure in \Cref{alg:maximalpartitioning}. Alternatively, the number of clusters can be selected using rank-based partitioning, as we describe in \Cref{app:rankonepartitioning} and empirically compare in \Cref{app:plateau}.

\paragraph{Unmediated Relevant Sets (ll. 23--27).}
As discussed in \Cref{subsec:maximalpartitioning}, the maximal partitioning can still lead to parametrically inconsistent relevant sets. Therefore, \model{dataRSE} only keeps the unmediated relevant sets $\set{P}_{\set{V}}\subseteq\set{P}$, which estimate $\set{P}^\ast_{\set{V}}$ (\Cref{def:unmediated}, Lines 23--24). If there are no unmediated relevant sets, \model{dataRSE} returns $\bot$ (Lines 25--27). The relevant variables of the discarded abstract variables return to being non-relevant and are resolved at the next recursive call, where their effects on the new abstract variables induced by $\set{P}_{\set{V}}$ are accounted for by the corresponding composed abstraction matrix. Given the exact composed abstraction matrix and maximal partitioning, it holds $\set{P}_{\set{V}}=\set{P}^\ast_{\set{V}}$. Hence, by \Cref{thm:maximalpartitioning}, whenever a parametrically consistent minimal sufficient extension satisfies \Cref{eq:mpcondition}, $\set{P}_{\set{V}}$ is a non-empty subset of its relevant sets.

\paragraph{Recursion (ll. 28--30).}
Finally, \model{dataRSE} adds the unmediated relevant sets to the relevant variables, $\rel{}'=\rel{}\cup\set{P}_{\set{V}}$, appends the fitted abstraction vectors $\vec{t}_h$, padded with zeros, to the abstraction coefficients (Lines 28--29), and recursively calls itself on $\rel{}'$ (Line 30). Given the exact composed abstraction matrix and maximal partitioning, $\set{P}_{\set{V}}$ is contained in any parametrically consistent minimal sufficient extension that satisfies \Cref{eq:mpcondition}. Such an extension remains a minimal sufficient extension of $\rel{}'$, so that \Cref{thm:maximalpartitioning} applies again at the next recursive call, provided that \Cref{eq:mpcondition} holds there as well. By induction, \model{dataRSE} then recovers the extension, as we formalize in the following corollary.

\begin{restatable}[\model{dataRSE} Correctness]{corollary}{datarsecorrectnesscor}%
\label{cor:datarsecorrectness}
Let $\rel{}$ be a graphically consistent set of relevant variables on a low-level ANM $\scm{L}$, and let $\rel{}'\in\mise{\rel{}}{\graph{G}_{\set{X}}}$ be parametrically consistent on $\scm{L}$, with $\scm{H}$ a strong $\mat{T}$-abstraction of $\scm{L}$ with relevant sets $\rel{}'$. Suppose that \Cref{eq:mpcondition} holds for $\scm{H}$ with $\rel{}$ replaced by every graphically consistent but not sufficient $\rel{}''$ such that $\rel{}\subseteq\rel{}''\subset\rel{}'$. Then, given the exact composed abstraction matrix and its maximal partitioning at each recursive call, \model{dataRSE} returns $\rel{}'$ and the abstraction $\mat{T}$ up to scaling. Moreover, at most one consistent minimal sufficient extension of $\rel{}$ satisfies these conditions.
\end{restatable}
\begin{proof}
    We report the proof in \Cref{app:datarsecorrectness}.
\end{proof}

\section{Related Work}%
\label{sec:related}

Causal abstraction learning methods aim to learn an abstract representation of a low-level model from data, focusing mainly on the function mapping low-level variables to abstract variables and, in some cases, on the abstract mechanisms. To this end, they require additional information, such as paired or unpaired high-level samples, low-level interventional data, an explicit map from low-level to high-level interventions, the abstract graph, a complete map from low-level to high-level variables, or a combination of these elements. To the best of our knowledge, \model{RSE} and \model{dataRSE} are the only methods that complete partial knowledge of the abstraction without any high-level information (\Cref{tab:related}).

Most existing methods define every abstract variable before learning, either explicitly through its relevant variables~\citep{zennaro2023jointly,dyer2024interventionally,xia2024neural} or implicitly through high-level samples~\citep{felekis2024causal,massidda2024learning,dacunto2025causal}. Notably, \model{LinSEPAL}~\citep{dacunto2025causal} shares our motivation by completing a partial map between low-level and high-level variables, but its high-level samples still determine all abstract variables. Closer to our setting, Targeted Causal Reduction~\citep{kekic2024targeted} only requires the abstraction of a known target, and \citet{zhu2024unsupervised} extend it to the case where no target is known. However, both fix the abstract graph, and thus the number of abstract variables, a priori, and require low-level interventional data. In contrast, \model{RSE} only requires the low-level graph and the relevant variables of a strict subset of abstract variables, and constructs the remaining abstract variables needed to ensure causal sufficiency together with the abstract graph they entail. Given observational low-level samples and the abstraction function of the provided abstract variables, \model{dataRSE} further recovers the abstraction function and the abstract model without any high-level information.%

Finally, causal abstraction bears similarities to existing research on Cluster DAGs~\citep{anand2023causal} and Partial Cluster DAGs~\citep{schooltink2025aligning}. These frameworks also define the existence of edges in a high-level graph according to the causal relations between low-level variables. However, linear causal abstraction~\citep{massidda2024learning} entails different graphical conditions and consequently different methods. We discuss the main differences between causal abstraction and Cluster DAGs in \Cref{app:graphcons}.

\DeclareRobustCommand{\gcirc}{\tikz[baseline=-0.5ex]\draw[scale=0.3, line width=1.5pt, color=green!80!black, fill=green!40!white] (0,0) circle (0.4);}
\DeclareRobustCommand{\rcirc}{\tikz[baseline=-0.5ex]\draw[scale=0.3, line width=1.5pt, color=red, fill=red!40!white] (0,0) circle (0.4);}
\DeclareRobustCommand{\ycirc}{\tikz[baseline=-0.5ex]\draw[scale=0.3, line width=2.0pt, color=yellow!80!black, fill=yellow!40!white] (0,0) circle (0.4);}
\DeclareRobustCommand{\textgcirc}{\raisebox{0.2ex}[0pt][0pt]{\gcirc}}
\DeclareRobustCommand{\textrcirc}{\raisebox{0.2ex}[0pt][0pt]{\rcirc}}
\DeclareRobustCommand{\textycirc}{\raisebox{0.2ex}[0pt][0pt]{\ycirc}}

\begin{table}
    \centering
    \begin{tabular}{lccccccc}
        \toprule
        & \multicolumn{3}{c}{\textbf{Low-Level}} & \multicolumn{2}{c}{\textbf{Abstraction}} & \multicolumn{2}{c}{\textbf{High-Level}} \\
        \textbf{Method}
        & $\graph{G}_{\set{X}}$
        & $\dist{\set{X}}$
        & $\dist{\set{X}}^{\set{I}}$
        & $\tau$ & $\rel{}$
        & $\graph{G}_{\set{Y}}$
        & $\dist{\set{Y}}$ \\
        \midrule
        \citet{zennaro2023jointly}&\rcirc&\rcirc&\rcirc&\gcirc&\rcirc&\rcirc&\rcirc\\
        \citet{kekic2024targeted}&&\rcirc&\rcirc&\ycirc&\ycirc&\rcirc&\gcirc\\
        \citet{zhu2024unsupervised}&\rcirc&\rcirc&\rcirc&\gcirc&\gcirc&\rcirc&\gcirc\\

        \model{COTA}~\citep{felekis2024causal}
        &\rcirc &\rcirc &\rcirc &\gcirc &\gcirc &\rcirc &\rcirc \\

        \citet{dyer2024interventionally}
        &&\rcirc &\rcirc &\rcirc &\rcirc &\rcirc &\gcirc \\
        \citet{xia2024neural}&&\rcirc&&\gcirc&\rcirc&\rcirc&\gcirc\\

        \model{Abs-LiNGAM}~\citep{massidda2024learning}&\gcirc&\rcirc&&\gcirc&\gcirc&\gcirc&\rcirc\\
        \model{LinSEPAL}~\citep{dacunto2025causal}&&\rcirc&&\gcirc&\ycirc&&\rcirc\\

        \model{RSE} (Ours)&\rcirc&&&&\ycirc&\gcirc&\\
        \model{dataRSE} (Ours)&\rcirc&\rcirc&&\ycirc&\ycirc&\gcirc&\gcirc\\
        \bottomrule
    \end{tabular}
    \caption{
    For each causal abstraction learning method, we highlight the required knowledge (in red \textrcirc), the outcome (in green \textgcirc), and the required \emph{partial} knowledge that the method completes (in yellow \textycirc). We denote as $\tau$ the abstraction function and as $\rel{}$ the knowledge of the low-level relevant variables constituting each abstract variable. We equate knowledge of a distribution $\dist{}$ with the availability of its samples, and denote as $\dist{\set{X}}^{\set{I}}$ the availability of low-level interventional samples. We detail our classification choices in \Cref{app:related}.
    }
    \label{tab:related}
\end{table}

\section{Experiments}%
\label{sec:experiments}

In this section, we evaluate on synthetic data how accurately \model{dataRSE} recovers, from a finite number of low-level samples, the relevant variables and the abstraction function of a reference abstraction%
\footnote{%
The code to reproduce our experiments is available at \url{https://github.com/rmassidda/dataRSE}. The results and plots of the reported runs are available at \url{https://doi.org/10.5281/zenodo.24581034}. All the experiments are conducted on a machine with an AMD EPYC 7543 32-Core processor.
}.
In particular, we show that, given the relevant variables and the abstraction function of the abstract treatment and outcome, \model{dataRSE} recovers the remaining relevant variables and the abstraction function of the reference abstraction up to scale and sign. Furthermore, we evaluate whether the learned abstraction supports the estimation of the causal effect between the abstract treatment and outcome.

\subsection{Data Generation}

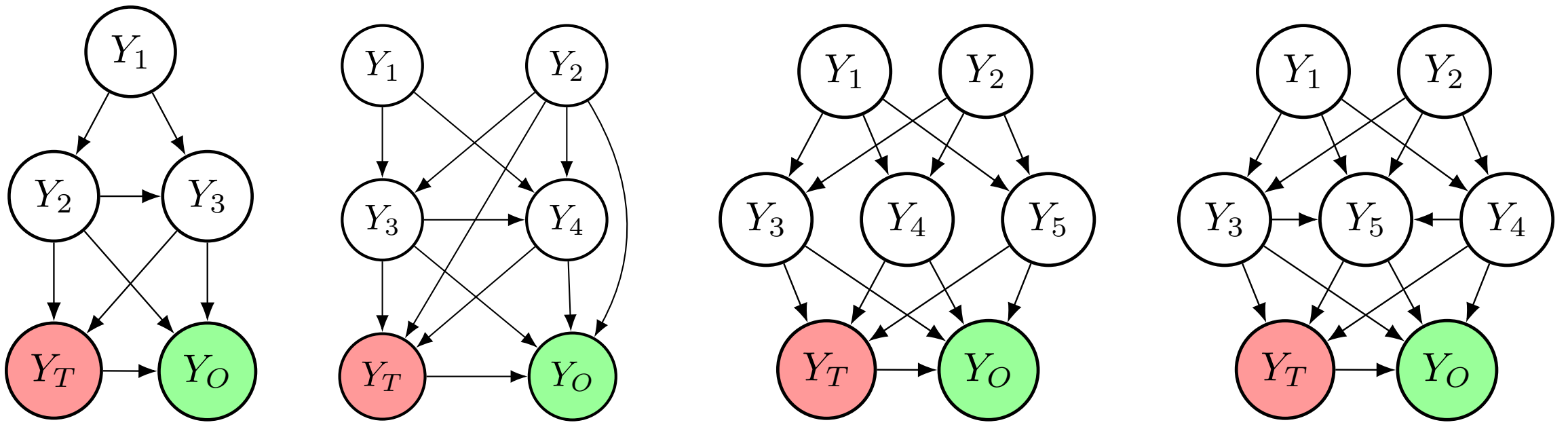
\begin{figure}[ht]
\centering
\begin{minipage}[b]{0.17\textwidth}
\centering\small\resizebox{\textwidth}{!}{\begin{tikzpicture}[node distance=1cm, minimum size=0.3cm]
  \node[draw, circle, thick] (Y1) {$Y_1$};
  \node[draw, circle, thick, below left=0.7cm and 0.1cm of Y1] (Y2) {$Y_2$};
  \node[draw, circle, thick, below right=0.7cm and 0.1cm of Y1] (Y3) {$Y_3$};
  \node[draw, circle, thick, fill=red!40!white, below=0.7cm of Y2] (Y4) {$Y_T$};
  \node[draw, circle, thick, fill=green!40!white, below=0.7cm of Y3] (Y5) {$Y_O$};
  \draw[-{Latex}] (Y1) -- (Y2);
  \draw[-{Latex}] (Y1) -- (Y3);
  \draw[-{Latex}] (Y2) -- (Y3);
  \draw[-{Latex}] (Y2) -- (Y4);
  \draw[-{Latex}] (Y2) -- (Y5);
  \draw[-{Latex}] (Y3) -- (Y4);
  \draw[-{Latex}] (Y3) -- (Y5);
  \draw[-{Latex}] (Y4) -- (Y5);
\end{tikzpicture}}%
\subcaption{Graph A}
\end{minipage}\hfill%
\begin{minipage}[b]{0.20\textwidth}
\centering\small\resizebox{\textwidth}{!}{\begin{tikzpicture}[node distance=1cm, minimum size=0.3cm]
  \node[draw, circle, thick] (Y1) {$Y_1$};
  \node[draw, circle, thick, right=of Y1] (Y2) {$Y_2$};
  \node[draw, circle, thick, below=0.7cm of Y1] (Y3) {$Y_3$};
  \node[draw, circle, thick, right=of Y3] (Y4) {$Y_4$};
  \node[draw, circle, thick, fill=red!40!white, below=0.7cm of Y3] (Y5) {$Y_T$};
  \node[draw, circle, thick, fill=green!40!white, right=of Y5] (Y6) {$Y_O$};
  \draw[-{Latex}] (Y1) -- (Y3);
  \draw[-{Latex}] (Y1) -- (Y4);
  \draw[-{Latex}] (Y2) -- (Y3);
  \draw[-{Latex}] (Y2) -- (Y4);
  \draw[-{Latex}] (Y2) -- (Y5);
  \draw[-{Latex}] (Y2) to[bend left=30] (Y6);
  \draw[-{Latex}] (Y3) -- (Y4);
  \draw[-{Latex}] (Y3) -- (Y5);
  \draw[-{Latex}] (Y3) -- (Y6);
  \draw[-{Latex}] (Y4) -- (Y5);
  \draw[-{Latex}] (Y4) -- (Y6);
  \draw[-{Latex}] (Y5) -- (Y6);
\end{tikzpicture}}%
\subcaption{Graph B}
\end{minipage}\hfill%
\begin{minipage}[b]{0.25\textwidth}
\centering\small\resizebox{\textwidth}{!}{\begin{tikzpicture}[node distance=0.9cm, minimum size=0.3cm]
  \node[draw, circle, thick] (Y1) {$Y_1$};
  \node[draw, circle, thick, right=0.4cm of Y1] (Y2) {$Y_2$};
  \node[draw, circle, thick, below left=0.7cm and 0.1cm of Y1] (Y3) {$Y_3$};
  \node[draw, circle, thick, right=0.4cm of Y3] (Y4) {$Y_4$};
  \node[draw, circle, thick, right=0.4cm of Y4] (Y5) {$Y_5$};
  \node[draw, circle, thick, fill=red!40!white, below left=0.7cm and 0.1cm of Y4] (Y6) {$Y_T$};
  \node[draw, circle, thick, fill=green!40!white, below right=0.7cm and 0.1cm of Y4] (Y7) {$Y_O$};
  \draw[-{Latex}] (Y1) -- (Y3);
  \draw[-{Latex}] (Y1) -- (Y4);
  \draw[-{Latex}] (Y1) -- (Y5);
  \draw[-{Latex}] (Y2) -- (Y3);
  \draw[-{Latex}] (Y2) -- (Y4);
  \draw[-{Latex}] (Y2) -- (Y5);
  \draw[-{Latex}] (Y3) -- (Y6);
  \draw[-{Latex}] (Y3) -- (Y7);
  \draw[-{Latex}] (Y4) -- (Y6);
  \draw[-{Latex}] (Y4) -- (Y7);
  \draw[-{Latex}] (Y5) -- (Y6);
  \draw[-{Latex}] (Y5) -- (Y7);
  \draw[-{Latex}] (Y6) -- (Y7);
\end{tikzpicture}}%
\subcaption{Graph C}
\end{minipage}\hfill%
\begin{minipage}[b]{0.25\textwidth}
\centering\small\resizebox{\textwidth}{!}{\begin{tikzpicture}[node distance=0.9cm, minimum size=0.3cm]
  \node[draw, circle, thick] (Y1) {$Y_1$};
  \node[draw, circle, thick, right=0.4cm of Y1] (Y2) {$Y_2$};
  \node[draw, circle, thick, below left=0.7cm and 0.1cm of Y1] (Y3) {$Y_3$};
  \node[draw, circle, thick, right=0.4cm of Y3] (Y5) {$Y_5$};
  \node[draw, circle, thick, right=0.4cm of Y5] (Y4) {$Y_4$};
  \node[draw, circle, thick, fill=red!40!white, below left=0.7cm and 0.1cm of Y5] (Y6) {$Y_T$};
  \node[draw, circle, thick, fill=green!40!white, below right=0.7cm and 0.1cm of Y5] (Y7) {$Y_O$};
  \draw[-{Latex}] (Y1) -- (Y3);
  \draw[-{Latex}] (Y1) -- (Y4);
  \draw[-{Latex}] (Y1) -- (Y5);
  \draw[-{Latex}] (Y2) -- (Y3);
  \draw[-{Latex}] (Y2) -- (Y4);
  \draw[-{Latex}] (Y2) -- (Y5);
  \draw[-{Latex}] (Y3) -- (Y5);
  \draw[-{Latex}] (Y3) -- (Y6);
  \draw[-{Latex}] (Y3) -- (Y7);
  \draw[-{Latex}] (Y4) -- (Y5);
  \draw[-{Latex}] (Y4) -- (Y6);
  \draw[-{Latex}] (Y4) -- (Y7);
  \draw[-{Latex}] (Y5) -- (Y6);
  \draw[-{Latex}] (Y5) -- (Y7);
  \draw[-{Latex}] (Y6) -- (Y7);
\end{tikzpicture}}%
\subcaption{Graph D}
\end{minipage}%
\caption{%
  We study the performance of \model{dataRSE} over the four abstract graph structures shown. For each experiment and abstract structure~$\graph{G}_{\set{Y}}$, we generate a high-level model $\scm{H}$ with graph $\graph{G}_{\set{Y}}$ and a low-level model $\scm{L}$ that is linearly abstracted by $\scm{H}$. To test the performance of \model{dataRSE} in different scenarios, we control the relevant set size $\size{\rel{Y}}$ and the block size $\size{\block{Y}}$, shared by all the abstract variables $Y$. We highlight in \textcolor{red!60!black}{red} the \textcolor{red!60!black}{treatment} variable~$Y_T$ and in \textcolor{green!60!black}{green} the \textcolor{green!60!black}{outcome} variable~$Y_O$ from which we extract the prior knowledge necessary for our algorithm.
}%
\label{fig:abstract-graphs}
\end{figure}

We generate synthetic data from four abstract graph structures (\Cref{fig:abstract-graphs}). Each abstract graph contains a treatment node $Y_T$ and an outcome node $Y_O$, which constitute the prior knowledge necessary for the \model{RSE} and \model{dataRSE} algorithms. All the selected structures contain two layers of confounders, \ie, confounders of the treatment-outcome pair and confounders of these confounders. They differ in the interaction among confounders (Graphs A, B, and D) and in the number of confounders (three in Graph A, four in Graph B, and five in Graphs C and D). Graph B additionally contains a confounder of both the treatment-outcome pair and of an upstream pair of confounders to highlight the scenario where \model{dataRSE} must keep only the unmediated relevant sets.

In each experiment, given an abstract graph $\graph{G}_{\set{Y}}$, we sample a high-level Additive Noise Model~$\scm{H}$, a linear abstraction function $\mat{T}$, and a low-level ANM~$\scm{L}$ such that $\scm{H}$ $\mat{T}$-abstracts $\scm{L}$. We refer to $\scm{H}$ and $\mat{T}$ as the \emph{ground truth}. By construction, the relevant sets are parametrically consistent on $\scm{L}$, and the abstract graphs are chosen so that they form a minimal sufficient extension of $\setdef{\rel{Y_T},\rel{Y_O}}$. Hence, the ground truth is the extension that \model{dataRSE} should return given the exact composed abstraction matrix and its maximal partitioning. We provide \model{dataRSE} only with the necessary prior knowledge for the algorithms, \ie, the low-level graph $\graph{G}_{\set{X}}$, the known sets of relevant variables of the abstract treatment $\rel{Y_T}$ and of the abstract outcome $\rel{Y_O}$, their abstraction functions $\vec{t}_T$ and $\vec{t}_O$, and a dataset of $n$ low-level samples $\mat{X}$ from the low-level model. Furthermore, we reject the ground truths whose abstract coefficients are close to violating the non-collinearity condition of \Cref{thm:maximalpartitioning} (\Cref{eq:mpcondition}), and we sample abstraction vectors with unit norm. We report the data generation procedure, including its pseudocode, in \Cref{app:datagen}.

\subsection{Evaluation Framework}%
\label{sec:eval-framework}

We measure the accuracy of \model{dataRSE} by checking whether the algorithm recovers the \emph{exact} sets of relevant variables of the ground-truth abstraction function. Then, given the recovered relevant variables $\relhat{}$, we compute the abstract graph~$\hat{\graph{G}}_{\set{Y}}$ (\Cref{alg:absgraph}). Given the learned abstraction $\hat{\mat{T}}$, we compute abstract realizations $\hat{Y}_i = \mat{X}_{\relhat{Y_i}}\hat{\vec{t}}_i$ for every abstract variable $Y_i\in\hat{\graph{G}}_{\set{Y}}$. Finally, we compute the optimal adjustment set~\citep{henckel2022graphical} to estimate the Total Causal Effect (TCE) of the abstract treatment $Y_T$ on the abstract outcome $Y_O$. We compute the reference ground-truth TCE over the ground-truth abstract ANM $\scm{H}$.

Furthermore, we study the performance of \model{dataRSE} in recovering the ground-truth abstraction function $\mat{T}$ by computing the absolute cosine similarity between the ground-truth abstraction coefficients $\vec{t}_{Y_i}$ and the learned abstraction coefficients $\hat{\vec{t}}_{Y_j}$ for every pair of abstract variables $Y_i\in\graph{G}_{\set{Y}}$ and $Y_j\in\hat{\graph{G}}_{\set{Y}}$, as in
\begin{equation}
    s_{ij} =
    \begin{cases}
    \left|
    \frac{\langle \vec{t}_{Y_i}, \hat{\vec{t}}_{Y_j} \rangle}
    {\norm{\vec{t}_{Y_i}}\norm{\hat{\vec{t}}_{Y_j}}}
    \right| & \text{if } \size{\rel{Y_i}} = \size{\relhat{Y_j}} \\
    0 & \text{otherwise,}
    \end{cases}
\end{equation}
where the similarity of two variables with different numbers of relevant variables is zero. We detail how we match the ground-truth and learned abstract variables and aggregate their similarities across runs in \Cref{app:matchabs}.

\subsection{Results}

\begin{figure}[ht]
\centering
\includegraphics[width=\textwidth]{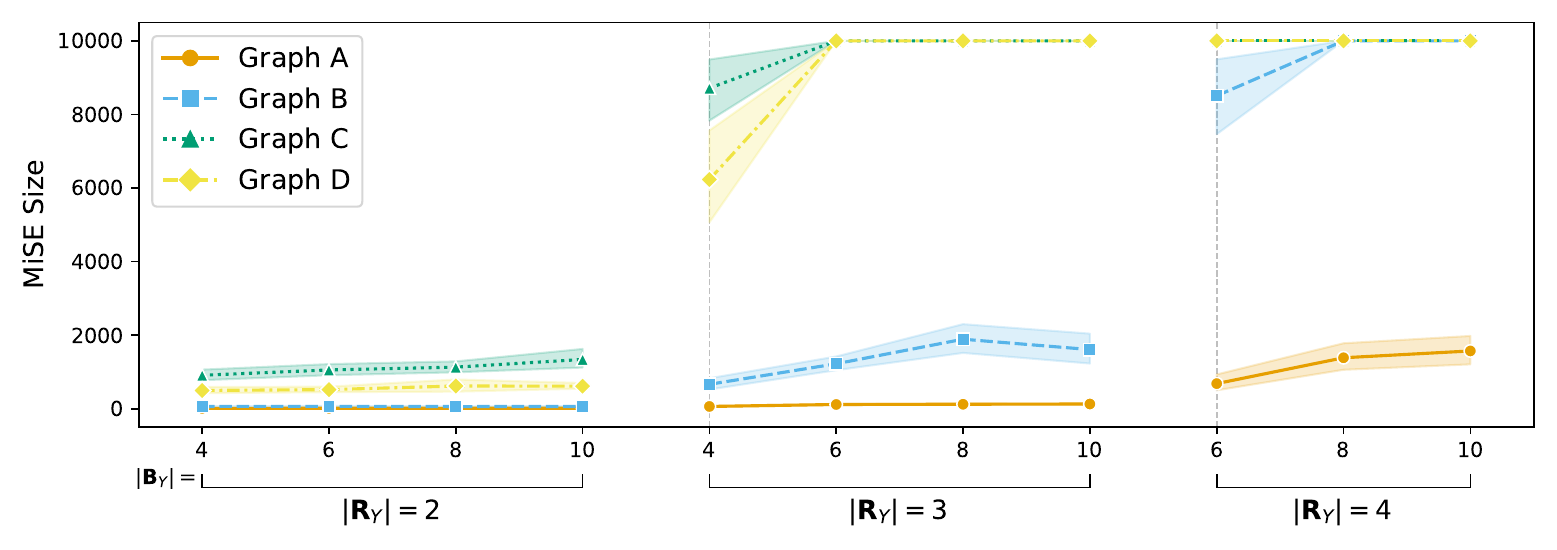}%
\caption{%
    For each abstract graph, we plot the size of the \model{MiSE} starting from a low-level model sampled with an increasing relevant set size $\size{\rel{Y}}$ and block size $\size{\block{Y}}$. The size of the \model{MiSE} depends on the abstract topology and grows quickly with the relevant set size. We compute the \model{MiSE} using \model{qRSE} over 15 random seeds and cap the algorithm at $10^4$ solutions.
}%
\label{fig:n_solutions-rse}
\end{figure}

\begin{figure}[htb]
\centering
\includegraphics[width=\textwidth]{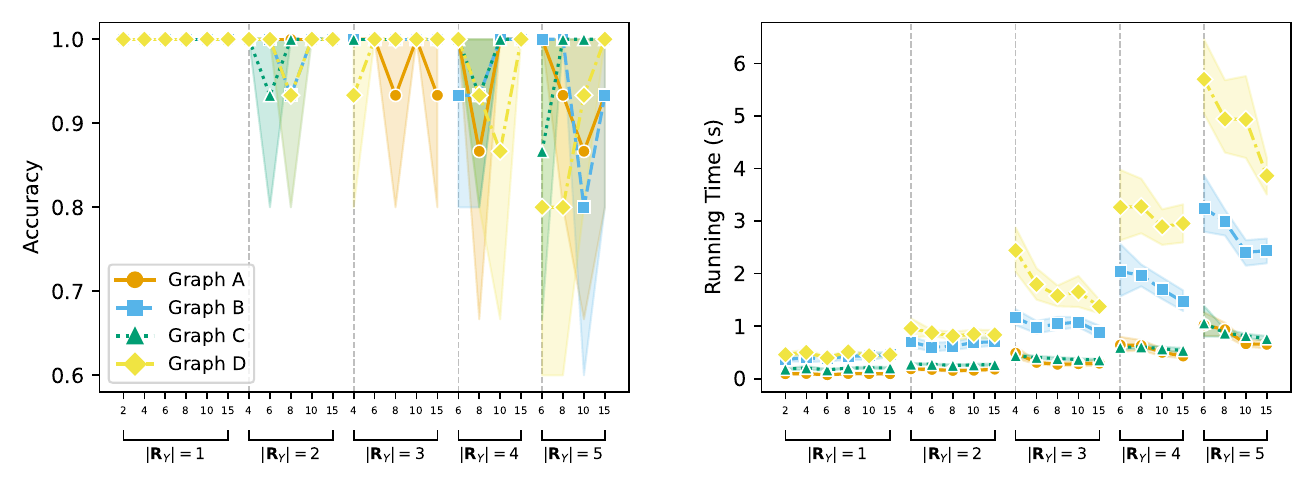}
\caption{%
    On the left, we report the accuracy of \model{dataRSE} in recovering the relevant variables $\rel{}$ from data for an increasing relevant set size and block size. We compute the accuracy by checking whether \model{dataRSE} recovers the exact set of relevant variables of the ground-truth abstraction function. On the right, we report the running time in seconds. We report results computed over 15 random seeds and $n=20000$ samples.
}%
\label{fig:size_data-rse}
\end{figure}

\begin{figure}[htb]
\centering
\includegraphics[width=\textwidth]{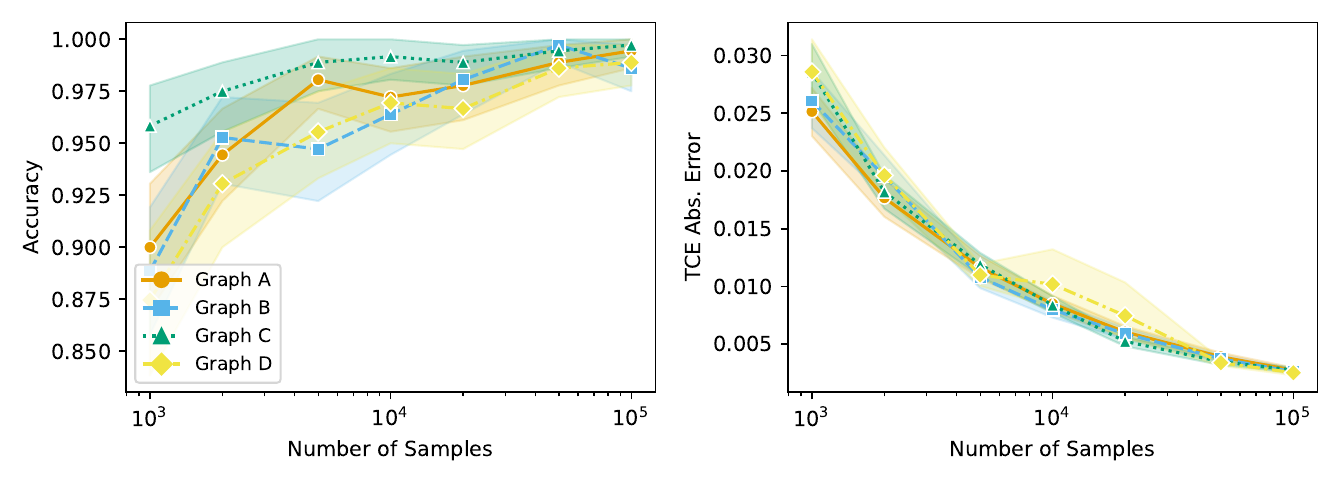}
\caption{%
    We evaluate the performance of \model{dataRSE} in recovering the relevant variables $\rel{}$ for an increasing number of samples over all the configurations on the relevant set size and block size. We report the accuracy in recovering the exact ground-truth relevant sets (Left) and the absolute error in estimating the TCE of $Y_T$ on $Y_O$ (Right). All results are averaged over 15 random seeds and all configurations on the relevant set size and block size; we report the failure rate and running time, with the same breakdown by graph, in \Cref{app:additional-experiments}.
}%
\label{fig:samples_data-rse}
\end{figure}

\begin{figure}[htb]
\centering
\includegraphics[width=0.48\textwidth]{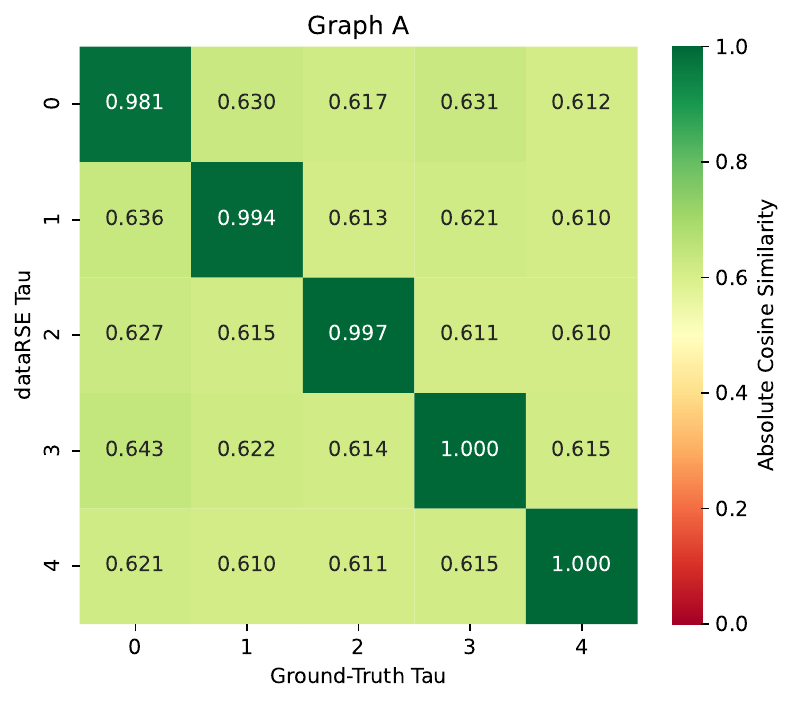}
\includegraphics[width=0.48\textwidth]{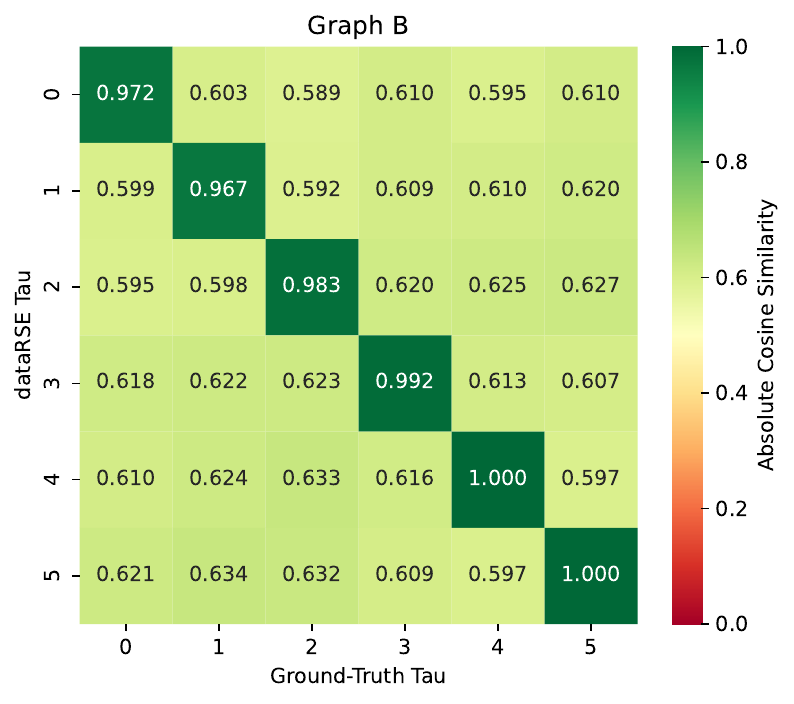}\\
\includegraphics[width=0.48\textwidth]{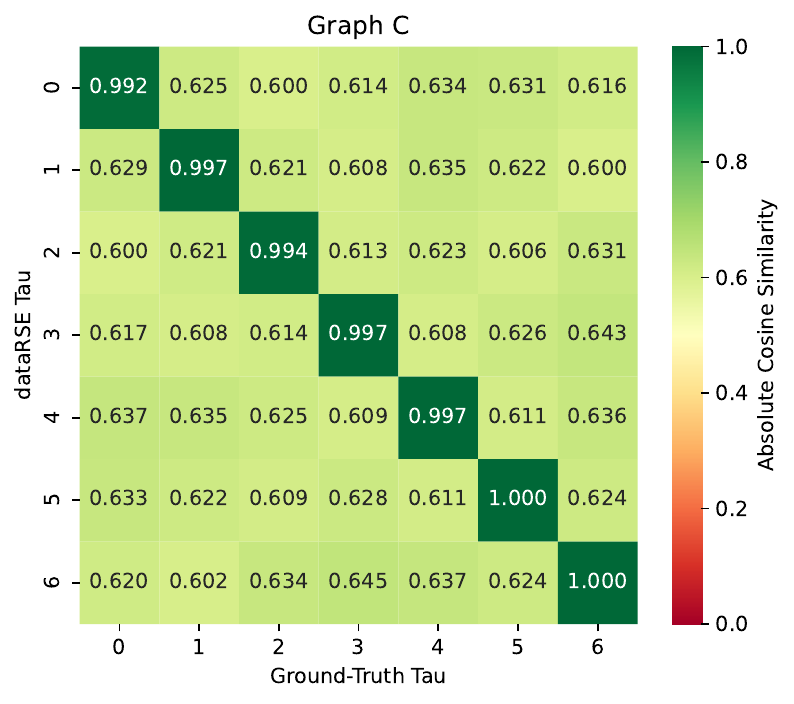}
\includegraphics[width=0.48\textwidth]{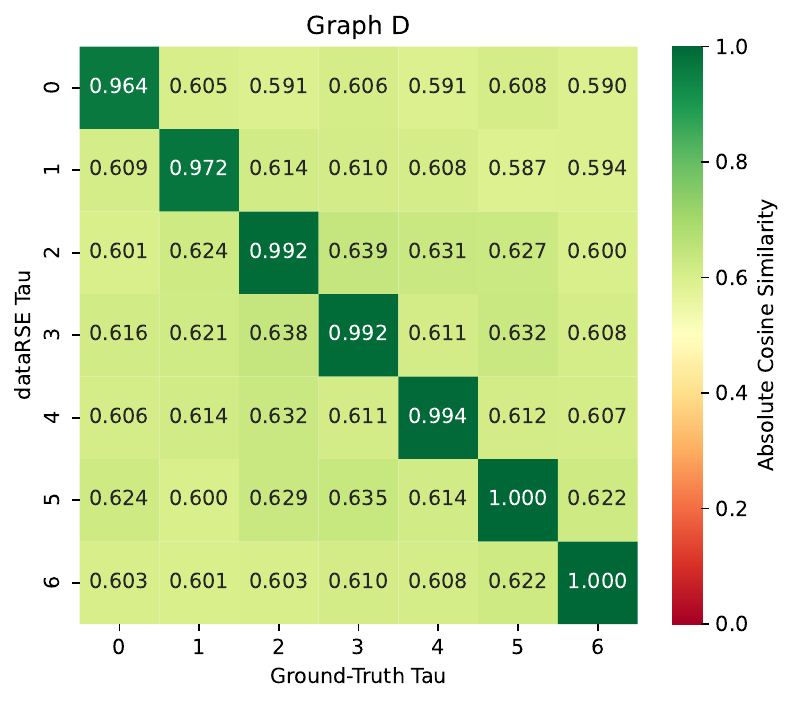}
\caption{%
	We report the cosine similarity between the ground-truth abstraction coefficients $\vec{t}_Y$ and the learned abstraction coefficients $\hat{\vec{t}}_Y$ for each abstract variable $Y\in\graph{G}_{\set{Y}}$. We compute the similarity by solving the linear sum assignment problem over the absolute cosine similarity matrix between the ground-truth and learned abstraction coefficients. We report results averaged over 15 random seeds and all configurations on the relevant set size and block size for $n=5000$ samples. The last two abstract variables of each graph are the treatment and the outcome, whose abstraction coefficients are given to \model{dataRSE}.
}%
\label{fig:tau_recovery}
\end{figure}

First, we study the size of the class of minimal sufficient extensions for an increasing relevant set size and an increasing block size (\Cref{fig:n_solutions-rse}). We compute the \model{MiSE} using \model{qRSE} capped at $10^4$ solutions. We observe that the number of minimal sufficient extensions quickly grows with the relevant set size. In fact, even small configurations reach the cap for the abstract graphs with five confounders, \ie, Graphs C and D in every configuration with $\size{\rel{Y}}\geq3$ and $\size{\block{Y}}\geq6$. Graph B, with four confounders, reaches the cap for $\size{\rel{Y}}=4$ and $\size{\block{Y}}\geq 8$. For the same relevant set size, the size of the \model{MiSE} also grows with the block size, although more slowly than with the relevant set size. The larger the block size, the larger the low-level graph, and consequently the more variables can become relevant. The size of the \model{MiSE} makes it impractical to enumerate all the minimal sufficient extensions even for small abstract graphs, and motivates the need for \model{dataRSE} to filter parametrically inconsistent solutions.

Despite the large number of minimal sufficient extensions, \model{dataRSE} correctly recovers the ground-truth relevant variables $\rel{}$ from the low-level model in the same settings where the \model{MiSE} contains up to $10^4$ candidate solutions (\Cref{fig:size_data-rse}). With $n=20000$ samples, its accuracy is at least $0.93$ for every abstract graph whenever $\size{\rel{Y}}\leq3$, and remains at least $0.80$ for the larger configurations with $\size{\rel{Y}}\geq4$. Configurations with more relevant variables per abstract variable are challenging, as they require estimating more coefficients of the composed abstraction matrix. The running time grows with the relevant set size and largely increases for Graphs B and D. We recall that whenever a new abstract variable causes another new one, \model{dataRSE} discards it (\Cref{alg:datarse}, Lines 23--24), and its relevant variables are partitioned again at the next recursive call. This behavior explains the time gap. Graph C has many confounders that do not interact with each other, and Graph A has only two interacting confounders, compared to the three in Graphs B and D.

Increasing the number of samples further reduces the errors in recovering the ground-truth relevant variables (\Cref{fig:samples_data-rse}) and reaches $0.99$ on average for $n=50000$ samples. In \Cref{app:additional-experiments}, we also report the failure rate, \ie, the fraction of runs in which \model{dataRSE} returns no solution, which is at most $0.003$ from $n=10000$ samples and vanishes from $n=50000$. With few samples, the rows of the estimated composed abstraction matrix $\hat{\mat{A}}$ that correspond to the same abstract variable are only approximately collinear, and the maximal partitioning may select more clusters than necessary. Therefore, almost all incorrect solutions are finer partitions of the ground truth, where \model{dataRSE} splits a ground-truth abstract variable into several abstract variables. However, as the number of samples grows, the BIC used by \model{dataRSE} empirically selects the correct number of clusters. Moreover, the absolute error in estimating the TCE drops from about $0.03$ at $n=1000$ to $0.003$ at $n=100000$. Since a finer partition still adjusts for all the relevant variables of the ground-truth confounders, the incorrect solutions still lead to small errors, although larger than those of the correct solutions. Therefore, a low error in estimating the TCE does not imply recovering the ground-truth abstraction. However, larger sample sizes lead to the recovery of the ground-truth high-level variables, which practitioners can then interpret and reason about beyond the treatment-outcome scenario.

Finally, we study the recovery of the coefficients of the abstraction function (\Cref{fig:tau_recovery}). In practice, even with a limited number of $n=5000$ samples, \model{dataRSE} mostly recovers the ground-truth abstraction coefficients up to scale and sign. As we report in \Cref{fig:app-tau-split-ms90} in \Cref{app:additional-experiments}, the small error is due to the runs that failed to recover the exact ground-truth relevant sets. Instead, whenever \model{dataRSE} recovers them, the learned abstraction vectors coincide with the ground-truth ones up to scale and sign, with an average absolute cosine similarity above $0.999$. Therefore, increasing the number of samples jointly improves the recovery of the sets of relevant variables and of the abstraction vectors.

In \Cref{app:additional-experiments}, we report several additional results. First, we relax the assumption of \Cref{thm:maximalpartitioning}, so that the ground truth can have coefficients close to violating the non-collinearity condition. As expected, \model{dataRSE} requires more samples to recover the ground truth, and its incorrect solutions can also merge the relevant sets of distinct abstract variables. Second, we consider further abstract graphs highlighting how the performance of \model{dataRSE} worsens due to the recursion depth more than to the number of confounders. Finally, we compare the BIC-based maximal partitioning of \Cref{alg:maximalpartitioning} with a rank-based partitioning that selects the number of clusters without fitting every candidate model. The rank-based strategy is consistently faster, but requires more samples to match the accuracy of \Cref{alg:maximalpartitioning}, especially on deeper abstract graphs.

\section{Conclusion}%
\label{sec:conclusion}

In this work, we tackled the problem of learning a causally sufficient high-level model over a causally sufficient low-level linear Additive Noise Model (ANM) from partial knowledge of the abstraction function. First, we considered the knowledge of a subset of abstract variables, defined by the set of low-level variables that they aggregate. Since a consistent abstract model might not be causally sufficient even when the low-level model is, we introduced the class of \emph{minimal sufficient extensions} (\model{MiSE}). The \model{MiSE} contains extensions of the provided partial knowledge that ensure causal sufficiency on the abstract model without introducing unneeded abstract variables. Next, we defined the Relevant Sufficiency Extender (\model{RSE}) algorithm to compute these extensions and proved its correctness and completeness.

Then, we focused on a scenario where we also know a linear abstraction map from the low-level variables to a strict subset of abstract variables. In this context, we studied how low-level samples can be used to filter the \model{MiSE} down to a \emph{parametrically consistent} extension, \ie, an extension that supports an interventionally consistent abstract model. Building on this, we extended \model{RSE} into \model{dataRSE}. We proved that at most one parametrically consistent minimal sufficient extension satisfies a non-collinearity condition on the abstract coefficients, and that, under the assumption of exact regression, \model{dataRSE} recovers it together with its abstraction coefficients up to scale. We validated \model{dataRSE} on synthetic causal abstractions satisfying this condition, where only the abstract treatment and outcome are provided. Our experiments show that \model{dataRSE} recovers the missing abstract variables and the abstraction function from a finite number of samples, and that the resulting smaller abstract model yields accurate estimates of the causal effect between the abstract treatment and outcome.

This work relies on the graphical characterization of linear abstraction between linear ANMs~\citep{massidda2024learning}. It is still an open question whether similar conditions also hold without the linearity assumption, and hence whether the concept of \model{MiSE} and the \model{RSE} and \model{dataRSE} algorithms could be extended to causal abstraction of non-linear models. Furthermore, both algorithms assume knowledge of the low-level graph, which restricts applications to scenarios where it is known by construction, \eg, artificial neural networks in mechanistic interpretability~\citep{bereska2024mechanistic}, complex simulation models, and engineered systems in general. Integrating causal discovery into causal abstraction learning is a non-trivial research direction that would enable more realistic applications. Finally, the running time of \model{dataRSE} grows with the number of new relevant variables, since the BIC-based maximal partitioning fits a model for each candidate number of clusters. The rank-based alternative described in \Cref{app:rankonepartitioning} avoids these fits, but requires more samples to reach the same accuracy (\Cref{app:plateau}). Combining the speed of the rank-based partitioning with the accuracy of the BIC-based one is an interesting future direction.

\bibliography{bibliography}

\newpage
\appendix
\onecolumn
\section{{T}-Abstractions}%
\label{app:t-abstraction}

\subsection{Assumptions}%
\label{app:assumptions}

Similarly to \citet{massidda2024learning}, we assume that both low-level and high-level causal models are faithful, \ie, in each model a conditional independence in the distribution implies a corresponding d-separation in the causal graph, and that the abstraction function does not yield cancelling paths from the low-level to the abstract variables.
\begin{restatable}[No Cancelling Paths]{assumption}{nocancellingpathsassumption}%
\label{assumption:nocancellingpaths}
    Let $\tau\colon\dom{\set{X}}\to\dom{\set{Y}}$ be a surjective function and
    $\scm{L} = (\set{X}, \set{E}, \set{f}, \dist{\set{E}})$ and $\scm{H} = (\set{Y}, \set{U}, \set{g}, \dist{\set{U}})$ be two linear ANMs. Then, we assume that $\scm{L}$ and $\scm{H}$ are faithful and that the joint SCM defined as
    \[
    \tau\circ\scm{L}=(\set{X}\cup\set{Y}, \set{E}, \set{f}\cup\setdef{\tau_j}_{Y_j\in\set{Y}}, \dist{\set{E}})
    \]
    does not have cancelling paths.
\end{restatable}

The non-cancellation of paths in the joint SCM can be intuitively interpreted as an extension of the faithfulness assumption to causal abstractions. The following example shows the pathological case in which the influence of a low-level variable is cancelled on an abstract variable, even when both the low-level and the high-level models are faithful.
\begin{example}[Cancelling Paths Assumption Violation]
    Let $\scm{L}=(\set{X}, \set{E}, \set{f}, \dist{\set{E}})$ be a linear ANM with the causal graph %
      \begin{center}
      \small
        \begin{tikzpicture}[node distance=1cm, minimum size=0.3cm]
        \node[draw, circle] (X1) {$X_1$};
        \node[draw, circle, left=of X1] (X2) {$X_2$};
        \node[draw, circle, right=of X1] (X3) {$X_3$};
        \draw[-{Latex}] (X1) -- (X2) node [midway, above] {$1$};
        \draw[-{Latex}] (X1) -- (X3) node [midway, above] {$1$};
      \end{tikzpicture}
      \end{center}
    where we assume all causal relations to
    have unit weight.
    Then, let $\scm{H}=(\set{Y},\set{U},\set{g},\dist{\set{U}})$ be a linear ANM with two disconnected variables $Y_1$ and $Y_2$, as in the causal graph %
      \begin{center}
      \begin{tikzpicture}[node distance=1.5cm]
        \node[draw, circle] (Y1) {$Y_1$};
        \node[draw, circle, right=of Y1] (Y2) {$Y_2$};
      \end{tikzpicture}
      \end{center}
    We can prove that,
    given the following transformation
    \begin{equation}
    \mat{T} = \begin{bmatrix}
    1&0&0\\
    0&1&-1\\
    \end{bmatrix}^\tr,
    \end{equation}
    $\scm{H}$ is a strong $\mat{T}$-abstraction of $\scm{L}$, \ie, it is consistent under all possible abstract interventions. Indeed, let $\mat{S}=\mat{T}$, \ie, $U_1=E_1$ and $U_2=E_2-E_3$. Then, denoting as $\bot$ the empty intervention, we check the consistency condition of a strong $\mat{T}$-abstraction (\Cref{def:tabstraction}) for the four abstract interventions $j$ by choosing the low-level interventions $i$ as follows:
    \begin{alignat*}{3}
        j&=\bot, &\quad i&=\bot, &\quad \scm{L}^i(\vec{e})\mat{T}&=(e_1,\, e_2-e_3)=\scm{H}^j(\vec{e}\mat{S}),\\
        j&=\operatorname{do}(Y_1\gets a), &\quad i&=\operatorname{do}(X_1\gets a), &\quad \scm{L}^i(\vec{e})\mat{T}&=(a,\, e_2-e_3)=\scm{H}^j(\vec{e}\mat{S}),\\
        j&=\operatorname{do}(Y_2\gets b), &\quad i&=\operatorname{do}(X_2\gets b, X_3\gets 0), &\quad \scm{L}^i(\vec{e})\mat{T}&=(e_1,\, b)=\scm{H}^j(\vec{e}\mat{S}),\\
        j&=\operatorname{do}(Y_1\gets a, Y_2\gets b), &\quad i&=\operatorname{do}(X_1\gets a, X_2\gets b, X_3\gets 0), &\quad \scm{L}^i(\vec{e})\mat{T}&=(a,\, b)=\scm{H}^j(\vec{e}\mat{S}).
    \end{alignat*}
    Without interventions on $X_2$ and $X_3$, the contributions of $X_1$ to $X_2$ and $X_3$ cancel in $Y_2$.
    Intuitively, this abstract model is not a faithful or interpretable representation of the low-level model.
    The issue arises because there is a cancelling path from $X_1$ to $Y_2$ through $X_2$ and $X_3$ in the joint SCM $\tau\circ\scm{L}$, even if both $\scm{L}$ and $\scm{H}$ satisfy the faithfulness assumption.
\end{example}

Faithfulness and the absence of cancelling paths concern the endogenous abstraction $\mat{T}$. We require an analogous condition on the exogenous abstraction $\mat{S}$ using a joint model over the low-level and the abstract exogenous terms, where each $U_j$ is a child of the low-level exogenous terms $E_k$ with $s_{kj}\neq 0$, \ie, the block exogenous terms $\set{E}_{\block{Y_j}}$.
\begin{restatable}[Exogenous Faithfulness]{assumption}{exogenousfaithfulnessassumption}%
\label{assumption:exogenousfaithfulness}
    Let $\scm{H}$ be a $\mat{T}$-abstraction of a causally sufficient linear ANM $\scm{L}=(\set{X}, \set{E}, \set{f}, \dist{\set{E}})$ with exogenous abstraction $\mat{S}$. Then, we assume that the joint distribution of $\set{E}\cup\set{U}$ is faithful to the graph with edges $E_k\to U_j$ for every $s_{kj}\neq 0$.
\end{restatable}

The assumption excludes cancellations among the contributions of different low-level exogenous terms to the abstract exogenous terms. We conjecture that this is implied neither by the faithfulness of $\scm{L}$ and $\scm{H}$ nor by the absence of cancelling paths. However, since we did not prove this result, we include it as an additional assumption of this work.

\subsection{Graphical Consistency}%
\label{app:graphcons}

Building on \citet{massidda2024learning}, we define the graphical conditions that characterize the sets of relevant variables of a consistent strong $\mat{T}$-abstraction. Similarly to \citet{anand2023causal,schooltink2025aligning}, we define the induced graph by studying $\rel{}$-direct paths between relevant variables of different abstract variables. Then, we define the relevant sets to be \emph{graphically consistent} if they satisfy the necessary graphical conditions of linear $\mat{T}$-abstraction. This requirement is stronger than similar conditions for Cluster DAGs~\citep{anand2023causal} and Partial Cluster DAGs~\citep{schooltink2025aligning}, as it requires multiple $\rel{}$-direct paths, which we show in the following example.

\begin{example}[Abstract Graphical Consistency]%
\label{ex:graphicalconsistency}
We provide an example to clarify how abstract graphical consistency differs from similar conditions for Cluster DAGs~(CDAGs; \citeauthor{anand2023causal}, \citeyear{anand2023causal}) and Partial Cluster DAGs~\citep{schooltink2025aligning}. CDAGs connect two clusters whenever there is an edge between their elements, while Partial CDAGs connect them whenever there is a ``mediated adjacency'', which can be interpreted as a $\rel{}$-direct path. In contrast, graphical consistency in $\mat{T}$-abstraction requires an outgoing $\rel{}$-direct path from each element of a cluster to at least one element of each child cluster.
\begin{center}
\resizebox{\textwidth}{!}{%
\small
\begin{tikzpicture}[node distance=1.5cm, minimum size=0.3cm]
\node[draw, circle, fill=red!40!white] (X1) {$X_1$};
\node[draw, circle, below of=X1, fill=red!40!white] (X2) {$X_2$};
\node[draw, circle, right of=X1, fill=green!40!white] (X3) {$X_3$};
\begin{pgfonlayer}{background}
\node[draw=red!60!black, very thick, rounded corners=3pt, fill=red!10!white, fit=(X1)(X2), inner sep=5pt] (Y1) {};
\node[draw=green!60!black, very thick, rounded corners=3pt, fill=green!10!white, fit=(X3), inner sep=5pt] (Y2) {};
\end{pgfonlayer}
\draw[-{Latex}] (X1) -- (X3);
\node[above of=X3, xshift=-0.75cm] {%
\begin{tikzpicture}
    \node[anchor=east] at (0.2,1.2) {CDAG: \cmark};
    \node[anchor=east] at (0.2,0.8) {Partial CDAG: \cmark};
    \node[anchor=east] at (0.2,0.4) {Abstraction: \xmark};
\end{tikzpicture}
};
\end{tikzpicture}
\hspace{0.25cm}
\begin{tikzpicture}[node distance=1.5cm, minimum size=0.3cm]
\node[draw, circle, fill=red!40!white] (X1) {$X_1$};
\node[draw, circle, below of=X1, fill=red!40!white] (X2) {$X_2$};
\node[draw, circle, right of=X1, fill=green!40!white] (X3) {$X_3$};
\begin{pgfonlayer}{background}
\node[draw=red!60!black, very thick, rounded corners=3pt, fill=red!10!white, fit=(X1)(X2), inner sep=5pt] (Y1) {};
\node[draw=green!60!black, very thick, rounded corners=3pt, fill=green!10!white, fit=(X3), inner sep=5pt] (Y2) {};
\end{pgfonlayer}
\draw[-{Latex}] (X1) -- (X3);
\draw[-{Latex}] (X2) -- (X3);
\node[above of=X3, xshift=-0.75cm] {%
\begin{tikzpicture}
    \node[anchor=east] at (0.2,1.2) {CDAG: \cmark};
    \node[anchor=east] at (0.2,0.8) {Partial CDAG: \cmark};
    \node[anchor=east] at (0.2,0.4) {Abstraction: \cmark};
\end{tikzpicture}
};
\end{tikzpicture}
\hspace{0.25cm}
\begin{tikzpicture}[node distance=1.5cm, minimum size=0.3cm]
\node[draw, circle, fill=red!40!white] (X1) {$X_1$};
\node[draw, circle, below of=X1, fill=red!40!white] (X2) {$X_2$};
\node[draw, circle, right of=X1] (X3) {$X_3$};
\node[draw, circle, right of=X3, fill=green!40!white] (X4) {$X_4$};
\begin{pgfonlayer}{background}
\node[draw=red!60!black, very thick, rounded corners=3pt, fill=red!10!white, fit=(X1)(X2), inner sep=5pt] (Y1) {};
\node[draw=green!60!black, very thick, rounded corners=3pt, fill=green!10!white, fit=(X4), inner sep=5pt] (Y2) {};
\end{pgfonlayer}
\draw[-{Latex}] (X1) -- (X3);
\draw[-{Latex}] (X3) -- (X4);
\node[above of=X4, xshift=-0.75cm] {%
\begin{tikzpicture}
    \node[anchor=east] at (0.2,1.2) {CDAG: \xmark};
    \node[anchor=east] at (0.2,0.8) {Partial CDAG: \cmark};
    \node[anchor=east] at (0.2,0.4) {Abstraction: \xmark};
\end{tikzpicture}
};
\end{tikzpicture}
\hspace{0.25cm}
\begin{tikzpicture}[node distance=1.5cm, minimum size=0.3cm]
\node[draw, circle, fill=red!40!white] (X1) {$X_1$};
\node[draw, circle, below of=X1, fill=red!40!white] (X2) {$X_2$};
\node[draw, circle, right of=X1] (X3) {$X_3$};
\node[draw, circle, right of=X3, fill=green!40!white] (X4) {$X_4$};
\begin{pgfonlayer}{background}
\node[draw=red!60!black, very thick, rounded corners=3pt, fill=red!10!white, fit=(X1)(X2), inner sep=5pt] (Y1) {};
\node[draw=green!60!black, very thick, rounded corners=3pt, fill=green!10!white, fit=(X4), inner sep=5pt] (Y2) {};
\end{pgfonlayer}
\draw[-{Latex}] (X1) -- (X3);
\draw[-{Latex}] (X3) -- (X4);
\draw[-{Latex}] (X2) -- (X3);
\node[above of=X4, xshift=-0.75cm] {%
\begin{tikzpicture}
    \node[anchor=east] at (0.2,1.2) {CDAG: \xmark};
    \node[anchor=east] at (0.2,0.8) {Partial CDAG: \cmark};
    \node[anchor=east] at (0.2,0.4) {Abstraction: \cmark};
\end{tikzpicture}
};
\end{tikzpicture}
}
\end{center}
\end{example}

\subsection{Causal Sufficiency}%
\label{app:sufficiency}

\begin{example}[Insufficient Relevant Sets]
    Let $\scm{L}$ be a linear ANM with the following graph
    \begin{center}
    \small
    \begin{tikzpicture}[node distance=1cm, minimum size=0.3cm]
    \node[draw, circle] (X1) {$X_1$};
    \node[draw, circle, left=of X1, yshift=-1cm, fill=red!40!white] (X2) {$X_2$};
    \node[draw, circle, right=of X1, yshift=-1cm, fill=green!40!white] (X3) {$X_3$};
    \draw[-{Latex}] (X1) -- (X2) node [midway, above] {$1$};
    \draw[-{Latex}] (X1) -- (X3) node [midway, above] {$1$};
    \draw[-{Latex}] (X2) -- (X3) node [midway, below] {$1$};
    \begin{pgfonlayer}{background}
    \node[draw=red!60!black, very thick, rounded corners=3pt, fill=red!10!white, fit=(X2), inner sep=4pt, label={[text=red!60!black]left:{\scriptsize$\rel{Y_1}$}}] {};
    \node[draw=green!60!black, very thick, rounded corners=3pt, fill=green!10!white, fit=(X3), inner sep=4pt, label={[text=green!60!black]right:{\scriptsize$\rel{Y_2}$}}] {};
    \end{pgfonlayer}
    \end{tikzpicture}
    \end{center}
    Then, we can abstract the model with two abstract variables $Y_1, Y_2$ such that $\rel{Y_1}=\setdef{X_2}$ and $\rel{Y_2}=\setdef{X_3}$, leading to the following \emph{overlapping} blocks $\block{Y_1}=\setdef{X_1, X_2}$ and $\block{Y_2}=\setdef{X_1, X_3}$, and the following abstract graph
    \begin{center}
    \small
    \begin{tikzpicture}[node distance=1cm, minimum size=0.3cm]
    \node[draw, circle, fill=red!40!white] (Y1) {$Y_1$};
    \node[draw, circle, right=of Y1, fill=green!40!white] (Y2) {$Y_2$};
    \draw[-{Latex}] (Y1) -- (Y2) node [midway, below] {$1$};
    \draw[{Latex}-{Latex}, bend left] (Y1) to (Y2);
    \end{tikzpicture}
    \end{center}
    The abstract model is provably a $\mat{T}$-abstraction of the low-level model for the transformation
    \begin{equation}
    \mat{T} = \begin{bmatrix}
    0 & 1 & 0\\
    0 & 0 & 1
    \end{bmatrix}^\tr.
    \end{equation}
    The causal effect $\dist{Y_2 \mid \operatorname{do}(Y_1)}$ is not identifiable from the abstract model. However, by using the corresponding low-level relevant variables, the causal effect $\dist{\rel{Y_2}\mid\operatorname{do}(\rel{Y_1})} = \dist{X_3\mid\operatorname{do}(X_2)}$ is identifiable, since $\setdef{X_1}$ is a valid adjustment set. The issue arises because the adjustment set $\setdef{X_1}$ is not abstracted.
\end{example}

In the previous example, the insufficient relevant sets admit a sufficient extension. However, this is not always the case, and the \model{MiSE} of a graphically consistent but insufficient set of relevant variables can be empty, as we show in the following example.

\begin{example}[Empty \model{MiSE}]%
\label{ex:suffandgraph}
Let $\scm{H}$ be an abstract SCM with four variables
and $\scm{L}$ be a causally sufficient low-level SCM with six variables.
We represent the low-level causal graph as follows,
where we employ colors to denote relevant variables
of distinct abstract variables, as in
$\rel{Y_1} = \setdef{X_1}$,
$\rel{Y_2} = \setdef{X_2, X_3}$,
$\rel{Y_3} = \setdef{X_5}$, and
$\rel{Y_4} = \setdef{X_6}$.
This choice of relevant variables is graphically consistent but not sufficient, since the low-level variable $X_4$ is shared by the blocks of $Y_3$ and $Y_4$. However, abstracting $X_4$ would break graphical consistency, since $X_3$ has a $\rel{}$-direct path towards $X_4$ while $X_2$ does not.

\begin{center}
\small
\begin{tikzpicture}[node distance=1.5cm, minimum size=0.3cm, scale=0.94, transform shape]
\node[draw, circle, fill=yellow!40!white] (X0) {$X_1$};
\node[draw, circle, right of=X0, fill=red!40!white] (X1) {$X_2$};
\node[draw, circle, right of=X1, fill=red!40!white] (X2) {$X_3$};
\node[draw, circle, right of=X2] (X3) {$X_4$};
\node[draw, circle, right of=X3, yshift=1cm, fill=green!40!white] (X4) {$X_5$};
\node[draw, circle, right of=X3, yshift=-1cm, fill=blue!40!white] (X5) {$X_6$};
\draw[-{Latex[length=2mm]}] (X0) -- (X1);
\draw[-{Latex[length=2mm]}] (X2) -- (X3);
\draw[-{Latex[length=2mm]}] (X3) -- (X4);
\draw[-{Latex[length=2mm]}] (X3) -- (X5);
\draw[-{Latex[length=2mm]}] (X1) to [bend left] (X4);
\draw[-{Latex[length=2mm]}] (X1) to [bend right] (X5);
\begin{pgfonlayer}{background}
\node[draw=yellow!60!black, very thick, rounded corners=3pt, fill=yellow!10!white, fit=(X0), inner sep=4pt, label={[text=yellow!60!black]above:{\scriptsize$\rel{Y_1}$}}] {};
\node[draw=red!60!black, very thick, rounded corners=3pt, fill=red!10!white, fit=(X1)(X2), inner sep=4pt, label={[text=red!60!black]above left:{\scriptsize$\rel{Y_2}$}}] {};
\node[draw=green!60!black, very thick, rounded corners=3pt, fill=green!10!white, fit=(X4), inner sep=4pt, label={[text=green!60!black]right:{\scriptsize$\rel{Y_3}$}}] {};
\node[draw=blue!60!black, very thick, rounded corners=3pt, fill=blue!10!white, fit=(X5), inner sep=4pt, label={[text=blue!60!black]right:{\scriptsize$\rel{Y_4}$}}] {};
\end{pgfonlayer}
\end{tikzpicture}
\end{center}
\end{example}

\newpage
\section{Classification of Related Work}%
\label{app:related}

We detail the classification of causal abstraction learning methods reported in \Cref{tab:related}.

\paragraph{Jointly Learning Consistent Abstractions.}
\citet{zennaro2023jointly} learn abstractions in a category-theoretic framework compatible with $\tau$-abstraction~\citep{beckers2019abstracting} that comprises a set of relevant low-level variables, a surjective map from relevant variables to high-level variables, and a collection of surjective maps on their values. Their problem statement assumes that both the low-level and the high-level SCMs are fully known, together with the set of relevant variables and their map to the high-level variables, and the method learns the maps between values minimizing the abstraction error. Therefore, we mark as required both graphs $\graph{G}_{\set{X}}$ and $\graph{G}_{\set{Y}}$, both distributions $\dist{\set{X}}$ and $\dist{\set{Y}}$, and the relevant variables $\rel{}$. Since the abstraction error is evaluated over a set of interventional distributions entailed by the known SCMs, we also mark the low-level interventional samples $\dist{\set{X}}^{\set{I}}$ as required, although they are entailed by the known SCMs rather than sampled. The only outcome of the method is therefore the value-abstraction function $\tau$. Notably, the method also assumes discrete variables with finite domains.

\paragraph{Targeted Causal Reduction.}
\citet{kekic2024targeted} explain a known low-level target phenomenon through a small number of learned high-level causes. The method requires samples from the observational and shift-interventional distributions of the low-level model ($\dist{\set{X}}$ and $\dist{\set{X}}^{\set{I}}$), but not the low-level graph $\graph{G}_{\set{X}}$. The abstraction function of the target, and thus its relevant variables, is known, while the abstraction functions of the high-level causes, their relevant variables, and the intervention map are learned, with a regularization term encouraging disjoint relevant sets. Hence, the method completes partial knowledge of both $\tau$ and $\rel{}$. The high-level graph $\graph{G}_{\set{Y}}$ is fixed a priori to independent causes pointing to the target, while the parameters of the high-level mechanisms, and thus $\dist{\set{Y}}$, are learned. Despite the different setting, Targeted Causal Reduction shares with \model{dataRSE} the idea of starting from a known abstract variable of interest and completing the abstraction around it.

\paragraph{Unsupervised Causal Abstraction.}
\citet{zhu2024unsupervised} extend Targeted Causal Reduction to the case where no target is known, combining the causal consistency loss with a projection loss from the Principal Component Analysis literature. The objective is defined over low-level observational and shift-interventional samples ($\dist{\set{X}}$ and $\dist{\set{X}}^{\set{I}}$). The candidate abstractions, and thus $\tau$ and $\rel{}$, are selected among the partitions formed by contiguous variables in the topological ordering of the low-level graph $\graph{G}_{\set{X}}$, which is therefore known. Furthermore, the abstraction function of the first abstract variable is set to the closed-form solution of Targeted Causal Reduction. This closed form holds only if the high-level graph $\graph{G}_{\set{Y}}$ consists of a single abstract cause pointing to an abstract target; therefore, the form of $\graph{G}_{\set{Y}}$ is also assumed. The parameters of the high-level model, and thus $\dist{\set{Y}}$, are learned jointly with $\tau$.

\paragraph{\model{COTA}.}
\citet{felekis2024causal} tackle causal abstraction learning as an optimal transport problem. The method assumes access to the graphs $\graph{G}_{\set{X}}$ and $\graph{G}_{\set{Y}}$, a finite set of interventions, a map between low-level and high-level interventions, and unpaired samples from the observational and interventional distributions of both models ($\dist{\set{X}}$, $\dist{\set{X}}^{\set{I}}$, and $\dist{\set{Y}}$). The high-level interventional samples and the intervention map are not reported in \Cref{tab:related}, as no other method in the table requires them. The method learns a stochastic map $\tau$ from low-level to high-level values without assuming a constructive abstraction, thus implicitly determining the relevant variables $\rel{}$.

\paragraph{Interventionally Consistent Surrogates.}
\citet{dyer2024interventionally} learn high-level surrogates of complex simulators, which are treated as implicit low-level models. The method does not exploit the low-level graph $\graph{G}_{\set{X}}$, but requires running the simulator under the interventions of interest ($\dist{\set{X}}$ and $\dist{\set{X}}^{\set{I}}$). The abstraction function $\tau$, and thus $\rel{}$, is pre-specified by domain experts, as it defines the aggregate quantities of interest. The method learns the intervention map jointly with the parameters of a differentiable surrogate ($\dist{\set{Y}}$), whose structure $\graph{G}_{\set{Y}}$ is fixed by the chosen parametric family. Since $\tau$ is known, high-level samples are obtained from low-level ones, while the learned surrogate provides the high-level model.

\paragraph{Neural Causal Abstractions.}
\citet{xia2024neural} define abstractions through a clustering of low-level variables, determining the relevant variables $\rel{}$ of each high-level variable, and a clustering of their domains, determining the values of high-level variables. Given a collection of low-level datasets ($\dist{\set{X}}$), the clustering of variables, and the cluster causal diagram $\graph{G}_{\set{Y}}$, they fit a neural causal model over the high-level variables ($\dist{\set{Y}}$) to identify and estimate high-level causal queries. Further, they learn $\tau$ through a representation of each cluster of variables, relaxing the knowledge of the clustering of domains. The method requires neither the low-level graph $\graph{G}_{\set{X}}$ nor interventional data $\dist{\set{X}}^{\set{I}}$, although interventional datasets extend the set of identifiable queries.

\paragraph{\model{Abs-LiNGAM}.}
\citet{massidda2024learning} learn linear abstractions between linear non-Gaussian models from a large set of low-level samples and a small set of paired low-level and high-level observational samples ($\dist{\set{X}}$ and $\dist{\set{Y}}$). From the paired samples, the method recovers $\tau$ by least squares, and hence $\rel{}$, and learns $\graph{G}_{\set{Y}}$ by causal discovery. Then, it exploits the constraints entailed by the learned abstraction to speed up the recovery of $\graph{G}_{\set{X}}$.

\paragraph{\model{LinSEPAL}.}
\citet{dacunto2025causal} require the low-level and high-level covariance matrices, estimated from unpaired observational samples ($\dist{\set{X}}$ and $\dist{\set{Y}}$), and partial prior knowledge on the support of the abstraction, denoting which low-level variables might constitute each high-level variable. From these, they learn the abstraction coefficients $\tau$ together with their support, thus completing $\rel{}$. Neither graph is required or learned.

\paragraph{\model{RSE} and \model{dataRSE}.}
Our \model{RSE} algorithm requires the low-level graph $\graph{G}_{\set{X}}$ and the relevant variables $\rel{}$ of a strict subset of abstract variables, and constructs the remaining abstract variables together with the abstract graph $\graph{G}_{\set{Y}}$ they entail. The \model{dataRSE} algorithm additionally requires observational low-level samples $\dist{\set{X}}$ and the abstraction function $\tau$ of the provided abstract variables, \ie, the corresponding columns of $\mat{T}$, and recovers the full $\tau$ together with the parameters of the abstract model, and thus $\dist{\set{Y}}$. Neither algorithm requires interventional data, high-level samples, or the abstract graph. Hence, they are the only methods in \Cref{tab:related} that complete partial knowledge of the abstraction without any high-level information.

\newpage
\section{RSE Execution}%
\label{app:rseexecution}
\begin{example}[RSE Execution]%
\label{ex:rseexecution}

Let $\graph{G}_{\set{X}}$ be a low-level graph over variables $\set{X}=\setdef{X_1, \ldots, X_6}$ and $\rel{}$ be the set of relevant variables with $\rel{Y_1}=\setdef{X_1}$ and $\rel{Y_2}=\setdef{X_2}$ for two abstract variables $\set{Y}=\setdef{Y_1, Y_2}$. The \model{RSE} algorithm selects the only jointly targeted set $\set{V}_1=\setdef{Y_1, Y_2}$ and computes the new relevant variables $\set{Z}_1=\setdef{X_3, X_4, X_6}$ (a.). The intersection $\block{Y_1}\cap\block{Y_2}$ also contains the variable $X_5$, which is not in $\set{Z}_1$ because it has no outgoing edge to a non-intersected variable.

Having selected three new relevant variables $\set{Z}_1$, the \model{RSE} algorithm explores all five possible partitions. The first three partitions (b., c., and d.) lead to graphically inconsistent sets of relevant variables and are therefore discarded. Instead, the fourth partition (e.) is graphically consistent, causally sufficient, and consequently included in the set of solutions.

Finally, the fifth partition (f.) is graphically consistent but not causally sufficient: \model{RSE} is therefore applied recursively. In this case, there is a unique intersected variable $X_5$, which is also the only new relevant variable of this recursive call. By adding $X_5$ to a new abstract variable, the resulting set of relevant variables (g.) is graphically consistent and causally sufficient. The algorithm returns two solutions (e. and g.), which coincide with the minimal sufficient extensions of $\rel{}$ on $\graph{G}_{\set{X}}$. In the figure, node colors denote relevant sets, and frames denote graphically inconsistent sets (red), solutions (green), and graphically consistent but not sufficient sets (black).

\begin{center}
\resizebox{\textwidth}{!}{\newcommand{\rsepanel}[9]{%
  \begin{scope}[shift={(#2,#3)}]
    \node[rnode, fill=#4] (#1-x1) at (0.0, 2.4) {$X_1$};
    \node[rnode, fill=#5] (#1-x2) at (1.4, 2.4) {$X_2$};
    \node[rnode, fill=#6] (#1-x3) at (0.0, 1.2) {$X_3$};
    \node[rnode, fill=#7] (#1-x4) at (1.4, 1.2) {$X_4$};
    \node[rnode, fill=#8] (#1-x5) at (0.0, 0.0) {$X_5$};
    \node[rnode, fill=#9] (#1-x6) at (1.4, 0.0) {$X_6$};
    \draw[redge] (#1-x1) -- (#1-x2);
    \draw[redge] (#1-x3) -- (#1-x1);
    \draw[redge] (#1-x3) -- (#1-x2);
    \draw[redge] (#1-x4) -- (#1-x1);
    \draw[redge] (#1-x4) -- (#1-x2);
    \draw[redge] (#1-x5) -- (#1-x3);
    \draw[redge] (#1-x5) -- (#1-x4);
    \draw[redge] (#1-x6) -- (#1-x4);
    \draw[redge] (#1-x6) to[bend right=35] (#1-x2);
  \end{scope}%
}
\newcommand{\rseframe}[3]{%
  \node[rframe, draw=#2, fit=(#1-x1)(#1-x2)(#1-x5)(#1-x6),
        label={[anchor=south west, inner sep=1pt]north west:{\emph{(#3.)}}}] (#1) {};%
}
\begin{tikzpicture}[
    rnode/.style={draw, circle, minimum size=0.75cm, inner sep=0pt},
    redge/.style={-{Latex[length=2mm]}},
    rframe/.style={line width=1.6pt, rounded corners=3pt, inner xsep=7pt, inner ysep=7pt},
]
  \def\cna{white}
  \def\cya{red!40!white}
  \def\cyb{green!40!white}
  \def\cra{blue!30!white}
  \def\crb{yellow!50!white}
  \def\crc{violet!30!white}
  \def\crd{orange!40!white}
  \rsepanel{a}{0.0}{0.0}{\cya}{\cyb}{\cna}{\cna}{\cna}{\cna}
  \rsepanel{b}{3.8}{2.2}{\cya}{\cyb}{\cra}{\cra}{\cna}{\cra}
  \rsepanel{c}{3.8}{-2.2}{\cya}{\cyb}{\cra}{\crb}{\cna}{\cra}
  \rsepanel{d}{7.0}{2.2}{\cya}{\cyb}{\cra}{\crb}{\cna}{\crb}
  \rsepanel{e}{7.0}{-2.2}{\cya}{\cyb}{\cra}{\cra}{\cna}{\crb}
  \rsepanel{f}{10.2}{0.0}{\cya}{\cyb}{\cra}{\crb}{\cna}{\crc}
  \rsepanel{g}{14.0}{0.0}{\cya}{\cyb}{\cra}{\crb}{\crd}{\crc}
  \rseframe{a}{black}{a}
  \rseframe{b}{red!60!black}{b}
  \rseframe{c}{red!60!black}{c}
  \rseframe{d}{red!60!black}{d}
  \rseframe{e}{green!60!black}{e}
  \rseframe{f}{black}{f}
  \rseframe{g}{green!60!black}{g}
  \node[below=2pt of a] {$\set{Z}=\setdef{X_3,X_4,X_6}$};
  \node[below=2pt of f] {$\set{Z}=\setdef{X_5}$};
  \node[draw, thick, dashed, inner xsep=10pt, inner ysep=16pt, fit=(b)(c)(d)(e)(f)] (box) {};
  \draw[-{Latex[length=3mm]}, very thick] (a.east) -- (box.west |- a.east);
  \draw[-{Latex[length=3mm]}, very thick] (f.east) -- (g.west);
\end{tikzpicture}
}
\end{center}
\end{example}

\clearpage
\section{Algorithms}\label{sec:algorithms}

\begin{algorithm}
\caption{Blocks from Relevant Variables}
\label{alg:blocks}
\begin{algorithmic}[1]
\Require Relevant sets $\rel{}$, low-level graph $\graph{G}_{\set{X}}$ on variables $\set{X}$
\Ensure Blocks $\block{}=\setdef{\block{Y_i}}_{Y_i\in\set{Y}}$

\State $\block{} \gets \emptyset$

\For{$\rel{Y_i} \in \rel{}$}
    \State Visited $\gets \rel{Y_i}$
    \State Stack $\gets \rel{Y_i}$

    \While{Stack $\neq \emptyset$}
        \State $X \gets$ Pop(Stack)

        \For{$X' \in \pa{X, \graph{G}_{\set{X}}}$}
            \If{$X' \notin \text{Visited} \wedge X' \notin \bigcup\rel{}$}
                \State Visited $\gets$ Visited $\cup \{X'\}$
                \State Stack $\gets$ Stack $\cup \{X'\}$
            \EndIf
        \EndFor
    \EndWhile

    \State $\block{Y_i} \gets \text{Sorted}( \text{Visited} )$
    \State $\block{} \gets \block{} \cup \setdef{\block{Y_i}}$
\EndFor

\State \Return $\block{}$
\end{algorithmic}
\end{algorithm}

\begin{algorithm}
\caption{Abstract Graph Construction}
\label{alg:absgraph}
\begin{algorithmic}[1]
\Require Relevant sets $\rel{}$, low-level graph $\graph{G}_{\set{X}}$ on variables $\set{X}$
\Ensure Abstract graph $\graph{G}_{\set{Y}}$ on variables $\set{Y}$

\State $\set{Y} \gets \setdef{1, \ldots, \size{\rel{}}}$
\State $\graph{G}_{\set{Y}} \gets$ directed graph with nodes $\set{Y}$
\State $\block{} \gets \operatorname{Blocks}(\rel{}, \graph{G}_{\set{X}})$ \Comment{\Cref{alg:blocks}}

\For{$Y_i\in\set{Y}$} \Comment{Iterate over abstract nodes $Y_i$}
    \For{$X\in\rel{Y_i}$} \Comment{Iterate over the relevant variables of $Y_i$}
        \For{$X'\in\ch{X,\graph{G}_{\set{X}}}$} \Comment{Iterate over the direct children of $X$}
            \For{$Y_j\in\targets{X'}{\block{}}$} \Comment{Abstract targets of $X'$, \ie{} blocks containing it}
                \If{$Y_j\neq Y_i$}
                    \State Add edge $(Y_i, Y_j)$ to $\graph{G}_{\set{Y}}$
                \EndIf
            \EndFor
        \EndFor
    \EndFor
\EndFor

\State \Return $\graph{G}_{\set{Y}}$
\end{algorithmic}
\end{algorithm}

\begin{algorithm}
\caption{Graphical Consistency}
\label{alg:graphcons}
\begin{algorithmic}[1]
\Require Relevant sets $\rel{}$, low-level graph $\graph{G}_{\set{X}}$ on variables $\set{X}$
\Ensure True if and only if $\rel{}$ is graphically consistent on $\graph{G}_{\set{X}}$

\State $\graph{G}_{\set{Y}} \gets \operatorname{AbstractGraph}(\rel{}, \graph{G}_{\set{X}})$
\For{$(Y_i \to Y_j)\in\graph{G}_{\set{Y}}$, $X \in \rel{Y_i}$} \Comment{Search a $\rel{}$-direct path from $X$ to $\rel{Y_j}$}
    \State HasPath $\gets$ False, Queue $\gets \setdef{X}$, Visited $\gets \setdef{X}$
    \While{Queue $\neq \emptyset$ and not HasPath}
        \State $X' \gets $ Pop(Queue)
        \For{$X'' \in \ch{X', \graph{G}_{\set{X}}}\setminus\text{Visited}$}
            \State Visited $\gets$ Visited $\cup\setdef{X''}$
            \If{$X'' \in \rel{Y_j}$}
                \State HasPath $\gets$ True
            \ElsIf{$X'' \notin \bigcup\rel{}$} \Comment{Only non-relevant variables extend the path}
                \State Queue $\gets$ Queue $\cup \setdef{X''}$
            \EndIf
        \EndFor
    \EndWhile
    \If{not HasPath}
        \State \Return False
    \EndIf
\EndFor
\State \Return True
\end{algorithmic}
\end{algorithm}

\begin{algorithm}
\caption{Jointly Targeted Sets}
\label{alg:intersections}
\begin{algorithmic}[1]
\Require Blocks $\block{}=\setdef{\block{Y_i}}_{Y_i\in\set{Y}}$
\Ensure Jointly targeted sets $\set{I}$

\State $\set{I} \gets \emptyset$
\State $\text{inverted} \gets \emptyset$ \Comment{Map variable $\to$ set of abstract variables}

\For{$\block{Y} \in \block{}$}
    \For{$X \in \block{Y}$}
        \State $\text{inverted}[X] \gets \text{inverted}[X] \cup \{Y\}$
    \EndFor
\EndFor

\For{$X \in \text{keys}(\text{inverted})$}
    \State $\set{S} \gets \text{inverted}[X]$
    \Comment{The exact abstract targets of $X$}
    \If{$\size{\set{S}} \geq 2$}
        \State $\set{I} \gets \set{I} \cup \{\set{S}\}$
    \EndIf
\EndFor

\State \Return $\set{I}$
\end{algorithmic}
\end{algorithm}

\clearpage
\section{Parametric Consistency of Minimal Sufficient Extensions}%
\label{app:paraminconsistency}

\subsection{Parametric Consistency of the Extensions}%
\label{app:miseparams}

In this subsection, we detail \Cref{ex:miseparams}. Let $\scm{L}$ be a linear ANM over the variables $X_1,\ldots,X_5$, where each of $X_1, X_2, X_3$ is a parent of both $X_4$ and $X_5$, and $X_4$ is a parent of $X_5$. The initial relevant sets $\rel{Y_T}=\setdef{X_4}$ and $\rel{Y_O}=\setdef{X_5}$ of an abstract treatment $Y_T$ and an abstract outcome $Y_O$ induce the set of new relevant variables $\set{Z}=\setdef{X_1, X_2, X_3}$. Then, we consider the following two minimal sufficient extensions:
\begin{itemize}
    \item $\rel{}'$, which adds $\rel{Y_1}'=\setdef{X_1}$ and $\rel{Y_2}'=\setdef{X_2, X_3}$;
    \item $\rel{}''$, which adds $\rel{Y_3}''=\setdef{X_1, X_2, X_3}$.
\end{itemize}

\begin{center}
\small
\begin{tikzpicture}[%
node distance=1.5cm,
scale=0.82, transform shape,
minimum size=0.3cm,
]
\node[draw, circle] (X1) {$X_1$};
\node[draw, circle, right of=X1] (X2) {$X_2$};
\node[draw, circle, right of=X2] (X3) {$X_3$};
\node[draw, circle, below of=X1, xshift=0.5cm, fill=blue!40!white] (X4) {$X_4$};
\node[draw, circle, below of=X3, xshift=-0.5cm, fill=red!40!white] (X5) {$X_5$};
\node[below of=X2, yshift=-1cm] (label) {\emph{a.} $\rel{}$ on $\scm{L}$};

\draw[-{Latex}] (X1) -- (X4);
\draw[-{Latex}] (X1) -- (X5);
\draw[-{Latex}] (X2) -- (X4);
\draw[-{Latex}] (X2) -- (X5);
\draw[-{Latex}] (X3) -- (X4);
\draw[-{Latex}] (X3) -- (X5);
\draw[-{Latex}] (X4) -- (X5);
\begin{pgfonlayer}{background}
\node[draw=blue!60!black, very thick, rounded corners=3pt, fill=blue!10!white, fit=(X4), inner sep=4pt, label={[text=blue!60!black]left:{\scriptsize$\rel{Y_T}$}}] {};
\node[draw=red!60!black, very thick, rounded corners=3pt, fill=red!10!white, fit=(X5), inner sep=4pt, label={[text=red!60!black]right:{\scriptsize$\rel{Y_O}$}}] {};
\end{pgfonlayer}
\end{tikzpicture}
\hspace{2em}
\begin{tikzpicture}[%
node distance=1.5cm,
scale=0.92, transform shape,
minimum size=0.3cm,
]
\node[draw, circle, fill=yellow!40!white] (X1) {$X_1$};
\node[draw, circle, right of=X1, fill=green!40!white] (X2) {$X_2$};
\node[draw, circle, right of=X2, fill=green!40!white] (X3) {$X_3$};
\node[draw, circle, below of=X1, xshift=0.5cm, fill=blue!40!white] (X4) {$X_4$};
\node[draw, circle, below of=X3, xshift=-0.5cm, fill=red!40!white] (X5) {$X_5$};
\node[below of=X2, yshift=-1cm] (label) {\emph{b.} $\rel{}'$ on $\scm{L}$};

\draw[-{Latex}] (X1) -- (X4);
\draw[-{Latex}] (X1) -- (X5);
\draw[-{Latex}] (X2) -- (X4);
\draw[-{Latex}] (X2) -- (X5);
\draw[-{Latex}] (X3) -- (X4);
\draw[-{Latex}] (X3) -- (X5);
\draw[-{Latex}] (X4) -- (X5);
\begin{pgfonlayer}{background}
\node[draw=yellow!60!black, very thick, rounded corners=3pt, fill=yellow!10!white, fit=(X1), inner sep=4pt, label={[text=yellow!60!black]above:{\scriptsize$\rel{Y_1}$}}] {};
\node[draw=green!60!black, very thick, rounded corners=3pt, fill=green!10!white, fit=(X2)(X3), inner sep=4pt, label={[text=green!60!black]above:{\scriptsize$\rel{Y_2}$}}] {};
\node[draw=blue!60!black, very thick, rounded corners=3pt, fill=blue!10!white, fit=(X4), inner sep=4pt, label={[text=blue!60!black]left:{\scriptsize$\rel{Y_T}$}}] {};
\node[draw=red!60!black, very thick, rounded corners=3pt, fill=red!10!white, fit=(X5), inner sep=4pt, label={[text=red!60!black]right:{\scriptsize$\rel{Y_O}$}}] {};
\end{pgfonlayer}
\end{tikzpicture}
\hspace{2em}
\small
\begin{tikzpicture}[%
node distance=1.5cm,
scale=0.92, transform shape,
minimum size=0.3cm,
]
\node[draw, circle, fill=orange!40!white] (X1) {$X_1$};
\node[draw, circle, right of=X1, fill=orange!40!white] (X2) {$X_2$};
\node[draw, circle, right of=X2, fill=orange!40!white] (X3) {$X_3$};
\node[draw, circle, below of=X1, xshift=0.5cm, fill=blue!40!white] (X4) {$X_4$};
\node[draw, circle, below of=X3, xshift=-0.5cm, fill=red!40!white] (X5) {$X_5$};
\node[below of=X2, yshift=-1cm] (label) {\emph{c.} $\rel{}''$ on $\scm{L}$};

\draw[-{Latex}] (X1) -- (X4);
\draw[-{Latex}] (X1) -- (X5);
\draw[-{Latex}] (X2) -- (X4);
\draw[-{Latex}] (X2) -- (X5);
\draw[-{Latex}] (X3) -- (X4);
\draw[-{Latex}] (X3) -- (X5);
\draw[-{Latex}] (X4) -- (X5);
\begin{pgfonlayer}{background}
\node[draw=orange!60!black, very thick, rounded corners=3pt, fill=orange!10!white, fit=(X1)(X2)(X3), inner sep=4pt, label={[text=orange!60!black]above:{\scriptsize$\rel{Y_3}$}}] {};
\node[draw=blue!60!black, very thick, rounded corners=3pt, fill=blue!10!white, fit=(X4), inner sep=4pt, label={[text=blue!60!black]left:{\scriptsize$\rel{Y_T}$}}] {};
\node[draw=red!60!black, very thick, rounded corners=3pt, fill=red!10!white, fit=(X5), inner sep=4pt, label={[text=red!60!black]right:{\scriptsize$\rel{Y_O}$}}] {};
\end{pgfonlayer}
\end{tikzpicture}
\end{center}

We fix the low-level parameters such that $\rel{}'$ is parametrically consistent by directly constructing a strong $\mat{T}$-abstraction and then show that $\rel{}''$ is not parametrically consistent. Let the coefficient matrix of $\scm{L}$ be
\begin{equation*}
    \mat{W}=
    \begin{bmatrix}
    0 & 0 & 0 & 1 & 2\\
    0 & 0 & 0 & 1 & 1\\
    0 & 0 & 0 & 2 & 2\\
    0 & 0 & 0 & 0 & 1\\
    0 & 0 & 0 & 0 & 0
    \end{bmatrix}.
\end{equation*}
Then, we prove that, given the following coefficient matrix $\mat{M}$ of an abstract ANM $\scm{H}$ over $Y_1, Y_2, Y_T, Y_O$, $\scm{H}$ is a strong $\mat{T}$-abstraction of $\scm{L}$ with relevant sets $\rel{}'$, where
\begin{equation*}
    \mat{M}=
    \begin{bmatrix}
    0 & 0 & 1 & 2\\
    0 & 0 & 1 & 1\\
    0 & 0 & 0 & 1\\
    0 & 0 & 0 & 0
    \end{bmatrix},
    \qquad
    \mat{T}=
    \begin{bmatrix}
    1 & 0 & 0 & 0\\
    0 & 1 & 0 & 0\\
    0 & 2 & 0 & 0\\
    0 & 0 & 1 & 0\\
    0 & 0 & 0 & 1
    \end{bmatrix}.
\end{equation*}

To prove the consistency condition of a strong $\mat{T}$-abstraction (\Cref{def:tabstraction}), we choose the exogenous abstraction $\mat{S}=\mat{T}$ and show that it holds for any abstract intervention $j$, which sets each $Y_k$ in a subset of $\setdef{Y_1,Y_2,Y_T,Y_O}$ to a value $v_k$. We choose the low-level intervention $i$ that sets $X_1\gets v_1$ if $j$ intervenes on $Y_1$, $X_2\gets v_2$ and $X_3\gets 0$ if on $Y_2$, $X_4\gets v_T$ if on $Y_T$, and $X_5\gets v_O$ if on $Y_O$. Then, denoting as $\bot$ the empty intervention and defining $c=e_2+2e_3$, for every realization $\vec{e}$ of the low-level exogenous terms, the last column reports the vector $\scm{L}^i(\vec{e})\mat{T}=\scm{H}^j(\vec{e}\mat{S})$ of the abstract variables $(Y_1,Y_2,Y_T,Y_O)$. For brevity, we only report the intervened variables in each intervention, whose values are the ones defined above:
    {\small
    \begin{alignat*}{3}
        j&=\bot, &\quad i&=\bot, &\quad &(e_1,\, c,\, e_1+c+e_4,\, 3e_1+2c+e_4+e_5),\\
        j&=\operatorname{do}(Y_1), &\quad i&=\operatorname{do}(X_1), &\quad &(v_1,\, c,\, v_1+c+e_4,\, 3v_1+2c+e_4+e_5),\\
        j&=\operatorname{do}(Y_2), &\quad i&=\operatorname{do}(X_2, X_3), &\quad &(e_1,\, v_2,\, e_1+v_2+e_4,\, 3e_1+2v_2+e_4+e_5),\\
        j&=\operatorname{do}(Y_T), &\quad i&=\operatorname{do}(X_4), &\quad &(e_1,\, c,\, v_T,\, 2e_1+c+v_T+e_5),\\
        j&=\operatorname{do}(Y_O), &\quad i&=\operatorname{do}(X_5), &\quad &(e_1,\, c,\, e_1+c+e_4,\, v_O),\\
        j&=\operatorname{do}(Y_1, Y_2), &\quad i&=\operatorname{do}(X_1, X_2, X_3), &\quad &(v_1,\, v_2,\, v_1+v_2+e_4,\, 3v_1+2v_2+e_4+e_5),\\
        j&=\operatorname{do}(Y_1, Y_T), &\quad i&=\operatorname{do}(X_1, X_4), &\quad &(v_1,\, c,\, v_T,\, 2v_1+c+v_T+e_5),\\
        j&=\operatorname{do}(Y_1, Y_O), &\quad i&=\operatorname{do}(X_1, X_5), &\quad &(v_1,\, c,\, v_1+c+e_4,\, v_O),\\
        j&=\operatorname{do}(Y_2, Y_T), &\quad i&=\operatorname{do}(X_2, X_3, X_4), &\quad &(e_1,\, v_2,\, v_T,\, 2e_1+v_2+v_T+e_5),\\
        j&=\operatorname{do}(Y_2, Y_O), &\quad i&=\operatorname{do}(X_2, X_3, X_5), &\quad &(e_1,\, v_2,\, e_1+v_2+e_4,\, v_O),\\
        j&=\operatorname{do}(Y_T, Y_O), &\quad i&=\operatorname{do}(X_4, X_5), &\quad &(e_1,\, c,\, v_T,\, v_O),\\
        j&=\operatorname{do}(Y_1, Y_2, Y_T), &\quad i&=\operatorname{do}(X_1, X_2, X_3, X_4), &\quad &(v_1,\, v_2,\, v_T,\, 2v_1+v_2+v_T+e_5),\\
        j&=\operatorname{do}(Y_1, Y_2, Y_O), &\quad i&=\operatorname{do}(X_1, X_2, X_3, X_5), &\quad &(v_1,\, v_2,\, v_1+v_2+e_4,\, v_O),\\
        j&=\operatorname{do}(Y_1, Y_T, Y_O), &\quad i&=\operatorname{do}(X_1, X_4, X_5), &\quad &(v_1,\, c,\, v_T,\, v_O),\\
        j&=\operatorname{do}(Y_2, Y_T, Y_O), &\quad i&=\operatorname{do}(X_2, X_3, X_4, X_5), &\quad &(e_1,\, v_2,\, v_T,\, v_O),\\
        j&=\operatorname{do}(Y_1, Y_2, Y_T, Y_O), &\quad i&=\operatorname{do}(X_1, X_2, X_3, X_4, X_5), &\quad &(v_1,\, v_2,\, v_T,\, v_O).
    \end{alignat*}}
Hence, $\scm{H}$ is a strong $\mat{T}$-abstraction of $\scm{L}$, and $\rel{}'$ is parametrically consistent.

\paragraph{Parametric inconsistency of $\rel{}''$.}
Let now $\scm{H}$ be any linear ANM with coefficient matrix $\mat{M}$ that is a strong $\mat{T}$-abstraction of $\scm{L}$ with relevant sets $\rel{}''$, and denote as $\alpha_i$ the coefficient of $X_i$ in $\mat{T}$. Writing samples as row vectors, it holds $\vec{x}=\vec{e}\mat{F}$ and $\vec{y}=\vec{u}\mat{G}$, where $\mat{F}=(\mat{I}-\mat{W})^\inv$ and $\mat{G}=(\mat{I}-\mat{M})^\inv$. Under the empty intervention, the consistency condition of \Cref{def:tabstraction} requires $\vec{y}=\vec{x}\mat{T}$ and $\vec{u}=\vec{e}\mat{S}$ for every $\vec{e}$, \ie,
\begin{equation}
\label{eq:linearconsistency}
    \mat{F}\mat{T}=\mat{S}\mat{G}.
\end{equation}
Since every low-level variable is relevant in $\rel{}''$, every block coincides with the corresponding relevant set. Hence, $\mat{S}$ has the same support as $\mat{T}$. Furthermore, the abstract graph induced by $\rel{}''$ has the edges $Y_3\to Y_T$, $Y_3\to Y_O$, and $Y_T\to Y_O$. Therefore, ordering the abstract variables as $Y_3, Y_T, Y_O$, we get
\begin{align}
    \underbrace{
    \begin{bmatrix}
    1 & 0 & 0 & 1 & 3\\
    0 & 1 & 0 & 1 & 2\\
    0 & 0 & 1 & 2 & 4\\
    0 & 0 & 0 & 1 & 1\\
    0 & 0 & 0 & 0 & 1
    \end{bmatrix}}_{\mat{F}}
    \underbrace{
    \begin{bmatrix}
    \alpha_1 & 0 & 0\\
    \alpha_2 & 0 & 0\\
    \alpha_3 & 0 & 0\\
    0 & \alpha_4 & 0\\
    0 & 0 & \alpha_5
    \end{bmatrix}}_{\mat{T}}
  &=
    \underbrace{
    \begin{bmatrix}
    s_{13} & 0 & 0\\
    s_{23} & 0 & 0\\
    s_{33} & 0 & 0\\
    0 & s_{4T} & 0\\
    0 & 0 & s_{5O}
    \end{bmatrix}}_{\mat{S}}
    \underbrace{
    \begin{bmatrix}
    1 & g_{3T} & g_{3O}\\
    0 & 1 & g_{TO}\\
    0 & 0 & 1
    \end{bmatrix}}_{\mat{G}}\\
    \begin{bmatrix}
    \alpha_1 & \alpha_4 & 3\alpha_5\\
    \alpha_2 & \alpha_4 & 2\alpha_5\\
    \alpha_3 & 2\alpha_4 & 4\alpha_5\\
    0 & \alpha_4 & \alpha_5\\
    0 & 0 & \alpha_5
    \end{bmatrix}
    &=
    \begin{bmatrix}
    s_{13} & s_{13}g_{3T} & s_{13}g_{3O}\\
    s_{23} & s_{23}g_{3T} & s_{23}g_{3O}\\
    s_{33} & s_{33}g_{3T} & s_{33}g_{3O}\\
    0 & s_{4T} & s_{4T}g_{TO}\\
    0 & 0 & s_{5O}
    \end{bmatrix}.
\end{align}

In particular, we consider the sub-block of the rows of $X_1, X_2, X_3$ and of the columns of $Y_T, Y_O$. Therefore, to satisfy $\mat{F}\mat{T}=\mat{S}\mat{G}$, the following equation needs to hold:
\begin{equation}
\label{eq:paramexample}
    \begin{bmatrix}
    \alpha_4 & 3\alpha_5\\
    \alpha_4 & 2\alpha_5\\
    2\alpha_4 & 4\alpha_5
    \end{bmatrix}
    =
    \begin{bmatrix}
    s_{13}\\
    s_{23}\\
    s_{33}
    \end{bmatrix}
    \begin{bmatrix}
    g_{3T} & g_{3O}
    \end{bmatrix}.
\end{equation}
However, since $\alpha_4,\alpha_5\neq 0$, the left-hand side has rank two, while the right-hand side has rank at most one. Hence, no choice of the abstraction coefficients and of the abstract parameters satisfies the equation, and consequently $\rel{}''$ is not parametrically consistent.

\subsection{Parametric Inconsistency of the Maximal Partitioning}%
\label{app:mpinconsistency}

In \Cref{subsec:maximalpartitioning}, we restrict the guarantees of \Cref{thm:maximalpartitioning} to the unmediated relevant sets $\set{P}^\ast_{\set{V}}$. The following example shows that this restriction is necessary, since the whole maximal partitioning $\set{P}^\ast$ might not be contained in any parametrically consistent minimal sufficient extension. Let $\rel{Y_T}=\setdef{X_T}$ and $\rel{Y_O}=\setdef{X_O}$ be the relevant sets of $\rel{}$ over the following linear ANM $\scm{L}$, where $w_{ij}\neq 0$ is the coefficient of the edge $X_i\to X_j$.
\begin{center}
  \small
  \begin{tikzpicture}[node distance=1.5cm, minimum size=0.3cm]
    \node[draw, circle] (XC) at (0,0) {$X_1$};
    \node[draw, circle] (XE) at (2.2,0) {$X_3$};
    \node[draw, circle, fill=green!40!white] (XO) at (4.4,0) {$X_O$};
    \node[draw, circle] (XCp) at (2.2,-1.6) {$X_2$};
    \node[draw, circle, fill=red!40!white] (XT) at (4.4,-1.6) {$X_T$};
    \draw[-{Latex}] (XC) -- (XE);
    \draw[-{Latex}] (XCp) -- (XE);
    \draw[-{Latex}] (XC) -- (XT);
    \draw[-{Latex}] (XCp) -- (XT);
    \draw[-{Latex}] (XE) -- (XT);
    \draw[-{Latex}] (XE) -- (XO);
    \draw[-{Latex}] (XT) -- (XO);
    \begin{pgfonlayer}{background}
      \node[draw=red!60!black, very thick, rounded corners=3pt, fill=red!10!white, fit=(XT), inner sep=4pt, label={[text=red!60!black]right:{\scriptsize$\rel{Y_T}$}}] {};
      \node[draw=green!60!black, very thick, rounded corners=3pt, fill=green!10!white, fit=(XO), inner sep=4pt, label={[text=green!60!black]right:{\scriptsize$\rel{Y_O}$}}] {};
    \end{pgfonlayer}
  \end{tikzpicture}
\end{center}

The only jointly targeted set is $\set{V}=\setdef{Y_T,Y_O}$, and the new relevant variables are $\set{Z}=\setdef{X_1,X_2,X_3}$, since each of them has the child $X_T$. Only $X_3$ has a $\rel{}$-direct path to $X_O$, so it cannot share a relevant set with $X_1$ or $X_2$. Hence, the \model{MiSE} contains two extensions:
\begin{enumerate}
  \item%
    $\rel{}'$, which adds $\rel{Y_1}'=\setdef{X_1}$, $\rel{Y_2}'=\setdef{X_3}$, and $\rel{Y_3}'=\setdef{X_2}$;
  \item%
    $\rel{}''$, which adds $\rel{Y_1}''=\setdef{X_1,X_2}$ and $\rel{Y_2}''=\setdef{X_3}$.
\end{enumerate}
Rescaling each variable of $\scm{L}$ by its abstraction coefficient yields a strong $\mat{T}$-abstraction with relevant sets $\rel{}'$, which is therefore always parametrically consistent. Instead, $\rel{}''$ is produced by the maximal partitioning, but it is not parametrically consistent in general.

\subsubsection*{Maximal Partitioning}

Both confounded variables are targeted by all the new relevant variables, \ie, $\set{Z}^{(T)}=\set{Z}^{(O)}=\set{Z}$. Since $Y_T=t_{TT}X_T$ and $Y_O=t_{OO}X_O$, by \Cref{thm:paramconsistency} the columns of $\mat{A}$ are given by the regressions
\begin{align*}
    Y_T &= t_{TT}w_{1T}X_1 + t_{TT}w_{2T}X_2 + t_{TT}w_{3T}X_3 + t_{TT}E_T,\\
    Y_O &= m_{TO}Y_T + t_{OO}w_{3O}X_3 + t_{OO}E_O,
\end{align*}
where $m_{TO}=t_{OO}w_{TO}/t_{TT}$. Hence, the composed abstraction matrix is
\begin{equation*}
    \mat{A}=
    \begin{bmatrix}
    t_{TT}w_{1T} & 0\\
    t_{TT}w_{2T} & 0\\
    t_{TT}w_{3T} & t_{OO}w_{3O}
    \end{bmatrix}.
\end{equation*}
Only the rows of $X_1$ and $X_2$ are collinear, so the maximal partitioning is $\set{P}^\ast=\setdef{\setdef{X_1,X_2},\setdef{X_3}}$, which extends $\rel{}$ to $\rel{}''$.

\subsubsection*{Parametric Inconsistency of $\rel{}''$}

Let $\scm{H}$ be a strong $\mat{T}$-abstraction of $\scm{L}$ with relevant sets $\rel{}''$ and abstract coefficients $\mat{M}$, where $t_{ij}$ is the abstraction coefficient of $X_i$ in $Y_j$. Since every block coincides with its relevant set, it holds $\mat{S}=\mat{T}$, as in \Cref{app:miseparams}. Then, comparing the coefficients of the exogenous terms in $Y_2$ and $Y_T$, computed through $\scm{H}$ and through $\scm{L}$, gives
\begin{equation*}
    m_{12}t_{11} = t_{32}w_{13},\quad
    m_{12}t_{21} = t_{32}w_{23},\quad
    m_{1T}t_{11} = t_{TT}w_{1T},\quad
    m_{1T}t_{21} = t_{TT}w_{2T},
\end{equation*}
where all the coefficients are non-zero. Therefore, $\rel{}''$ is parametrically consistent only if
\begin{equation}
\label{eq:mpexample}
    \frac{w_{1T}}{w_{2T}}
    =
    \frac{t_{11}}{t_{21}}
    =
    \frac{w_{13}}{w_{23}}.
\end{equation}
Conversely, if \Cref{eq:mpexample} holds, the coefficients $t_{11}=w_{13}$, $t_{21}=w_{23}$, $t_{32}=t_{TT}=t_{OO}=1$, $m_{12}=1$, $m_{1T}=w_{1T}/w_{13}$, $m_{2T}=w_{3T}$, $m_{2O}=w_{3O}$, and $m_{TO}=w_{TO}$ define a strong $\mat{T}$-abstraction, where interventions on $Y_1$ set $X_1\gets v_1/w_{13}$ and $X_2\gets 0$. However, \Cref{eq:mpexample} does not hold in general, \eg, for $w_{1T}=w_{13}=w_{23}=1$ and $w_{2T}=2$. Then, $\rel{}'$ is the only parametrically consistent minimal sufficient extension, and it does not contain $\setdef{X_1,X_2}\in\set{P}^\ast$.

\subsubsection*{\model{dataRSE} Recursive Call}

In the abstract graph induced by $\rel{}''$, $Y_1$ has the child $Y_2\notin\set{V}$, while both children $Y_T$ and $Y_O$ of $Y_2$ are in $\set{V}$. Hence, the unmediated relevant sets are $\set{P}^\ast_{\set{V}}=\setdef{\setdef{X_3}}\subseteq\rel{}'$, as guaranteed by \Cref{thm:maximalpartitioning}. Indeed, for $\rel{}'$ it holds $\set{N}_{\set{V}}=\setdef{Y_2}$, and \Cref{eq:mpcondition} holds since $m'_{2O}\neq 0$, whereas $\vec{m}'_{1,\set{V}}=(m'_{1T},0)$ and $\vec{m}'_{3,\set{V}}=(m'_{3T},0)$. Therefore, \model{dataRSE} only adds $\setdef{X_3}$ and recursively calls itself. At the second call, the new relevant variables $X_1$ and $X_2$ confound $Y_T$ and $Y_2$, and the composed abstraction matrix is
\begin{equation*}
    \mat{A}=
    \begin{bmatrix}
    t_{TT}w_{1T} & t_{32}w_{13}\\
    t_{TT}w_{2T} & t_{32}w_{23}
    \end{bmatrix},
\end{equation*}
whose rows are collinear if and only if \Cref{eq:mpexample} holds. Both induced variables only have children in $\setdef{Y_T,Y_2}$, hence they are unmediated.
\begin{itemize}
  \item%
    If \Cref{eq:mpexample} does not hold, the maximal partitioning separates $X_1$ and $X_2$, and \model{dataRSE} returns $\rel{}'$.
  \item%
    If \Cref{eq:mpexample} holds, it merges them, and \model{dataRSE} returns $\rel{}''$. In this case, $\rel{}'$ is also parametrically consistent, but it violates \Cref{eq:mpcondition} at the second call, since $\vec{m}'_{1,\set{V}}=(m'_{1T},m'_{12})$ and $\vec{m}'_{3,\set{V}}=(m'_{3T},m'_{32})$ are collinear. Hence, as stated by \Cref{cor:datarsecorrectness}, \model{dataRSE} returns $\rel{}''$.
\end{itemize}
In both cases, \model{dataRSE} returns a parametrically consistent minimal sufficient extension, although, when \Cref{eq:mpexample} does not hold, none of them contains the maximal partitioning of the first call.

\newpage
\section{Proofs}%
\label{sec:proofs}

All the results in this section hold under the assumptions of \Cref{app:t-abstraction} and under the clean low-level graph assumption of \Cref{sec:mise}, which we restate here for convenience.

\nocancellingpathsassumption*

\exogenousfaithfulnessassumption*

\cleanassumption*

Furthermore, every abstract graph induced by a graphically consistent set of relevant variables on an acyclic low-level graph is acyclic (\Cref{prop:acyclicity}).

\subsection{Proposition~\ref{prop:causalsufficiency}}%
\label{proof:causalsufficiency}

\causalsufficiencyprop*
\begin{proof}
We recall that $\scm{H}$ is causally sufficient if its ADMG has no bidirected edges, \ie, if its exogenous terms are pairwise marginally independent (\Cref{sec:background}). In a $\mat{T}$-abstraction, the exogenous term $U_j$ of an abstract variable $Y_j$ is a linear combination of the low-level exogenous terms of its block, \ie, $u_j=\vec{e}_{\block{Y_j}}\vec{s}_j$, where $\vec{s}_j$ has non-zero entries. Furthermore, since $\scm{L}$ is causally sufficient, the low-level exogenous terms $\set{E}$ are mutually independent.

\paragraph{$\rel{}$-Sufficiency $\implies$ $\scm{H}$-Sufficiency.}
If the blocks are pairwise disjoint, the abstract exogenous terms are functions of disjoint subsets of the mutually independent low-level exogenous terms. Hence, they are mutually independent and, in particular, pairwise independent. Therefore, $\scm{H}$ is causally sufficient.

\paragraph{$\scm{H}$-Sufficiency $\implies$ $\rel{}$-Sufficiency.}
Let $Y_1\neq Y_2$ be two abstract variables with intersecting blocks, and let $X_k\in\block{Y_1}\cap\block{Y_2}$. Since $\vec{s}_1$ and $\vec{s}_2$ have non-zero entries on the whole blocks, the low-level exogenous term $E_k$ is a common parent of $U_1$ and $U_2$ in the graph of \Cref{assumption:exogenousfaithfulness}. Hence, $U_1$ and $U_2$ are d-connected through $E_k$ and, by \Cref{assumption:exogenousfaithfulness}, they are not marginally independent. Hence, $\scm{H}$ has the bidirected edge $Y_1\leftrightarrow Y_2$ and is not causally sufficient. This direction extends the corresponding result of \citet{massidda2024learning}.

Finally, the two directions show that, under \Cref{assumption:exogenousfaithfulness}, the pairwise and the mutual independence of the abstract exogenous terms are equivalent, since both are equivalent to the disjointness of the blocks.
\end{proof}

\subsection{Lemma~\ref{lemma:singletarget}}%
\label{proof:singletarget}

\singletargetlemma*
\begin{proof}
Let $\rel{}'\in\mise{\rel{}}{\graph{G}_{\set{X}}}$ induce the abstract variables $\set{Y}'\supseteq\set{Y}$, the abstract graph $\graph{G}_{\set{Y}'}$, and the blocks $\block{}'$, and let $X\in\set{X}$ with $\targets{X}{\block{}}=\setdef{Y}$. If $X\in\rel{Y}$, then $X\in\rel{Y}'$ and $\targets{X}{\block{}'}=\setdef{Y}$. Hence, we assume that $X$ is not relevant in $\rel{}$ and, by contradiction, that $\targets{X}{\block{}'}\neq\setdef{Y}$.

\paragraph{Proof strategy.}
Under $\rel{}$, every variable that $X$ reaches before hitting a relevant variable only leads to $\rel{Y}$. If the targets of $X$ change in $\rel{}'$, some new abstract variables must have been built over these variables. We show that these new variables can only point to $Y$ or to each other. Hence, removing them still produces a smaller sufficient extension, contradicting the minimality of $\rel{}'$.

\paragraph{Notation.}
We write $X_a\rdirect X_b$ if there is a $\rel{}$-direct path from $X_a$ to $X_b$, and $\bigcup\rel{}$ for the set of variables relevant in $\rel{}$, and similarly for $\rel{}'$. Throughout the proof, $Q$ denotes an arbitrary abstract variable and $X_r$ a variable relevant to it. A variable in $\rel{Q}$ lies only in the block of $Q$, while for a non-relevant variable $X_a\notin\bigcup\rel{}$ and any $Q\in\set{Y}$ it holds (\Cref{sec:background})
\begin{equation}
\label{eq:blockpath}
    Q\in\targets{X_a}{\block{}}
    \iff
    \exists X_r\in\rel{Q}.\ X_a\rdirect X_r,
\end{equation}
and similarly for $\rel{}'$. Since $\graph{G}_{\set{X}}$ is acyclic, concatenation of directed paths is a directed path.

\paragraph{Variables reached by $X$ only lead to $Y$.}
Let $\set{S}_X$ be the set of variables $X_a$ such that some directed path from $X$ to $X_a$ contains no variable of $\bigcup\rel{}$, including its endpoints. Intuitively, $\set{S}_X$ is what $X$ can reach before hitting a relevant variable. Any $\rel{}$-direct path leaving $\set{S}_X$ can be prefixed with the path from $X$, so it can only end where $X$ ends, \ie, in $\rel{Y}$. Formally, for every $X_a\in\set{S}_X$ and every $Q\in\set{Y}$,
\begin{align*}
    X_a\rdirect X_r\in\rel{Q}
    &\implies X\rdirect X_r
    && \text{(prepend the path from $X$ to $X_a$)}\\
    &\implies Q\in\targets{X}{\block{}}=\setdef{Y}
    && \text{(\Cref{eq:blockpath})}\\
    &\implies Q = Y.
\end{align*}
Hence, $\targets{X_a}{\block{}}=\setdef{Y}$, where $\supseteq$ holds since the graph is clean (\Cref{assumption:clean}). Applying the same reasoning to $X_a$, it holds that $\set{S}_X$ is closed over its children until reaching $\rel{Y}$:
\begin{equation}
\label{eq:reachchildren}
    X_a\in\set{S}_X\ \land\ X_c\in\ch{X_a}
    \implies
    X_c\in\rel{Y}\ \lor\ X_c\in\set{S}_X.
\end{equation}

\paragraph{In $\rel{}'$, abstract targets of $X$ only lead to $Y$ or to new variables built over them.}
Let $\set{N}=\setdef{W\in\set{Y}'\setminus\set{Y}\mid\rel{W}'\cap\set{S}_X\neq\emptyset}$ be the new abstract variables induced by $\rel{}'$ and built over at least one variable of $\set{S}_X$. A $\rel{}'$-direct path starting from a variable $X_a\in\set{S}_X$ stays in $\set{S}_X$ until it hits a variable relevant in $\rel{}'$. This variable is either in an existing relevant set $\rel{Y}$, or it is a variable of $\set{S}_X$ that has become relevant in the extension. Then, its corresponding abstract variable is in $\set{N}$. Formally, for every $X_a\in\set{S}_X$ and every $Q\in\set{Y}'$,
\begin{align}
\label{eq:reachnew}
    X_a\rpdirect X_r\in\rel{Q}'
    &\implies \text{the path stays in $\set{S}_X$ before $X_r$}
    && \text{(\Cref{eq:reachchildren})}\nonumber\\
    &\implies X_r\in\rel{Y}\ \lor\ X_r\in\set{S}_X\cap\rel{Q}'
    && \text{(\Cref{eq:reachchildren})}\nonumber\\
    &\implies Q\in\setdef{Y}\cup\set{N}
    && \text{($\set{S}_X\cap\textstyle\bigcup\rel{}=\emptyset$)}.
\end{align}
In the first step, the intermediate variables of the path are not relevant in $\rel{}'$, hence not in $\rel{Y}\subseteq\bigcup\rel{}'$, so \Cref{eq:reachchildren} keeps them in $\set{S}_X$.

\paragraph{If the targets of $X$ change, some new variable is built over $\set{S}_X$.}
Since we prove the lemma by contradiction assuming that the targets of $X$ change, some variable that $X$ reaches must have become relevant in $\rel{}'$. Otherwise, the $\rel{}'$-direct paths from $X$ would be exactly its $\rel{}$-direct paths, which only lead to $\rel{Y}$. Formally, we prove this statement by showing that $\set{N}=\emptyset$ implies $\targets{X}{\block{}'}=\setdef{Y}$. Assume $\set{N}=\emptyset$. Since $\set{S}_X\cap\bigcup\rel{}=\emptyset$, no variable of $\set{S}_X$, including $X$, is relevant in $\rel{}'$. Then, for every $Q\in\set{Y}'$,
\begin{align*}
    Q\in\targets{X}{\block{}'}
    &\implies X\rpdirect X_r\in\rel{Q}'
    && \text{(\Cref{eq:blockpath})}\\
    &\implies Q=Y
    && \text{(\Cref{eq:reachnew}, $\set{N}=\emptyset$)},\\[4pt]
    Y\in\targets{X}{\block{}}
    &\implies X\rdirect X_r\in\rel{Y}
    && \text{(\Cref{eq:blockpath})}\\
    &\implies X\rpdirect X_r\in\rel{Y}'
    && \text{($\set{N}=\emptyset$)}\\
    &\implies Y\in\targets{X}{\block{}'}
    && \text{(\Cref{eq:blockpath})}.
\end{align*}
In the fourth step, the intermediate variables of the path are in $\set{S}_X$, and none of them is relevant in $\rel{}'$ since $\set{N}=\emptyset$. Hence, $\targets{X}{\block{}'}=\setdef{Y}$, contradicting our assumption. Therefore, $\set{N}\neq\emptyset$.

\paragraph{New variables in $\set{N}$ only point to $Y$ or to each other.}
Let $W\in\set{N}$ be a generic new abstract variable that, by construction, has at least one relevant variable in $\set{S}_X$. Therefore, this variable $X_a\in\rel{W}'$ by \Cref{eq:reachnew} only reaches $Y$ and $\set{N}$. Graphical consistency requires \emph{every} relevant variable of $W$, including this one, to reach each abstract child of $W$. Hence, $W$ cannot have other children without breaking graphical consistency. Formally, for every $W\in\set{N}$ and every $Q\in\set{Y}'$,
\begin{align}
\label{eq:reachabschildren}
    W\to Q\in\graph{G}_{\set{Y}'}
    &\implies X_a\rpdirect X_r\in\rel{Q}' \text{ for some } X_a\in\rel{W}'\cap\set{S}_X
    && \text{(\Cref{def:wellformed})}\nonumber\\
    &\implies Q\in\setdef{Y}\cup\set{N}
    && \text{(\Cref{eq:reachnew})}.
\end{align}

\paragraph{Removing the new variables in $\set{N}$ preserves sufficiency.}
Up to now, we proved that if the abstract targets of $X$ change, then there must be some new abstract variables $\set{N}$ built over the low-level variables $\set{S}_X$ that $X$ reaches before hitting the relevant set $\rel{Y}$. We now conclude the proof by showing that removing these new variables preserves sufficiency, contradicting the minimality of $\rel{}'$.
Let $\rel{}''=\rel{}'\setminus\setdef{\rel{W}'}_{W\in\set{N}}$ with blocks $\block{}''$. Since we proved that $\set{N}$ is not empty, it holds $\rel{}\subseteq\rel{}''\subset\rel{}'$. Removing $\set{N}$ opens new paths through its variables. However, by \Cref{eq:reachabschildren} any such path leaves $\set{N}$ towards $Y$ or towards another variable of $\set{N}$, so the only new target it can create is $Y$, and only for variables that were previously targeting $\set{N}$.
To prove the sufficiency of $\rel{}''$, we show that any non-relevant variable has exactly one abstract target. Formally, let $X_v$ be a variable that is not relevant in $\rel{}''$. Since $\rel{}'$ is sufficient, $X_v$ has a unique abstract target in $\rel{}'$. Furthermore, $X_v$ has at least one abstract target in $\rel{}''$, since the graph is clean (\Cref{assumption:clean}) and $\rel{}\subseteq\rel{}''$. Hence, it suffices to show that $X_v$ has at most one abstract target in $\rel{}''$. We distinguish two mutually exclusive cases.
\begin{enumerate}
\item%
$\targets{X_v}{\block{}'}=\setdef{W}$ with $W\in\set{N}$.
In this case, the variable $X_v$ had as abstract target one of the new abstract variables $W\in\set{N}$, which is now removed in $\rel{}''$. Therefore, in $\rel{}'$, every path from $X_v$ is captured by a variable of $\set{N}$. Once $\set{N}$ is removed, the path continues, but by graphical consistency the variables of $\set{N}$ only point to $Y$ or to each other (\Cref{eq:reachabschildren}), so the path can only end in $\rel{Y}$.
\item%
$\targets{X_v}{\block{}'}=\setdef{Q}$ with $Q\notin\set{N}$.
In $\rel{}'$, no path from $X_v$ is captured by a variable of $\set{N}$, so the removal of $\set{N}$ does not affect the paths from $X_v$. Therefore, in $\rel{}''$ the variable $X_v$ still has $Q$ as its unique abstract target.
\end{enumerate}

In both cases, $X_v$ has a unique abstract target in $\rel{}''$. Hence, $\rel{}''$ is a sufficient set of relevant variables with $\rel{}\subseteq\rel{}''\subset\rel{}'$, contradicting the minimality of $\rel{}'$ (\Cref{def:mise}). Therefore, $\targets{X}{\block{}'}=\setdef{Y}$.
\end{proof}

\subsection{Lemma~\ref{lemma:gc-extension}}%
\label{proof:gc-extension}

\gcextensionlemma*
\begin{proof}
We prove \Cref{lemma:gc-extension} by contradiction, building on \Cref{prop:prata-lemma}. Let $\rel{}$ be a graphically inconsistent set of relevant variables over the graph $\graph{G}_{\set{X}}$. We assume the existence of an extension $\rel{}'=\rel{}\cup\set{P}$ such that $\rel{}'$ is graphically consistent on $\graph{G}_{\set{X}}$. Due to \Cref{prop:prata-lemma}, if we remove the subset $\set{P}$ from $\rel{}'$ then $\rel{}$ must be graphically consistent. This contradicts the hypothesis of \Cref{lemma:gc-extension} that $\rel{}$ is not graphically consistent. Therefore, no extension $\rel{}'$ of a graphically inconsistent set of relevant variables $\rel{}$ can be graphically consistent on any graph $\graph{G}_{\set{X}}$.
\end{proof}

\subsection{Lemma~\ref{lemma:newrelevants}}%
\label{proof:newrelevants}

\newrelevantslemma*
\begin{proof}
Throughout the proof, we write $\set{V}$ and $\set{Z}$ for $\set{V}_i$ and $\set{Z}_i$. Let $\rel{}'\in\mise{\rel{}}{\graph{G}_{\set{X}}}$, and let $\block{}$ and $\block{}'$ be the blocks induced by $\rel{}$ and $\rel{}'$, respectively. Since $\rel{}'$ is sufficient, the blocks $\block{}'$ are pairwise disjoint (\Cref{prop:causalsufficiency}). We recall that a relevant variable lies only in the block of its own abstract variable, and that a non-relevant variable lies in the block of $Y$ if and only if it has a $\rel{}$-direct path to some variable in $\rel{Y}$, and similarly for $\rel{}'$.

Let $X\in\set{Z}$. Since $X\in\bigcap_{Y\in\set{V}}\block{Y}$ and $\size{\set{V}}\geq 2$, $X$ is in multiple blocks, so it is not relevant in $\rel{}$. Assume by contradiction that $X$ is not relevant in $\rel{}'$ either. By \Cref{def:newrelevants}, $X$ has a child $X'$ with a unique abstract target $\targets{X'}{\block{}}=\setdef{Y}$.

\paragraph{$X$ is in the block of $Y$ in $\rel{}'$.}
By target stability (\Cref{lemma:singletarget}), the child $X'$ of $X$ in a unique block will still be in a unique block, namely $\block{Y}'$. Therefore, $X$ has a $\rel{}'$-direct path to a relevant variable in $\rel{Y}$. Since $X$ is not relevant in $\rel{}'$, it holds $X\in\block{Y}'$.

\paragraph{$X$ is in the block of $W\neq Y$ in $\rel{}'$.}
Since $X$ lies in the blocks of all variables in $\set{V}$ and $\size{\set{V}}\geq 2$, there exists an abstract variable $Y_2\in\set{V}$ with $Y_2\neq Y$ such that $X\in\block{Y_2}$. Therefore, there exists a $\rel{}$-direct path $\vec{p}=(X,\ldots,X_n)$ with $X_n\in\rel{Y_2}$. Let $X''$ be the first variable on $\vec{p}$ after $X$ that is relevant in $\rel{}'$, which always exists since existing relevant sets are not altered, thus $X_n\in\rel{Y_2}=\rel{Y_2}'$. Let $W\in\set{Y}'$ be the abstract variable such that $X''\in\rel{W}'$. The subpath of $\vec{p}$ from $X$ to $X''$ is $\rel{}'$-direct and, since $X$ is not relevant in $\rel{}'$, it holds $X\in\block{W}'$. Notably, $W\neq Y$, since either $X''$ is not relevant in $\rel{}$ and therefore cannot be in $\rel{Y}=\rel{Y}'$, or if $X''$ is relevant in $\rel{}$ then it is in $\rel{Y_2}=\rel{Y_2}'$ with $Y_2\neq Y$.

\paragraph{Contradiction.}
We proved that in $\rel{}'$ the variable $X$ is in the block of $Y$ and in the block of $W\neq Y$. However, this contradicts the sufficiency of $\rel{}'$, which requires the blocks $\block{}'$ to be disjoint. Therefore, $X$ must be relevant in $\rel{}'$. Since $X$ is not relevant in $\rel{}$ and extensions do not alter existing relevant sets, $X\in\rel{Y}'$ for some new abstract variable $Y\in\set{Y}'\setminus\set{Y}$.
\end{proof}

\subsection{Theorem~\ref{thm:rse}}%
\label{app:rse}

\rsethm*
\begin{proof}
Throughout the proof, we write $\set{V}$ and $\set{Z}$ for the jointly targeted set $\set{V}_i$ selected by \rse{} and its new relevant variables $\set{Z}_i$, and $\set{Z}^{(k)}$ for the new relevant variables selected at the $k$-th step of a path.

\paragraph{Termination.} Each call of \rse{} on a graphically consistent but not sufficient $\rel{}$ selects a minimum-size jointly targeted set $\set{V}$, for which the new relevant variables $\set{Z}$ are non-empty (\Cref{lemma:nonemptyz}) and not relevant in $\rel{}$. Hence, every recursive call $\model{RSE}(\rel{}\cup\set{P},\graph{G}_{\set{X}})$ with $\set{P}\in\partitions{\set{Z}}$ has strictly more relevant low-level variables than its caller. Since there are at most $\size{\set{X}}$ low-level variables and each call iterates over finitely many partitions, the recursion tree is finite. We remark that the minimum-size selection is necessary, since for a jointly targeted set that is not of minimum size, $\set{Z}$ can be empty, and \rse{} would then call itself on the same input.

\paragraph{Correctness.} We prove the correctness of the algorithm by showing that
\begin{equation}
    \rel{}'\in\model{RSE}(\rel{}, \graph{G}_{\set{X}})
    \implies
    \rel{}'\in\mise{\rel{}}{\graph{G}_{\set{X}}}.
\end{equation}

First, we notice that if $\rel{}$ is not graphically consistent on $\graph{G}_{\set{X}}$, \model{RSE} returns an empty set, trivially satisfying the property. Furthermore, if the set of relevant variables $\rel{}$ is both graphically consistent and sufficient on $\graph{G}_{\set{X}}$, \model{RSE} returns a set containing only $\rel{}$. From \Cref{def:mise}, it follows directly that a graphically consistent and sufficient set of relevant variables is in its own \model{MiSE}. Consequently, \Cref{thm:rse} holds.

Therefore, we focus on the general scenario where $\rel{}$ is graphically consistent but not sufficient on $\graph{G}_{\set{X}}$. At each call, \rse{} identifies a jointly targeted set and creates one or more new abstract variables by selecting new low-level relevant variables. \rse{} defines a computational graph that branches for every decision to add one or more abstract variables, as shown in \Cref{ex:rseexecution}. Formally, for each $\rel{}'\in\model{RSE}(\rel{},\graph{G}_{\set{X}})$, we can identify one or more paths in the computational graph, such that
\begin{equation}
    \rel{}^{(1)} \subset
    \rel{}^{(2)} \subset
    \ldots\subset
    \rel{}^{(n)},
\end{equation}
where $\rel{}^{(1)}=\rel{}$ and $\rel{}^{(n)}=\rel{}'$. Therefore, $\rel{}'\supseteq\rel{}$ holds by construction.

A set of relevant variables is returned by \rse{} only if it is sufficient and graphically consistent. Consequently, to prove that $\rel{}'$ is in the \model{MiSE} of $\rel{}$ on $\graph{G}_{\set{X}}$, we only have to prove \emph{minimality}. In other words, we must show that any $\rel{}''$ such that $\rel{}\subseteq\rel{}''\subset\rel{}'$ is not sufficient on $\graph{G}_{\set{X}}$.

By construction of \rse{}, each step of the path satisfies $\rel{}^{(k+1)}=\rel{}^{(k)}\cup\set{P}^{(k)}$, where $\rel{}^{(k)}$ is graphically consistent but not sufficient, and $\set{P}^{(k)}\in\partitions{\set{Z}^{(k)}}$ is a partition of the new relevant variables $\set{Z}^{(k)}$ computed by $\model{RSE}(\rel{}^{(k)},\graph{G}_{\set{X}})$. Since the new relevant variables $\set{Z}^{(k)}$ are not relevant in $\rel{}^{(k)}$, every variable in $\set{Z}^{(k)}$ is relevant in $\rel{}'$ only for the abstract variable induced by its own element of $\set{P}^{(k)}$.

Assume by contradiction that there exists a sufficient set of relevant variables $\rel{}''$ such that $\rel{}\subseteq\rel{}''\subset\rel{}'$. Since $\rel{}''\neq\rel{}'$, some relevant set introduced along the path is missing from $\rel{}''$. Let $k^\ast$ be the first step of the path such that $\set{P}^{(k^\ast)}\not\subseteq\rel{}''$, and let $\set{P}\in\set{P}^{(k^\ast)}$ be a relevant set missing from $\rel{}''$. By the choice of $k^\ast$, it holds $\rel{}^{(k^\ast)}\subseteq\rel{}''$.

Among the sufficient sets of relevant variables $\rel{}^\ast$ such that $\rel{}^{(k^\ast)}\subseteq\rel{}^\ast\subseteq\rel{}''$, which include $\rel{}''$, we choose one that is minimal with respect to inclusion. Such a set exists since there are finitely many candidates. We show that $\rel{}^\ast\in\mise{\rel{}^{(k^\ast)}}{\graph{G}_{\set{X}}}$ by checking \Cref{def:mise}:
(i.) $\rel{}^\ast\supseteq\rel{}^{(k^\ast)}$ by construction;
(ii.) $\rel{}^\ast$ is sufficient by construction, and it is graphically consistent by \Cref{prop:prata-lemma}, since $\rel{}^\ast\subseteq\rel{}'$ and $\rel{}'$ is graphically consistent;
(iii.) any $\rel{}^{(k^\ast)}\subseteq\rel{}'''\subset\rel{}^\ast$ is not sufficient, since $\rel{}^{(k^\ast)}$ is not sufficient and by the inclusion-minimality of $\rel{}^\ast$.

Since $\rel{}^{(k^\ast)}$ is graphically consistent, \Cref{lemma:newrelevants} applies to $\rel{}^{(k^\ast)}$ and $\rel{}^\ast$: every variable in $\set{Z}^{(k^\ast)}$ is relevant in $\rel{}^\ast$. However, the variables in $\set{P}\subseteq\set{Z}^{(k^\ast)}$ are relevant in $\rel{}'$ only for the abstract variable induced by $\set{P}$, which is missing from $\rel{}''$. Since $\rel{}^\ast\subseteq\rel{}''\subseteq\rel{}'$, no variable in $\set{P}$ is relevant in $\rel{}^\ast$, and since $\set{P}$ is non-empty this leads to a contradiction. Therefore, no $\rel{}''$ such that $\rel{}\subseteq\rel{}''\subset\rel{}'$ is sufficient, and $\rel{}'\in\mise{\rel{}}{\graph{G}_{\set{X}}}$.

\paragraph{Completeness.}
If $\rel{}$ is sufficient and graphically consistent, then \rse{} returns $\setdef{\rel{}}$, which is indeed the only minimal sufficient extension of $\rel{}$. Similarly, if $\rel{}$ is not graphically consistent, the algorithm returns the empty set, which again coincides with the \model{MiSE} of $\rel{}$ on $\graph{G}_{\set{X}}$.

We prove the general case by defining the sequence $\rel{}^{(1)}, \rel{}^{(2)}, \ldots$ as
\begin{equation}
\rel{}^{(i)} =
\begin{cases}
    \rel{} & \text{if } i=1\\
    \rel{}^{(i-1)}\cup\set{P}^{(i-1)} & \text{if } i>1
\end{cases}
\end{equation}
where $\set{P}^{(i)}$ is one of the partitions produced by $\model{RSE}(\rel{}^{(i)}, \graph{G}_{\set{X}})$. We separately prove in \Cref{lemma:steprse} that $\set{P}^{(i)}$ always exists and is non-empty whenever $\rel{}'\in\mise{\rel{}^{(i)}}{\graph{G}_{\set{X}}}$ and $\rel{}'\neq\rel{}^{(i)}$. We consequently prove the existence of the sequence by showing that $\rel{}'\in\mise{\rel{}^{(i)}}{\graph{G}_{\set{X}}}$.

\begin{description}
\item[Base case] Let $\rel{}^{(1)}=\rel{}$. By hypothesis, $\rel{}'\in\mise{\rel{}}{\graph{G}_{\set{X}}}=\mise{\rel{}^{(1)}}{\graph{G}_{\set{X}}}$.
\item[Inductive step] Assume $\rel{}'\in\mise{\rel{}^{(i)}}{\graph{G}_{\set{X}}}$ for some $i\geq 1$. If $\rel{}^{(i)}=\rel{}'$, the sequence stops at $n=i$. Otherwise, we can apply \Cref{lemma:steprse} to $\rel{}^{(i)}$ and $\rel{}'$, which guarantees the existence of a non-empty $\set{P}^{(i)}\subseteq\rel{}'\setminus\rel{}^{(i)}$, and hence of $\rel{}^{(i+1)}=\rel{}^{(i)}\cup\set{P}^{(i)}$. Since $\rel{}^{(i)}\subset\rel{}^{(i+1)}\subseteq\rel{}'$ and $\rel{}'\in\mise{\rel{}^{(i)}}{\graph{G}_{\set{X}}}$, we prove in \Cref{lemma:incmise} that $\rel{}'\in\mise{\rel{}^{(i+1)}}{\graph{G}_{\set{X}}}$, completing the inductive step.
\end{description}

Since $\set{P}^{(i)}$ is non-empty at every step and the sequence is strictly increasing within the finite set $\rel{}'\setminus\rel{}^{(1)}$, there exists a finite integer $n$ such that $\rel{}^{(n)}=\rel{}'$. Having defined the finite sequence $\rel{}^{(i)}$, we show by induction on $i\in\setdef{1,\ldots,n}$ that $\model{RSE}(\rel{}^{(i)}, \graph{G}_{\set{X}})\subseteq\model{RSE}(\rel{}, \graph{G}_{\set{X}})$.

\begin{description}
\item[Base case] Trivially, $\model{RSE}(\rel{}^{(1)}, \graph{G}_{\set{X}})=\model{RSE}(\rel{}, \graph{G}_{\set{X}})$.
\item[Inductive step] Assume $\model{RSE}(\rel{}^{(i)}, \graph{G}_{\set{X}})\subseteq\model{RSE}(\rel{}, \graph{G}_{\set{X}})$ for some $1\leq i<n$. Since $\set{P}^{(i)}$ is one of the partitions explored by \model{RSE} at $\rel{}^{(i)}$, by construction of the algorithm $\model{RSE}(\rel{}^{(i+1)}, \graph{G}_{\set{X}})\subseteq\model{RSE}(\rel{}^{(i)}, \graph{G}_{\set{X}})$, and therefore $\model{RSE}(\rel{}^{(i+1)}, \graph{G}_{\set{X}})\subseteq\model{RSE}(\rel{}, \graph{G}_{\set{X}})$.
\end{description}

At $i=n$, $\rel{}^{(n)}=\rel{}'$ is graphically consistent and sufficient, and we have shown above that $\model{RSE}(\rel{}^{(n)}, \graph{G}_{\set{X}})\subseteq\model{RSE}(\rel{}, \graph{G}_{\set{X}})$. Therefore, $\rel{}'\in\model{RSE}(\rel{}, \graph{G}_{\set{X}})$.
\end{proof}

\subsection{Lemma~\ref{lemma:qxconstraint}}%
\label{app:qxconstraint}

\qxconstraintlemma*
\begin{proof}
Throughout the proof, we write $\set{V}$ and $\set{Z}$ for $\set{V}_i$ and $\set{Z}_i$. Intuitively, two variables that are relevant to the same abstract variable must reach the same abstract variables in $\rel{}'$ (\Cref{def:wellformed}). The set $\set{Q}_X$ collects the existing abstract variables that $X$ reaches through a child in a unique block, and these are exactly the existing abstract variables that $X$ reaches in $\rel{}'$ through its children. Hence, $\set{Q}_{X_1}$ and $\set{Q}_{X_2}$ must coincide.

Formally, let $X_1, X_2\in\set{Z}$ with $X_1,X_2\in\rel{Y}'$ for some $Y\in\set{Y}'$. Since $X_1, X_2$ are not relevant in $\rel{}$, $Y$ is a new abstract variable, and in particular $Y\neq Y_k$ for every $Y_k\in\set{Y}$. By symmetry, it suffices to show $\set{Q}_{X_1}\subseteq\set{Q}_{X_2}$. For every $Y_k\in\set{Y}$,
\begin{align*}
    Y_k\in\set{Q}_{X_1}
    &\implies \exists X_r\in\rel{Y_k}.\ X_1\rpdirect X_r
    && \text{(\Cref{lemma:singletarget})}\\
    &\implies Y\to Y_k\in\graph{G}_{\set{Y}'}
    && \text{(\Cref{def:wellformed})}\\
    &\implies \exists X_r'\in\rel{Y_k}.\ X_2\rpdirect X_r'
    && \text{(\Cref{def:wellformed})}\\
    &\implies Y_k\in\set{Q}_{X_2}
    && \text{(\Cref{lemma:uniquechild})}.
\end{align*}
In the first step, the child of $X_1$ with the unique abstract target $Y_k$ still leads only to $\rel{Y_k}$ in $\rel{}'$, as in the proof of \Cref{lemma:newrelevants}. The third step uses that graphical consistency requires every relevant variable of $Y$ to reach each abstract child of $Y$. In the last step, the child of $X_2$ on the $\rel{}'$-direct path has the unique abstract target $Y_k$.
\end{proof}

\subsection{Correctness and Completeness of \model{qRSE}}%
\label{app:rsestar}

\begin{corollary}[\model{qRSE} Correctness and Completeness]%
\label{cor:rsestar}
    Let $\rel{}$ be a graphically consistent set of relevant variables on a low-level graph $\graph{G}_{\set{X}}$. Then, $\model{qRSE}(\rel{},\graph{G}_{\set{X}})=\mise{\rel{}}{\graph{G}_{\set{X}}}$.
\end{corollary}
\begin{proof}
\model{qRSE} differs from \model{RSE} only in the partitions of the new relevant variables $\set{Z}_i$ explored at each recursive call (Lines 11--14 of \Cref{alg:rse}). Since the groups $\set{M}_1,\ldots,\set{M}_k$ of variables sharing the same set $\set{Q}_X$ partition $\set{Z}_i$, every element of $\partitions{\set{M}_1}\times\cdots\times\partitions{\set{M}_k}$ corresponds to a partition of $\set{Z}_i$.

\paragraph{Termination and correctness.} Every recursive call of \model{qRSE} is also a recursive call of \model{RSE} on the same input. Hence, the recursion tree of \model{qRSE} is a subtree of the one of \model{RSE}, which is finite (\Cref{thm:rse}), and by induction on its depth it holds $\model{qRSE}(\rel{},\graph{G}_{\set{X}})\subseteq\model{RSE}(\rel{},\graph{G}_{\set{X}})=\mise{\rel{}}{\graph{G}_{\set{X}}}$.

\paragraph{Completeness.} Let $\rel{}'\in\mise{\rel{}}{\graph{G}_{\set{X}}}$. In the proof of completeness of \Cref{thm:rse}, we built a sequence $\rel{}=\rel{}^{(1)}\subset\cdots\subset\rel{}^{(n)}=\rel{}'$ with $\rel{}^{(i+1)}=\rel{}^{(i)}\cup\set{P}^{(i)}$, where $\rel{}'\in\mise{\rel{}^{(i)}}{\graph{G}_{\set{X}}}$ (\Cref{lemma:incmise}) and $\set{P}^{(i)}\subseteq\rel{}'\setminus\rel{}^{(i)}$ is a partition of the new relevant variables $\set{Z}^{(i)}$ computed at $\rel{}^{(i)}$ (\Cref{lemma:steprse}). Each $\rel{}^{(i)}$ is graphically consistent, since $\rel{}^{(i)}\subseteq\rel{}'$ (\Cref{prop:prata-lemma}). We show that \model{qRSE} explores $\set{P}^{(i)}$ at $\rel{}^{(i)}$. Every element of $\set{P}^{(i)}$ is the relevant set $\rel{Y}'$ of some abstract variable $Y$ of $\rel{}'$. By \Cref{lemma:qxconstraint}, applied to $\rel{}^{(i)}$ and $\rel{}'$, all the variables in $\rel{Y}'$ share the same set $\set{Q}_X$, and thus belong to the same group $\set{M}_j$. Since $\set{P}^{(i)}$ covers $\set{Z}^{(i)}$, the elements of $\set{P}^{(i)}$ contained in each group $\set{M}_j$ form a partition of $\set{M}_j$. Hence, $\set{P}^{(i)}$ belongs to the Cartesian product explored by \model{qRSE} at $\rel{}^{(i)}$, and $\model{qRSE}(\rel{}^{(i+1)},\graph{G}_{\set{X}})\subseteq\model{qRSE}(\rel{}^{(i)},\graph{G}_{\set{X}})$. By induction on $i$, as in the proof of \Cref{thm:rse}, it follows that $\rel{}'\in\model{qRSE}(\rel{}^{(n)},\graph{G}_{\set{X}})\subseteq\model{qRSE}(\rel{},\graph{G}_{\set{X}})$.
\end{proof}

\subsection{Theorem \ref{thm:paramconsistency}}%
\label{app:paramconsistency}

\paramconsistencythm*
\begin{proof}
Let $\rel{}'\in\mise{\rel{}}{\graph{G}_{\set{X}}}$ be parametrically consistent on $\scm{L}$. Then, there exist an abstraction function $\mat{T}$ with relevant sets $\rel{}'$ and a linear ANM $\scm{H}'$ with coefficients $\mat{M}$ over the abstract variables $\set{Y}'$ such that $\scm{H}'$ is a strong $\mat{T}$-abstraction of $\scm{L}$. To prove the theorem, we are interested in the structural equations of the abstract variables~$\set{V}$ that are confounded in $\rel{}$. We can write the structural equation of each $Y_j\in\set{V}$ in $\scm{H}'$ as
\begin{equation}
    Y_j = \sum_{Y_i\in\pa{Y_j, \graph{G}_{\set{Y}'}}} m_{ij} Y_i + U_j,
\end{equation}
where $\graph{G}_{\set{Y}'}$ is the abstract graph induced by $\rel{}'$ and $U_j$ is the exogenous term of $Y_j$ in $\scm{H}'$, whose independence from the other exogenous terms is guaranteed by the sufficiency of $\rel{}'$.

\paragraph{Old and new parents of $Y_j$.}
We split the parents of $Y_j$ according to whether they are introduced by the extension or whether they were already induced by the starting set of relevant variables $\rel{}$. Formally, we denote as $\set{C} = \pa{Y_j, \graph{G}_{\set{Y}'}} \cap \set{Y}$ the \emph{old} parents, and as $\set{D} = \pa{Y_j, \graph{G}_{\set{Y}'}} \setminus \set{C}$ the \emph{new} parents. The old parents are a subset of the parents of $Y_j$ in the abstract graph $\graph{G}_{\set{Y}}$ induced by $\rel{}$, \ie, $\set{C}\subseteq\pa{Y_j, \graph{G}_{\set{Y}}}$. Indeed, since every variable relevant in $\rel{}$ is also relevant in $\rel{}'$, every $\rel{}'$-direct path is also $\rel{}$-direct, and the $\rel{}'$-direct path witnessing an edge $Y_i\to Y_j$ of $\graph{G}_{\set{Y}'}$, with $Y_i\in\set{Y}$, also witnesses the same edge in $\graph{G}_{\set{Y}}$. However, the inclusion might be strict, since a $\rel{}$-direct path might cross a variable that is relevant in $\rel{}'$, but not in $\rel{}$. Therefore, we can write
\begin{align}
    Y_j &= \sum_{Y_i\in\set{C}} m_{ij} Y_i + \sum_{Y_h\in\set{D}} m_{hj} Y_h + U_j\\
    &= \sum_{Y_i\in\pa{Y_j, \graph{G}_{\set{Y}}}} m_{ij} Y_i + \sum_{Y_h\in\set{D}} m_{hj} Y_h + U_j,
\end{align}
where we set $m_{ij}=0$ for every $Y_i\in\pa{Y_j, \graph{G}_{\set{Y}}}\setminus\set{C}$.

\paragraph{Proof structure.}
In the remainder of the proof, we first show that the relevant sets of the abstract variables in $\set{D}$ are \emph{necessarily} built over a subset of the new relevant variables $\set{Z}^{(j)}$ targeting $Y_j$, then derive the composed abstraction matrix~$\mat{A}$, and finally prove the independence of the exogenous term. We denote as $\block{}$ and $\block{}'$ the blocks induced by $\rel{}$ and $\rel{}'$, respectively. Since $\rel{}'$ is sufficient, the blocks $\block{}'$ are pairwise disjoint (\Cref{prop:causalsufficiency}).

\paragraph{Relevant variables of the new parents.}
Let $Y_h\in\set{D}$ and let $X\in\rel{Y_h}'$ be any of its relevant variables. Since $\rel{}'$ is graphically consistent, $X$ has at least one $\rel{}'$-direct path to some variable in $\rel{Y_j}$. We choose one of these paths and denote as $X'$ the child of $X$ on this path. We show that $X'$ has a unique abstract target in $\rel{}$, namely $Y_j$, and that $X$ is intersected in $\rel{}$.
\begin{enumerate}[label=(\roman*)]
\item We show that $\targets{X'}{\block{}}=\setdef{Y_j}$. Since the chosen path is $\rel{}'$-direct and ends in $\rel{Y_j}$, with $Y_j\in\set{Y}$, this follows from \Cref{lemma:uniquechild}.
\item Since the $\rel{}'$-direct path from $X$ to $\rel{Y_j}$ is also $\rel{}$-direct, it holds $Y_j\in\targets{X}{\block{}}$. If $\targets{X}{\block{}}=\setdef{Y_j}$, by target stability (\Cref{lemma:singletarget}) it would hold $\targets{X}{\block{}'}=\setdef{Y_j}$, contradicting $X\in\rel{Y_h}'\subseteq\block{Y_h}'$. Hence, $\size{\targets{X}{\block{}}}\geq 2$, and $\set{V}_i=\targets{X}{\block{}}\in\set{I}$.
\end{enumerate}

\paragraph{New parents are built over $\set{Z}^{(j)}$.}
By (ii), $X\in\bigcap_{Y\in\set{V}_i}\block{Y}$ and, by (i), $X$ has a child $X'$ with a unique abstract target. Hence, $X\in\set{Z}_i\subseteq\set{Z}$ (\Cref{def:newrelevants}). Furthermore, since $Y_j\in\set{V}_i=\targets{X}{\block{}}$, it holds $X\in\set{Z}^{(j)}$. Since $X$ is arbitrary, it holds $\rel{Y_h}'\subseteq\set{Z}^{(j)}\subseteq\set{Z}$ for every $Y_h\in\set{D}$. In other words, the relevant variables of the new parents of $Y_j$ are only drawn from the new relevant variables of the jointly targeted sets containing $Y_j$.

\paragraph{Composed abstraction matrix.}
Since $\scm{H}'$ is a $\mat{T}$-abstraction of $\scm{L}$, each new parent can be written as $Y_h=\sum_{X_l\in\rel{Y_h}'}t_{lh}X_l$. Since $\rel{Y_h}'\subseteq\set{Z}^{(j)}$ for every $Y_h\in\set{D}$ and relevant sets are disjoint, it holds $t_{lh}=0$ for every $X_l\in\set{Z}^{(j)}\setminus\rel{Y_h}'$. Furthermore, we set $m_{hj}=0$ for every new variable $Y_h\in\set{Y}'\setminus\set{Y}$ that is not in $\set{D}$. Hence, we can extend the inner sum to $\set{Z}^{(j)}$ and exchange the order of summation, obtaining
\begin{align}
    Y_j
    &= \sum_{Y_i\in\pa{Y_j, \graph{G}_{\set{Y}}}} m_{ij} Y_i + \sum_{Y_h\in\set{Y}'\setminus\set{Y}} m_{hj} \sum_{X_l\in\set{Z}^{(j)}} t_{lh} X_l + U_j\\
    &= \sum_{Y_i\in\pa{Y_j, \graph{G}_{\set{Y}}}} m_{ij} Y_i + \sum_{X_l\in\set{Z}^{(j)}} X_l \sum_{Y_h\in\set{Y}'\setminus\set{Y}} t_{lh} m_{hj} + U_j\\
    &= \sum_{Y_i\in\pa{Y_j, \graph{G}_{\set{Y}}}} m_{ij} Y_i + \sum_{X_l\in\set{Z}^{(j)}} X_l a_{lj} + U_j,
\end{align}
where $a_{lj}=\sum_{Y_h\in\set{Y}'\setminus\set{Y}} t_{lh} m_{hj}$. Finally, we show that the composed abstraction matrix $\mat{A}$, defined by \Cref{eq:composedabstraction} for every $X_l\in\set{Z}$ and $Y_j\in\set{V}$, is consistent with the above equation, \ie, $a_{lj}=0$ for every $X_l\in\set{Z}\setminus\set{Z}^{(j)}$. If $X_l$ is relevant in $\rel{}'$ for some new variable $Y_h$, then $Y_h\notin\set{D}$, since $\rel{Y_h}'\subseteq\set{Z}^{(j)}$ for every $Y_h\in\set{D}$, hence $m_{hj}=0$. Otherwise, $t_{lh}=0$ for every new variable $Y_h$. In both cases, it holds $a_{lj}=0$.

\paragraph{Independence of the exogenous term.}
We show that $U_j$ is jointly independent of the abstract parents $\pa{Y_j,\graph{G}_{\set{Y}}}$ and of each low-level variable in $\set{Z}^{(j)}$, which are the regressors of $Y_j$ in \Cref{eq:paramconsistency}. Since $\scm{H}'$ is causally sufficient and acyclic, $U_j$ is independent of every non-descendant of $Y_j$ in $\graph{G}_{\set{Y}'}$. However, the regressors are not necessarily parents of $Y_j$ in $\graph{G}_{\set{Y}'}$. Indeed, $\pa{Y_j,\graph{G}_{\set{Y}}}$ might contain old variables whose edge towards $Y_j$ is not in $\graph{G}_{\set{Y}'}$, \ie, the ones with $m_{ij}=0$, and $\set{Z}^{(j)}$ might contain relevant variables of new abstract variables that are ancestors, but not parents, of $Y_j$. Therefore, we denote as $\set{S}_j=\set{Z}^{(j)}\cup\bigcup_{Y_i\in\pa{Y_j,\graph{G}_{\set{Y}}}}\rel{Y_i}$ the low-level variables that determine the regressors of $Y_j$, and we show that every $X\in\set{S}_j$ is relevant to a proper ancestor of $Y_j$ in $\graph{G}_{\set{Y}'}$. To this end, we show that directed paths between blocks of $\rel{}'$ follow the abstract graph $\graph{G}_{\set{Y}'}$, and then apply this result to the variables in $\set{S}_j$.
\begin{enumerate}[label=(\alph*)]
\item Whenever there exists a directed path from a variable in $\block{Y_k}'$ to a variable in $\block{Y_g}'$, then either $Y_k=Y_g$ or $Y_k$ is an ancestor of $Y_g$ in $\graph{G}_{\set{Y}'}$. By definition of blocks, we can extend the path to a directed path ending in $\rel{Y_g}'$. Since the blocks $\block{}'$ are disjoint, the first variable on the path that is relevant in $\rel{}'$ belongs to $\rel{Y_k}'$. We split the path at the variables that are relevant in $\rel{}'$, obtaining a sequence of $\rel{}'$-direct sub-paths. Each sub-path between relevant variables of different abstract variables witnesses an edge of $\graph{G}_{\set{Y}'}$, and the claim follows.
\item Let $X\in\set{S}_j$. If $X\in\rel{Y_i}$ for some $Y_i\in\pa{Y_j,\graph{G}_{\set{Y}}}$, then $X$ is relevant to $Y_i$ also in $\rel{}'$ and, by graphical consistency of $\rel{}$, $X$ has a directed path to $\rel{Y_j}$. If $X\in\set{Z}^{(j)}$, then $X$ is relevant in $\rel{}'$ for some new abstract variable $Y_h$ (\Cref{lemma:newrelevants}) and, since $Y_j\in\targets{X}{\block{}}$, $X$ has a $\rel{}$-direct path to $\rel{Y_j}$. In both cases, $X\in\rel{Y_k}'\subseteq\block{Y_k}'$ for some $Y_k\neq Y_j$ with a directed path to $\rel{Y_j}$. Hence, by (a), $Y_k$ is a proper ancestor of $Y_j$ in $\graph{G}_{\set{Y}'}$.
\end{enumerate}
Therefore, the abstract variables $Y_k$ whose relevant variables determine the regressors of $Y_j$ are proper ancestors of $Y_j$ in $\graph{G}_{\set{Y}'}$ and, by acyclicity, non-descendants of $Y_j$. Hence, $U_j$ is jointly independent of them. Finally, by the absence of cancelling paths (\Cref{assumption:nocancellingpaths}), every low-level exogenous term affecting a relevant variable $X\in\rel{Y_k}'$ also affects $Y_k=\sum_{X_l\in\rel{Y_k}'}t_{lk}X_l$. Hence, the independence of $U_j$ from $Y_k$ extends to each of its relevant variables, and $U_j$ is jointly independent of $\pa{Y_j,\graph{G}_{\set{Y}}}$ and of the low-level variables $\set{Z}^{(j)}$.
\end{proof}

\subsection{Corollary \ref{cor:collinear}}%
\label{app:collinear}

\collinearcor*
\begin{proof}
Let $\rel{}'\in\mise{\rel{}}{\graph{G}_{\set{X}}}$ be parametrically consistent on $\scm{L}$ with abstraction function $\mat{T}$ and abstract coefficients $\mat{M}$. By \Cref{thm:paramconsistency}, the composed abstraction matrix satisfies $a_{lj}=\sum_{Y_h\in\set{Y}'\setminus\set{Y}} t_{lh} m_{hj}$ for every $X_l\in\set{Z}$ and $Y_j\in\set{V}$.

Without loss of generality, let $X_1, X_2\in\set{Z}$ be two low-level variables in the relevant set of the same new abstract variable $Y_p\in\set{Y}'\setminus\set{Y}$. Then, in the composed abstraction matrix, it holds
\begin{align}
    a_{1j} &= \sum_{Y_h\in\set{Y}'\setminus\set{Y}}t_{1h} m_{hj} = t_{1p} m_{pj}\\
    a_{2j} &= \sum_{Y_h\in\set{Y}'\setminus\set{Y}}t_{2h} m_{hj} = t_{2p} m_{pj},
\end{align}
since $t_{1p}$ and $t_{2p}$ are the only entries of the sum different from zero given that each low-level variable is a member of at most one set of relevant variables. Since both $X_1$ and $X_2$ are relevant to $Y_p$, it holds $t_{1p}\neq 0$ and $t_{2p}\neq 0$ and, consequently, $\mat{A}_{1,:}=(t_{1p}/t_{2p})\,\mat{A}_{2,:}$ with a non-zero scalar $t_{1p}/t_{2p}$. Hence, the rows of low-level variables in the same relevant set are collinear.
\end{proof}

\subsection{Proposition \ref{prop:rankone}}%
\label{app:rankone}

\rankoneprop*
\begin{proof}
We first show the factorization of the submatrix $\mat{A}_{\rel{Y_h}',\set{V}}$, then that both factors are non-zero, and finally the uniqueness of the factorization.

\paragraph{Factorization.}
By \Cref{thm:paramconsistency}, it holds $a_{lj}=\sum_{Y_{h'}\in\set{Y}'\setminus\set{Y}} t_{lh'}m_{h'j}$ for every $X_l\in\set{Z}$ and $Y_j\in\set{V}$. Let $X_l\in\rel{Y_h}'$. Since the relevant sets of $\rel{}'$ are disjoint, $Y_h$ is the only abstract variable with $t_{lh}\neq 0$. Hence, only one entry of the sum is non-zero, and $a_{lj}=t_{lh}m_{hj}$ for every $Y_j\in\set{V}$. Stacking these entries for every $X_l\in\rel{Y_h}'$ and $Y_j\in\set{V}$, we obtain $\mat{A}_{\rel{Y_h}',\set{V}}=\vec{t}_h\vec{m}_{h,\set{V}}^\top$.

\paragraph{Non-zero factors.}
The abstraction vector $\vec{t}_h$ is non-zero, since $t_{lh}\neq 0$ for every relevant variable $X_l\in\rel{Y_h}'$. To show that $\vec{m}_{h,\set{V}}$ is non-zero, let $X\in\rel{Y_h}'\subseteq\set{Z}$. Since $X\in\set{Z}_i$ for some $i$ (\Cref{def:newrelevants}), $X$ is intersected in $\rel{}$ and has a child $X'$ with a unique abstract target $\targets{X'}{\block{}}=\setdef{Y_j}$, with $Y_j\in\targets{X}{\block{}}\subseteq\set{V}$. By target stability (\Cref{lemma:singletarget}), it holds $\targets{X'}{\block{}'}=\setdef{Y_j}$, hence $X'$ is either relevant to $Y_j$ or has a $\rel{}'$-direct path to $\rel{Y_j}$. In both cases, the path from $X$ through $X'$ to $\rel{Y_j}$ is $\rel{}'$-direct, and $Y_h\to Y_j$ is an edge of $\graph{G}_{\set{Y}'}$. By faithfulness, $m_{hj}\neq 0$. Notice that the remaining entries of $\vec{m}_{h,\set{V}}$ might be zero, whenever $Y_h$ is not a parent of the corresponding confounded variable.

\paragraph{Uniqueness.}
Since both $\vec{t}_h$ and $\vec{m}_{h,\set{V}}$ are non-zero, $\mat{A}_{\rel{Y_h}',\set{V}}$ has rank one. Let $\mat{A}_{\rel{Y_h}',\set{V}}=\vec{t}\vec{m}^\top$ be any other factorization. Then, the column space of $\mat{A}_{\rel{Y_h}',\set{V}}$ is spanned by both $\vec{t}_h$ and $\vec{t}$, hence $\vec{t}=c\,\vec{t}_h$ for some $c\neq 0$. Consequently, $\vec{t}_h(c\,\vec{m}-\vec{m}_{h,\set{V}})^\top=\mat{0}$ and, since $\vec{t}_h\neq\vec{0}$, it holds $\vec{m}=\vec{m}_{h,\set{V}}/c$.
\end{proof}

\subsection{Theorem \ref{thm:maximalpartitioning}}%
\label{app:maximalpartitioning}

\maximalpartitioningthm*

\begin{proof}
    Throughout the proof, $\mat{T}$ and $\mat{M}$ denote the abstraction function and the abstract coefficients of the strong $\mat{T}$-abstraction $\scm{H}$ satisfying \Cref{eq:mpcondition}. We denote as $\set{N}=\set{Y}'\setminus\set{Y}$ the new abstract variables, as $\rel{}''=\rel{}\cup\set{P}^\ast$ the extension induced by the maximal partitioning, and, given $\set{P}\in\set{P}^\ast$, as $Y_{\set{P}}$ the abstract variable induced by $\set{P}$ in $\graph{G}''$. We recall that $\set{P}^\ast_{\set{V}}$ collects the sets $\set{P}\in\set{P}^\ast$ such that $\ch{Y_{\set{P}},\graph{G}''}\subseteq\set{V}$ (\Cref{def:unmediated}). Since every new relevant variable must be relevant in $\rel{}'$ (\Cref{lemma:newrelevants}, applied to each $\set{Z}_i$) and is not relevant in $\rel{}$, it holds $\set{Z}\subseteq\bigcup_{Y\in\set{N}}\rel{Y}'$. Consequently, every variable relevant in $\rel{}''$ is also relevant in $\rel{}'$. Furthermore, for every $\set{P}\in\set{P}^\ast$, it holds $\ch{Y_{\set{P}},\graph{G}''}\cap\set{Y}\subseteq\set{V}$. Indeed, an edge from $Y_{\set{P}}$ to some $Y_k\in\set{Y}$ is witnessed by an $\rel{}''$-direct, hence $\rel{}$-direct, path from some $X\in\set{P}\subseteq\set{Z}$ to $\rel{Y_k}$, hence $Y_k\in\targets{X}{\block{}}\subseteq\set{V}$. We prove the theorem in four steps.

    \paragraph{Step 1: New variables with children in $\set{Y}$ are in $\set{N}_{\set{V}}$.}
    Let $Y_s\in\set{N}$ with $\ch{Y_s,\graph{G}_{\set{Y}'}}\subseteq\set{Y}$, and let $X\in\rel{Y_s}'$. We show that $\ch{Y_s,\graph{G}_{\set{Y}'}}\subseteq\set{V}$, \ie, $Y_s\in\set{N}_{\set{V}}$, and that $\rel{Y_s}'\subseteq\set{Z}$. First, $\ch{Y_s,\graph{G}_{\set{Y}'}}\neq\emptyset$. Since the graph is clean (\Cref{assumption:clean}), $X$ has a $\rel{}$-direct path to some variable in $\rel{}$. Taking the last node of this path in $\rel{Y_s}'$ and the first node in $\cup\rel{}'$ after it, we obtain a $\rel{}'$-direct path from $\rel{Y_s}'$ to the relevant set of some $Y\neq Y_s$, \ie, an edge $Y_s\to Y$ in $\graph{G}_{\set{Y}'}$. Second, since $\rel{}\subseteq\rel{}'$, every $\rel{}'$-direct path from $X$ to some $\rel{Y_j}$, with $Y_j\in\ch{Y_s,\graph{G}_{\set{Y}'}}\subseteq\set{Y}$, is also $\rel{}$-direct, hence $\ch{Y_s,\graph{G}_{\set{Y}'}}\subseteq\targets{X}{\block{}}$. If $\size{\targets{X}{\block{}}}=1$, by target stability (\Cref{lemma:singletarget}) we would have $\targets{X}{\block{}'}=\targets{X}{\block{}}\subseteq\set{Y}$, contradicting $X\in\rel{Y_s}'\subseteq\block{Y_s}'$. Hence, $\size{\targets{X}{\block{}}}\geq 2$, \ie, $X$ is intersected, $\targets{X}{\block{}}\in\set{I}$, and $\ch{Y_s,\graph{G}_{\set{Y}'}}\subseteq\set{V}$. Therefore, $Y_s\in\set{N}_{\set{V}}$ is a parent of at least one confounded variable, and, as shown in the proof of \Cref{thm:paramconsistency} (paragraph ``New parents are built over $\set{Z}^{(j)}$''), its relevant set is contained in the new relevant variables, \ie, $\rel{Y_s}'\subseteq\set{Z}$. Furthermore, $\vec{m}_{s,\set{V}}\neq\vec{0}$, since the abstract coefficients of the edges of $\graph{G}_{\set{Y}'}$ are non-zero.

    \paragraph{Step 2: The relevant set of each variable in $\set{N}_{\set{V}}$ is in $\set{P}^\ast$.}
    Let $Y_s\in\set{N}_{\set{V}}$. By \Cref{prop:rankone}, the row of each $X_l\in\rel{Y_s}'$ is $\vec{a}_{l,\set{V}}=t_{ls}\vec{m}_{s,\set{V}}$, with $t_{ls}\neq 0$, hence the rows of $\rel{Y_s}'$ are mutually collinear. Any other $X_{l'}\in\set{Z}\setminus\rel{Y_s}'$ is relevant in $\rel{}'$ for some $Y_{h'}\in\set{N}$ with $Y_{h'}\neq Y_s$, hence $\vec{a}_{l',\set{V}}=t_{l'h'}\vec{m}_{h',\set{V}}$, with $t_{l'h'}\neq 0$. Since $\rel{Y_{h'}}'$ intersects $\set{Z}_i$ for some $i$, it holds $\rel{Y_{h'}}'\subseteq\set{Z}_i\subseteq\set{Z}$ (\Cref{lemma:newrelclosure}). By \Cref{eq:mpcondition}, $\vec{m}_{s,\set{V}}\not\propto\vec{m}_{h',\set{V}}$, hence $\vec{a}_{l,\set{V}}\not\propto\vec{a}_{l',\set{V}}$. Therefore, by definition of maximal partitioning, $\rel{Y_s}'\in\set{P}^\ast$. Consequently, every $\set{P}\in\set{P}^\ast$ that is not the relevant set of a variable in $\set{N}_{\set{V}}$ only contains relevant variables of the new variables in $\set{N}\setminus\set{N}_{\set{V}}$.

    \paragraph{Step 3: The relevant sets of the variables in $\set{N}_{\set{V}}$ are unmediated.}
    Let $Y_s\in\set{N}_{\set{V}}$ and assume by contradiction that $Y_{\rel{Y_s}'}$ has a child $Y_{\set{Q}}$ in $\graph{G}''$ for some $\set{Q}\in\set{P}^\ast$, $\set{Q}\neq\rel{Y_s}'$. Then, there exists a $\rel{}''$-direct path from some $X\in\rel{Y_s}'$ to some $X'\in\set{Q}$. Its intermediate nodes are not relevant in $\rel{}''$, hence they are neither in $\cup\rel{}$ nor in $\set{Z}\supseteq\rel{Y_s}'$. Let $X''$ be the first node after $X$ on the path that is relevant in $\rel{}'$, which exists since $X'\in\set{Z}$ is relevant in $\rel{}'$. Then, $X''\in\rel{Y}'$ for some $Y\in\set{N}\setminus\setdef{Y_s}$, and the sub-path from $X$ to $X''$ is $\rel{}'$-direct. Hence, $Y_s\to Y$ is an edge of $\graph{G}_{\set{Y}'}$ with $Y\notin\set{V}$, contradicting $Y_s\in\set{N}_{\set{V}}$. Hence, $Y_{\rel{Y_s}'}$ has no new children in $\graph{G}''$ and, by the above, all its children are in $\set{V}$. Therefore, $\setdef{\rel{Y}'\mid Y\in\set{N}_{\set{V}}}\subseteq\set{P}^\ast_{\set{V}}$.

    \paragraph{Step 4: The remaining relevant sets are mediated.}
    Let $\set{P}\in\set{P}^\ast$ be a set that is not the relevant set of a variable in $\set{N}_{\set{V}}$, and let $X\in\set{P}$. By Step 2, $X\in\rel{Y_{g_0}}'$ for some $Y_{g_0}\in\set{N}\setminus\set{N}_{\set{V}}$. Since $\graph{G}_{\set{Y}'}$ is acyclic and finite, and each variable in $\set{N}\setminus\set{N}_{\set{V}}$ has a child in $\set{N}$, there exists a directed path $Y_{g_0}\to Y_{g_1}\to\ldots\to Y_{g_m}$ in $\graph{G}_{\set{Y}'}$ with $Y_{g_1},\ldots,Y_{g_m}\in\set{N}$ and $Y_{g_m}\in\set{N}_{\set{V}}$. By graphical consistency of $\rel{}'$ (\Cref{def:wellformed}), every relevant variable of $Y_{g_i}$ has a $\rel{}'$-direct path to some relevant variable of $Y_{g_{i+1}}$. By concatenating these paths, we obtain a directed path $\pi$ from $X$ to some variable in $\rel{Y_{g_m}}'$, whose nodes are either non-relevant in $\rel{}'$ or relevant to some $Y_{g_i}\in\set{N}$, hence not in $\cup\rel{}$. Let $X_1$ be the last node of $\pi$ in $\set{P}$, and $X_2$ the first node after $X_1$ that is relevant in $\rel{}''$, which exists since $\rel{Y_{g_m}}'\subseteq\set{Z}$ (Step 1). By construction, $X_2\in\set{Q}$ for some $\set{Q}\in\set{P}^\ast$ with $\set{Q}\neq\set{P}$, and the sub-path from $X_1$ to $X_2$ is $\rel{}''$-direct. Hence, $Y_{\set{P}}\to Y_{\set{Q}}$ is an edge of $\graph{G}''$ with $Y_{\set{Q}}\notin\set{V}$, and $\set{P}\notin\set{P}^\ast_{\set{V}}$.

    \paragraph{Conclusion.}
    By Steps 3 and 4, it holds $\set{P}^\ast_{\set{V}}=\setdef{\rel{Y}'\mid Y\in\set{N}_{\set{V}}}$, hence $\rel{}\cup\set{P}^\ast_{\set{V}}\subseteq\rel{}'$. Finally, since $\rel{}$ is not sufficient, $\rel{}'\neq\rel{}$ and $\set{N}\neq\emptyset$. Since $\graph{G}_{\set{Y}'}$ is acyclic, at least one variable in $\set{N}$ has no children in $\set{N}$, \ie, all its children are in $\set{Y}$. By Step 1, it is in $\set{N}_{\set{V}}$, hence $\set{N}_{\set{V}}\neq\emptyset$ and $\set{P}^\ast_{\set{V}}\neq\emptyset$.
\end{proof}

\subsection{Corollary \ref{cor:datarsecorrectness}}%
\label{app:datarsecorrectness}

\datarsecorrectnesscor*

\begin{proof}
    We proceed by induction on the number of new abstract variables $\size{\rel{}'}-\size{\rel{}}$. If $\rel{}$ is sufficient, then $\rel{}$ is a graphically consistent and sufficient subset of $\rel{}'$, hence $\rel{}'=\rel{}$ by minimality of $\rel{}'$, and \model{dataRSE} returns $\rel{}$ (\Cref{alg:datarse}, Lines 6--8). Otherwise, by \Cref{thm:maximalpartitioning}, \model{dataRSE} extends $\rel{}$ to $\rel{}''=\rel{}\cup\set{P}^\ast_{\set{V}}$ (Lines 22--29) and recursively calls itself on $\rel{}''$, with $\rel{}\subset\rel{}''\subseteq\rel{}'$. By \Cref{prop:prata-lemma}, $\rel{}''$ is graphically consistent, and, if $\rel{}''\neq\rel{}'$, by \Cref{lemma:incmise} it holds $\rel{}'\in\mise{\rel{}''}{\graph{G}_{\set{X}}}$. Furthermore, by \Cref{prop:rankone}, the block of the exact composed abstraction matrix corresponding to each unmediated relevant set $\rel{Y_h}'$ is the rank-one matrix $\vec{t}_h\vec{m}_{h,\set{V}}^\top$. Hence, its first left singular vector, which initializes the abstraction vector of $Y_h$ in \Cref{alg:maximalpartitioning}, is $\vec{t}_h$ up to a non-zero scaling. Since the corresponding model fits the confounded variables exactly, this initialization minimizes the least-squares objective, and the alternating least-squares refinement keeps it. Therefore, the abstraction function of $\rel{}''$ is the restriction of $\mat{T}$ to the columns of $\set{Y}\cup\set{N}_{\set{V}}$, up to a non-zero scaling of the columns of $\set{N}_{\set{V}}$. Rescaling an abstract variable rescales the corresponding column of the composed abstraction matrix of the following calls, which preserves the collinearity of its rows and \Cref{eq:mpcondition}. Hence, $\rel{}''$ satisfies the same assumptions as $\rel{}$. Since $\size{\rel{}'}-\size{\rel{}''}<\size{\rel{}'}-\size{\rel{}}$, the result follows by the inductive hypothesis. Finally, the variant of \model{dataRSE} with exact composed abstraction matrices and maximal partitionings is deterministic, since both only depend on $\scm{L}$, on the current set of relevant variables, and on its abstraction coefficients, which are in turn determined at each call up to the convention on scale and sign. Therefore, if two parametrically consistent minimal sufficient extensions of $\rel{}$ satisfy the hypotheses, the variant returns both of them, and they coincide.
\end{proof}

\section{Additional Theoretical Results}%
\label{subsec:addlemmas}

\subsection{Proposition~\ref{prop:acyclicity}}

Requiring a $\rel{}$-direct path from \emph{each} relevant variable (\Cref{def:wellformed}) guarantees that the abstract graph induced by a graphically consistent set of relevant variables is acyclic.

\begin{proposition}[Acyclicity of the Abstract Graph]%
\label{prop:acyclicity}
    Let $\rel{}$ be a graphically consistent set of relevant variables on an acyclic low-level graph $\graph{G}_{\set{X}}$. Then, the abstract graph $\graph{G}_{\set{Y}}$ induced by $\rel{}$ (\Cref{def:wellformed}) is acyclic.
\end{proposition}
\begin{proof}
Intuitively, graphical consistency requires \emph{every} relevant variable of an abstract variable to reach the relevant variables of each of its abstract children. Following an abstract cycle would then produce an endless chain of low-level descendants, which a finite acyclic graph cannot contain.

Formally, assume by contradiction that $\graph{G}_{\set{Y}}$ contains a cycle $Y_1\to Y_2\to\cdots\to Y_k\to Y_1$, and extend the indices cyclically, \ie, $Y_{i+k}=Y_i$. Let $X_1\in\rel{Y_1}$. For each $i\geq 1$, since $Y_i\to Y_{i+1}\in\graph{G}_{\set{Y}}$ and $X_i\in\rel{Y_i}$, graphical consistency guarantees a $\rel{}$-direct path from $X_i$ to some $X_{i+1}\in\rel{Y_{i+1}}$. Since $Y_i\neq Y_{i+1}$ and relevant sets are disjoint, $X_{i+1}\neq X_i$, \ie, $X_{i+1}$ is a proper descendant of $X_i$. Hence, $X_1, X_2, \ldots$ is an infinite sequence in which each variable is a proper descendant of all the previous ones. Since $\graph{G}_{\set{X}}$ is acyclic, all these variables are distinct, which contradicts the finiteness of $\set{X}$.
\end{proof}

The universal quantifier of \Cref{def:wellformed} is essential. For instance, let $X_1\to X_m\to X_2$ with $\rel{Y}=\setdef{X_1,X_2}$ and $\rel{W}=\setdef{X_m}$. The induced abstract graph contains the cycle $Y\to W\to Y$, but $\rel{}$ is not graphically consistent, since $X_2$ has no path to $X_m$.

\subsection{Proposition~\ref{prop:prata-lemma}}

\begin{restatable}{proposition}{pratalemmaprop}%
\label{prop:prata-lemma}
Let $\rel{}'$ be a graphically consistent set of relevant variables over a graph $\graph{G}_{\set{X}}$. Then, any non-empty subset $\rel{}\subseteq\rel{}'$ is graphically consistent on $\graph{G}_{\set{X}}$.
\end{restatable}
\begin{proof}
    Let $\set{Y}'$ be the set of abstract variables induced by $\rel{}'$ on $\graph{G}_{\set{X}}$ and $\graph{G}_{\set{Y}'}$ be the corresponding abstract graph. Given a generic variable $Y\in\set{Y}'$, we consider the set of relevant variables $\rel{}=\rel{}'\setminus\rel{Y}'$ obtained by removing its relevant variables.

    Intuitively, removing $\rel{Y}'$ does not only remove the edges of $Y$ from the abstract graph, but it can also create new $\rel{}$-direct paths that cross the variables of $\rel{Y}'$, and hence new abstract edges. Each such path splits at the variables of $\rel{Y}'$ into $\rel{}'$-direct segments, so every new edge $Y_1\to Y_2$ corresponds to the abstract path $Y_1\to Y\to Y_2$ in $\graph{G}_{\set{Y}'}$, and graphical consistency of $\rel{}'$ provides the required paths through $\rel{Y}'$.

    Formally, let $Y_1\to Y_2$ be an edge of the abstract graph induced by $\rel{}$, witnessed by a $\rel{}$-direct path $\vec{p}$ from some variable in $\rel{Y_1}$ to some variable in $\rel{Y_2}$. We show that every $X_1\in\rel{Y_1}$ has a $\rel{}$-direct path to $\rel{Y_2}$. We first remark that, since the variables relevant in $\rel{}$ are also relevant in $\rel{}'$, every $\rel{}'$-direct path is also $\rel{}$-direct. We distinguish two cases.
    \begin{itemize}
    \item If $\vec{p}$ does not contain any variable of $\rel{Y}'$, then it is $\rel{}'$-direct and $Y_1\to Y_2\in\graph{G}_{\set{Y}'}$. By graphical consistency of $\rel{}'$, $X_1$ has a $\rel{}'$-direct, hence $\rel{}$-direct, path to $\rel{Y_2}$.
    \item Otherwise, the subpath of $\vec{p}$ up to its first variable in $\rel{Y}'$ is $\rel{}'$-direct, and so is the subpath from its last variable in $\rel{Y}'$. Hence, $Y_1\to Y$ and $Y\to Y_2$ are edges of $\graph{G}_{\set{Y}'}$. By graphical consistency of $\rel{}'$, $X_1$ has a $\rel{}'$-direct path to some $X\in\rel{Y}'$, and $X$ has a $\rel{}'$-direct path to some $X_2\in\rel{Y_2}$. Since $\graph{G}_{\set{X}}$ is acyclic, their concatenation is a directed path from $X_1$ to $X_2$, whose intermediate variables are either not relevant in $\rel{}'$ or equal to $X$, which is not relevant in $\rel{}$. Hence, it is $\rel{}$-direct.
    \end{itemize}

    Since removing any abstract variable $Y$ from $\rel{}'$ leads to a graphically consistent set of relevant variables, by induction the same holds for any subset $\set{P}\subset\rel{}'$ of relevant sets.
\end{proof}

\subsection{Auxiliary Results on Abstract Targets}

The following four lemmas support the completeness of \rse{} (\Cref{lemma:steprse}). Throughout, we use two elementary facts. First, a non-relevant variable $X$ reaches a relevant variable through one of its children, and a relevant child $X_c\in\rel{Y}$ lies only in the block of $Y$, \ie, $\targets{X_c}{\block{}}=\setdef{Y}$. Hence, for every non-relevant $X$,
\begin{equation}
\label{eq:targetsunion}
    \targets{X}{\block{}}=\bigcup_{X_c\in\ch{X}}\targets{X_c}{\block{}}.
\end{equation}
Second, since the low-level graph is acyclic, the concatenation of directed paths is a directed path.

\begin{restatable}[Non-Empty New Relevant Variables]{lemma}{nonemptyzlemma}%
\label{lemma:nonemptyz}
    Let $\rel{}$ be a graphically consistent but not sufficient set of relevant variables on a low-level graph $\graph{G}_{\set{X}}$, and let $\set{V}_i\in\set{I}$ be a jointly targeted set of minimum size (\Cref{def:jointtargets}). Then, the set of new relevant variables $\set{Z}_i$ on $\set{V}_i$ is non-empty.
\end{restatable}
\begin{proof}
Throughout the proof, we write $\set{V}$ and $\set{Z}$ for $\set{V}_i$ and $\set{Z}_i$. Let $\set{S}=\setdef{X\in\set{X}\mid\targets{X}{\block{}}=\set{V}}$ be the low-level variables whose abstract targets are exactly $\set{V}$. The set $\set{S}$ is non-empty, since $\set{V}\in\set{I}$, and it is contained in $\bigcap_{Y\in\set{V}}\block{Y}$. Furthermore, since $\size{\set{V}}\geq 2$, the variables in $\set{S}$ lie in multiple blocks and are therefore not relevant.

First, there exists a variable $X\in\set{S}$ with a child outside of $\set{S}$. Indeed, let $X$ be the last variable of $\set{S}$ in a topological order of $\graph{G}_{\set{X}}$. Since $X$ is not relevant and has at least one abstract target, it has at least one child (\Cref{eq:targetsunion}), and no child of $X$ is in $\set{S}$.

Second, if $\set{V}$ has minimum size, any such child $X_c\notin\set{S}$ lies in a unique block. By \Cref{eq:targetsunion}, the abstract targets of $X_c$ are among those of its parent, \ie, $\targets{X_c}{\block{}}\subseteq\set{V}$, and, since $X_c\notin\set{S}$, they are a proper subset of $\set{V}$. If $X_c$ had at least two abstract targets, then $\targets{X_c}{\block{}}$ would be an element of $\set{I}$ smaller than $\set{V}$, contradicting the minimality of $\size{\set{V}}$. Hence, since the graph is clean (\Cref{assumption:clean}), $X_c$ has exactly one abstract target.

Therefore, $X$ lies in the blocks of all variables in $\set{V}$ and has a child with a unique abstract target, \ie, $X\in\set{Z}$ (\Cref{def:newrelevants}).
\end{proof}

The minimum size of $\set{V}_i$ is necessary. For instance, let $X_0\to X_1$, $X_0\to X_2$, $X_1\to X_{r_1}$, $X_1\to X_{r_2}$, $X_2\to X_{r_2}$, $X_2\to X_{r_3}$, with $\rel{Y_i}=\setdef{X_{r_i}}$. For $\set{V}_i=\setdef{Y_1,Y_2,Y_3}$, it holds $\bigcap_{Y\in\set{V}_i}\block{Y}=\setdef{X_0}$, whose children are both intersected, hence $\set{Z}_i=\emptyset$.

\begin{restatable}[Entering an Existing Abstract Variable]{lemma}{uniquechildlemma}%
\label{lemma:uniquechild}
    Let $\rel{}$ be a set of relevant variables on a low-level graph $\graph{G}_{\set{X}}$ and $\rel{}'\supseteq\rel{}$ a sufficient extension. Let $(X, X_1, \ldots, X_n)$ be a $\rel{}'$-direct path with $X_n\in\rel{Y}$ for some $Y\in\set{Y}$. Then, $\targets{X_1}{\block{}}=\setdef{Y}$.
\end{restatable}
\begin{proof}
If $X_1=X_n$, then $X_1\in\rel{Y}$ and the claim holds. Otherwise, $X_1$ is not relevant in $\rel{}'$, hence not relevant in $\rel{}$. The subpath $(X_1,\ldots,X_n)$ is $\rel{}'$-direct and consequently $\rel{}$-direct. Hence, $X_1\in\block{Y}'$ and $Y\in\targets{X_1}{\block{}}$. Assume by contradiction that $X_1$ also has a $\rel{}$-direct path to some $X_k\in\rel{Y_k}$ with $Y_k\neq Y$. Let $X''$ be the first variable after $X_1$ on this path that is relevant in $\rel{}'$, which exists since $X_k$ is relevant in $\rel{}'$, and let $W\in\set{Y}'$ be the abstract variable such that $X''\in\rel{W}'$. Note that $X''$ may be relevant to a new abstract variable, so that $W\neq Y_k$ in general. The subpath from $X_1$ to $X''$ is $\rel{}'$-direct, hence $X_1\in\block{W}'$. So, $X_1$ is in the blocks of both $W$ and $Y$ induced by $\rel{}'$. Due to the sufficiency of $\rel{}'$ the blocks are disjoint, hence $W=Y$ and $X''\in\rel{Y}'=\rel{Y}$. Then, $X''$ is relevant in $\rel{}$ and lies on a $\rel{}$-direct path, so it is its endpoint, \ie, $X''=X_k\in\rel{Y_k}$. Therefore, $X_k\in\rel{Y}\cap\rel{Y_k}$ with $Y\neq Y_k$, contradicting the disjointness of relevant sets. Therefore, $X_1$ cannot have a $\rel{}$-direct path to any other relevant set, and $\targets{X_1}{\block{}}=\setdef{Y}$.
\end{proof}

\begin{restatable}[Target Homogeneity]{lemma}{homogeneitylemma}%
\label{lemma:homogeneity}
    Let $\rel{}$ be a set of relevant variables on a low-level graph $\graph{G}_{\set{X}}$ and $\rel{}'\supseteq\rel{}$ a graphically consistent extension inducing the abstract variables $\set{Y}'$. Then, for every new abstract variable $W\in\set{Y}'\setminus\set{Y}$ and every $X,X'\in\rel{W}'$, it holds $\targets{X}{\block{}}=\targets{X'}{\block{}}$, where $\block{}$ are the blocks induced by $\rel{}$.
\end{restatable}
\begin{proof}
Intuitively, under $\rel{}$ a variable of $\rel{W}'$ reaches an existing abstract variable by hopping across variables that are relevant only in $\rel{}'$. Each hop corresponds to an edge of the abstract graph $\graph{G}_{\set{Y}'}$ and, by graphical consistency, every variable of a relevant set can perform the same hops. Hence, the targets under $\rel{}$ depend only on $W$.

Formally, we denote as $\set{D}_W$ the set of $Q\in\set{Y}$ such that $\graph{G}_{\set{Y}'}$ contains a directed walk $W=W_0\to W_1\to\cdots\to W_k\to Q$, with $k\geq 0$ and $W_1,\ldots,W_k\in\set{Y}'\setminus\set{Y}$. We show that $\targets{X}{\block{}}=\set{D}_W$ for every $X\in\rel{W}'$. We recall that variables that are relevant to a new abstract variable are not relevant in $\rel{}$, since relevant sets are disjoint.
\begin{itemize}
\item[($\subseteq$)] Let $Q\in\targets{X}{\block{}}$ and let $\vec{p}$ be a $\rel{}$-direct path from $X$ to $\rel{Q}$. Let $X=X_0, X_1,\ldots,X_k$ be the variables of $\vec{p}$ that are relevant in $\rel{}'$, excluding its endpoint, in order. Since $\vec{p}$ is $\rel{}$-direct, each $X_i$ is relevant to a new abstract variable $W_i$. The subpaths of $\vec{p}$ between consecutive variables are $\rel{}'$-direct. Hence, $W_i\to W_{i+1}\in\graph{G}_{\set{Y}'}$ whenever $W_i\neq W_{i+1}$, and $W_k\to Q\in\graph{G}_{\set{Y}'}$. Removing repeated consecutive variables yields a walk from $W$ to $Q$ through new abstract variables, hence $Q\in\set{D}_W$.
\item[($\supseteq$)] Let $W=W_0\to\cdots\to W_k\to Q$ be a walk as in the definition of $\set{D}_W$, and let $X_0=X$. By graphical consistency of $\rel{}'$ (\Cref{def:wellformed}), each $X_i\in\rel{W_i}'$ has a $\rel{}'$-direct path to some $X_{i+1}\in\rel{W_{i+1}}'$, and $X_k$ has a $\rel{}'$-direct path to some $X_r\in\rel{Q}'=\rel{Q}$. The concatenation of these paths is a directed path from $X$ to $X_r$ whose intermediate variables are either not relevant in $\rel{}'$ or relevant to a new abstract variable, and therefore not relevant in $\rel{}$. Hence, it is $\rel{}$-direct and $Q\in\targets{X}{\block{}}$.
\end{itemize}
Since $\set{D}_W$ depends only on $W$, all variables of $\rel{W}'$ have the same abstract targets under $\rel{}$.
\end{proof}

\begin{lemma}[Closure of New Relevant Variables]%
\label{lemma:newrelclosure}
    Let $\rel{}$ be a graphically consistent set of relevant variables on a low-level graph $\graph{G}_{\set{X}}$, let $\rel{}'\in\mise{\rel{}}{\graph{G}_{\set{X}}}$ inducing abstract variables $\set{Y}'\supseteq\set{Y}$, and let $\set{Z}_i$ be the new relevant variables on a jointly targeted set $\set{V}_i\in\set{I}$. Then, for every new abstract variable $W\in\set{Y}'\setminus\set{Y}$ such that $\rel{W}'\cap\set{Z}_i\neq\emptyset$, it holds $\rel{W}'\subseteq\set{Z}_i$.
\end{lemma}
\begin{proof}
Throughout the proof, we write $\set{V}$ and $\set{Z}$ for $\set{V}_i$ and $\set{Z}_i$.
Let $W\in\set{Y}'\setminus\set{Y}$ with a relevant variable $X\in\rel{W}'\cap\set{Z}$, and let $X'\in\rel{W}'$ be any other relevant variable of $W$. We show that $X'\in\set{Z}$. Intuitively, $X$ and $X'$ are relevant to the same abstract variable, and graphical consistency forces them to behave alike: they have the same abstract targets under $\rel{}$, and they reach the same abstract variables in $\rel{}'$. Hence, $X'$ satisfies, like $X$, both properties that define a new relevant variable (\Cref{def:newrelevants}).
\begin{enumerate}[label=(\roman*)]
\item \emph{$X'$ is in the blocks of all variables in $\set{V}$.} By target homogeneity (\Cref{lemma:homogeneity}), $X$ and $X'$ have the same abstract targets under $\rel{}$. Hence, $\targets{X'}{\block{}}=\targets{X}{\block{}}\supseteq\set{V}$.
\item \emph{$X'$ has a child with a unique abstract target.} Since $X\in\set{Z}$, it has a child $X_c$ whose unique abstract target is some $Y\in\set{Y}$. By target stability (\Cref{lemma:singletarget}), $X_c$ still leads only to $\rel{Y}$ in $\rel{}'$, so $X$ has a $\rel{}'$-direct path to $\rel{Y}$ through $X_c$, as shown in the proof of \Cref{lemma:newrelevants}. Since $W\neq Y$, this path is an edge $W\to Y$ of the abstract graph $\graph{G}_{\set{Y}'}$. Graphical consistency requires every relevant variable of $W$ to reach $\rel{Y}$, so $X'$ also has a $\rel{}'$-direct path to $\rel{Y}$. By \Cref{lemma:uniquechild}, the first step of this path is a child of $X'$ whose unique abstract target is $Y$.
\end{enumerate}
Therefore, $\rel{W}'\subseteq\set{Z}$.
\end{proof}

\subsection{Lemma~\ref{lemma:steprse}}

\begin{restatable}[Step-Completeness of \rse{}]{lemma}{steprselemma}%
\label{lemma:steprse}
    Let $\rel{}$ be a graphically consistent set of relevant variables on a low-level graph $\graph{G}_{\set{X}}$. Then, for any $\rel{}'\in\mise{\rel{}}{\graph{G}_{\set{X}}}$ such that $\rel{}'\neq\rel{}$, there exists a non-empty subset of relevant sets $\set{P}\subseteq\rel{}'\setminus\rel{}$ such that $\set{P}\in\partitions{\set{Z}_i}$, where $\set{Z}_i$ is the set of new relevant variables on the jointly targeted set $\set{V}_i$ selected by $\model{RSE}(\rel{}, \graph{G}_{\set{X}})$.
\end{restatable}
\begin{proof}
Throughout the proof, we write $\set{V}$ and $\set{Z}$ for $\set{V}_i$ and $\set{Z}_i$. At input $\rel{}$, \rse{} computes the new relevant variables $\set{Z}$ and calls itself on $\rel{}\cup\set{P}$ for every partition $\set{P}$ of $\set{Z}$. Hence, \rse{} can reach $\rel{}'$ only if one of these branches is compatible with $\rel{}'$, \ie, only if the new relevant sets of $\rel{}'$ that contain variables of $\set{Z}$ form a partition of $\set{Z}$. This requires two conditions:
\begin{enumerate}[label=(\alph*)]
\item every variable of $\set{Z}$ is relevant in $\rel{}'$ for some new abstract variable;
\item a new relevant set of $\rel{}'$ that contains a variable of $\set{Z}$ contains only variables of $\set{Z}$.
\end{enumerate}
Condition (a) is \Cref{lemma:newrelevants}, and condition (b) is \Cref{lemma:newrelclosure}.

\paragraph{Construction of $\set{P}$.}
Since $\rel{}\subset\rel{}'$, the minimality of $\rel{}'$ (\Cref{def:mise}) implies that $\rel{}$ is not sufficient. Hence, some low-level variable lies in at least two blocks of $\rel{}$, and the set $\set{I}$ of jointly targeted sets (\Cref{def:jointtargets}) is non-empty. Therefore, \rse{} does not return at Lines 6--8 of \Cref{alg:rse}, but selects a minimum-size jointly targeted set $\set{V}\in\set{I}$ at Line 9 and computes the new relevant variables $\set{Z}$ on $\set{V}$ at Line 10. By \Cref{lemma:nonemptyz}, $\set{Z}\neq\emptyset$. We collect the new relevant sets of $\rel{}'$ that contain at least one new relevant variable,
\begin{equation}
    \set{P}=\setdef{\rel{W}'\mid W\in\set{Y}'\setminus\set{Y},\ \rel{W}'\cap\set{Z}\neq\emptyset}\subseteq\rel{}'\setminus\rel{}.
\end{equation}
The elements of $\set{P}$ are pairwise disjoint, since relevant sets are disjoint. Their union contains $\set{Z}$ by condition (a), and $\set{P}$ is non-empty since $\set{Z}$ is. Hence, $\set{P}$ is a partition of $\set{Z}$ as soon as each of its elements is contained in $\set{Z}$. By \Cref{lemma:newrelclosure}, every element of $\set{P}$ is contained in $\set{Z}$, which is condition (b), and $\set{P}\in\partitions{\set{Z}}$.
\end{proof}

\subsection{Lemma~\ref{lemma:incmise}}

\begin{restatable}[\model{MiSE} Decomposition]{lemma}{incmiselemma}%
\label{lemma:incmise}
    Let $\rel{}$ be a graphically consistent
    but not sufficient set of relevant variables
    on a low-level graph $\graph{G}_{\set{X}}$.
    Then,
    for any minimal sufficient extension
    $\rel{}'$ of $\rel{}$
    and any set $\rel{}''$
    such that $\rel{}\subseteq\rel{}''\subset\rel{}'$, it holds that $\rel{}'\in\mise{\rel{}''}{\graph{G}_{\set{X}}}$.
\end{restatable}
\begin{proof}
We prove the result by contradiction. Suppose that there exists a set $\rel{}''$ such that $\rel{}\subseteq\rel{}''\subset\rel{}'$ but $\rel{}'\not\in\mise{\rel{}''}{\graph{G}_{\set{X}}}$. Then, $\rel{}'$ must be either (i.) not graphically consistent on $\graph{G}_{\set{X}}$, or (ii.) not sufficient on $\graph{G}_{\set{X}}$, or (iii.) not minimal with respect to $\rel{}''$ on $\graph{G}_{\set{X}}$. The first two cases are impossible since $\rel{}'$ is in the \model{MiSE} of $\rel{}$ on $\graph{G}_{\set{X}}$, and therefore it is graphically consistent and sufficient. The third case is also impossible since if $\rel{}'$ were not minimal with respect to $\rel{}''$, there would exist a set $\rel{}'''$ such that $\rel{}''\subseteq\rel{}'''\subset\rel{}'$ that is sufficient and graphically consistent on $\graph{G}_{\set{X}}$. However, since $\rel{}\subseteq\rel{}'''\subset\rel{}'$, this exhibits a sufficient and graphically consistent set strictly between $\rel{}$ and $\rel{}'$, contradicting the fact that $\rel{}'$ is minimal with respect to $\rel{}$, \ie, that $\rel{}'\in\mise{\rel{}}{\graph{G}_{\set{X}}}$.
\end{proof}

\newpage
\section{Additional Experimental Details}%

\subsection{Data Generation}%
\label{app:datagen}

Given an abstract graph $\graph{G}_{\set{Y}}$, we generate a ground truth in four steps (\Cref{alg:datagen}). In all of our experiments, we control the relevant set size $\size{\rel{Y}}$ and the block size $\size{\block{Y}}$ to study how they affect the performance of \model{dataRSE}. All the abstract variables $Y$ share the same relevant set size $\size{\rel{Y}}$ and block size $\size{\block{Y}}$. First, we assign to each abstract variable $Y$ a block $\block{Y}$ of low-level variables, a subset of which we promote to the relevant set $\rel{Y}\subset\block{Y}$. We also add a block of ignored low-level variables, not belonging to any abstract variable.

Second, we sample the low-level graph $\graph{G}_{\set{X}}$. Within each block, we start from a complete DAG and remove random edges until at most $\rho_e\size{\block{Y}}$ edges remain, where $\rho_e$ is the \emph{edge ratio}. We only remove an edge if every non-relevant variable of the block stays connected to a relevant one, and we stop after $100$ failed removals. For each abstract edge $Y_i\to Y_j$, every relevant variable of $Y_i$ gets an edge towards a random variable of $\block{Y_j}$. We then add random edges from $\rel{Y_i}$ to $\block{Y_j}$ until their number reaches $\lfloor\rho_o\size{\block{Y_j}}\rfloor$, where $\rho_o$ is the \emph{output ratio}. Finally, each block sends $\lfloor\rho_o\size{\block{Y}}\rfloor$ random edges to the ignored block, whose variables are connected by a random DAG.

Third, we sample the abstract coefficients $\mat{M}$ from the uniform distribution over the set $(-3.0, -0.5)\cup(0.5,3.0)$. Then, for each abstract variable $Y$, we sample a linear abstraction function $\vec{t}_Y\in\real^{\size{\rel{Y}}}$. We sample it either from the same uniform distribution or from the normal distribution $\mathcal{N}(0,1)$ normalized to unit norm. In both cases, we resample it until all its entries are at least $0.1$ in absolute value.

Finally, given $\graph{G}_{\set{X}}$, $\mat{M}$, and $\mat{T}$, we apply the $\mat{T}$-concretization sampling algorithm~\citep{massidda2024learning} to sample the coefficients $\mat{W}$ of the low-level ANM~$\scm{L}$. We then draw the low-level samples $\mat{X}$ by sampling the exogenous terms of $\scm{L}$ from $\mathcal{N}(0,1)$ and propagating them through its structural equations.

\begin{algorithm}%
\caption{Synthetic Data Generation}%
\label{alg:datagen}
\begin{algorithmic}[1]
\Require
\Statex Abstract graph~$\graph{G}_{\set{Y}}$ with treatment~$Y_T$ and outcome~$Y_O$, number of low-level samples~$n$
\Statex Relevant set size~$\size{\rel{Y}}$, block size~$\size{\block{Y}}$, edge ratio~$\rho_e$, and output ratio~$\rho_o$
\Statex Maximum similarity threshold~$\gamma$, normalized abstraction flag~$\texttt{normalized}$
\Ensure
\Statex Low-level samples~$\mat{X}$ and graph~$\graph{G}_{\set{X}}$
\Statex Relevant sets~$\rel{Y_T},\rel{Y_O}$ and abstraction coefficients~$\vec{t}_T,\vec{t}_O$
\Repeat
\Comment{Resample the whole ground truth}
    \State $\block{}, \rel{} \gets$ Sample blocks of size $\size{\block{Y}}$, an ignored block, and relevant sets of size $\size{\rel{Y}}$
    \State $\graph{G}_{\set{X}} \gets$ Sample intra-block edges with ratio $\rho_e$ and inter-block edges with ratio $\rho_o$
    \State $\mat{M} \gets$ Sample $\graph{G}_{\set{Y}}$ coefficients from $\mathcal{U}\left((-3.0,-0.5)\cup(0.5,3.0)\right)$
    \For{$Y\in\graph{G}_{\set{Y}}$}
    \Comment{Iterate over all abstract variables}
        \Repeat
            \If{not $\texttt{normalized}$}
                \State $\vec{t}_Y \gets$ Sample from $\mathcal{U}\left((-3.0,-0.5)\cup(0.5,3.0)\right)^{\size{\rel{Y}}}$
                \Comment{Uniform coefficients}
            \Else
                \State $\vec{t}_Y \gets$ Sample from $\mathcal{N}(0,1)^{\size{\rel{Y}}}$
                \State $\vec{t}_Y \gets \vec{t}_Y / \norm{\vec{t}_Y}$
                \Comment{Normalized Gaussian coefficients}
            \EndIf
        \Until{every entry of $\vec{t}_Y$ is at least $0.1$ in absolute value}
    \EndFor
    \State $\mat{T} \gets [\vec{t}_Y]_{Y\in\graph{G}_{\set{Y}}}$
    \Comment{Collect abstraction functions}
    \State $\mat{W} \gets \operatorname{Concretize}(\graph{G}_{\set{X}}, \mat{M}, \mat{T})$
    \Comment{$\mat{T}$-concretization sampling~\citep{massidda2024learning}}
\Until{$\operatorname{MaxSim}(\mat{M}, \graph{G}_{\set{Y}}) < \gamma$}
\Comment{Enforce the condition of \Cref{thm:maximalpartitioning}}
\State $\scm{L} \gets$ Low-level ANM with graph $\graph{G}_{\set{X}}$ and coefficients $\mat{W}$
\State $\mat{E} \gets$ Sample $n$ i.i.d. realizations from $\mathcal{N}(0,1)$
\Comment{Gaussian exogenous noise}
\State $\mat{X} \gets \scm{L}(\mat{E})$
\Comment{Generate low-level endogenous samples}
\State \Return $\mat{X}, \graph{G}_{\set{X}}, \rel{Y_T}, \rel{Y_O}, \vec{t}_T, \vec{t}_O$
\end{algorithmic}
\end{algorithm}

\paragraph{Enforcing the Condition of \Cref{thm:maximalpartitioning}.} \Cref{thm:maximalpartitioning} requires a non-collinearity condition on the outgoing coefficients of the new abstract variables (\Cref{eq:mpcondition}). For each new abstract variable whose children are all confounded, its outgoing coefficients towards the confounded variables must not be collinear with those of any other new abstract variable. Since \model{dataRSE} recursively extends the set of relevant variables, this condition must hold at every recursive call, not only at the first one. Two coefficient rows sampled independently from a continuous distribution are collinear with probability zero. However, near-collinear rows still yield an ill-conditioned composed abstraction matrix~$\hat{\mat{A}}$ and degrade the accuracy of \model{dataRSE}. Therefore, we enforce the condition with an explicit margin $\gamma$ on a similarity score $\operatorname{MaxSim}(\mat{M}, \graph{G}_{\set{Y}})$, which we define below.

We compute $\operatorname{MaxSim}(\mat{M}, \graph{G}_{\set{Y}})$ by mirroring the recursion of \model{dataRSE} over $\graph{G}_{\set{Y}}$. Hence, it makes no assumption on the structure of $\graph{G}_{\set{Y}}$, e.g., it allows skip connections. We initialize the set of known abstract variables as $\set{K}\gets\setdef{Y_T, Y_O}$. At every step, the targets of an unknown abstract variable $Y\notin\set{K}$ are the known variables reachable from $Y$ through directed paths whose intermediate nodes are all unknown. They are the abstract counterpart of the abstract targets of its relevant variables. Then, we compute:
\begin{itemize}
    \item the confounded variables $\set{V}$, \ie, the union of the targets of the unknown variables with at least two targets;
    \item the new parents $\set{H}$, \ie, the unknown variables with at least two targets and at least one child in $\set{K}$, whose relevant variables are the new relevant variables $\set{Z}$;
    \item the variables $\set{N}_{\set{V}}=\setdef{Y\in\set{H}\mid\ch{Y,\graph{G}_{\set{Y}}}\subseteq\set{V}}$ whose children are all confounded.
\end{itemize}
For every $Y_s\in\set{N}_{\set{V}}$ and every other $Y_h\in\set{H}$, we compute the absolute cosine similarity between the outgoing coefficient rows of $\mat{M}$ restricted to the columns of $\set{V}$, \ie,
\begin{equation*}
    \operatorname{sim}(Y_s, Y_h)
    =
    \left|
        \frac{\vec{m}_{s,\set{V}}}{\norm{\vec{m}_{s,\set{V}}}}
        \cdot
        \frac{\vec{m}_{h,\set{V}}}{\norm{\vec{m}_{h,\set{V}}}}
    \right|.
\end{equation*}
By \Cref{prop:rankone}, this similarity coincides with the absolute cosine similarity between the corresponding rows of the composed abstraction matrix~$\mat{A}$. We do not check pairs of new parents that are both outside $\set{N}_{\set{V}}$. Indeed, \model{dataRSE} discards their merges and resolves them at a later step, where they are checked as members of $\set{N}_{\set{V}}$. Then, as in \model{dataRSE}, we only add the variables in $\set{N}_{\set{V}}$ to the known variables, \ie, $\set{K}\gets\set{K}\cup\set{N}_{\set{V}}$. We repeat until no unknown variable has at least two targets or $\set{N}_{\set{V}}$ is empty. Finally, we define $\operatorname{MaxSim}(\mat{M}, \graph{G}_{\set{Y}})$ as the maximum similarity over the checked pairs at every recursion step.

Whenever $\operatorname{MaxSim}(\mat{M}, \graph{G}_{\set{Y}}) \geq \gamma$, we reject the sampled ground truth and resample it, up to $100$ times. Although $\operatorname{MaxSim}$ only depends on $\mat{M}$, each trial repeats all four steps above. If no ground truth satisfies the threshold within this budget, we keep the last sampled one and report a warning. We did not encounter this fallback in the reported configurations.

Since abstraction vectors are only determined up to scale (\Cref{prop:rankone}), we fix their scale. With the $\texttt{normalized}$ flag set to true, every abstraction vector $\vec{t}_Y$ has unit norm. Otherwise, the norm of $\vec{t}_Y$ grows with the size of its relevant set. In both cases, the variance of each abstract variable still depends on the covariance of its relevant variables.

We report the results in the main paper with the $\texttt{normalized}$ flag set to true and $\gamma=0.9$. In \Cref{app:additional-experiments}, we additionally report results with the $\texttt{normalized}$ flag set to false, which are less accurate with few samples and converge to the same accuracy as the number of samples grows. We also report results without rejection sampling, which we denote as $\gamma=1.00$.

\subsection{Estimating the Maximal Partitioning}%
\label{app:mpestimation}

\Cref{alg:maximalpartitioning} estimates the maximal partitioning of the new relevant variables $\set{Z}$ from the estimated composed abstraction matrix $\hat{\mat{A}}$ (\Cref{alg:datarse}, Line 22). It normalizes the rows of $\hat{\mat{A}}$ and computes their absolute cosine distance, which is zero between collinear rows (Lines 1--2). Then, for each $k\in\setdef{1,\ldots,p}$, with $p=\size{\set{Z}}$, it clusters the rows into $k$ clusters $\set{P}_1,\ldots,\set{P}_k$ (Line 4). To this end, it uses agglomerative clustering with complete linkage \citep{hastie2009elements,mullner2011modern}, as implemented in scikit-learn \citep{pedregosa2011scikit}. Each cluster $\set{P}_h$ induces a new abstract variable with abstraction vector $\vec{t}_h\in\real^{\size{\set{P}_h}}$. We initialize $\vec{t}_h$ with the first left singular vector of the submatrix $\hat{\mat{A}}_{\set{P}_h,:}$ (Line 6, \Cref{prop:rankone}).

\begin{algorithm}%
\caption{Maximal Partitioning}%
\label{alg:maximalpartitioning}
\begin{algorithmic}[1]
\Require Composed abstraction matrix~$\hat{\mat{A}}\in\real^{p\times q}$, realizations of the new relevant variables~$\mat{X}_{\set{Z}}\in\real^{n\times p}$, realizations of the known parents~$\mat{Y}_{\set{K}}\in\real^{n\times\size{\set{K}}}$, realizations of the confounded variables~$\mat{Y}_{\set{V}}\in\real^{n\times q}$, new relevant variables targeting each confounded variable~$\setdef{\set{Z}^{(j)}}_{Y_j\in\set{V}}$, abstract graph~$\graph{G}_{\set{Y}}$
\Ensure Partition $\set{P}=\setdef{\set{P}_1,\ldots,\set{P}_{k^\ast}}$ of $\set{Z}$ and abstraction coefficients $\mat{T}\in\real^{p\times k^\ast}$ of the induced variables
\State $\bar{\mat{A}} \gets \operatorname{RowNormalize}(\hat{\mat{A}})$
\Comment{Unit-norm rows}
\State $\mat{D} \gets \mat{1} - \left|\bar{\mat{A}}\bar{\mat{A}}^\tr\right|\in\real^{p\times p}$
\Comment{Distance matrix of the $p$ rows}
\For {$k\in\setdef{1, \ldots, p}$}
\Comment{Iterate over an \emph{increasing} number of clusters}
    \State $\set{P}^{(k)} \gets \operatorname{AgglomerativeClustering}(\mat{D}, k)$
    \Comment{$k$ clusters $\set{P}_1,\ldots,\set{P}_k$}
    \For {$h\in\setdef{1, \ldots, k}$}
        \State $\vec{t}_h \gets$ first left singular vector of $\hat{\mat{A}}_{\set{P}_h,:}$
        \Comment{Rank-one initialization (\Cref{prop:rankone})}
    \EndFor
    \State $\setdef{\vec{t}_h}_{h=1}^k, \vec{\sigma}^{2}, d^{(k)} \gets \operatorname{Fit}(\set{P}^{(k)}, \setdef{\vec{t}_h}_{h=1}^k)$
    \Comment{\Cref{alg:reducedrank}}
    \State $\operatorname{BIC}^{(k)} \gets n\sum_{j=1}^{q}\log\sigma^2_j + \log(n)\, d^{(k)}$
    \Comment{$d^{(k)}$: free parameters of the $k$ variables}
    \State $\mat{T}^{(k)} \gets [\vec{t}_1\mid\ldots\mid\vec{t}_k]\in\real^{p\times k}$
    \Comment{$\vec{t}_h$ padded with zeros outside $\set{P}_h$}
\EndFor
\State $k^\ast \gets \arg\min_{k} \operatorname{BIC}^{(k)}$
\State \Return $\left(\set{P}^{(k^\ast)}, \mat{T}^{(k^\ast)}\right)$
\end{algorithmic}
\end{algorithm}

Therefore, for each candidate partition $\set{P}^{(k)}$, \Cref{alg:maximalpartitioning} fits the model
\begin{equation*}
    Y_j = \sum_{Y_i\in\pa{Y_j,\graph{G}_{\set{Y}}}} m_{ij} Y_i + \sum_{h=1}^{k} m_{hj}\, \vec{x}_{\set{P}_h}\vec{t}_h + U_j,
    \qquad Y_j\in\set{V},
\end{equation*}
where $\vec{x}_{\set{P}_h}$ is the row vector of the low-level variables in $\set{P}_h$, and $m_{hj}=0$ whenever no variable in $\set{P}_h$ targets $Y_j$, \ie, whenever $\set{P}_h\cap\set{Z}^{(j)}=\emptyset$ (Line 8). The coefficients of the variables in $\set{P}_h$ on the confounded variables factor as $\vec{t}_h\vec{m}_{h,\set{V}}^\top$. Hence, the fit is a reduced-rank regression with a rank-one constraint on the block of each cluster, a fixed zero pattern, and unconstrained parent coefficients.

Each abstraction vector $\vec{t}_h$ is shared by all the regressions in which its induced variable appears. Therefore, \Cref{alg:reducedrank} alternates between two steps, starting from the rank-one initialization. First, it regresses each $Y_j$ on its parents and on the realizations $\mat{X}_{\set{P}_h}\vec{t}_h$ of the induced variables by least squares (Lines 3--6). Second, it updates all the abstraction vectors $\vec{t}_h$ by weighted least squares (Lines 10--11). The weights $1/\sigma_j^2$ are given by the residual variance of each $Y_j$, so that the update approximates the maximum-likelihood fit. It runs for at most $50$ iterations, or until the relative improvement of $\sum_j\log\sigma_j^2$ falls below $10^{-8}$ (Lines 7--9).%

\begin{algorithm}%
\caption{Fit of a Candidate Partition}%
\label{alg:reducedrank}
\begin{algorithmic}[1]
\Require Partition $\set{P}=\setdef{\set{P}_1,\ldots,\set{P}_k}$ of $\set{Z}$, initial abstraction vectors $\setdef{\vec{t}_h}_{h=1}^k$, realizations of the new relevant variables~$\mat{X}_{\set{Z}}$, of the known parents~$\mat{Y}_{\set{K}}$, and of the confounded variables~$\mat{Y}_{\set{V}}$, new relevant variables targeting each confounded variable~$\setdef{\set{Z}^{(j)}}_{Y_j\in\set{V}}$, abstract graph~$\graph{G}_{\set{Y}}$
\Ensure Fitted abstraction vectors $\setdef{\vec{t}_h}_{h=1}^k$, residual variances $\vec{\sigma}^{2}$, and number of free parameters $d$
\State $\set{H}_j \gets \setdef{h \mid \set{P}_h\cap\set{Z}^{(j)}\neq\emptyset}$ for every $Y_j\in\set{V}$
\Comment{Clusters targeting $Y_j$}
\For{$i\in\setdef{1,\ldots,50}$}
    \For{$Y_j\in\set{V}$}
    \Comment{Regression step}
        \State $\setdef{m_{hj}}_{h\in\set{H}_j}, \vec{m}_{\pa{Y_j}} \gets \operatorname{LeastSquares}\big(\vec{y}_j \sim [\mat{X}_{\set{P}_h}\vec{t}_h]_{h\in\set{H}_j} \mid \mat{Y}_{\pa{Y_j}}\big)$
        \State $\sigma_j^2 \gets$ mean squared residual of $\vec{y}_j$
    \EndFor
    \If{$i>1$ and the relative improvement of $\sum_{j}\log\sigma_j^2$ is below $10^{-8}$}
        \State \textbf{break}
    \EndIf
    \State $\setdef{\vec{t}_h}_{h=1}^k \gets \arg\min \sum_{Y_j\in\set{V}} \frac{1}{\sigma_j^2}\norm{\vec{y}_j - \mat{Y}_{\pa{Y_j}}\vec{m}_{\pa{Y_j}} - \sum_{h\in\set{H}_j} m_{hj}\mat{X}_{\set{P}_h}\vec{t}_h}^2$
    \Comment{Abstraction step}
    \State $\vec{t}_h \gets \vec{t}_h/\norm{\vec{t}_h}$ for every $h\in\setdef{1,\ldots,k}$
\EndFor
\State Flip the sign of each $\vec{t}_h$ so that its largest entry in absolute value is positive
\State $d \gets \sum_{h=1}^{k}\left(\size{\set{P}_h}-1\right) + \sum_{Y_j\in\set{V}}\size{\set{H}_j}$
\Comment{Free parameters of the $k$ induced variables}
\State \Return $\setdef{\vec{t}_h}_{h=1}^k, \vec{\sigma}^{2}, d$
\end{algorithmic}
\end{algorithm}

Finally, \Cref{alg:maximalpartitioning} in lines 9 and 12 selects the partition minimizing
\begin{equation*}
    \operatorname{BIC}^{(k)}=n\sum_{j}\log\sigma^2_j+\log(n)\,d^{(k)},
    \qquad
    d^{(k)}=\sum_{h=1}^{k}\left(\size{\set{P}_h}-1\right)+\size{\setdef{(h,j)\mid\set{P}_h\cap\set{Z}^{(j)}\neq\emptyset}}.
\end{equation*}
Here, $d^{(k)}$ counts the free parameters of the $k$ induced variables, as computed by \Cref{alg:reducedrank} (Line 14). Each abstraction vector $\vec{t}_h$ has $\size{\set{P}_h}$ entries, one of which is fixed by the scale shared with its outgoing coefficients $m_{hj}$. The parent coefficients are not counted, since their number does not depend on $k$. Finally, \Cref{alg:maximalpartitioning} returns the fitted abstraction vectors of the selected partition (Line 13).

\subsection{Maximal Partitioning with Rank-One Factors}%
\label{app:rankonepartitioning}

In preliminary versions of this work, we experimented with a variant of \Cref{alg:maximalpartitioning} that selects the number of clusters $k^\ast$ without fitting the candidate models. We refer to it as rank-based partitioning (\Cref{alg:maximalpartitioning-plateau}). For each $k$, rank-based partitioning computes the residuals of the confounded variables directly from the rank-one factors of the blocks of $\hat{\mat{A}}$. To this end, it subtracts two terms from $\mat{Y}_{\set{V}}$. The first is the effect of the known parents, through the parent coefficients $\hat{\vec{m}}_j$ fitted jointly with $\hat{\mat{A}}$ (\Cref{alg:datarse}, Line 19) and stacked into $\hat{\mat{M}}$. The second is the effect of the induced abstract variables, through $\vec{t}_h\vec{m}_{h,\set{V}}^\top$. Then, it evaluates the log-likelihood of the residuals, which we denote as $\ell^{(k)}$. It selects the first $k$ of the last run of consecutive values of $k$ whose improvement $\left|\ell^{(k+1)}-\ell^{(k)}\right|$ is below a threshold $\delta$, which we set to $\delta=0.25$. If there is no such run, it selects $k^\ast=p$.

Since it avoids fitting each candidate model, rank-based partitioning is faster than \Cref{alg:maximalpartitioning}, but we found it to be less sample-efficient in practice, as we show in \Cref{app:plateau}.%

\begin{algorithm}%
\caption{Rank-Based Partitioning}%
\label{alg:maximalpartitioning-plateau}
\begin{algorithmic}[1]
\Require Composed abstraction matrix~$\hat{\mat{A}}\in\real^{p\times q}$, parent coefficients~$\hat{\mat{M}}\in\real^{(\size{\set{K}}+q)\times q}$ stacking the $\hat{\vec{m}}_j$ of \Cref{alg:datarse}, with zeros for non-parents, realizations of the new relevant variables~$\mat{X}_{\set{Z}}\in\real^{n\times p}$, realizations of the known parents~$\mat{Y}_{\set{K}}\in\real^{n\times\size{\set{K}}}$, realizations of the confounded variables~$\mat{Y}_{\set{V}}\in\real^{n\times q}$, threshold~$\delta$
\Ensure Partition $\set{P}=\setdef{\set{P}_1,\ldots,\set{P}_{k^\ast}}$ of $\set{Z}$ and abstraction coefficients $\mat{T}\in\real^{p\times k^\ast}$ of the induced variables
\State $\bar{\mat{A}} \gets \operatorname{RowNormalize}(\hat{\mat{A}})$
\Comment{Unit-norm rows}
\State $\mat{D} \gets \mat{1} - \left|\bar{\mat{A}}\bar{\mat{A}}^\tr\right|\in\real^{p\times p}$
\Comment{Distance matrix of the $p$ rows}
\For {$k\in\setdef{1, \ldots, p}$}
\Comment{Iterate over an \emph{increasing} number of clusters}
    \State $\set{P}^{(k)} \gets \operatorname{AgglomerativeClustering}(\mat{D}, k)$
    \Comment{$k$ clusters $\set{P}_1,\ldots,\set{P}_k$}
    \For {$h\in\setdef{1, \ldots, k}$}
        \State $\vec{u}_1, \sigma_1, \vec{v}_{1} \gets \operatorname{SVD}(\hat{\mat{A}}_{\set{P}_h,:})$
        \Comment{First singular triplet}
        \State $\vec{t}_h \gets \vec{u}_1$, $\vec{m}_{h,\set{V}} \gets \sigma_1 \vec{v}_{1}$
        \Comment{Rank-one factors (\Cref{prop:rankone})}
    \EndFor
    \State $\mat{U} \gets \mat{Y}_{\set{V}} - [\mat{Y}_{\set{K}}\mid\mat{Y}_{\set{V}}]\hat{\mat{M}} - \sum_{h=1}^{k}\mat{X}_{\set{P}_h} \vec{t}_h\vec{m}_{h,\set{V}}^\top$
    \Comment{Residuals of the rank-one model}
    \State $\ell^{(k)} \gets -2\sum_{j=1}^{q}\log\left(\frac{1}{n}\norm{\mat{U}_{:,j}}^2\right)$
    \Comment{Log-likelihood, rescaled by $4/n$, up to constants}
    \State $\mat{T}^{(k)} \gets [\vec{t}_1\mid\ldots\mid\vec{t}_k]\in\real^{p\times k}$
    \Comment{$\vec{t}_h$ padded with zeros outside $\set{P}_h$}
\EndFor
\State $k^\ast \gets$ first $k$ of the last run of consecutive $k$ with $\left|\ell^{(k+1)}-\ell^{(k)}\right|<\delta$, or $p$ if none
\State \Return $\left(\set{P}^{(k^\ast)}, \mat{T}^{(k^\ast)}\right)$
\end{algorithmic}
\end{algorithm}

\subsection{Hyperparameters}%
\label{app:hyperparams}

Unless stated otherwise, every experiment in \Cref{sec:experiments} uses the main configuration of \Cref{app:datagen}. That is, it uses the collinearity margin $\gamma=0.9$, the $\texttt{normalized}$ flag set to true, an edge ratio $\rho_e=0.5$, and an output ratio $\rho_o=0.2$. Each configuration is repeated $15$ times, each with an independently sampled ground truth and a fresh draw of exogenous noise. We sweep the sample size, the relevant set size, and the block size over
\begin{gather*}
    n\in\{1000, 2000, 5000, 10000, 20000, 50000, 100000\},\\
    \size{\rel{Y}}\in\{1,2,3,4,5\},\quad
    \size{\block{Y}}\in\{2,4,6,8,10,15\},
\end{gather*}
restricted to the pairs with $\size{\block{Y}}>\size{\rel{Y}}$.%

When recovering the composed abstraction matrix $\hat{\mat{A}}$, \model{dataRSE} runs on mean-centered data, with ridge regularization $\lambda=10^{-5}$ on standardized regressors (\Cref{alg:datarse}, Lines 17--21). At every recursive call, we only use the single candidate partition $\set{P}^{(k^\ast)}$ selected by \Cref{alg:maximalpartitioning}. We keep its unmediated relevant sets, \ie, the clusters whose induced variables only have children in $\set{V}$ (\Cref{alg:datarse}, Lines 23--24).

\subsection{Matching of the Abstraction Coefficients}%
\label{app:matchabs}

To report the recovery of the abstraction function (\Cref{fig:tau_recovery}), we first align the learned abstract variables with the ground-truth ones in each run. If the recovered graph $\hat{\graph{G}}_{\set{Y}}$ has more abstract variables than the ground-truth graph $\graph{G}_{\set{Y}}$, we keep the $\size{\set{Y}}$ learned variables with the highest similarity to some ground-truth variable. Otherwise, if it has fewer abstract variables, we pad the learned abstraction coefficients with zero vectors, which always have zero similarity. In particular, whenever \model{dataRSE} returns no solution, we only keep the provided abstraction coefficients of the treatment and the outcome. Then, we solve the linear sum assignment problem~\citep{jonker1987shortest} over the resulting square similarity matrix $\mat{S}=[s_{ij}]_{i,j}$. This gives the optimal matching between ground-truth and learned variables. We permute the learned variables so that matched pairs lie on the diagonal. Finally, we average the aligned similarity matrices across runs. Hence, the diagonal reports the average similarity of each ground-truth variable with its matched learned variable. Instead, the off-diagonal entries report the average similarity of unmatched pairs.

\newpage
\section{Additional Experiments}%
\label{app:additional-experiments}

In this section, we discuss additional results on different data generation settings (\Cref{app:additional-experiments-gamma}), additional abstract graphs (\Cref{app:additional-experiments-graphs}), and strategies to estimate the maximal partitioning (\Cref{app:plateau}). We summarize in \Cref{tab:app-figure-configs} the configuration of every figure in \Cref{sec:experiments} of the main body and in this appendix.

\begin{table}[htb]
\centering
\small
\caption{Configuration of every figure. All results are computed over $15$ random seeds. In the last column, ``All'' means that results are averaged over every configuration of the relevant set size $\size{\rel{Y}}$ and block size $\size{\block{Y}}$, ``Grid'' means that each configuration is reported separately, and $\{2,4\}\times\{4,8\}$ means that results are averaged over $\size{\rel{Y}}\in\{2,4\}$ and $\size{\block{Y}}\in\{4,8\}$. Metrics: accuracy (Acc.), running time (Time), absolute TCE error (TCE), failure rate (Fail.), and cosine similarity of the abstraction coefficients (Cos.).}
\label{tab:app-figure-configs}
\resizebox{\textwidth}{!}{%
\begin{tabular}{lllcccc}
\hline
Figure & Metrics & Graphs & $\gamma$ & $\texttt{normalized}$ & $n$ & $(\size{\rel{Y}},\size{\block{Y}})$\\
\hline
\multicolumn{7}{l}{\emph{Main results (\Cref{sec:experiments})}}\\
Fig.~\ref{fig:n_solutions-rse} & \model{MiSE} size & A--D & -- & -- & -- & Grid\\
Fig.~\ref{fig:size_data-rse} & Acc., Time & A--D & $0.9$ & true & $2\cdot10^4$ & Grid\\
Fig.~\ref{fig:samples_data-rse} & Acc., TCE & A--D & $0.9$ & true & $10^3$--$10^5$ & All\\
Fig.~\ref{fig:tau_recovery} & Cos. & A--D & $0.9$ & true & $5\cdot10^3$ & All\\
\hline
\multicolumn{7}{l}{\emph{Collinearity margin and normalization (\Cref{app:additional-experiments-gamma})}}\\
Fig.~\ref{fig:app-ms90} & Acc., Time, TCE, Fail. & A--D & $0.9$ & true & $10^3$--$10^5$ & All\\
Fig.~\ref{fig:app-ms90-combo} & Acc., Time & A--D & $0.9$ & true & $10^3$, $5\cdot10^3$, $10^4$ & Grid\\
Fig.~\ref{fig:app-ms90-ntn} & Acc., Time, TCE, Fail. & A--D & $0.9$ & false & $10^3$--$10^5$ & All\\
Fig.~\ref{fig:app-ms100} & Acc., Time, TCE, Fail. & A--D & $1.00$ & true & $10^3$--$10^5$ & All\\
Fig.~\ref{fig:app-ms100-combo} & Acc., Time & A--D & $1.00$ & true & $10^3$, $5\cdot10^3$, $10^4$ & Grid\\
Fig.~\ref{fig:app-tau-ms100} & Cos. & A--D & $1.00$ & true & $5\cdot10^3$ & All\\
Fig.~\ref{fig:app-tau-split-ms90} & Cos. & A--D & $0.9$ & true & $5\cdot10^3$ & All\\
Fig.~\ref{fig:app-tau-split-ms100} & Cos. & A--D & $1.00$ & true & $5\cdot10^3$ & All\\
\hline
\multicolumn{7}{l}{\emph{Additional graphical structures (\Cref{app:additional-experiments-graphs})}}\\
Fig.~\ref{fig:app-ijk} & Acc., TCE & E--H & $0.9$ & true & $10^3$--$10^5$ & All\\
Fig.~\ref{fig:app-ijk-full} & Acc., Time, TCE, Fail. & E--H & $0.9$ & true & $10^3$--$10^5$ & All\\
Fig.~\ref{fig:app-ijk-tau} & Cos. & E--H & $0.9$ & true & $5\cdot10^3$ & All\\
\hline
\multicolumn{7}{l}{\emph{Comparison with rank-based partitioning (\Cref{app:plateau})}}\\
Fig.~\ref{fig:app-compare-rule-abcd} & Acc., TCE, Time & A--D & $0.9$ & true & $10^3$--$10^5$ & $\{2,4\}\times\{4,8\}$\\
Fig.~\ref{fig:app-compare-rule} & Acc., TCE, Time & E--H & $0.9$ & true & $10^3$--$10^5$ & $\{2,4\}\times\{4,8\}$\\
\hline
\end{tabular}%
}
\end{table}

\subsection{Collinearity Margin and Normalization}%
\label{app:additional-experiments-gamma}

The main results in \Cref{sec:experiments} are computed with the $\texttt{normalized}$ flag set to true and collinearity margin $\gamma=0.9$. We first report full results of this main configuration, including the running time and failure rate omitted from \Cref{fig:samples_data-rse} in the main body. We then report the same experiments with the $\texttt{normalized}$ flag set to false, \ie, with abstraction coefficients sampled uniformly from $(-3.0,-0.5)\cup(0.5,3.0)$, and without the collinearity margin, \ie, with $\gamma=1.00$, which disables the rejection sampling check of \Cref{app:datagen}.

\begin{figure}[htb]
\centering
\includegraphics[width=\textwidth]{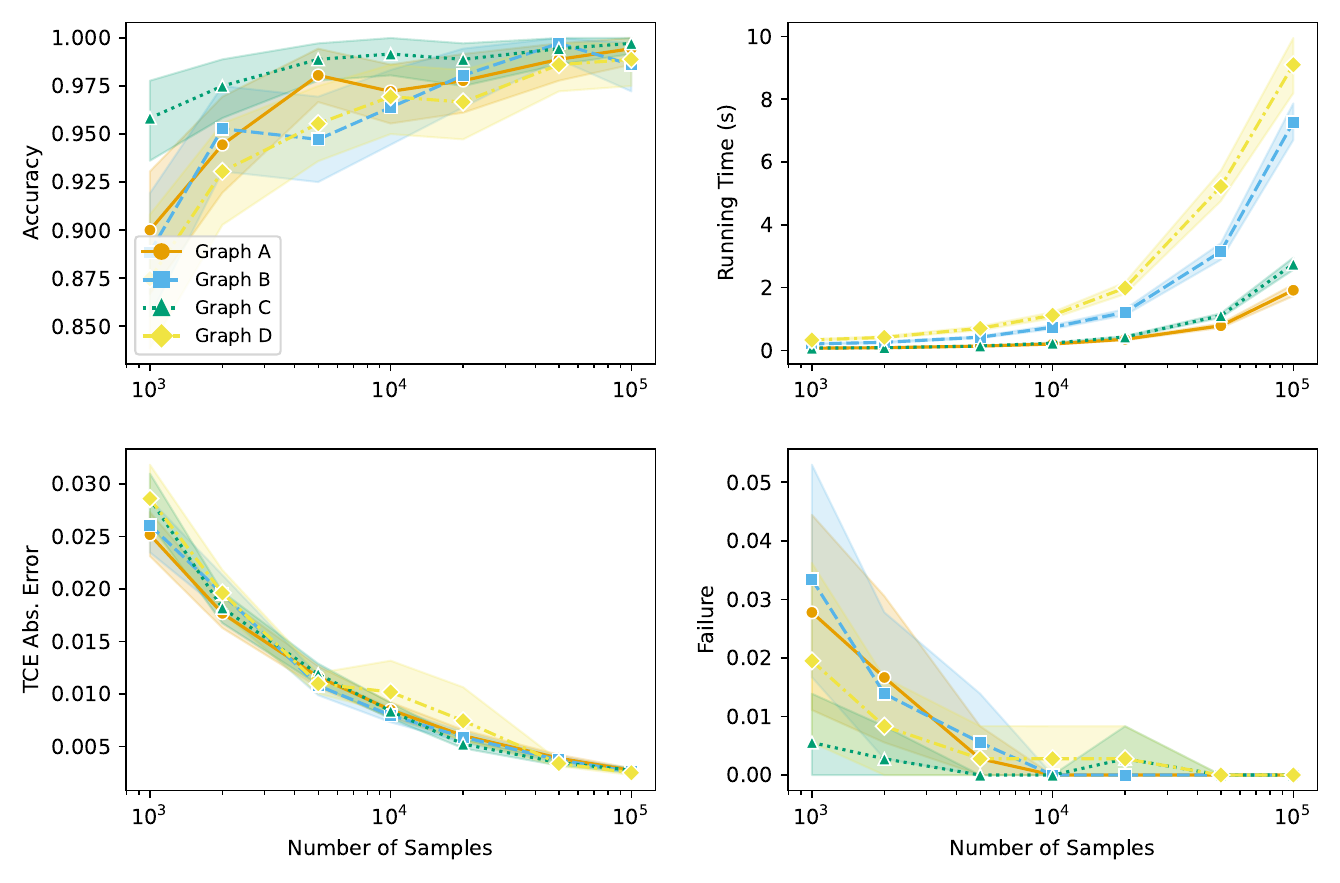}
\caption{%
    Full breakdown of the main configuration of \Cref{fig:samples_data-rse} ($\gamma=0.9$, $\texttt{normalized}$ true): accuracy, running time, TCE error, and failure rate, with all abstract graphs overlaid as in \Cref{fig:samples_data-rse}. The failure rate is at most $0.033$ for every graph at $n=10^3$, at most $0.003$ from $n=10^4$, and vanishes from $n=5\cdot10^4$. The running time grows with the number of samples, since each least-squares fit of the maximal partitioning involves all the $n$ samples. As in \Cref{fig:size_data-rse}, it is highest for Graphs B and D (about $7$ and $9$ seconds at $n=10^5$, against $2$ and $3$ seconds for Graphs A and C), which combine more confounders with interactions among them (\Cref{sec:experiments}).
}%
\label{fig:app-ms90}
\end{figure}

\begin{figure}[htb]
\centering
\includegraphics[width=\textwidth]{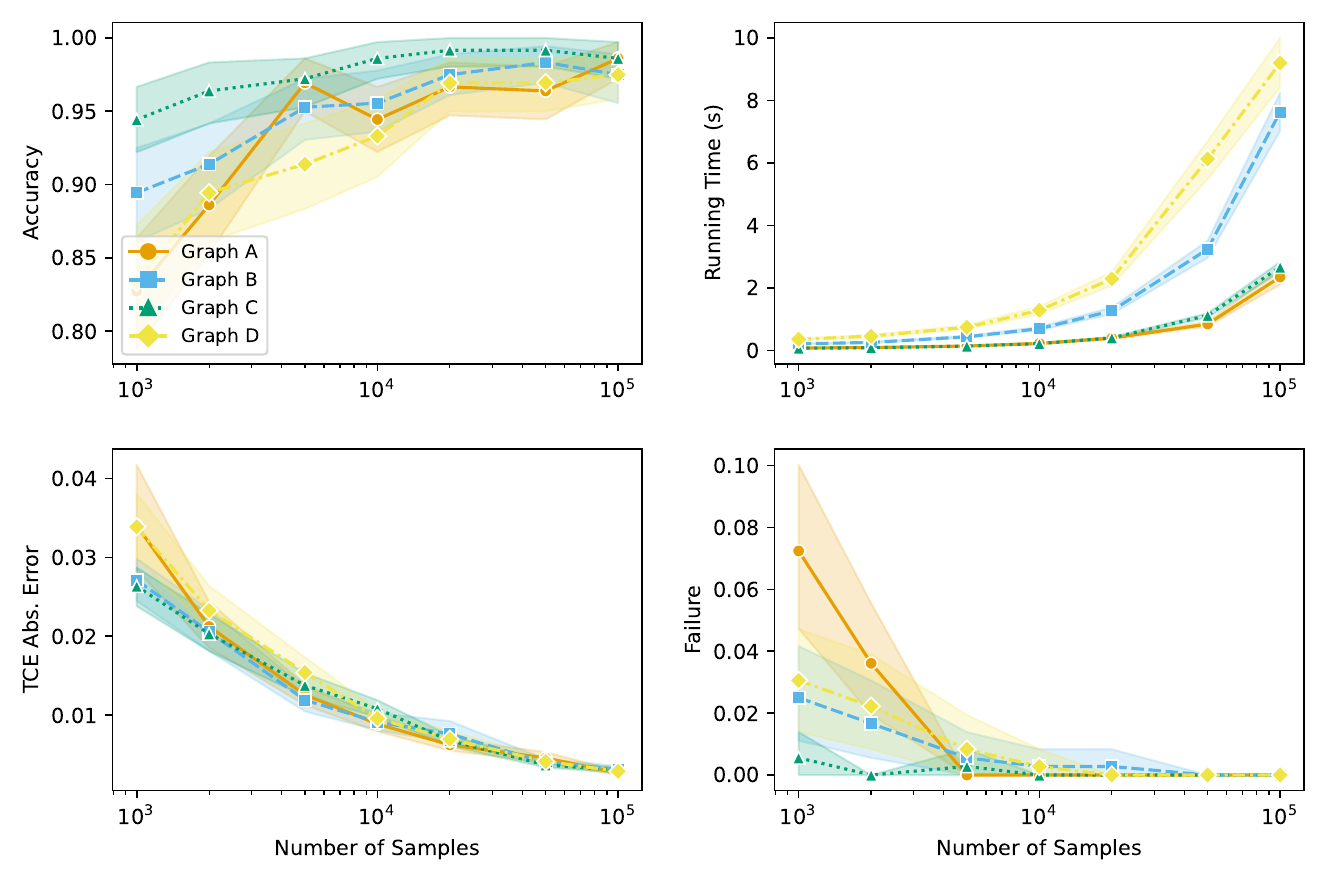}
\caption{%
    Same setting as \Cref{fig:app-ms90} ($\gamma=0.9$), with the $\texttt{normalized}$ flag set to false. With few samples, the accuracy is lower than in \Cref{fig:app-ms90} for Graphs A and D ($0.83$ for both at $n=10^3$, against $0.90$ and $0.87$), and the failure rate of Graph A reaches $0.07$. The gap closes as $n$ grows, and the accuracy reaches $0.98$--$0.99$ at $n=10^5$, with the same TCE error as in \Cref{fig:app-ms90}.
}%
\label{fig:app-ms90-ntn}
\end{figure}

The $\texttt{normalized}$ flag mainly affects the low-data regime. With abstraction coefficients sampled uniformly, the accuracy at $n=10^3$ is lower for some graphs, especially for the configurations with more relevant variables, while the two settings converge as the number of samples grows (compare \Cref{fig:app-ms90} with \Cref{fig:app-ms90-ntn}).

\begin{figure}[htb]
\centering
\includegraphics[width=\textwidth]{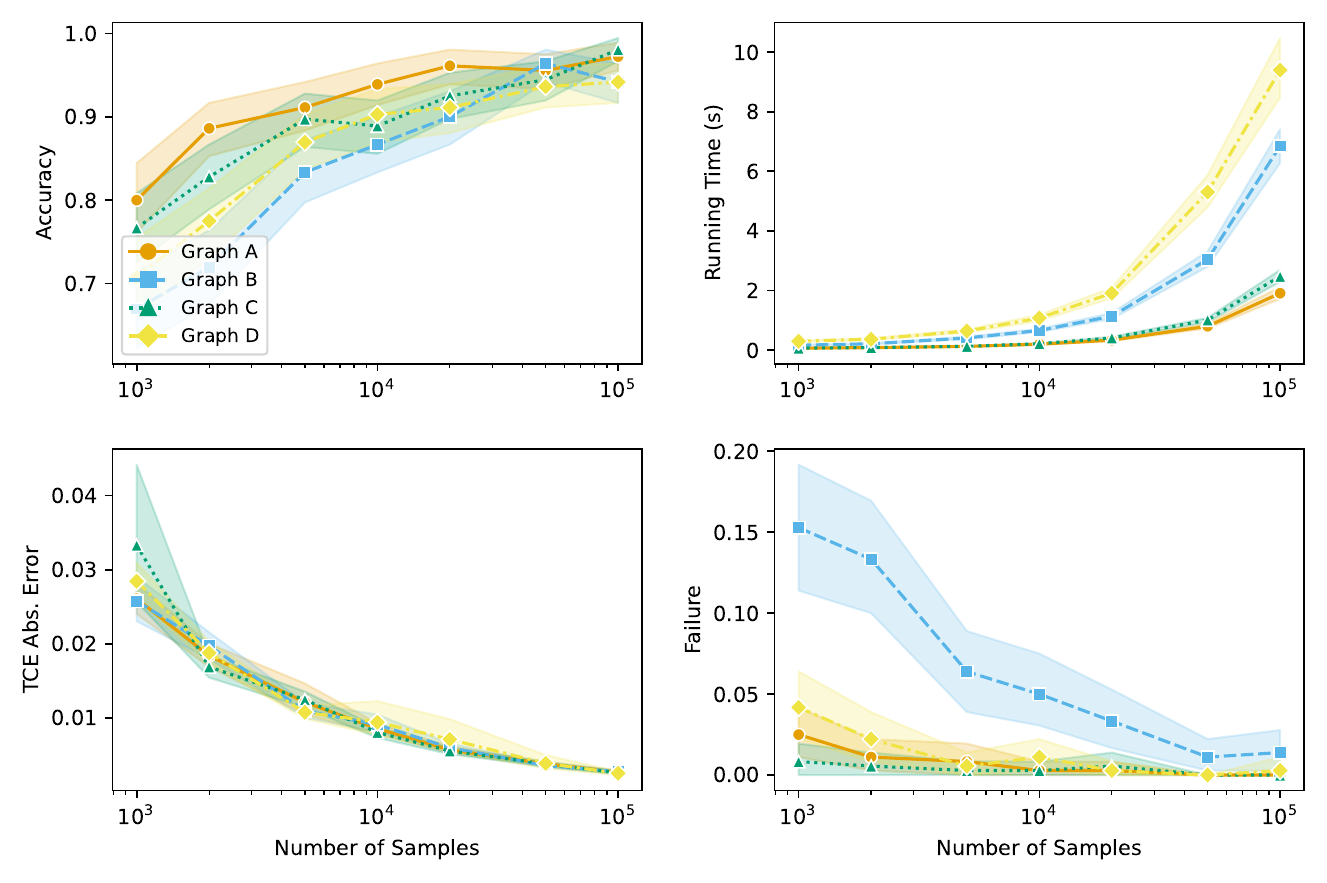}
\caption{%
    Same setting as \Cref{fig:app-ms90}, with $\gamma=1.00$ (\ie, no rejection sampling) and the $\texttt{normalized}$ flag set to true. Disabling the collinearity margin substantially degrades the results with few samples: at $n=10^3$, the accuracy ranges from $0.67$ for Graph B to $0.80$ for Graph A, and the failure rate of Graph B reaches $0.15$. However, the accuracy recovers as the number of samples grows, reaching $0.94$--$0.98$ at $n=10^5$, and the failure rate of Graph B decreases to $0.01$. Graph C, which is the most accurate graph under every other configuration in this section, drops to $0.77$ at $n=10^3$ and recovers to $0.98$ at $n=10^5$. Without the margin, some sampled instances draw coefficients close to violating the non-collinearity condition of \Cref{thm:maximalpartitioning}, so that more samples are needed to distinguish the corresponding rows of the composed abstraction matrix.
}%
\label{fig:app-ms100}
\end{figure}

Instead, removing the collinearity margin entirely ($\gamma=1.00$) substantially degrades the accuracy with few samples, and Graph C in particular drops from being the most accurate graph ($0.96$) to $0.77$ at $n=10^3$ (compare \Cref{fig:app-ms100} with \Cref{fig:app-ms90}). Nevertheless, the accuracy recovers as the number of samples grows. Since exactly collinear coefficients have probability zero, the margin does not determine whether the ground truth can be recovered, but how many samples are needed to recover it. Near-collinear coefficients, which the rejection sampling procedure of \Cref{app:datagen} excludes, require a more accurate estimate of the composed abstraction matrix.

\begin{figure}[htb]
\centering
\includegraphics[width=0.48\textwidth]{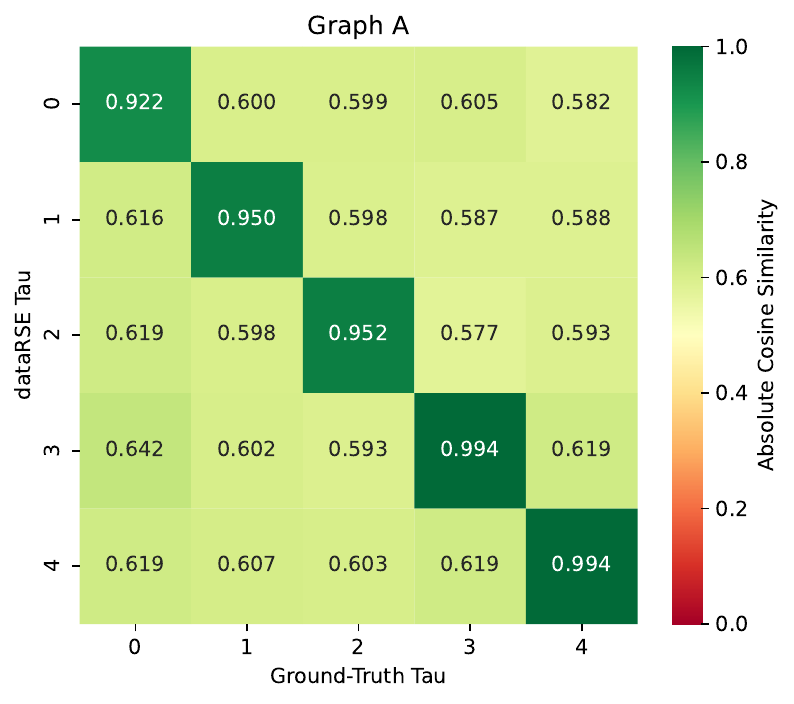}
\includegraphics[width=0.48\textwidth]{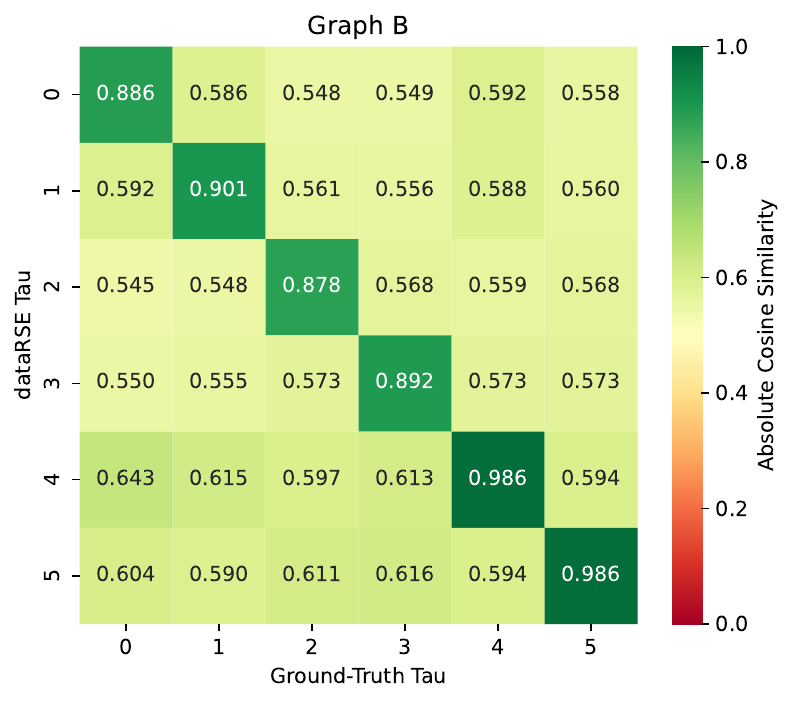}\\
\includegraphics[width=0.48\textwidth]{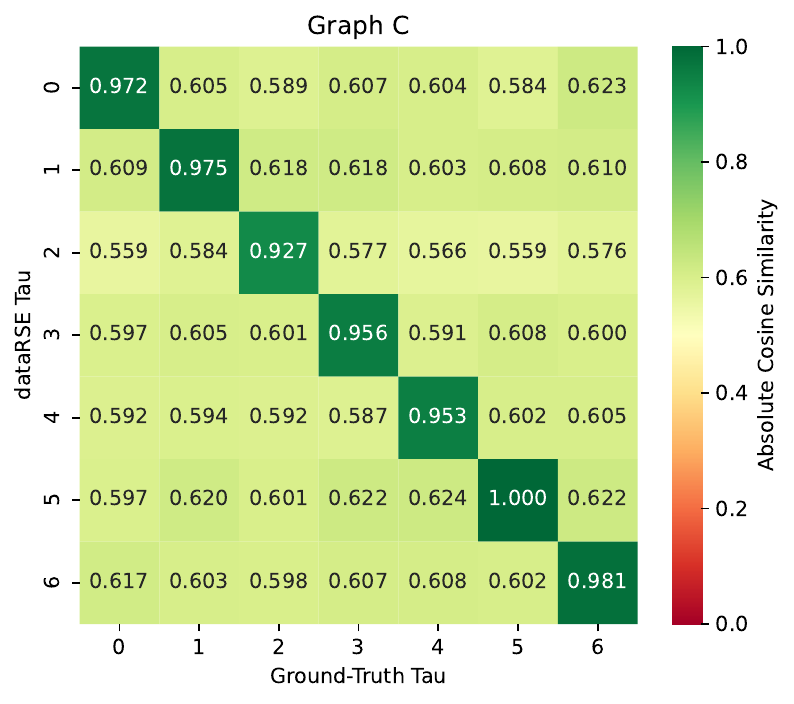}
\includegraphics[width=0.48\textwidth]{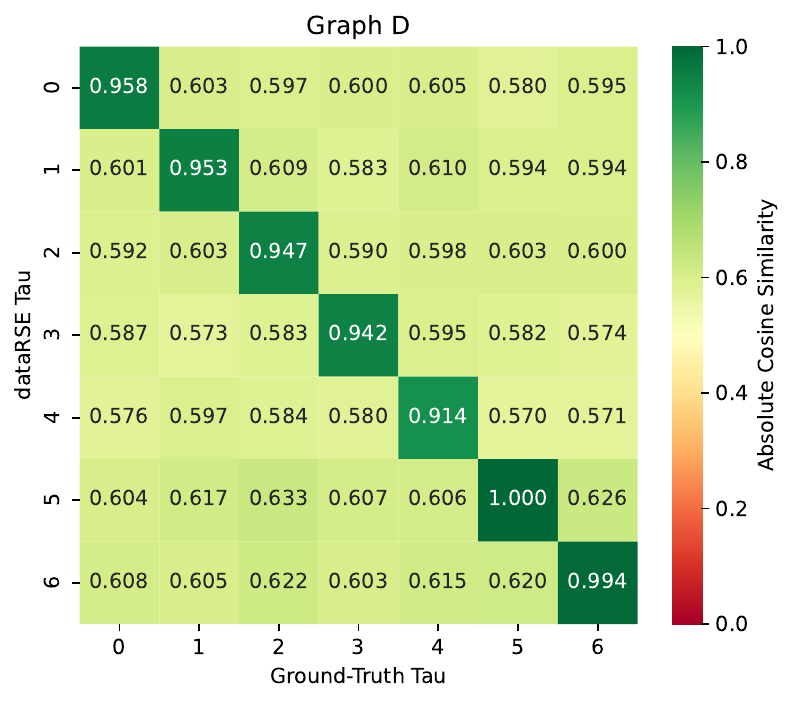}
\caption{%
    Same setting as \Cref{fig:tau_recovery}, with $\gamma=1.00$ (\ie, no rejection sampling) and the $\texttt{normalized}$ flag set to true. Consistent with the accuracy degradation of \Cref{fig:app-ms100}, the diagonal is weaker than with the collinearity margin (e.g., between $0.93$ and $1.0$ for Graph C, against at least $0.99$ at $\gamma=0.9$, \Cref{fig:tau_recovery}), while the off-diagonal entries (about $0.56$--$0.62$) remain at the similarity level of independent abstraction vectors. The weaker diagonal is entirely due to the incorrect solutions, since the correct ones recover the abstraction vectors up to scale and sign.
}%
\label{fig:app-tau-ms100}
\end{figure}

\begin{figure}[htb]
\centering
\includegraphics[width=0.24\textwidth]{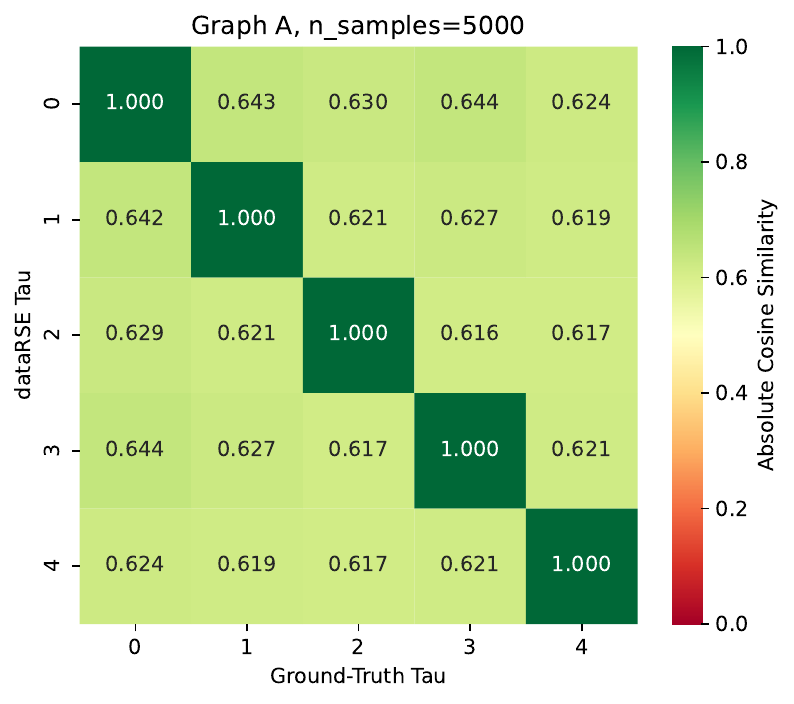}
\includegraphics[width=0.24\textwidth]{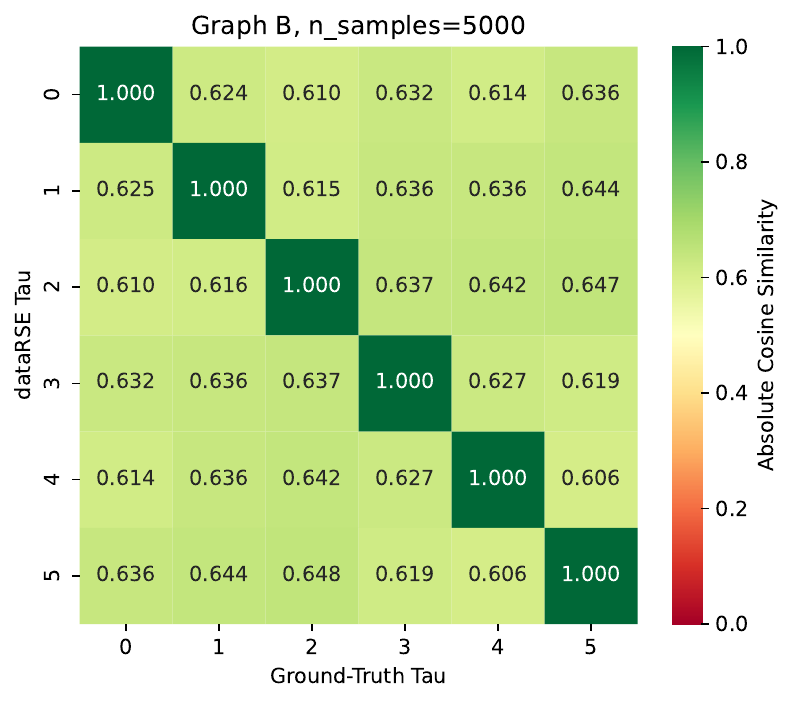}
\includegraphics[width=0.24\textwidth]{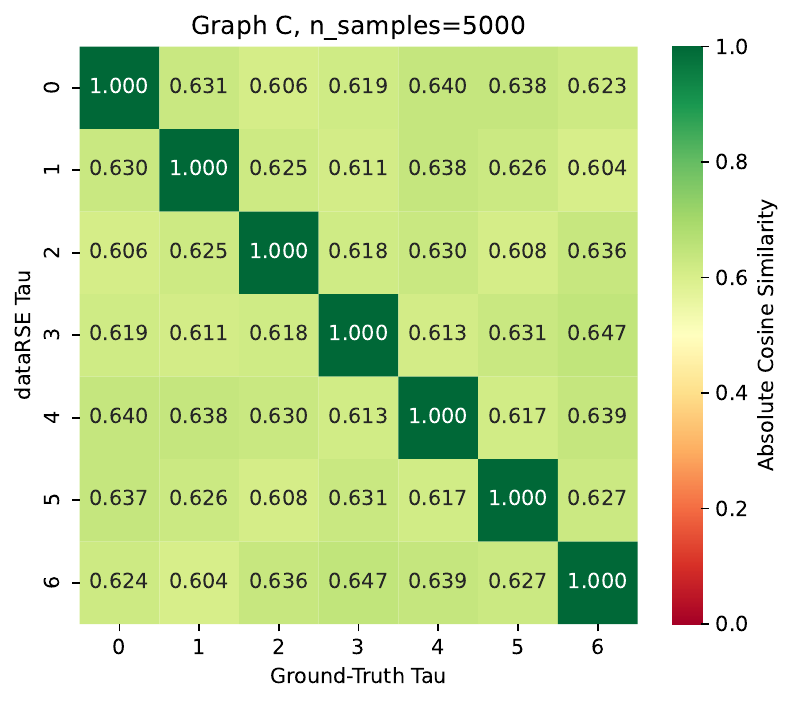}
\includegraphics[width=0.24\textwidth]{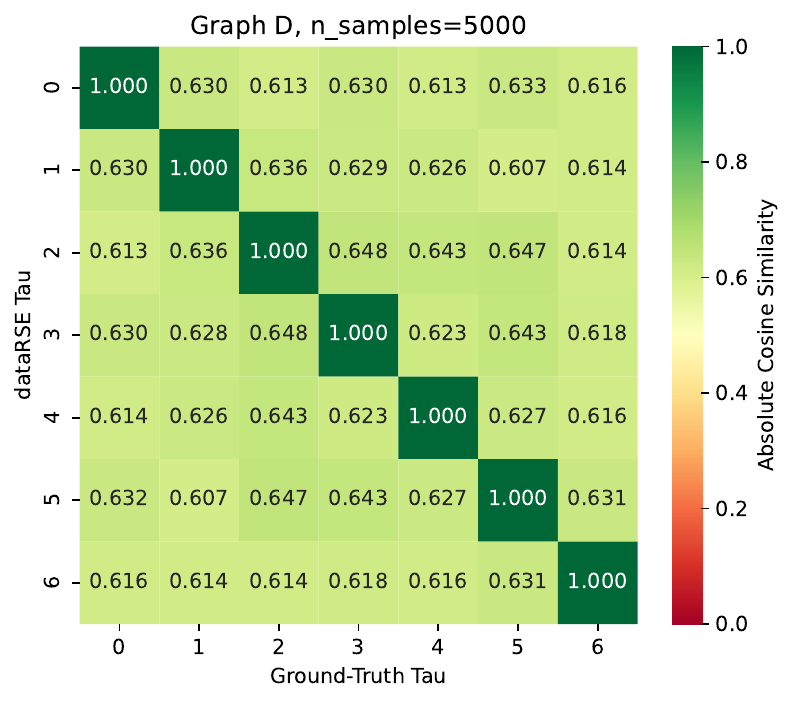}\\
\includegraphics[width=0.24\textwidth]{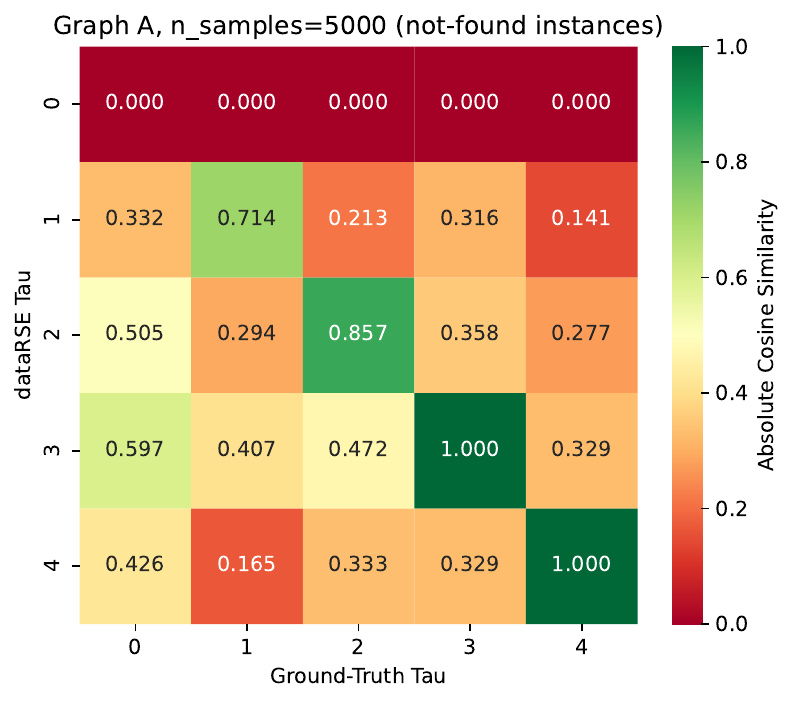}
\includegraphics[width=0.24\textwidth]{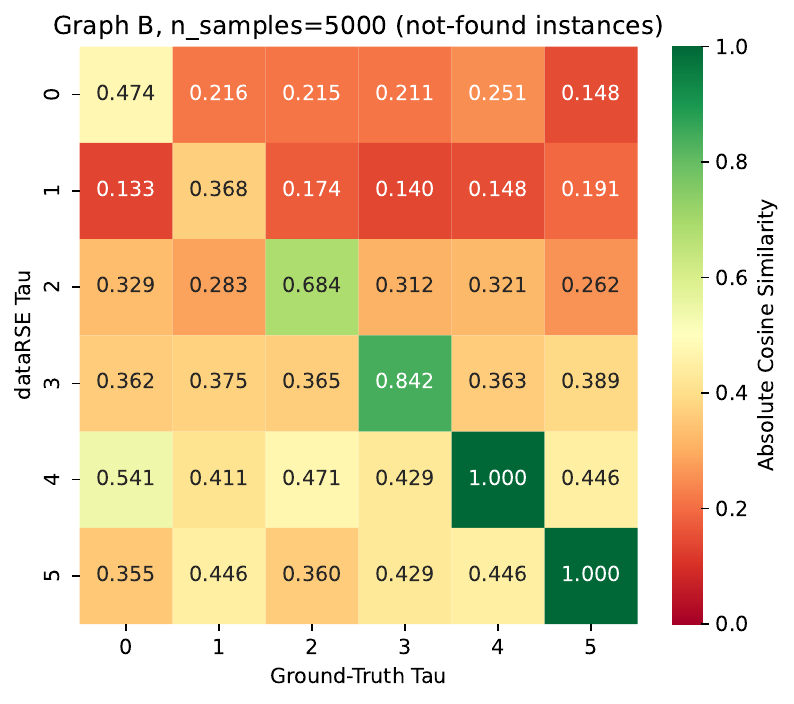}
\includegraphics[width=0.24\textwidth]{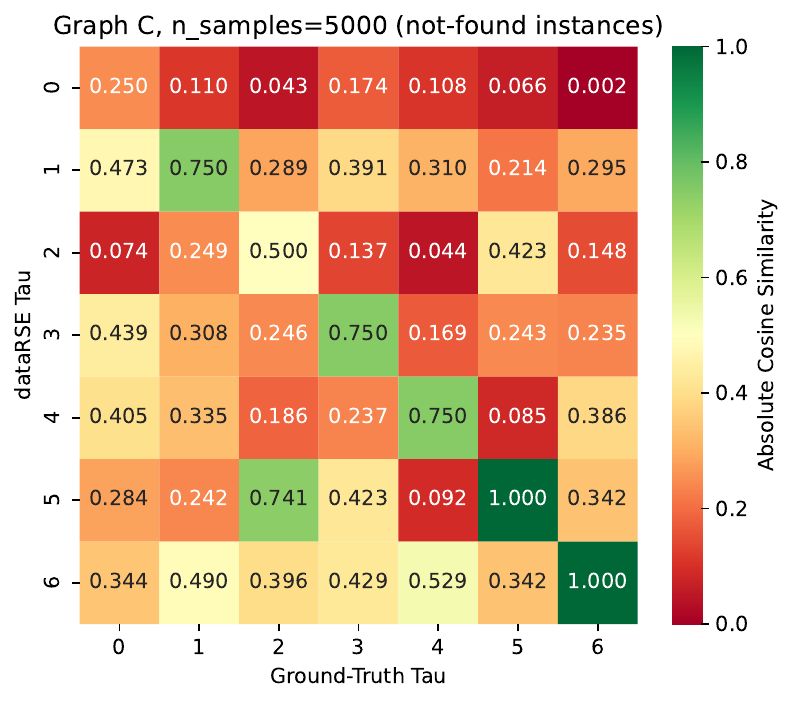}
\includegraphics[width=0.24\textwidth]{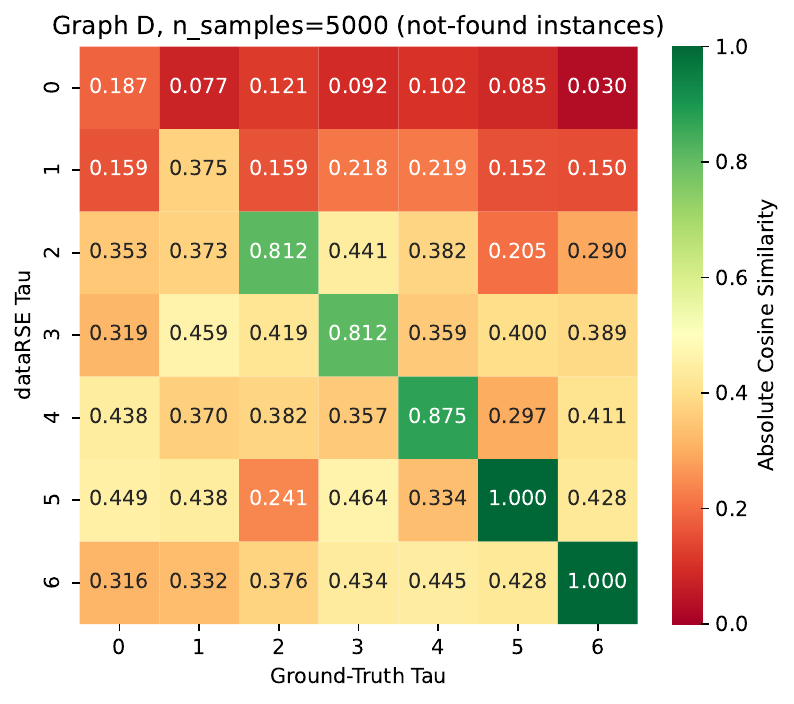}
\caption{%
    Same setting as \Cref{fig:tau_recovery} ($\gamma=0.9$, $n=5\cdot10^3$), restricted to the runs in which \model{dataRSE} recovers the ground-truth relevant variables (top) and to the remaining runs (bottom), for Graphs A--D from left to right. The incorrect runs are $7$, $19$, $4$, and $16$ out of $360$ for Graphs A, B, C, and D, respectively ($359$ for Graph D, whose ground truth could not be sampled in one run). In the correct runs, the learned abstraction vectors coincide with the ground-truth ones up to scale and sign (average absolute cosine similarity above $0.999$). In the incorrect runs, the average similarity on the diagonal, excluding the treatment and the outcome, ranges between $0.52$ and $0.61$, and some abstract variables are still recovered while others are not.
}%
\label{fig:app-tau-split-ms90}
\end{figure}

\begin{figure}[htb]
\centering
\includegraphics[width=0.24\textwidth]{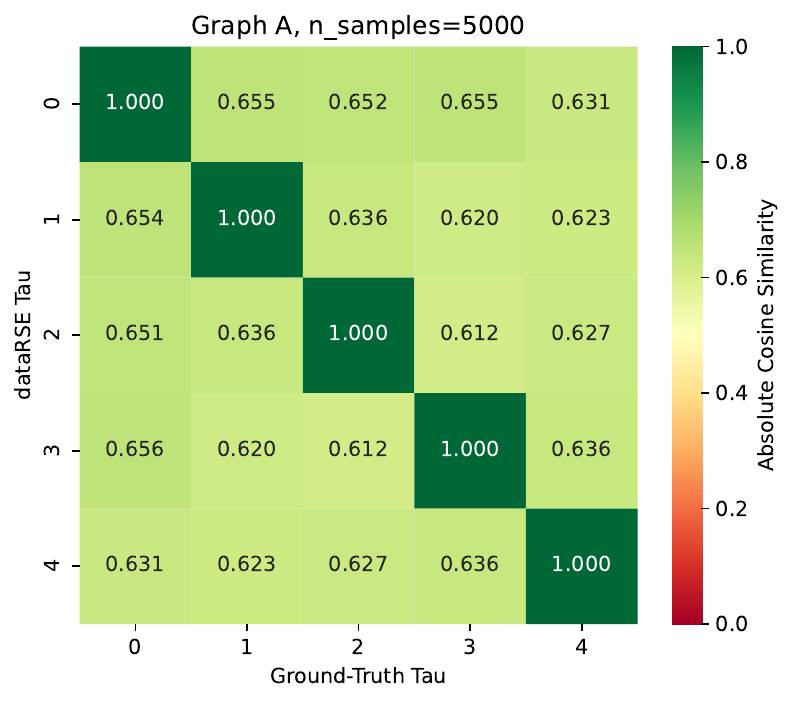}
\includegraphics[width=0.24\textwidth]{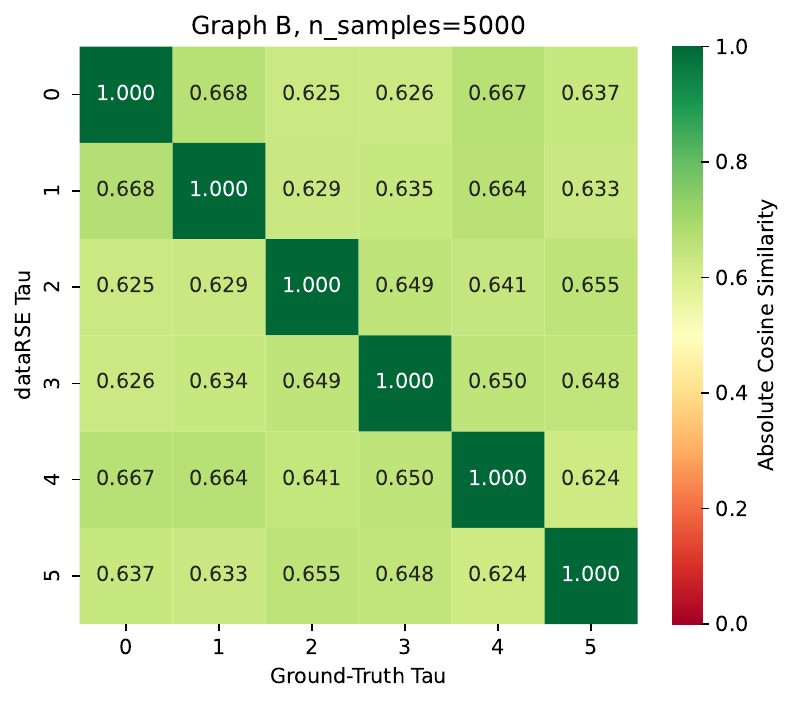}
\includegraphics[width=0.24\textwidth]{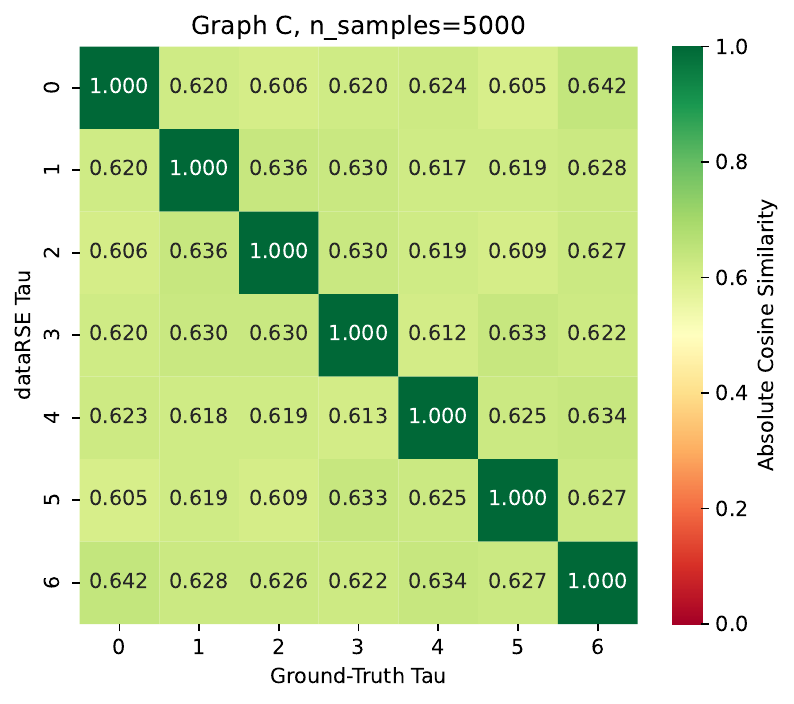}
\includegraphics[width=0.24\textwidth]{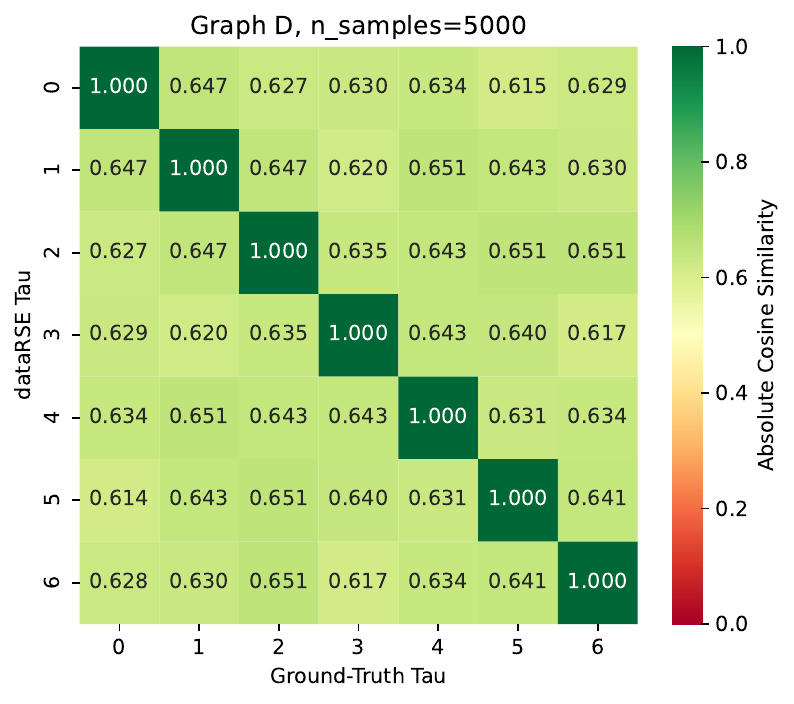}\\
\includegraphics[width=0.24\textwidth]{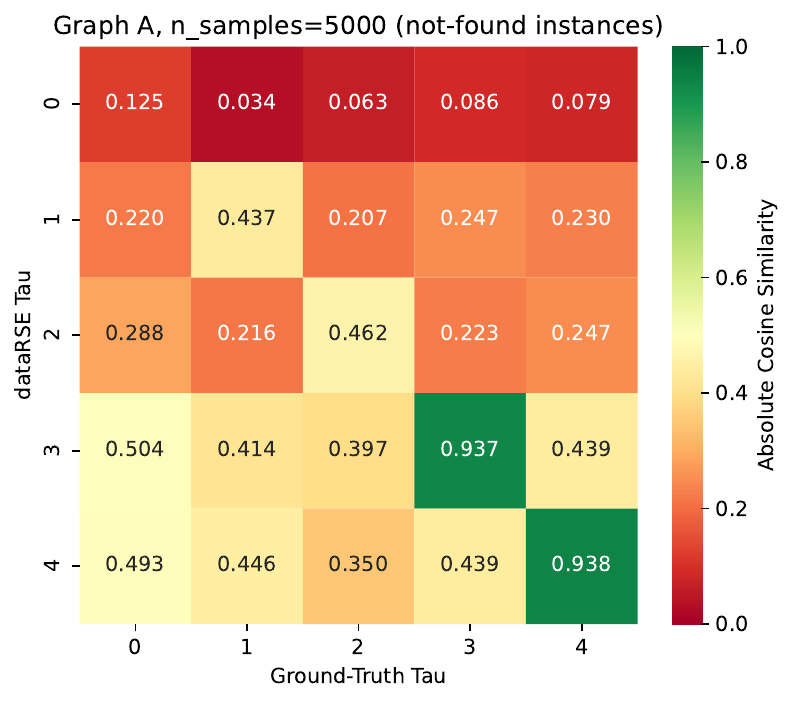}
\includegraphics[width=0.24\textwidth]{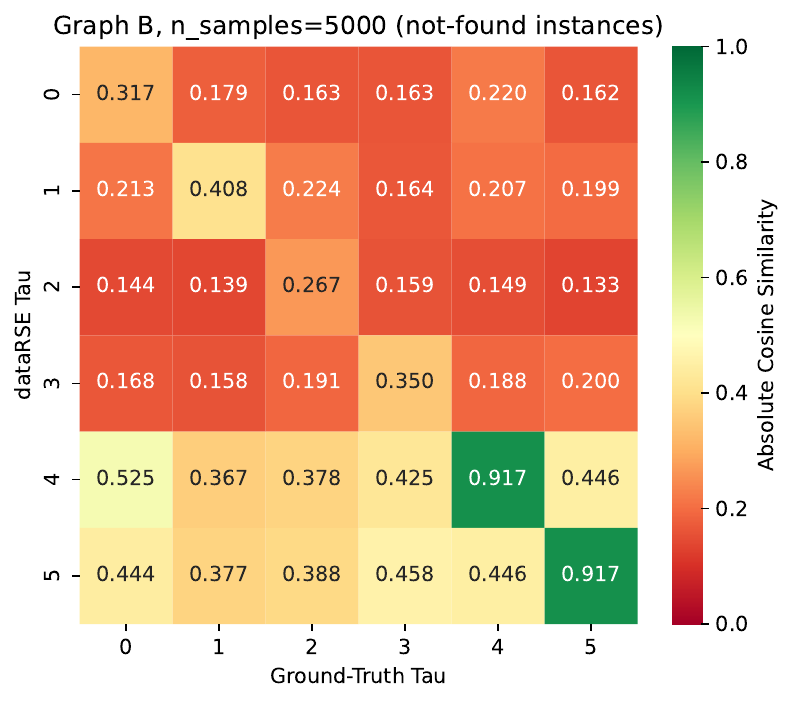}
\includegraphics[width=0.24\textwidth]{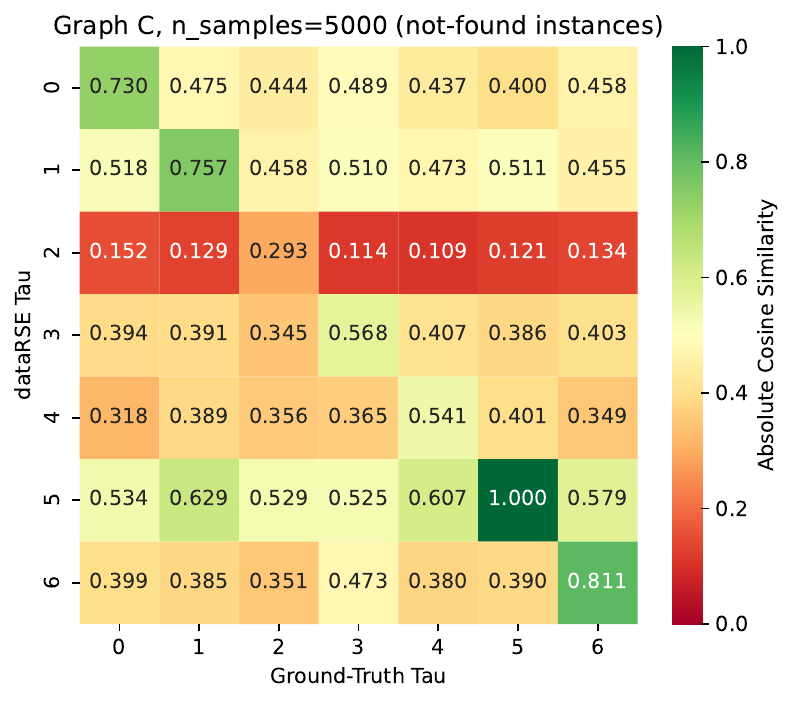}
\includegraphics[width=0.24\textwidth]{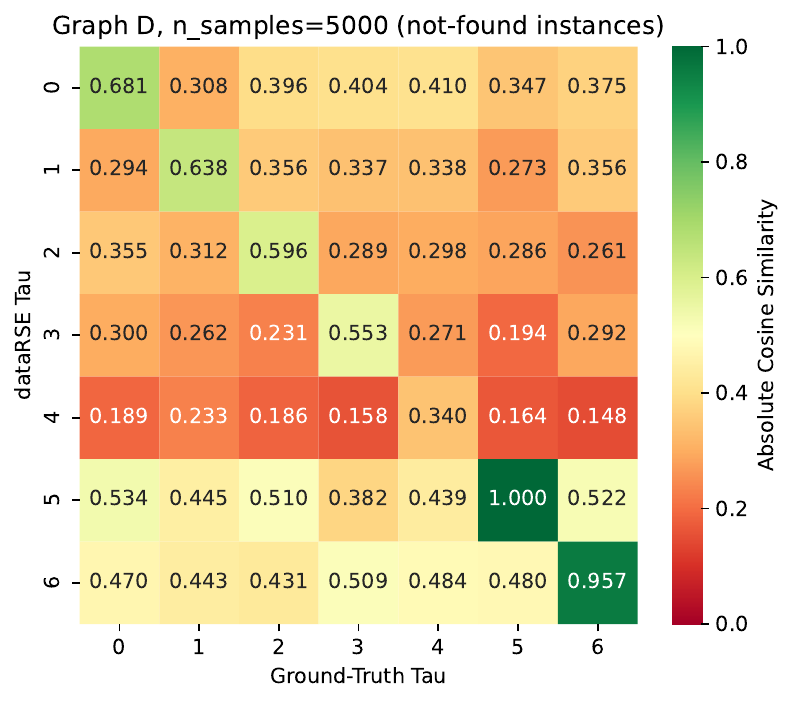}
\caption{%
    Same setting as \Cref{fig:app-tau-ms100} ($\gamma=1.00$, $n=5\cdot10^3$), restricted to the runs in which \model{dataRSE} recovers the ground-truth relevant variables (top) and to the remaining runs (bottom), for Graphs A--D from left to right. The incorrect runs are $32$, $60$, $37$, and $47$ out of $360$ for Graphs A, B, C, and D, respectively. As in \Cref{fig:app-tau-split-ms90}, the correct runs recover the abstraction vectors up to scale and sign (average absolute cosine similarity above $0.999$), whereas the average similarity on the diagonal of the incorrect runs, excluding the treatment and the outcome, ranges between $0.34$ and $0.58$.
}%
\label{fig:app-tau-split-ms100}
\end{figure}

Finally, we report the recovery of the abstraction coefficients without the collinearity margin, at $n=5\cdot10^3$ samples (\Cref{fig:app-tau-ms100}). Disabling the margin weakens the diagonal with respect to \Cref{fig:tau_recovery}, consistent with the lower accuracy for the same configuration, between $0.83$ and $0.91$ (\Cref{fig:app-ms100}). As with the margin (\Cref{fig:app-tau-split-ms90}), whenever \model{dataRSE} recovers the ground-truth relevant variables, the learned abstraction vectors coincide with the ground-truth ones up to scale and sign, with an average absolute cosine similarity above $0.999$. Instead, the incorrect solutions reach an average similarity between $0.34$ and $0.58$, which is not higher than the similarity between independent abstraction vectors (\Cref{fig:app-tau-split-ms100}). Therefore, the weaker diagonal is entirely due to the more frequent incorrect solutions, and the diagonal remains high only because most runs recover the ground-truth relevant variables.

\subsection{Additional Graphical Structures}%
\label{app:additional-experiments-graphs}

\begin{figure}[ht]
\centering
\begin{minipage}[b]{0.22\textwidth}
\centering\small\resizebox{\textwidth}{!}{\begin{tikzpicture}[node distance=0.8cm, minimum size=0.3cm]
  \node[draw, circle, thick] (Y4) {$Y_4$};
  \node[draw, circle, thick, right=0.4cm of Y4] (Y5) {$Y_5$};
  \node[draw, circle, thick, right=0.4cm of Y5] (Y6) {$Y_6$};
  \node[draw, circle, thick, fill=red!40!white, below=0.7cm of Y4] (Y7) {$Y_T$};
  \node[draw, circle, thick, fill=green!40!white, below=0.7cm of Y6] (Y8) {$Y_O$};
  \node[draw, circle, thick, above=0.7cm of Y6] (Y2) {$Y_2$};
  \node[draw, circle, thick, above=0.7cm of Y4] (Y3) {$Y_3$};
  \node[draw, circle, thick, above=1.7cm of Y5] (Y1) {$Y_1$};
  \draw[-{Latex}] (Y1) -- (Y2);
  \draw[-{Latex}] (Y1) -- (Y3);
  \draw[-{Latex}] (Y2) -- (Y3);
  \draw[-{Latex}] (Y2) -- (Y4);
  \draw[-{Latex}] (Y2) -- (Y5);
  \draw[-{Latex}] (Y2) -- (Y6);
  \draw[-{Latex}] (Y3) -- (Y4);
  \draw[-{Latex}] (Y3) -- (Y5);
  \draw[-{Latex}] (Y3) -- (Y6);
  \draw[-{Latex}] (Y4) to[bend right=65] (Y6);
  \draw[-{Latex}] (Y5) -- (Y6);
  \draw[-{Latex}] (Y4) -- (Y7);
  \draw[-{Latex}] (Y4) -- (Y8);
  \draw[-{Latex}] (Y5) -- (Y7);
  \draw[-{Latex}] (Y5) -- (Y8);
  \draw[-{Latex}] (Y6) -- (Y7);
  \draw[-{Latex}] (Y6) -- (Y8);
  \draw[-{Latex}] (Y7) -- (Y8);
\end{tikzpicture}}%
\subcaption{Graph E}
\end{minipage}\hfill%
\begin{minipage}[b]{0.29\textwidth}
\centering\small\resizebox{\textwidth}{!}{\begin{tikzpicture}[node distance=0.8cm, minimum size=0.3cm]
  \node[draw, circle, thick] (Y1) {$Y_1$};
  \node[draw, circle, thick, right=0.4cm of Y1] (Y2) {$Y_2$};
  \node[draw, circle, thick, below left=0.7cm and 0.7cm of Y1] (Y3) {$Y_3$};
  \node[draw, circle, thick, right=0.4cm of Y3] (Y4) {$Y_4$};
  \node[draw, circle, thick, right=0.4cm of Y4] (Y5) {$Y_5$};
  \node[draw, circle, thick, right=0.4cm of Y5] (Y6) {$Y_6$};
  \node[draw, circle, thick, fill=red!40!white, below=0.7cm of Y4] (Y7) {$Y_T$};
  \node[draw, circle, thick, fill=green!40!white, below=0.7cm of Y5] (Y8) {$Y_O$};
  \draw[-{Latex}] (Y1) -- (Y3);
  \draw[-{Latex}] (Y1) -- (Y4);
  \draw[-{Latex}] (Y1) -- (Y5);
  \draw[-{Latex}] (Y1) -- (Y6);
  \draw[-{Latex}] (Y2) -- (Y3);
  \draw[-{Latex}] (Y2) -- (Y4);
  \draw[-{Latex}] (Y2) -- (Y5);
  \draw[-{Latex}] (Y2) -- (Y6);
  \draw[-{Latex}] (Y3) -- (Y7);
  \draw[-{Latex}] (Y3) -- (Y8);
  \draw[-{Latex}] (Y4) -- (Y7);
  \draw[-{Latex}] (Y4) -- (Y8);
  \draw[-{Latex}] (Y5) -- (Y7);
  \draw[-{Latex}] (Y5) -- (Y8);
  \draw[-{Latex}] (Y6) -- (Y7);
  \draw[-{Latex}] (Y6) -- (Y8);
\end{tikzpicture}}%
\subcaption{Graph F}
\end{minipage}\hfill%
\begin{minipage}[b]{0.22\textwidth}
\centering\small\resizebox{\textwidth}{!}{\begin{tikzpicture}[node distance=0.8cm, minimum size=0.3cm]
  \node[draw, circle, thick] (Y3) {$Y_3$};
  \node[draw, circle, thick, right=0.4cm of Y3] (Y4) {$Y_4$};
  \node[draw, circle, thick, right=0.4cm of Y4] (Y5) {$Y_5$};
  \node[draw, circle, thick, fill=red!40!white, below=0.7cm of Y3] (Y6) {$Y_T$};
  \node[draw, circle, thick, fill=green!40!white, below=0.7cm of Y5] (Y7) {$Y_O$};
  \node[draw, circle, thick, above=0.7cm of Y3] (Y1) {$Y_1$};
  \node[draw, circle, thick, above=0.7cm of Y5] (Y2) {$Y_2$};
  \draw[-{Latex}] (Y1) -- (Y3);
  \draw[-{Latex}] (Y1) -- (Y4);
  \draw[-{Latex}] (Y2) -- (Y3);
  \draw[-{Latex}] (Y2) -- (Y4);
  \draw[-{Latex}] (Y2) to[bend left=22] (Y6);
  \draw[-{Latex}] (Y2) to[bend left=25] (Y7);
  \draw[-{Latex}] (Y3) -- (Y6);
  \draw[-{Latex}] (Y3) -- (Y7);
  \draw[-{Latex}] (Y4) -- (Y6);
  \draw[-{Latex}] (Y4) -- (Y7);
  \draw[-{Latex}] (Y5) -- (Y6);
  \draw[-{Latex}] (Y5) -- (Y7);
\end{tikzpicture}}%
\subcaption{Graph G}
\end{minipage}\hfill%
\begin{minipage}[b]{0.22\textwidth}
\centering\small\resizebox{\textwidth}{!}{\begin{tikzpicture}[node distance=0.8cm, minimum size=0.3cm]
  \node[draw, circle, thick] (Y1) {$Y_1$};
  \node[draw, circle, thick, right=0.4cm of Y1] (Y2) {$Y_2$};
  \node[draw, circle, thick, right=0.4cm of Y2] (Y3) {$Y_3$};
  \node[draw, circle, thick, fill=red!40!white, below=0.7cm of Y1] (Y4) {$Y_T$};
  \node[draw, circle, thick, fill=green!40!white, below=0.7cm of Y3] (Y5) {$Y_O$};
  \draw[-{Latex}] (Y1) -- (Y4);
  \draw[-{Latex}] (Y2) -- (Y4);
  \draw[-{Latex}] (Y1) to[bend left=45] (Y3);
  \draw[-{Latex}] (Y2) -- (Y3);
  \draw[-{Latex}] (Y3) -- (Y4);
  \draw[-{Latex}] (Y3) -- (Y5);
  \draw[-{Latex}] (Y4) -- (Y5);
\end{tikzpicture}}%
\subcaption{Graph H}
\end{minipage}%
\caption{%
    Four additional abstract graph structures. Graph E is \emph{deep}: a 3-level recursion with a within-layer interaction edge at each of the first two confounder layers (\Cref{fig:abstract-graphs}'s Graph D only has one such layer). Graph F is \emph{wide}: a 4-way confounder set directly on $Y_T,Y_O$ (Graphs A--D have at most 3), confounded in turn by a further 2-way set. Graph G is \emph{compound}: it combines Graph F's 4-way width with Graph B's known-parents mechanism, so that one of the four first-level confounders must be adjusted for as a known parent $\set{K}$ when resolving a second, deeper confounder of a subset of its siblings. Graph H reproduces the example of \Cref{app:mpinconsistency}, where the maximal partitioning of the first recursive call is not contained in any parametrically consistent minimal sufficient extension: the effects of the confounders $Y_1$ and $Y_2$ on $Y_O$ are entirely mediated by $Y_3$, so that their rows of the composed abstraction matrix are collinear at the first recursive call.
}%
\label{fig:app-graphs-ijk}
\end{figure}
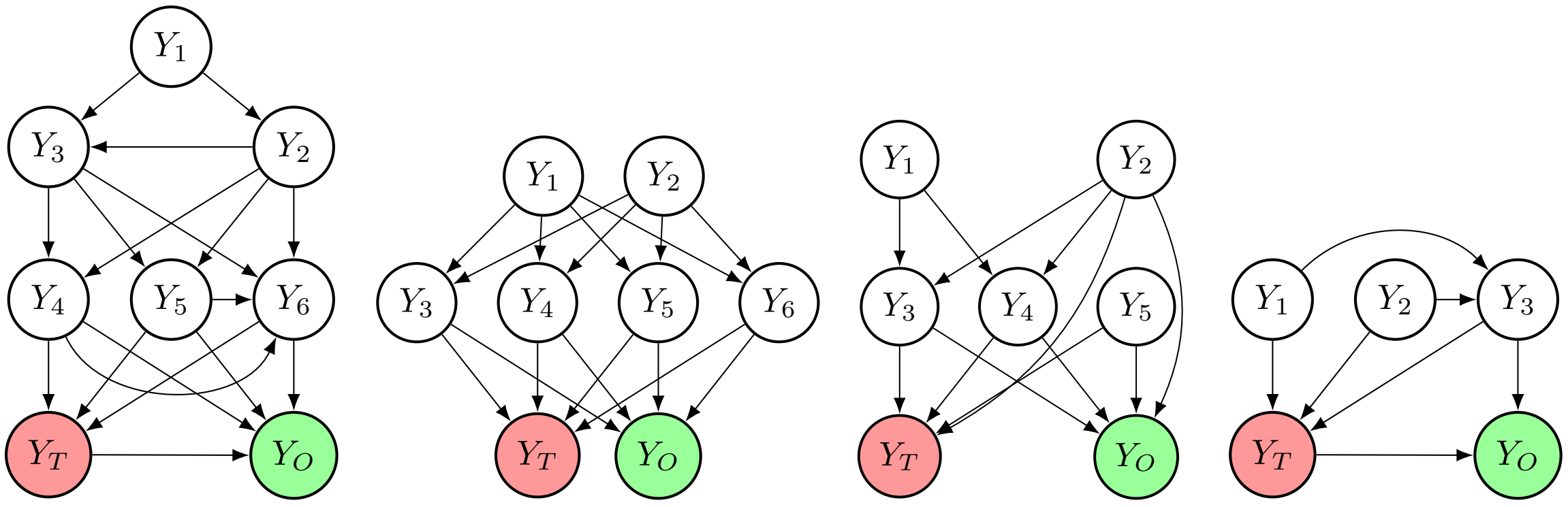

We evaluate \model{dataRSE} on four additional abstract graphs, Graphs E--H (\Cref{fig:app-graphs-ijk}), in the main configuration of \Cref{sec:experiments}. Each of them stresses a different source of complexity with respect to Graphs A--D, and Graph H reproduces the example of \Cref{app:mpinconsistency}, where the maximal partitioning of the first recursive call is not contained in any parametrically consistent minimal sufficient extension.

\begin{figure}[htb]
\centering
\includegraphics[width=\textwidth]{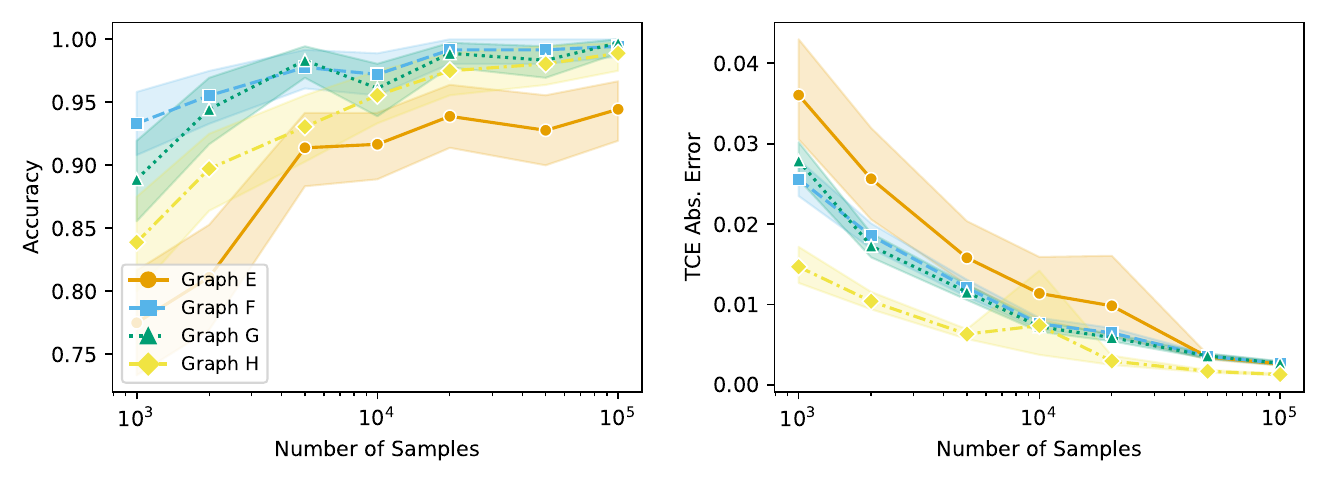}
\caption{%
    Accuracy and TCE error of \model{dataRSE} on Graphs E--H (\Cref{fig:app-graphs-ijk}), for an increasing number of samples, in the same format as \Cref{fig:samples_data-rse}.
}%
\label{fig:app-ijk}
\end{figure}

\begin{figure}[htb]
\centering
\includegraphics[width=\textwidth]{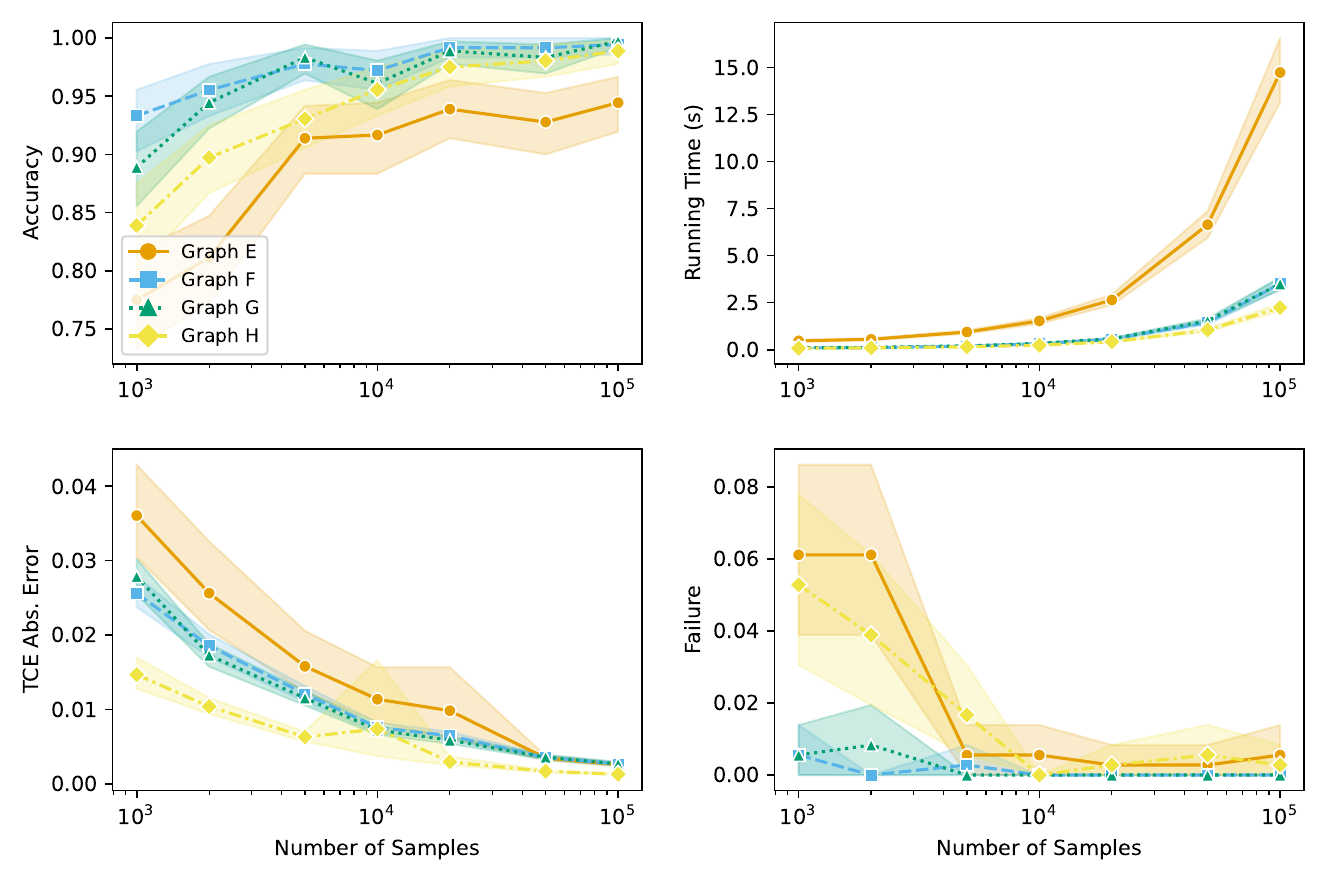}
\caption{%
    Full breakdown of Graphs E--H (\Cref{fig:app-graphs-ijk}), in the same format as \Cref{fig:app-ms90}: accuracy, running time, TCE error, and failure rate. Graphs E and H have a non-negligible failure rate at $n=10^3$ ($0.06$ and $0.05$), which is at most $0.006$ from $n=10^4$, while Graphs F and G fail in only $3$ and $5$ runs, respectively. Graph E also has the highest running time (about $15$ seconds at $n=10^5$, against $2$--$4$ seconds for the other graphs), since it requires about six recursive calls, against three for the other graphs.
}%
\label{fig:app-ijk-full}
\end{figure}

The deep Graph E is harder than the Graph F and the Graph G (\Cref{fig:app-ijk,fig:app-ijk-full}). At $n=10^3$, the accuracy of Graph E is $0.78$, against $0.93$ and $0.89$ for F and G, and it reaches $0.92$ at $n=10^4$ and $0.94$ at $n=10^5$, whereas F and G are already at $0.96$--$1.00$ from $n=10^4$. Instead, the four confounders of Graph F cost little, since they only add collinearity constraints within a single recursive call, rather than an additional call. Graph G, which also requires adjusting for a known parent in $\set{K}$, behaves similarly to Graph F rather than to Graph E. Hence, the accuracy of \model{dataRSE} depends on the recursion depth more than on the number of confounders.

\model{dataRSE} also recovers Graph H, with an accuracy of $0.84$ at $n=10^3$, $0.96$ at $n=10^4$, and $0.99$ at $n=10^5$: by keeping only the unmediated relevant sets (\Cref{alg:datarse}, Lines 23--24), it first recovers the mediator $Y_3$ and only then separates the confounders $Y_1$ and $Y_2$, whose rows of the composed abstraction matrix are collinear at the first recursive call. For all four graphs, the TCE error decreases from at most $0.04$ at $n=10^3$ to at most $0.003$ at $n=10^5$. As for Graphs A--D, most incorrect solutions are finer partitions of the ground truth, from about two thirds of them for Graph F to more than $90\%$ for Graphs E and H.

\subsection{Comparison with Rank-Based Partitioning}%
\label{app:plateau}

\begin{figure}[htb]
\centering
\includegraphics[width=\textwidth]{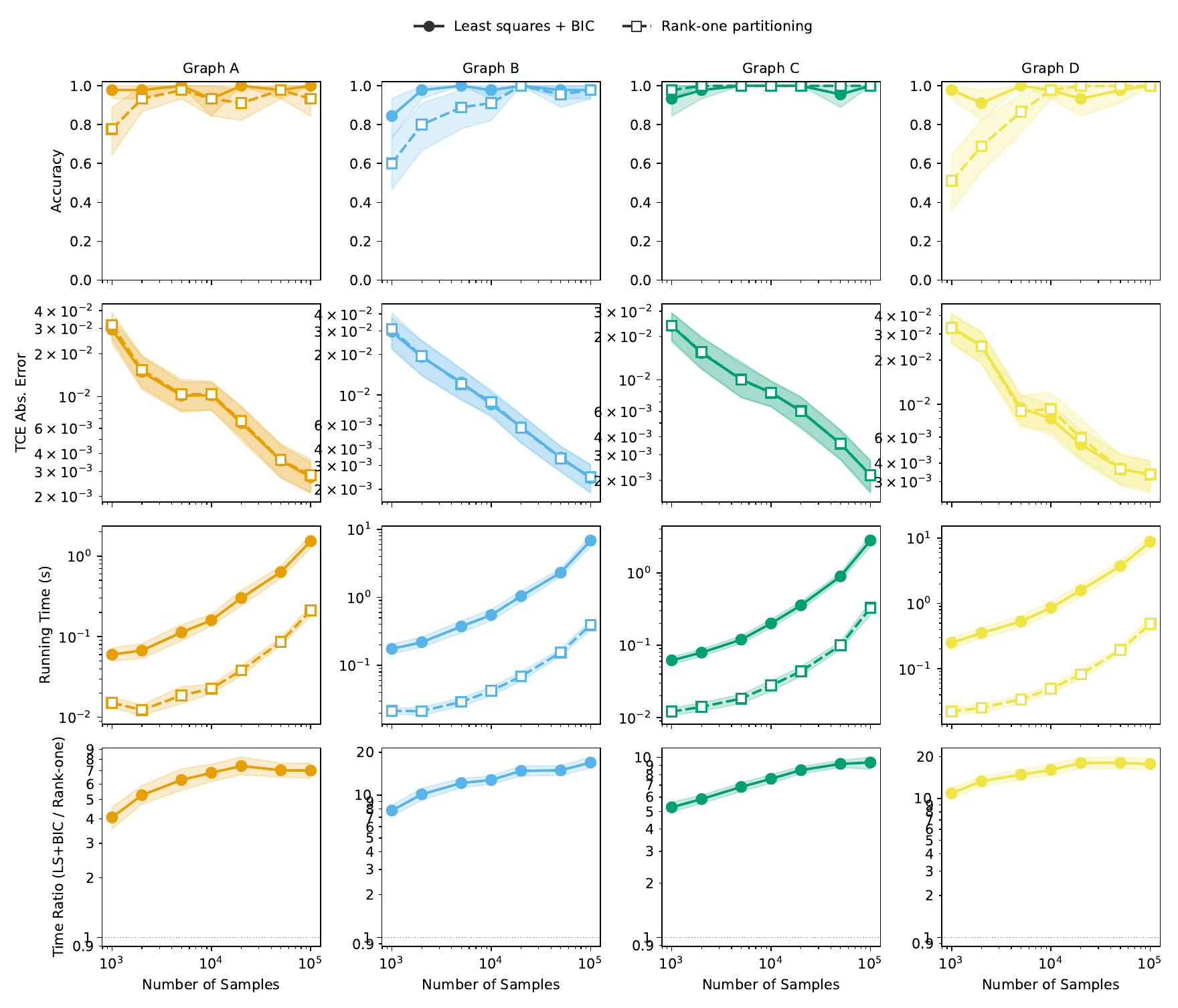}
\caption{%
    Comparison between \Cref{alg:maximalpartitioning} (solid) and rank-based partitioning (\Cref{alg:maximalpartitioning-plateau}, dashed), on Graphs A--D. For an increasing number of samples, we report the accuracy in recovering the exact ground-truth relevant sets, the absolute error in estimating the TCE of $Y_T$ on $Y_O$, the running time in seconds, and the ratio between the running times of the two rules on the same data. Both rules run on the same ground truths and samples.
}%
\label{fig:app-compare-rule-abcd}
\end{figure}

\begin{figure}[htb]
\centering
\includegraphics[width=\textwidth]{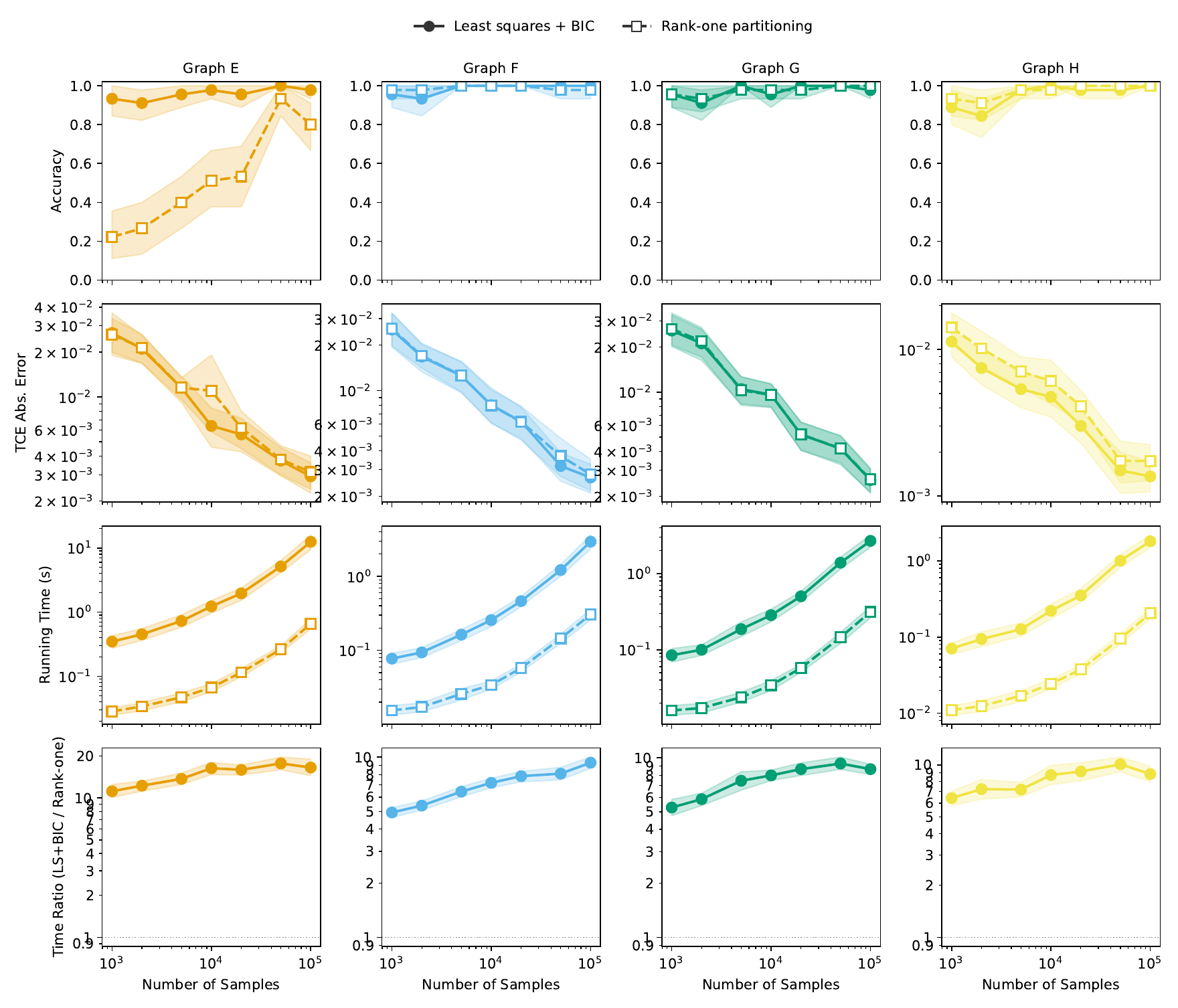}
\caption{%
    Comparison between \Cref{alg:maximalpartitioning} (solid) and rank-based partitioning (\Cref{alg:maximalpartitioning-plateau}, dashed), on Graphs E--H. For an increasing number of samples, we report the accuracy in recovering the exact ground-truth relevant sets, the absolute error in estimating the TCE of $Y_T$ on $Y_O$, the running time in seconds, and the ratio between the running times of the two rules on the same data. Both rules run on the same ground truths and samples.
}%
\label{fig:app-compare-rule}
\end{figure}

We compare rank-based partitioning (\Cref{app:rankonepartitioning}) with \Cref{alg:maximalpartitioning} by running both of them on the same ground truths and low-level samples, with $\size{\rel{Y}}\in\{2,4\}$, $\size{\block{Y}}\in\{4,8\}$, $\gamma=0.9$, and $15$ random seeds (\Cref{fig:app-compare-rule-abcd,fig:app-compare-rule}). We consider all Graphs A--H.

Rank-based partitioning is consistently faster, with running times on average between $4$ and $18$ times lower than \Cref{alg:maximalpartitioning} on the same data, and the gap generally grows with the number of samples. Given a large amount of data, it is as accurate as \Cref{alg:maximalpartitioning} on most graphs, e.g., from $n=10^4$ for Graphs A, C, and D, with accuracy between $0.91$ and $1.0$, and at every sample size for Graphs F--H, where it is even slightly more accurate on Graph H at the smallest sample sizes ($0.93$ against $0.89$ at $n=10^3$). However, with few samples, its accuracy drops on the graphs whose confounded variables are strongly determined by their regressors. At $n=10^3$, it reaches an accuracy of $0.60$ and $0.51$ on Graphs B and D, against $0.84$ and $0.98$ for \Cref{alg:maximalpartitioning}. On the deep Graph E, on the restricted grid of configurations of this comparison, its accuracy remains between $0.22$ and $0.53$ up to $n=2\cdot10^4$, and it reaches $0.93$ at $n=5\cdot10^4$ and $0.80$ at $n=10^5$, whereas \Cref{alg:maximalpartitioning} already reaches $0.93$ at $n=10^3$.

Despite these differences in accuracy, the two procedures lead to almost the same TCE error on every graph, e.g., between $0.011$ and $0.033$ at $n=10^3$ for both of them, which further shows that the TCE error alone does not distinguish correct from incorrect abstractions. Moreover, rank-based partitioning fails more often, up to $13\%$ of the runs on Graph E, against at most $2\%$ for \Cref{alg:maximalpartitioning}. Both procedures recover Graph H, since both keep only the unmediated relevant sets (\Cref{alg:datarse}, Lines 23--24). Overall, rank-based partitioning represents a viable, faster alternative only when the number of available samples is large relative to the complexity of the abstract graph.

\begin{figure}[htb]
\centering
\includegraphics[width=\textwidth]{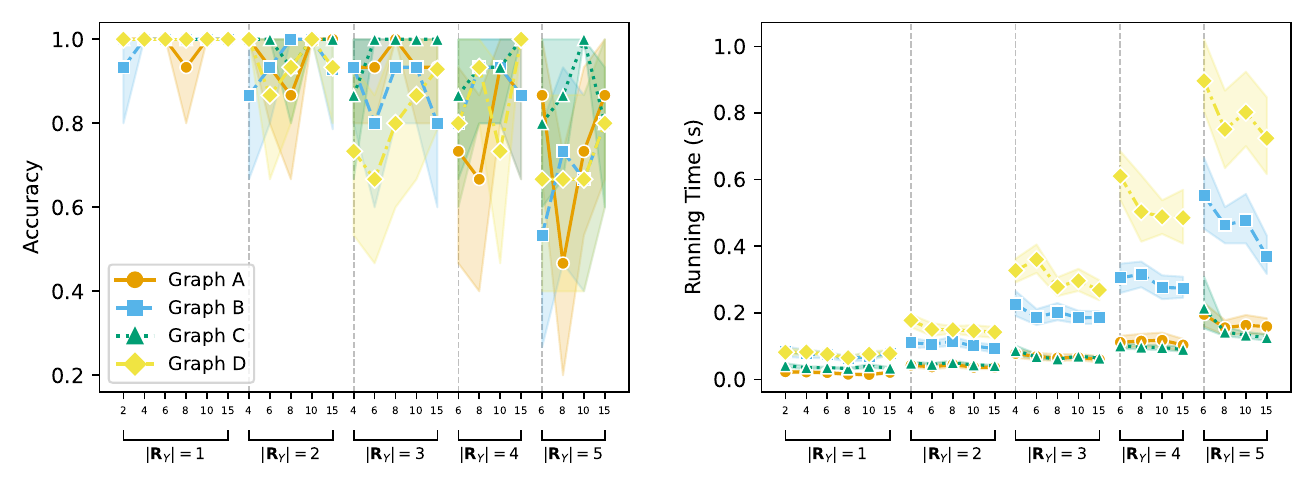}\\
\includegraphics[width=\textwidth]{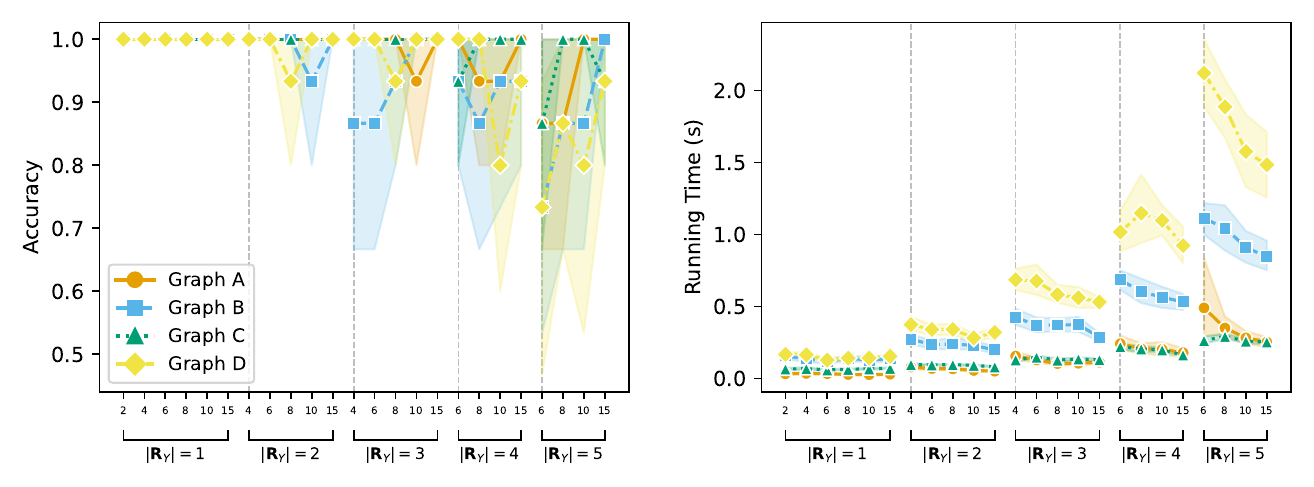}\\
\includegraphics[width=\textwidth]{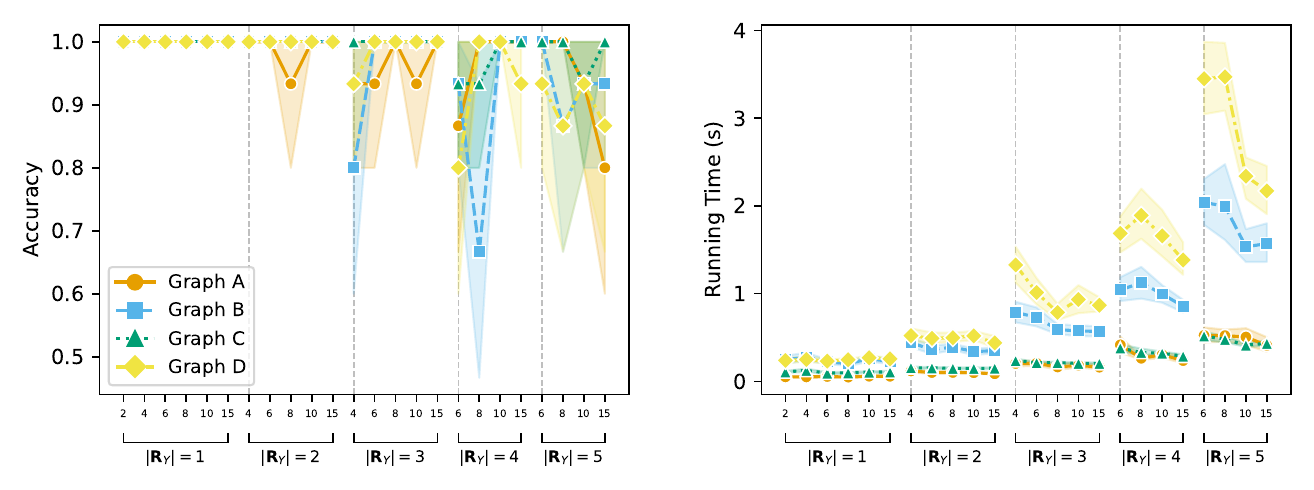}
\caption{%
    Accuracy and running time of \model{dataRSE} on the main configuration ($\gamma=0.9$, $\texttt{normalized}$ true), broken down by $(\size{\rel{Y}},\size{\block{Y}})$ combination as in \Cref{fig:size_data-rse}, at $n=10^3$ (top), $n=5\cdot10^3$ (center), and $n=10^4$ (bottom) samples. Accuracy degrades with $\size{\rel{Y}}$ at fixed $n$ (e.g., on average over the block sizes, from $0.99$--$1.00$ at $\size{\rel{Y}}=1$ down to $0.68$--$0.87$ at $\size{\rel{Y}}=5$, for $n=10^3$), and that degradation is progressively recovered as $n$ grows from $10^3$ to $10^4$, consistent with \Cref{fig:samples_data-rse}.
}%
\label{fig:app-ms90-combo}
\end{figure}

\begin{figure}[htb]
\centering
\includegraphics[width=\textwidth]{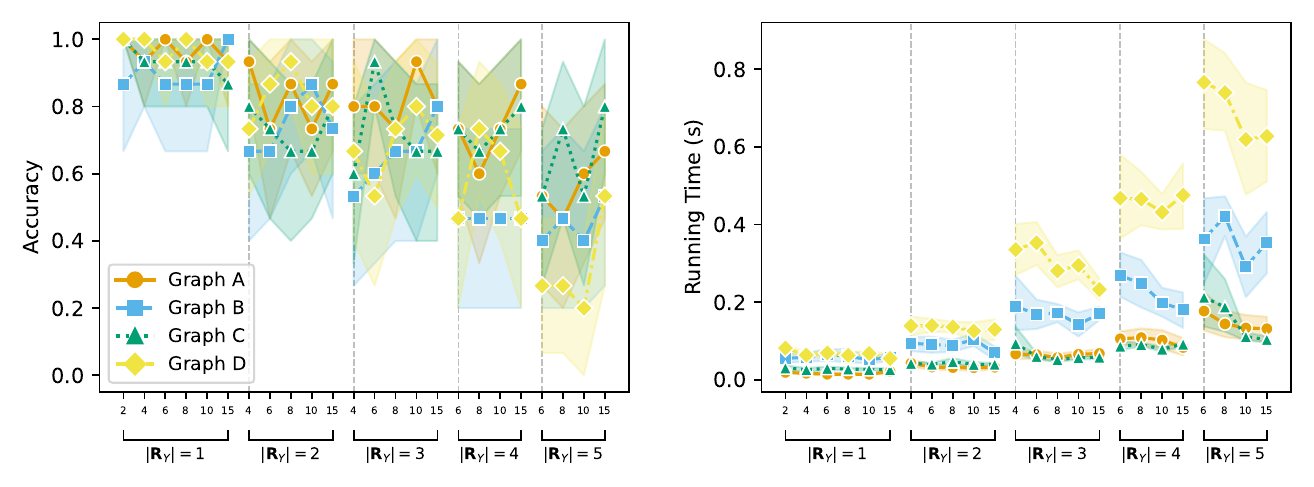}\\
\includegraphics[width=\textwidth]{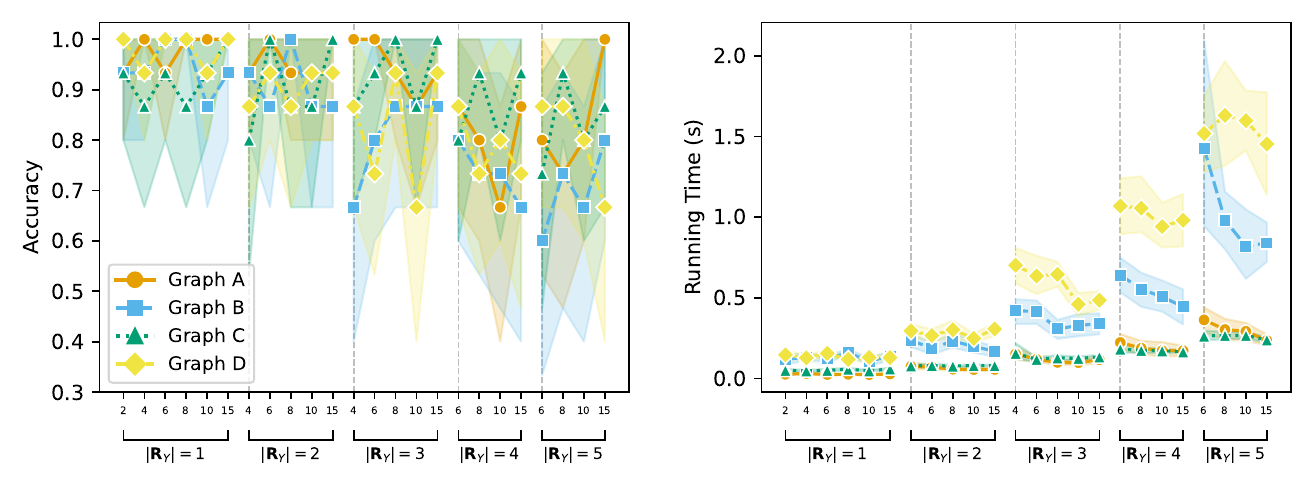}\\
\includegraphics[width=\textwidth]{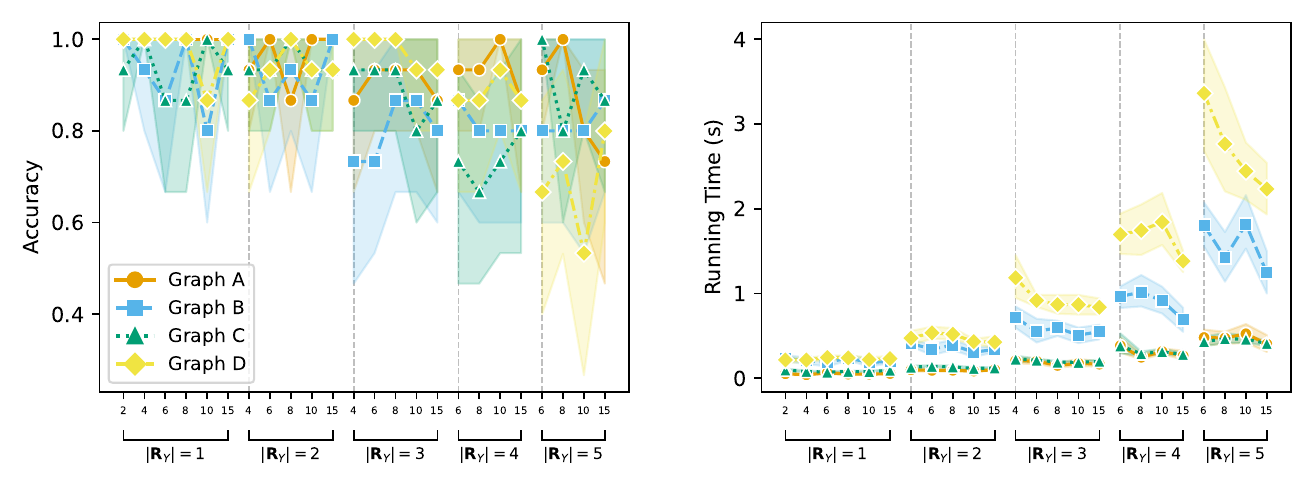}
\caption{%
    Same setting as \Cref{fig:app-ms90-combo}, with $\gamma=1.00$ (\ie, no rejection sampling). Unlike \Cref{fig:app-ms90-combo}, accuracy is degraded even at $\size{\rel{Y}}=1$ (on average over the block sizes, $0.90$--$0.97$ at $n=10^3$, against $0.99$--$1.00$ in \Cref{fig:app-ms90-combo}), and much more for larger relevant sets ($0.32$--$0.65$ at $\size{\rel{Y}}=5$). The degradation is recovered more slowly as $n$ grows from $10^3$ to $10^4$, consistent with \Cref{fig:app-ms100}.
}%
\label{fig:app-ms100-combo}
\end{figure}

\begin{figure}[htb]
\centering
\includegraphics[width=0.48\textwidth]{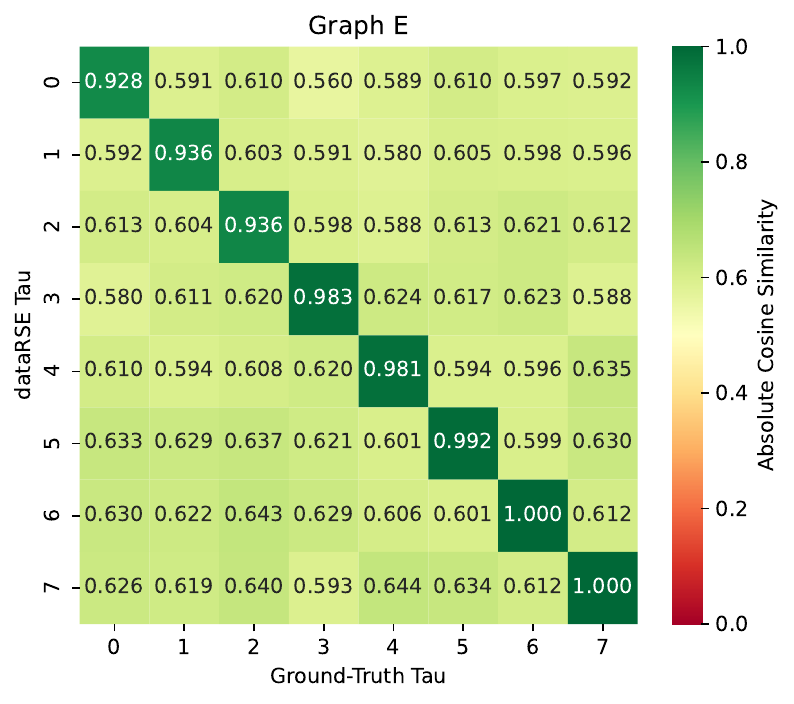}
\includegraphics[width=0.48\textwidth]{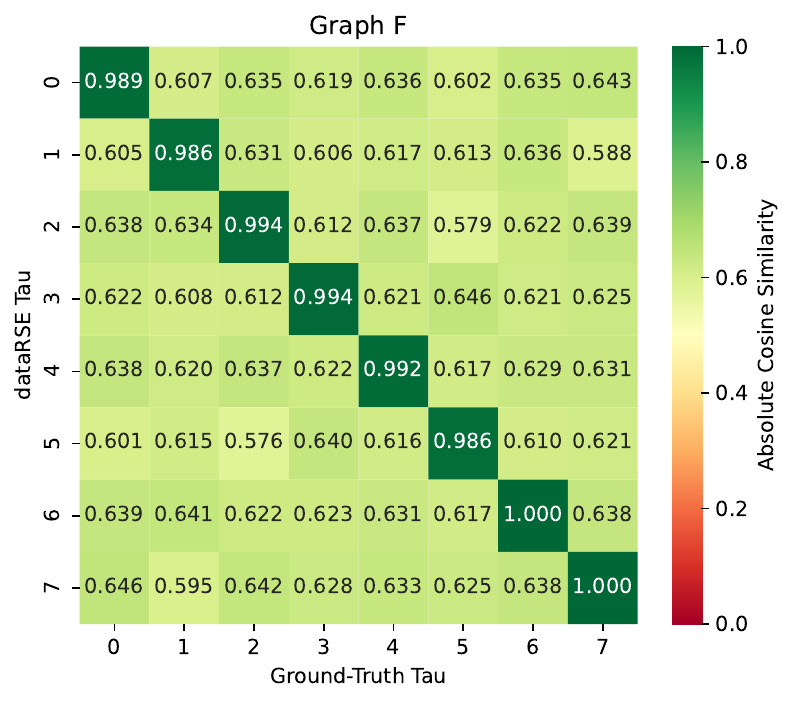}\\
\includegraphics[width=0.48\textwidth]{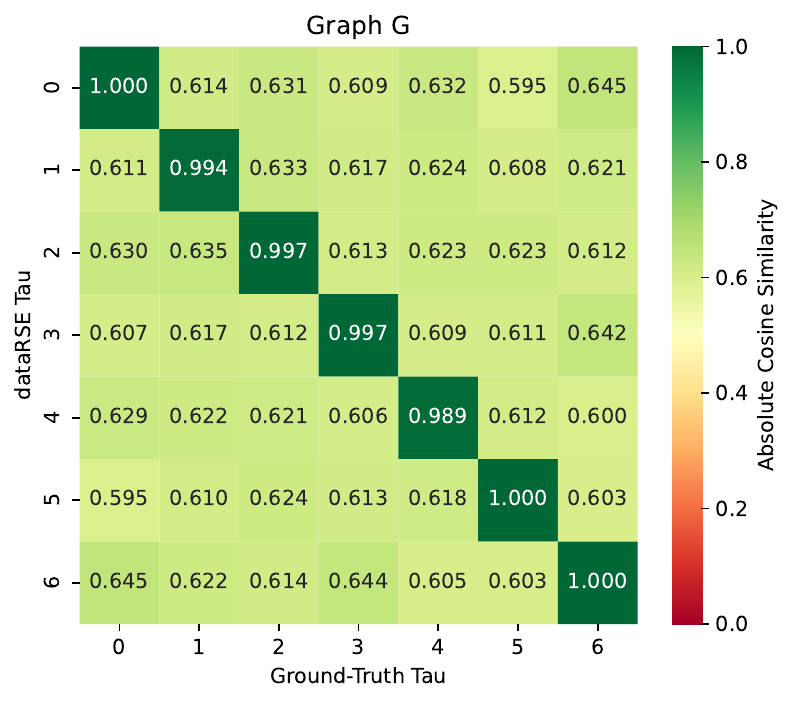}
\includegraphics[width=0.48\textwidth]{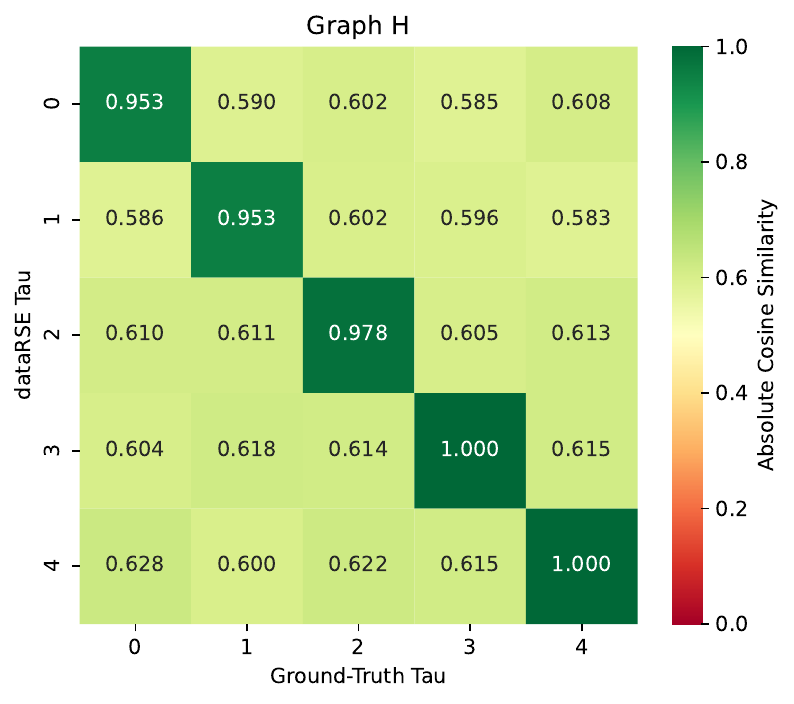}
\caption{%
    Abstraction-coefficient recovery for Graphs E--H at $n=5\cdot10^3$ samples, in the same format as \Cref{fig:tau_recovery}. Consistent with \Cref{fig:app-ijk,fig:app-ijk-full}, Graph E's diagonal is slightly weaker for the abstract variables recovered in its deepest recursive calls ($0.93$--$0.94$, against $0.98$--$1.0$ for the others), whereas Graphs F and G, which need fewer recursive calls, reach $0.98$--$1.0$ for every abstract variable, and Graph H reaches $0.95$--$1.0$. As for Graphs A--D, the runs that recover the ground-truth relevant variables also recover the abstraction vectors up to scale and sign, hence the weaker entries of the diagonal are due to the incorrect solutions, while the off-diagonal entries are at the similarity level of independent abstraction vectors.
}%
\label{fig:app-ijk-tau}
\end{figure}

\end{document}